\documentclass{article}
\usepackage{iclr2027_conference,times}

\usepackage{amsmath,amsfonts,bm}

\def\eqref#1{equation~\ref{#1}}

\def\1{\bm{1}}

\DeclareMathAlphabet{\mathsfit}{\encodingdefault}{\sfdefault}{m}{sl}
\SetMathAlphabet{\mathsfit}{bold}{\encodingdefault}{\sfdefault}{bx}{n}

\usepackage{hyperref}
\usepackage{url}
\usepackage{booktabs}
\usepackage{pifont}
\usepackage{tabularx}
\usepackage{xcolor}
\usepackage{colortbl}
\usepackage{graphicx}
\usepackage{wrapfig}
\usepackage{flafter}
\usepackage{needspace}
\usepackage{placeins}
\usepackage{enumitem}
\usepackage{multirow}
\usepackage[most]{tcolorbox}

\newcommand{\dataset}{WorldAuditBench}

\iclrfinalcopy
\definecolor{preprintblue}{HTML}{0057A8}
\definecolor{abstractbackground}{HTML}{F5F8FA}
\hypersetup{colorlinks=true,linkcolor=preprintblue,citecolor=preprintblue,urlcolor=preprintblue,
 pdftitle={WorldAuditBench: Interactive 3D World Auditing with Multimodal Agents},
 pdfauthor={Ziyan Jiang, Jingbo Yang, Jiabao Ji, Yujian Liu, Qiucheng Wu, Tommi Jaakkola, Yang Zhang, Shiyu Chang}}
\renewcommand{\headrulewidth}{0.4pt}
\fancypagestyle{preprintfirst}{%
 \fancyhf{}
 \fancyhead[R]{\small Preprint.}
 \fancyfoot[C]{\thepage}
 \renewcommand{\headrulewidth}{0.4pt}}
\newcommand{\preprinttitle}{%
 \thispagestyle{preprintfirst}
 \begin{center}
 {\fontsize{17}{20}\selectfont\bfseries
 \raisebox{-0.15em}{\includegraphics[height=1.05em]{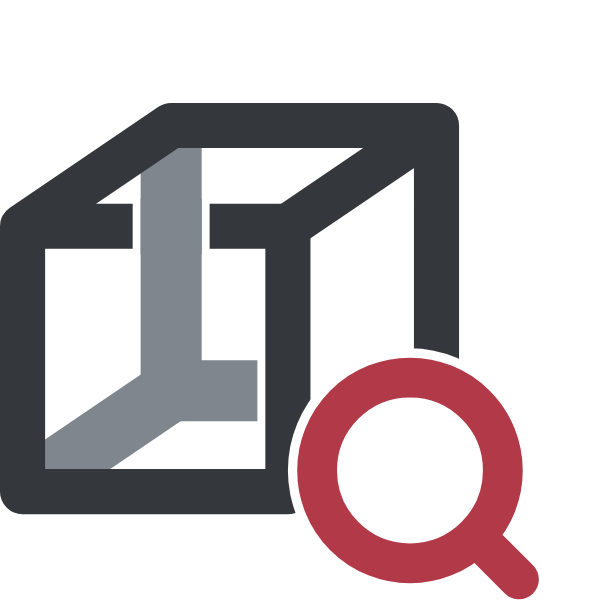}}WorldAuditBench: Interactive 3D World\\
 Auditing with Multimodal Agents\par}
 \medskip
 {\normalsize\bfseries
 Ziyan Jiang\textsuperscript{1*}\quad Jingbo Yang\textsuperscript{1*}\quad Jiabao Ji\textsuperscript{1*}\quad Yujian Liu\textsuperscript{1}\\[4pt]
 Qiucheng Wu\textsuperscript{1}\quad Tommi Jaakkola\textsuperscript{2}\quad Yang Zhang\textsuperscript{3}\quad Shiyu Chang\textsuperscript{1}\par}
 \smallskip
 {\small \textsuperscript{1}UC Santa Barbara\quad \textsuperscript{2}MIT CSAIL\quad \textsuperscript{3}MIT-IBM Watson AI Lab\par}
 {\footnotesize \textsuperscript{*}Equal contribution.\par}
 \end{center}%
 \begingroup
 \renewcommand{\thefootnote}{\fnsymbol{footnote}}%
 \footnotetext[1]{Correspondence:
 \texttt{\textcolor{preprintblue}{\{}\href{mailto:ziyanjiang@ucsb.edu}{ziyanjiang}\textcolor{preprintblue}{, }\href{mailto:jingbo@ucsb.edu}{jingbo}\textcolor{preprintblue}{, }\href{mailto:jiabaoji@ucsb.edu}{jiabaoji}\textcolor{preprintblue}{\}@ucsb.edu}}.}%
 \endgroup}
\newenvironment{preprintabstract}{%
 \begin{tcolorbox}[colback=abstractbackground,colframe=abstractbackground,
 boxrule=0pt,arc=3pt,left=10pt,right=10pt,top=8pt,bottom=9pt,
 before skip=8pt,after skip=12pt]
 {\centering\large\bfseries Abstract\par}\smallskip
 \small\noindent\ignorespaces
}{%
 \par\medskip
 {\centering\normalsize\href{https://ucsb-nlp-chang.github.io/WorldAuditBench/}{\textbf{Project Page}}\par}
 \end{tcolorbox}}

\begin{document}
\preprinttitle
\begin{preprintabstract}
As interactive 3D worlds are increasingly used to study intelligent behavior, it becomes important to develop efficient pipelines for identifying anomalies in these simulated environments, such as floating objects, traversable walls, or objects inconsistent with the surrounding scene. Multimodal AI systems, including vision-language models (VLMs) and vision-language-action models (VLAs), have shown potential for automating this task. However, 3D world auditing is complex, requiring the close coupling of two distinct capabilities: \textbf{action}, to navigate the 3D world and search for anomalies systematically and efficiently; and \textbf{visual reasoning}, to understand the environment and identify anomalies from multimodal observations. It remains largely unexplored whether multimodal agents can effectively couple these two capabilities---using visual reasoning to identify potential anomalies while taking actions to validate them. In this paper, we introduce \textbf{\dataset{}}, a benchmark for 3D world auditing comprising 213 anomaly tasks across 13 environments built with Unreal Engine 5 and Three.js, spanning five anomaly families. We evaluate five frontier models under a fixed exploration budget using two auditing paradigms: VLA-based exploration followed by VLM-based anomaly identification, and an end-to-end VLM agent in which visual reasoning directly guides action selection. Across the evaluated models and two paradigms, success rates range from 6.6\% to 42.3\%, substantially below human performance (83.4\%). Through the task of world auditing, \dataset{} provides a testbed for studying how multimodal agents couple action and visual reasoning in interactive 3D environments, while highlighting current limitations in their ability to gather and interpret evidence during exploration.
\end{preprintabstract}
\suppressfloats[t] 

\begin{figure}[!t]
\centering
\includegraphics[width=1.0\linewidth]{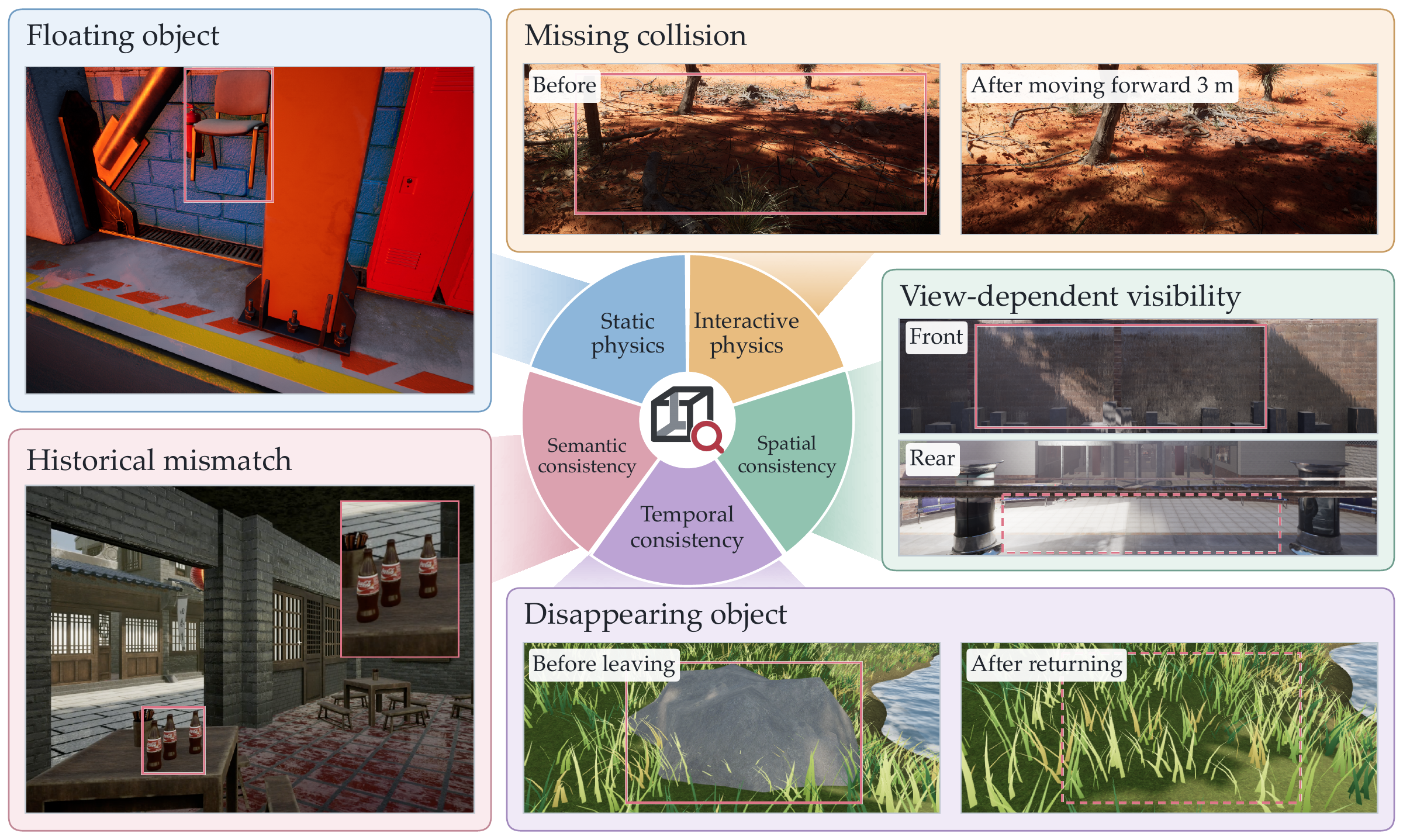}
\caption{\textbf{Anomaly taxonomy of \dataset{} (Section~\ref{sec:taxonomy}).} Representative cases illustrate five families of physical, spatial, temporal, and semantic inconsistencies.}
\label{fig:taxonomy}
\vspace{-12pt}
\end{figure}

\section{Introduction}

Interactive 3D worlds allow human players and agents to explore, interact with objects, and complete tasks in realistic simulated environments. Recent work has expanded their use in embodied AI, from navigation and manipulation to open-ended exploration \citep{yang2024holodeck,xie2025vid2sim,lee2025dynscene,yang2025threedmem,ye2025simworld}. Yet a world can look convincing while still containing anomalies. For example, a chair may float above the floor, a solid wall may allow an agent to walk through it, or an object may disappear after the agent looks away. Recent works highlight such visual and physical defects, while others develop methods to improve object placement, collision handling, and physical consistency \citep{yang2024physcene,zhou2026layoutdreamer,jia2025cluttergen,lin2025pat3d,taesiri2025videogameqa}. Such anomalies matter wherever interactive worlds are used to study intelligent behavior. For example, an agent that succeeds at certain tasks by walking through a broken wall should not be seen as reliable in navigation. Checking these environments therefore requires more than inspecting how they look, since some defects become apparent only through interaction or repeated exploration in the environments. This raises a natural question: \emph{Can multimodal agents autonomously identify anomalies in a 3D world, \textit{i.e.}, perform \textbf{world auditing}?}

World auditing requires two skills: \textbf{action}, to navigate the 3D world and search for anomalies systematically and efficiently, and \textbf{visual reasoning}, to understand the scene and identify anomalies from observations. 
More importantly, these two skills are deeply coupled.
The agent must search large spaces within a limited exploration budget while distinguishing real defects from unusual but valid appearances or behavior. An observation may reveal a possible anomaly without providing enough evidence to confirm it. The agent must then choose an action to further investigate and decide whether a real defect is present. Auditing therefore requires both skills to work together: visual reasoning guides what to test, and action provides further evidence for the diagnosis.
For example, the fence in Figure~\ref{fig:taxonomy} \textit{Missing Collision Anomaly} appears to be a solid barrier, but its collision defect becomes apparent only through interaction. The agent must navigate toward the fence and attempt to cross it. Comparing views in the same direction before and after moving straight ahead by 3\,m reveals that the agent passed through the fence without being blocked. This process connects visual reasoning and action: recognizing the barrier establishes an expectation about blocked movement, moving forward tests it, and the resulting observation provides evidence of the collision failure.

Existing benchmarks evaluate visual reasoning over fixed observations or interactive task completion~\citep{taesiri2025videogameqa,yang2025embodiedbench,majumdar2024openeqa}, while interactive anomaly search covers out-of-place objects~\citep{pardyl2025flysearch}. Comprehensive evaluation of world auditing remains underexplored: agents must act to gather evidence and identify diverse anomalies, including those revealed only through interaction or comparison across viewpoints and time.

\ifdefined\preprinttitle
\begin{table}[!t]
\caption{\textbf{Comparison with related benchmarks and simulation environments.} \dataset{} evaluates interactive anomaly identification across 15 categories grouped into five families.}
\label{tab:benchmark_comparison}
\centering\scriptsize
\setlength{\tabcolsep}{3.5pt}
\renewcommand{\arraystretch}{1.08}
\newcommand{\benchyes}{\textcolor[HTML]{28766E}{\ding{51}}}
\newcommand{\benchno}{\textcolor{black!45}{\ding{55}}}
\begin{tabularx}{0.95\linewidth}{@{}l>{\raggedright\arraybackslash}Xcc>{\raggedright\arraybackslash}p{0.20\linewidth}@{}}
\toprule
\textbf{Work} & \textbf{Task family} & \shortstack{\textbf{World}\\\textbf{interaction}} & \shortstack{\textbf{Anomaly}\\\textbf{identification}} & \textbf{Anomaly coverage} \\
\midrule
\multicolumn{5}{@{}l}{\textit{Offline visual benchmarks}} \\
Video-MME~\citeyearpar{fu2025videomme} & Question answering & \benchno & \benchno & --- \\
GlitchBench~\citeyearpar{taesiri2024glitchbench} & Game testing & \benchno & \benchyes & Game glitches \\
VideoGameQA-Bench~\citeyearpar{taesiri2025videogameqa} & Game testing & \benchno & \benchyes & Game glitches \\
VideoGlitchBench~\citeyearpar{zheng2026videoglitchbench} & Game testing & \benchno & \benchyes & Game glitches \\
\midrule
\multicolumn{5}{@{}l}{\textit{Interactive environments and benchmarks}} \\
SimWorld~\citeyearpar{ren2025simworld} & Task execution & \benchyes & \benchno & --- \\
UnrealZoo~\citeyearpar{zhong2025unrealzoo} & Task execution & \benchyes & \benchno & --- \\
EmbodiedBench~\citeyearpar{yang2025embodiedbench} & Task execution & \benchyes & \benchno & --- \\
OpenEQA (active)~\citeyearpar{majumdar2024openeqa} & Question answering & \benchyes & \benchno & --- \\
FlySearch~\citeyearpar{pardyl2025flysearch} & Target search & \benchyes & \benchyes & Out-of-place objects \\
\midrule
\rowcolor[HTML]{EDF5F3}
\multicolumn{1}{@{}>{\columncolor[HTML]{EDF5F3}[0pt][\tabcolsep]}l}{\textbf{\dataset{} (ours)}} & World auditing & \benchyes & \benchyes & \multicolumn{1}{>{\columncolor[HTML]{EDF5F3}[\tabcolsep][0pt]\raggedright\arraybackslash}p{0.20\linewidth}@{}}{15 categories (\S\ref{sec:taxonomy})} \\
\bottomrule
\end{tabularx}
\end{table}

\fi



To fill this gap, we introduce \dataset{}, comprising 213 anomaly tasks across 13 environments implemented in Unreal Engine and Three.js. As illustrated in Figure~\ref{fig:taxonomy}, these environments cover indoor, city, historical, industrial, and natural settings. We organize anomalies into five families: static physics, interactive physics, spatial consistency, temporal consistency, and semantic consistency. Together, these families cover both visible defects and anomalies that require interaction or repeated observation to detect. Each task pairs an anomaly with its expected normal behavior, relevant scene context, and an evaluation rubric. To solve a task, the auditor explores the scene from a first-person perspective and submits a report describing the anomaly and supporting observations. Table~\ref{tab:benchmark_comparison} compares \dataset{} with prior benchmarks and simulation environments, highlighting its focus on interactive anomaly identification across diverse anomaly families.

To study how action and visual reasoning work together in world auditing, we evaluate two auditor setups: \ding{192} a \textit{single VLM agent auditor}, which uses a single multimodal LLM to inspect observations, select actions, and report anomalies; and \ding{193} a \textit{two-stage VLA--VLM agent auditor}, which uses a VLA model to explore the environment and then passes the resulting trajectory to a VLM for offline anomaly analysis. The key difference is whether anomaly analysis can guide exploration process: the VLM agent can adjust its actions to investigate suspected anomalies, while the two-stage auditor analyzes a fixed trajectory, resembling a conventional video QA setup.

Across the 213 evaluation tasks, \textit{VLM agents} achieve success rates of 28.2\% to 42.3\%, compared with 6.6\% to 17.4\% for \textit{two-stage VLA--VLM auditors}. These results suggest that using observations to guide further exploration helps agents gather evidence to verify suspected defects. However, both approaches fall well below the human success rate of 83.4\%, leaving a substantial gap in their ability to audit interactive worlds.

To better understand this gap, we evaluate five model backbones and conduct controlled ablations on environment initialization and in-context examples. Further analysis identifies difficulties in both exploration and the use of past observations. The evaluated VLA models struggle to reach anomalies in large 3D worlds, limiting the evidence available for subsequent analysis. For VLM agents, the memory tool that lets them store and retrieve observations during exploration substantially improves anomaly discovery. These findings suggest that world auditing benefits from both exploration guided by ongoing analysis and access to evidence gathered earlier in the trajectory.

Our contributions are listed below:
\begin{itemize}[leftmargin=*]
\item \textbf{A comprehensive benchmark for interactive world auditing.} We introduce \dataset{}, with 213 tasks across 13 environments, covering static physics, interactive physics, spatial consistency, temporal consistency, and semantic consistency. These tasks evaluate anomaly identification through visual inspection, interaction, and repeated observation.
\item \textbf{A detailed analysis of world auditing agents.} We compare VLM agents with two-stage VLA--VLM auditors to examine their exploration and reasoning abilities. Our analysis further highlights exploration failures in VLA models and the benefits of memory control for VLM agents, while documenting a substantial performance gap relative to human auditors.
\end{itemize}

\ifdefined\preprinttitle\else

\fi

\section{Related Work}
\textbf{Visual Understanding and Anomaly Detection.} Existing works mainly evaluate agents' multimodal abilities using static visual assets. For example, Video-MME focuses on prerecorded videos \citep{fu2025videomme}, while VideoGameBunny studies game video clips \citep{taesiri2025videogamebunny}. GlitchBench, VideoGameQA-Bench, and VideoGlitchBench further evaluate glitch recognition, localization, description, and reporting from images or videos \citep{taesiri2024glitchbench,taesiri2025videogameqa,zheng2026videoglitchbench}. However, these benchmarks provide observations to the model in advance and mainly test whether agents can answer questions reliably based on available visual evidence. In contrast, our work focuses on the more challenging setting of interactive world auditing, where agents must actively gather the evidence needed to expose and verify anomalies.

\textbf{Interactive Embodied Reasoning and Anomaly Search.}
Existing works also evaluate agents' embodied abilities in interactive environments. For example, AI2-THOR~\citep{kolve2017ai2thor}, Habitat~\citep{savva2019habitat}, UnrealCV~\citep{qiu2016unrealcv}, SimWorld~\citep{ren2025simworld}, and UnrealZoo~\citep{zhong2025unrealzoo} provide 3D environments for embodied perception and interaction, while EmbodiedBench~\citep{yang2025embodiedbench}, VisualAgentBench~\citep{liu2024visualagentbench}, and OpenEQA~\citep{majumdar2024openeqa} evaluate agents on interactive task execution and information gathering. 
More closely related to our work, FlySearch~\citep{pardyl2025flysearch} includes an anomaly-search task that requires agents to locate out-of-place objects in 3D environments. However, they mainly focus on task completion or locating visually identifiable anomalies. In contrast, our work studies anomalies that may require agents to actively test physical properties or compare observations across viewpoints and time, making evidence gathering itself an essential part of anomaly identification.

\section{\dataset{} Dataset}
\label{sec:dataset}

This section introduces \dataset{} and its construction. Section~\ref{subsec:overview} provides an overview, Section~\ref{sec:taxonomy} presents our anomaly taxonomy, and Section~\ref{sec:data-collection} describes the data collection process.

\subsection{Dataset Overview}
\label{subsec:overview}

\dataset{} consists of 213 tasks built on 13 interactive 3D environments, each task implanted with a \textbf{single} pre-defined, labeled anomaly. An agent explores and interacts with the environment, and its goal is to correctly identify the anomaly based on its observations. The environments span five scene types (indoor, urban, historical, industrial, and natural) and are built with Unreal Engine 5\footnote{\url{https://www.unrealengine.com}} (126 tasks) or Three.js\footnote{\url{https://threejs.org}} (87 tasks). Appendix~\ref{app:environments} provides further details.

\subsection{Anomaly Taxonomy}
\label{sec:taxonomy}

Our anomaly taxonomy draws on game-bug studies that distinguish failures visible in a single state from those requiring observations over time~\citep{lewis2010taxonomy}, and describe failures in object position, collision, rendering, persistence, and state transitions~\citep{truelove2021fix,butt2023taxonomy}. We group anomalies by their observable effects into five families (Figure~\ref{fig:taxonomy}): 






\begin{itemize}[leftmargin=*, nosep]
    \item \textbf{Static physics:} Anomalies in which static objects disobey physics, involving object support, overlap, or scale;
    \item \textbf{Interactive physics:} Anomalies in which interactions between objects or between the agent and the environment disobey physics, involving obstruction and objects' responses to contact;
    \item \textbf{Spatial consistency:} Anomalies in which objects are inconsistent across positions or viewpoints, such as appearance changes in an object as the observer moves;
    \item \textbf{Temporal consistency:} Anomalies in which objects display unexpected changes over time, such as changes in existence, attributes, and behavior;
    \item \textbf{Semantic consistency:} Anomalies in which object configurations are inconsistent with the semantic context, involving the intended functionality of the object and historical context.
\end{itemize}

\begin{figure}[!t]
\centering
\includegraphics[width=\linewidth]{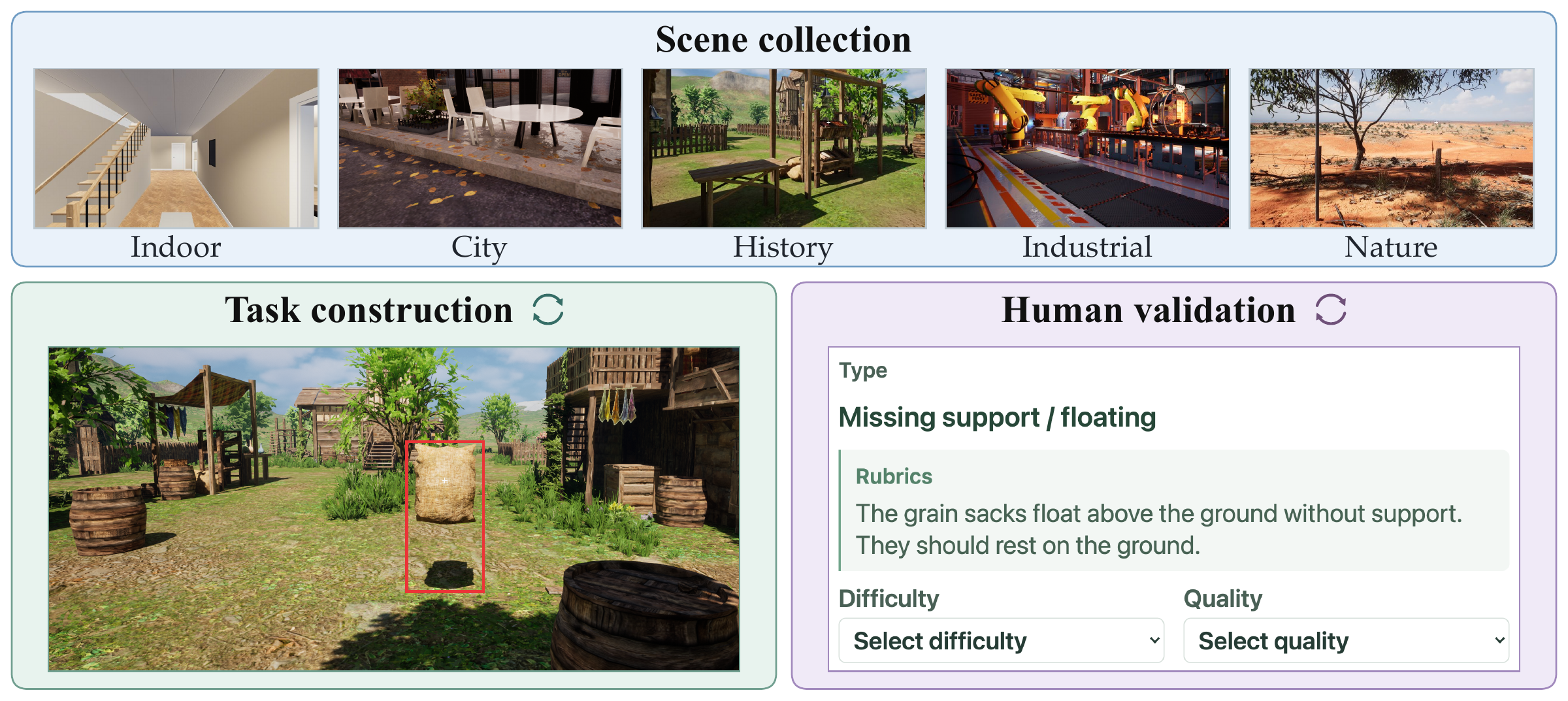}
\caption{\textbf{Task construction in \dataset{}.} We collect scenes, introduce anomalies, and iteratively refine tasks through human review of their quality.}
\label{fig:data-collection}
\end{figure}

The five families of anomaly are further divided into fifteen categories. Appendix~\ref{app:full-examples} provides full definitions of each category and some examples. Anomaly families differ in the evidence required to detect them: some are visible in a single view, such as those in the static physics family, while others can only be exposed through deliberate actions, such as interacting with objects, changing viewpoints, or revisiting a location after a period of time. Together, these families form a comprehensive test of planning, memory, and scene understanding.

\subsection{Task Construction}
\label{sec:data-collection}
To construct tasks based on the aforementioned taxonomy, we follow three steps: scene collection, task construction, and human validation (Figure~\ref{fig:data-collection}).

\textbf{Scene Collection.} We collect 13 interactive 3D environments built with Unreal Engine 5 and Three.js from publicly available sources, covering indoor, urban, historical, industrial, and natural settings. These environments range from furnished rooms and urban streets to historical markets and outdoor landscapes. For larger environments, we manually select several bounded regions as individual scenes, seeking distinct spatial layouts and scene content, yielding 27 scenes in total. Appendix~\ref{app:environments} presents the environments and representative views.

\textbf{Task Construction.} All tasks are hand-crafted by the authors. To construct a task, we first select a bug-free scene, where we further define permissible regions for exploration, the starting position and orientation of the agent, and available interactions. Based on the taxonomy, we then introduce an anomaly by perturbing the scene, such as changing object placement, collision, appearance, responses to interaction, or behavior over time. For semantic anomalies, we rearrange objects in ways that conflict with their intended function or add objects that do not fit the scene's historical setting. Appendix~\ref{app:construction} details the construction procedure. Each task is annotated with the ground-truth anomaly category along with an evaluation rubric describing the target anomaly and its expected normal appearance or behavior.
Appendix~\ref{app:annotations} presents an example.

\textbf{Human Validation.}
To verify task quality, judges who did not participate in the task construction process, are asked to explore the scene in each task to check whether the target anomaly can be observed and verified and whether it matches the annotations. They rate task quality as Pass, Fail, or Uncertain. Fail or Uncertain judgments lead to further revision and review. This process repeats until the task meets the acceptance requirement: \textit{Rated Pass by at least \textbf{two judges} on the same task version}. Appendix~\ref{app:human-annotation} describes the review protocol and annotation interface.

\newcommand{\auditorModesFigure}{%
\begin{figure}[!t]
\centering
\includegraphics[width=\linewidth]{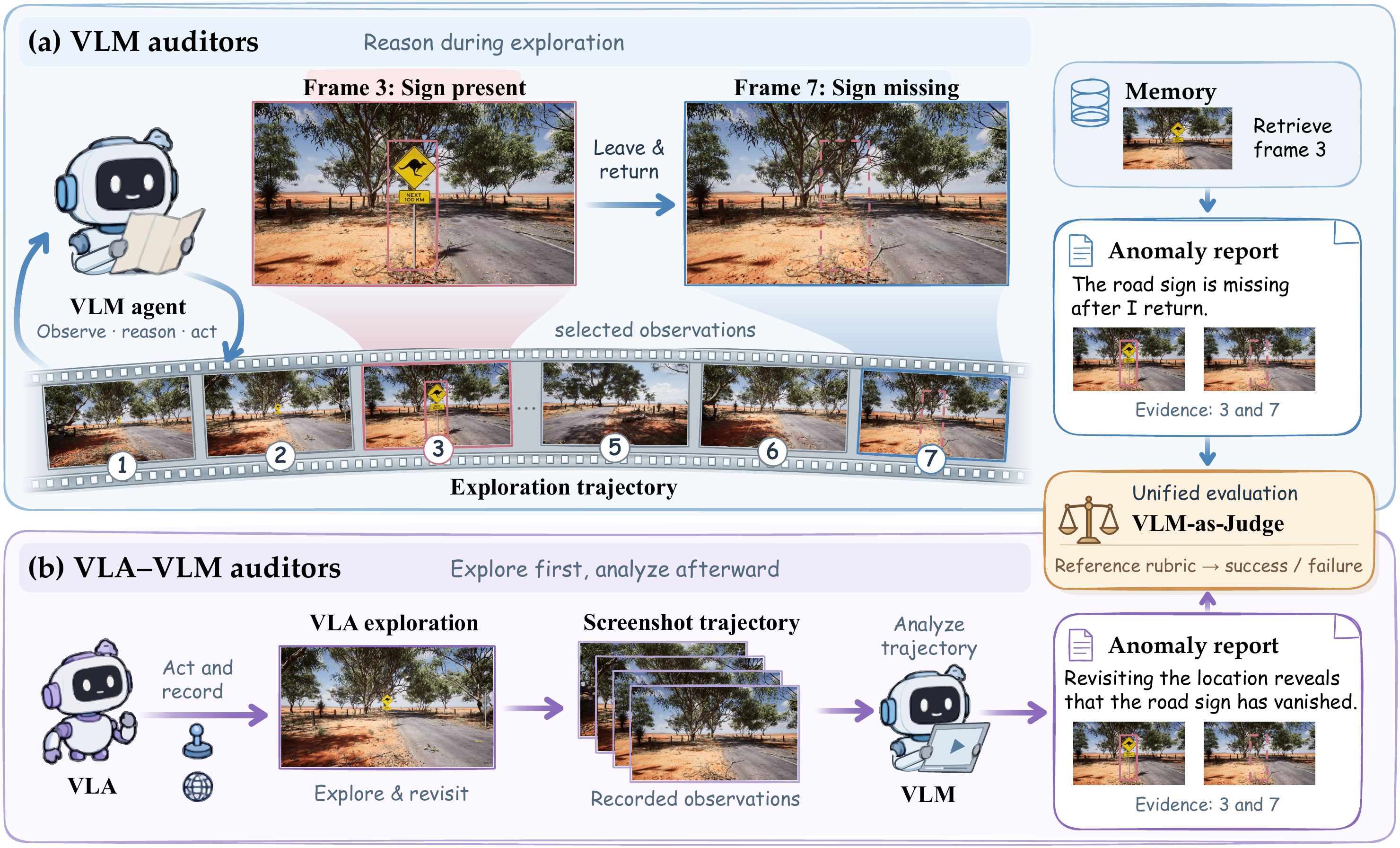}
\caption{\textbf{Two auditing paradigms.} (a) VLM agents reason and use tools during exploration. (b) VLA--VLM auditors analyze trajectories after VLA exploration. Both use the same judge.}
\label{fig:auditor-modes}
\vspace{-6pt}
\end{figure}
}

\section{Auditing Paradigms}
\label{sec:auditing-paradigms}

To study how the coupling of \textbf{action} and \textbf{visual reasoning} affects 3D world auditing, we evaluate two paradigms: VLM-only and VLA--VLM (Figure~\ref{fig:auditor-modes}). We first define the task for the auditing agent in Section~\ref{sec:task-definition}, and then describe these two paradigms in Sections~\ref{subsec:vlm} and \ref{subsec:vla-vlm}.

\ifdefined\preprinttitle
\auditorModesFigure
\fi

\subsection{Task Definition}\label{sec:task-definition}

At the start of a task, the auditor receives an initial input $x$ containing auditing instructions, task context, and an initial observation. The task context includes a scene description, an anomaly-type hint, and a matching in-context example. The initial observation is the first-person RGB view from the starting pose. From this input, the auditor explores the environment under a fixed budget and produces a final anomaly report containing the identified anomaly and supporting observation images during its exploration and interaction within the 3D world.
Each task in the main evaluation contains \textit{only one} target anomaly, and success is determined by whether the report identifies it.
In both paradigms, the agent submits an anomaly report and evidence to the VLM judge, which evaluates them against the task's evaluation rubric, such as identifying a solid wall that the agent can pass through. Appendices~\ref{app:case-output} and~\ref{app:judge} illustrate an example report and its evaluation process.

To formalize this interaction, we represent the trajectory of an auditor agent as
\begin{equation}
\tau=(x,a_0,o_1,\ldots,a_{T-1},o_T,y),
\label{eq}
\end{equation}
where $a_t$ is an environment action, $o_{t+1}$ is the resulting observation, and $y$ is the final anomaly report. As with the initial observation, each subsequent observation contains a first-person RGB view and available interaction hints, \textit{e.g.}, that the agent can open or close a door. The auditor selects each action based on the initial input and the preceding actions and observations. 




\ifdefined\preprinttitle\else
\auditorModesFigure
\fi

\subsection{VLM-Only Paradigm}\label{subsec:vlm}
In the VLM-only paradigm, a VLM auditor receives the initial input $x$ (Appendix~\ref{app:case-input}). At step $t$, it reasons over the trajectory collected so far, including the current observation $o_t$, to choose the next environment action $a_t$. A suspected anomaly can therefore guide where the agent moves, which viewpoint it examines, or which interaction it attempts. Visual reasoning and action remain coupled as the trajectory is collected.

The agent uses \textbf{environment actions} to navigate, change its view, interact with objects, and observe changes over time; \textbf{memory tools} to retrieve earlier frames, review history, and save notes; and \textbf{anomaly-reporting tools} to record or revise findings and link supporting observations. Figure~\ref{fig:auditor-modes}(a) illustrates revisiting a roadside location, retrieving an earlier observation, and reporting a missing sign using the matched before-and-after views as evidence. Tool definitions are shown in Table~\ref{tab:tools}.

\subsection{VLA--VLM Paradigm}\label{subsec:vla-vlm}
The VLA--VLM paradigm separates exploration from anomaly analysis (Figure~\ref{fig:auditor-modes}(b)). A VLA model first generates environment actions $a_t$ and collects a trajectory of observations. Due to the limitations of existing VLA models, no explicit reasoning is produced during exploration. A VLM then analyzes the recorded observations together with the task context to identify anomalies and produce a report with supporting evidence. In our experiments, we use fixed recorded trajectories recorded by the same VLA model for second-stage analysis, which is conducted by different VLM backbones.
Because the VLM analyzes the trajectory only after exploration is complete, its intermediate observations cannot guide further actions to investigate potential anomalies, which may limit its auditing performance.





\begin{table}[!t]
\centering
\begingroup
\caption{\textbf{Auditing performance and cost on \dataset{}.} SR (\%) and cost (\$/task). SP: static physics; IP: interactive physics; SpC: spatial consistency; TC: temporal consistency; SeC: semantic consistency. \textbf{Bold} and \underline{underline} mark the best and second-best results.}
\label{tab:main}
\centering\small
\setlength{\tabcolsep}{4pt}
\renewcommand{\arraystretch}{1.04}
\newcommand{\mainscore}[1]{\makebox[2em][r]{#1}}
\newcommand{\maincost}[1]{\makebox[2.5em][r]{#1}}

\begin{tabular}{@{}l*{5}{>{\centering\arraybackslash}p{2.7em}}>{\columncolor{black!3}}c c@{}}
\toprule
Auditor & SP & IP & SpC & TC & SeC & \textbf{Overall} $\uparrow$ & Cost $\downarrow$ \\
\midrule
\rowcolor{black!4}\multicolumn{1}{@{}>{\columncolor{black!4}[0pt][\tabcolsep]}l}{Human} & \mainscore{89.8} & \mainscore{81.3} & \mainscore{76.3} & \mainscore{78.1} & \mainscore{88.6} & \mainscore{83.4} & \multicolumn{1}{>{\columncolor{black!4}[\tabcolsep][0pt]}c@{}}{N/A} \\
\midrule
\multicolumn{8}{@{}l}{\textbf{VLM auditors}} \\
\addlinespace[2pt]
\makebox[1.3em][c]{\raisebox{-0.15em}{\includegraphics[height=1.05em]{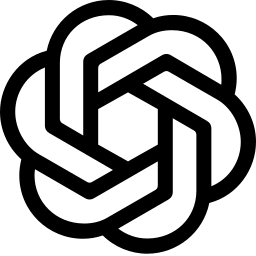}}}\ GPT-6 Astra & \mainscore{\textbf{59.3}} & \mainscore{\textbf{29.3}} & \mainscore{\textbf{45.1}} & \mainscore{\textbf{12.5}} & \mainscore{\textbf{68.2}} & \mainscore{\textbf{42.3}} & \maincost{2.955} \\
\makebox[1.3em][c]{\raisebox{-0.15em}{\includegraphics[height=1.05em]{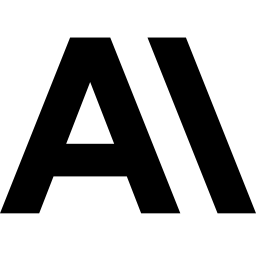}}}\ Claude Opus 5 & \mainscore{35.6} & \mainscore{19.5} & \mainscore{\underline{29.4}} & \mainscore{\textbf{12.5}} & \mainscore{\underline{50.0}} & \mainscore{28.2} & \maincost{2.445} \\
\makebox[1.3em][c]{\raisebox{-0.15em}{\includegraphics[height=1.05em]{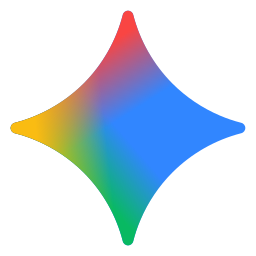}}}\ Gemini 3.8 Flash & \mainscore{\underline{49.2}} & \mainscore{\underline{22.0}} & \mainscore{\underline{29.4}} & \mainscore{\textbf{12.5}} & \mainscore{\underline{50.0}} & \mainscore{\underline{32.4}} & \maincost{1.382} \\
\makebox[1.3em][c]{\raisebox{-0.15em}{\includegraphics[height=1.05em]{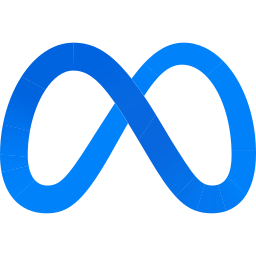}}}\ Muse Spark 1.3 & \mainscore{22.0} & \mainscore{12.2} & \mainscore{13.7} & \mainscore{\underline{5.0}} & \mainscore{22.7} & \mainscore{15.0} & \maincost{\underline{0.791}} \\
\makebox[1.3em][c]{\raisebox{-0.15em}{\includegraphics[height=1.05em]{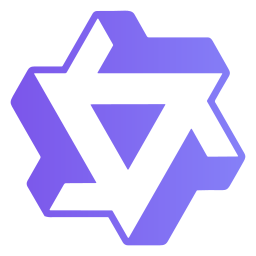}}}\ Qwen 3.8 Flash & \mainscore{11.9} & \mainscore{7.3} & \mainscore{3.9} & \mainscore{2.5} & \mainscore{22.7} & \mainscore{8.5} & \maincost{\textbf{0.077}} \\
\midrule
\multicolumn{8}{@{}l}{\textbf{VLA--VLM auditors}} \\
\addlinespace[2pt]
\makebox[1.3em][c]{\raisebox{-0.15em}{\includegraphics[height=1.05em]{assets/logos/openai.png}}}\ GPT-6 Astra & \mainscore{\underline{20.3}} & \mainscore{17.1} & \mainscore{\textbf{25.5}} & \mainscore{0.0} & \mainscore{\underline{13.6}} & \mainscore{\underline{16.4}} & \maincost{0.281} \\
\makebox[1.3em][c]{\raisebox{-0.15em}{\includegraphics[height=1.05em]{assets/logos/anthropic.png}}}\ Claude Opus 5 & \mainscore{\textbf{22.0}} & \mainscore{\underline{22.0}} & \mainscore{11.8} & \mainscore{\underline{2.5}} & \mainscore{\underline{13.6}} & \mainscore{15.0} & \maincost{0.727} \\
\makebox[1.3em][c]{\raisebox{-0.15em}{\includegraphics[height=1.05em]{assets/logos/gemini-color.png}}}\ Gemini 3.8 Flash & \mainscore{\underline{20.3}} & \mainscore{\textbf{24.4}} & \mainscore{\underline{17.6}} & \mainscore{\textbf{5.0}} & \mainscore{\textbf{18.2}} & \mainscore{\textbf{17.4}} & \maincost{0.296} \\
\makebox[1.3em][c]{\raisebox{-0.15em}{\includegraphics[height=1.05em]{assets/logos/meta-color.png}}}\ Muse Spark 1.3 & \mainscore{5.1} & \mainscore{9.8} & \mainscore{9.8} & \mainscore{0.0} & \mainscore{9.1} & \mainscore{6.6} & \maincost{\underline{0.112}} \\
\makebox[1.3em][c]{\raisebox{-0.15em}{\includegraphics[height=1.05em]{assets/logos/qwen-color.png}}}\ Qwen 3.8 Flash & \mainscore{18.6} & \mainscore{\underline{22.0}} & \mainscore{11.8} & \mainscore{0.0} & \mainscore{4.5} & \mainscore{12.7} & \maincost{\textbf{0.039}} \\
\bottomrule
\end{tabular}

\endgroup
\end{table}

\section{Experiments}
\label{sec:experiments}
\begingroup
\makeatletter\setlength{\@fptop}{0pt}\makeatother

\subsection{Experimental Setup and Metrics}

\textbf{Agent Implementation.}
We evaluate GPT-6 Astra, Claude Opus 5, Gemini 3.8 Flash, Muse Spark 1.3, and Qwen 3.8 Flash under the two auditing paradigms in Section~\ref{sec:auditing-paradigms}. We run GPT-6 Astra with Codex CLI~\citep{openai2026codex}, Claude Opus 5 with Claude Code~\citep{anthropic2026claudecode}, Gemini 3.8 Flash with Gemini CLI~\citep{google2026geminicli}, Muse Spark 1.3 with OpenCode~\citep{anomalyco2026opencode}, and Qwen 3.8 Flash with Qwen Code~\citep{qwen2026code}. These harnesses manage model calls and conversation history, while a shared Model Context Protocol (MCP) interface~\citep{mcp2025specification} exposes the auditing tools in Table~\ref{tab:tools}. Each comparison thus evaluates a model together with its harness. In the VLM-only paradigm, VLM auditors receive a budget of 40 environment actions. In the VLA--VLM paradigm, we use Open-P2P 1.2B~\citep{yue2026scaling} as the VLA model to collect 60 seconds of simulated exploration, with observations sampled every 0.5 seconds. Recorded trajectories are shared across VLM backbones; the two paradigms use different exploration budgets.

\ifdefined\preprinttitle\else

\fi

\textbf{Human Baseline.}
Besides AI agents, we evaluate human performance on this task. Our human evaluation involves approximately ten computer science PhD students. Each task is evaluated by two participants. Participants receive the same task information as the models, including the scene description, auditing instructions, and in-context examples (Appendix~\ref{app:case-input}). They have up to ten minutes per task to explore the environment and report their findings. Human reports are evaluated by the same VLM judge against the same evaluation rubrics as model reports.

\textbf{Metrics.}
Following the VLM-as-a-judge paradigm, we use GPT-6 Astra to assess each report and its evidence against the task's evaluation rubric. We report success rate (SR), the percentage of tasks for which the target anomaly is correctly identified. Appendix~\ref{app:judge} provides the judge prompt and an example evaluation. Table~\ref{tab:judge-agreement} reports agreement between the VLM judge and human judgments.


\subsection{Main Results}
Table~\ref{tab:main} shows that auditing remains challenging across five backbones. Among VLM auditors, GPT-6 Astra achieves the highest SR at 42.3\%, followed by Gemini at 32.4\% and Claude at 28.2\%, while Muse and Qwen reach 15.0\% and 8.5\%. Across the five anomaly families, static physics and semantic consistency are relatively easy for VLM auditors, as the relevant physical or contextual inconsistency can often be identified from a single frame. Spatial and temporal consistency require comparing observations across viewpoints or time, and receive lower scores. Temporal consistency is particularly challenging, with SR at or below 12.5\% for every model and paradigm. Appendix~\ref{app:qualitative} illustrates difficulties in gathering before-and-after evidence and identifying the target change.

\setlength{\columnsep}{10pt}
\setlength{\intextsep}{4pt}
\begin{wrapfigure}{r}{0.5\textwidth}
\centering
\includegraphics[width=\linewidth]{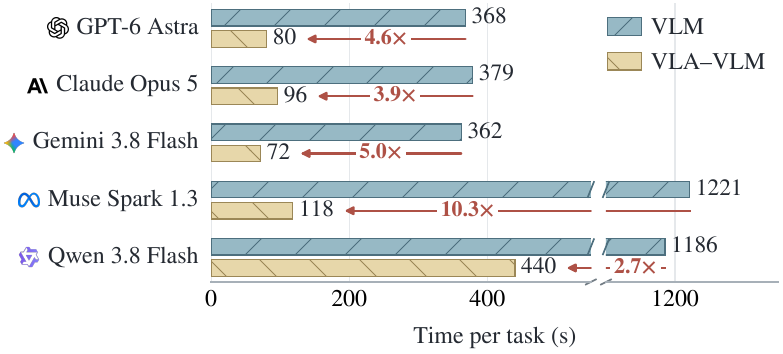}
\caption{\small\raggedright\textbf{Auditing time per task.}}
\label{fig:auditing-time}
\end{wrapfigure}
Most models achieve higher SR with interactive VLM auditing than with VLA--VLM analysis of fixed trajectories. Interactive auditors can choose additional viewpoints and actions as they reason, whereas VLA--VLM auditors analyze fixed trajectories. Section~\ref{sec:ablation-discovery} examines anomaly reach and recognition under these evidence-gathering procedures. Figure~\ref{fig:auditing-time} compares the auditing time of the two paradigms. Given prerecorded trajectories, VLA--VLM analysis is 2.7--10.3$\times$ faster than interactive VLM auditing and has lower inference costs, these efficiency gains come at the cost of lower SR, reflecting a trade-off between efficient trajectory analysis and adaptive evidence gathering.

\FloatBarrier
\endgroup

\section{Ablation Studies}
\label{sec:ablations}

\subsection{Factors Affecting Auditing Performance}
\label{sec:ablation-factors}
\begin{wraptable}[14]{r}{0.50\textwidth}
\vspace{-6pt}
\setlength{\abovecaptionskip}{2pt}
\setlength{\belowcaptionskip}{2pt}
\caption{\small\raggedright\textbf{Auditing factors.} SR (\%), $\Delta$ (pp).}
\label{tab:factors}
\centering
\begingroup
\definecolor{abblue}{HTML}{355C7D}
\definecolor{abgreen}{HTML}{287568}
\definecolor{abred}{HTML}{AA5858}
\definecolor{abgray}{HTML}{77818A}
\small
\setlength{\tabcolsep}{2pt}
\setlength{\aboverulesep}{1.6pt}
\setlength{\belowrulesep}{1.6pt}
\renewcommand{\arraystretch}{1.05}
\newcommand{\abdelta}[2]{\textcolor{#1}{\small$(#2)$}}
\newcommand{\abup}[1]{\abdelta{abgreen}{+#1}}
\newcommand{\abdown}[1]{\abdelta{abred}{-#1}}
\newcommand{\abgroup}[1]{\addlinespace[2pt]\multicolumn{5}{@{}l@{}}{\textcolor{abblue}{\textit{#1}}}\\[-0.6pt]}
\begin{tabularx}{\linewidth}{@{}>{\raggedright\arraybackslash}Xr@{\hspace{2pt}}l@{\hspace{8pt}}r@{\hspace{2pt}}l@{}}
\toprule
Setting & \multicolumn{1}{c@{\hspace{2pt}}}{\makebox[0pt][c]{VLM}} & & \multicolumn{1}{@{}c@{\hspace{2pt}}}{\makebox[0pt][c]{VLA--VLM}} & \\
\midrule
\textbf{Default} & \textbf{33.3} & & \textbf{5.6} & \\
\abgroup{(a) Starting distance}
\hspace{4pt}Near ($1/3$) & 38.1 & \abup{4.8} & 7.1 & \abup{1.6} \\
\abgroup{(b) Exploration budget}
\hspace{4pt}$0.5\times$ & 23.0 & \abdown{10.3} & 4.0 & \abdown{1.6} \\
\hspace{4pt}$1.5\times$ & 34.9 & \abup{1.6} & 7.1 & \abup{1.6} \\
\abgroup{(c) Task guidance}
\hspace{4pt}Without ICL & 31.0 & \abdown{2.4} & 5.6 & \abdelta{abgray}{0.0} \\
\hspace{4pt}Without ICL \& type hint & 15.9 & \abdown{17.5} & 2.4 & \abdown{3.2} \\
\bottomrule
\end{tabularx}
\endgroup
\vspace{-5pt}
\end{wraptable}

We study how starting distance, exploration budget, and task guidance affect auditing with Gemini 3.8 Flash on 126 Unreal tasks (Table~\ref{tab:factors}).

\textbf{Starting Distance.} We move the starting position closer to the target anomaly, reducing the navigable route distance to approximately one third of its original length while keeping the agent's initial orientation unchanged. Moving closer increases the VLM-only success rate from 33.3\% to 38.1\% and the VLA--VLM success rate from 5.6\% to 7.1\%. These results indicate that auditing becomes more challenging when agents start farther from the anomaly.

\textbf{Exploration Budget.} We express exploration budgets relative to each approach's default ($1\times$): 40 steps for VLM and 60 seconds for VLA--VLM. The $0.5\times$ and $1.5\times$ settings correspond to 20 and 60 steps for VLM, and 30 and 90 seconds for VLA--VLM. At $0.5\times$, success rate drops from 33.3\% to 23.0\% for VLM and from 5.6\% to 4.0\% for VLA--VLM. At $1.5\times$, success rate increases to 34.9\% for VLM and 7.1\% for VLA--VLM.

\textbf{Task Guidance.} By default, agents receive an anomaly-type hint and a matching in-context example. We first remove thein-context example while retaining the anomaly-type hint. VLM success rate decreases from 33.3\% to 31.0\%, while VLA--VLM success rate remains at 5.6\%. Further removing the type hint reduces success rates to 15.9\% for VLM and 2.4\% for VLA--VLM. The anomaly-type hint matters far more than the in-context example, suggesting that knowing what to look for is critical for directing exploration.

\FloatBarrier

\subsection{Auditing Multiple Anomalies In a Scene}
\label{sec:ablation-multiple}
\ifdefined\preprinttitle
\def\multiplefigurewidth{0.44\textwidth}
\else
\def\multiplefigurewidth{0.5\textwidth}
\fi
\begin{wrapfigure}{R}{\multiplefigurewidth}
\vspace{-6pt}
\centering
\setlength{\abovecaptionskip}{3pt}
\includegraphics[width=\linewidth]{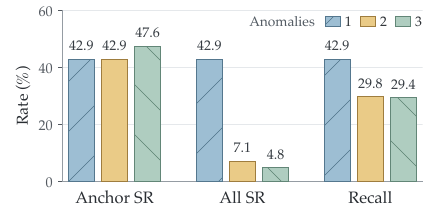}
\caption{\small\textbf{Auditing multiple anomalies.}}
\label{fig:ablation-multiple}
\end{wrapfigure}
To test auditing with coexisting anomalies, we select 21 anchor tasks, three from each of seven Unreal Engine 5 environments, and evaluate the Gemini 3.8 Flash VLM auditor with a 40-step budget. The original anomaly is called the \textit{anchor anomaly}. For each anchor, we randomly introduce one or two additional anomalies on distinct objects to form two- and three-anomaly conditions, sampling two configurations per anchor for each condition, yielding 42 runs per condition. Agents are given the total number of anomalies, but only the anchor anomaly's type and in-context example. We report the success rate in identifying the anchor anomaly (anchor SR), the success rate in identifying every anomaly (all-anomaly SR), and the fraction of anomalies identified (target recall).

Figure~\ref{fig:ablation-multiple} shows that anchor SR remains similar across anomaly counts: 42.9\% with one or two anomalies and 47.6\% with three. Complete discovery remains difficult: the agent identifies every target in only 3/42 two-anomaly runs (7.1\%) and 2/42 three-anomaly runs (4.8\%), with target recall of 29.8\% and 29.4\%, respectively. Under a fixed exploration budget, identifying the anchor often leaves other coexisting anomalies undiscovered.

\subsection{Anomaly Exposure and Identification}
\label{sec:ablation-discovery}

As discussed in Section~\ref{sec:auditing-paradigms}, VLM auditing couples action with visual reasoning. Using GPT-6 Astra on 213 tasks, we examine two aspects of its advantage over VLA--VLM auditing: \ding{182} \textbf{anomaly exposure}, whether exploration reveals the anomaly, and \ding{183} \textbf{anomaly identification upon exposure}, whether the agent identifies the anomaly once it is revealed. We assess exposure using geometric coverage, based on the agent's position and orientation, and human-rated exposure, based on exploration videos. For identification, we compare success rates on exposed tasks under both measures. 


\begin{wrapfigure}{r}{0.52\textwidth}
\vspace{-6pt}
\centering
\setlength{\abovecaptionskip}{1pt}
\setlength{\belowcaptionskip}{0pt}
\includegraphics[width=\linewidth,trim=0 4pt 0 0,clip]{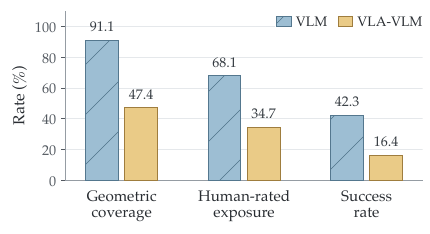}
\caption{\small\raggedright\textbf{Anomaly exposure and success.}}
\label{fig:ablation-exposure}
\end{wrapfigure}
\textbf{Anomaly Exposure.} Geometric coverage is 91.1\% for the VLM agent and 47.4\% for the VLA explorer, while human-rated exposure is 68.1\% and 34.7\%, respectively (Figure~\ref{fig:ablation-exposure}). For spatial and temporal consistency anomalies, human-rated exposure requires contrasting viewpoints or before-and-after states, whereas geometric coverage requires only one qualifying frame. The higher exposure of the VLM agent suggests that reasoning and planning help direct exploration toward relevant parts of the scene. Given in-context examples, the VLM agent can reason about what an anomaly may look like, plan actions to investigate likely locations, and update its plan based on new observations. The VLA--VLM paradigm separates exploration from anomaly analysis, preventing this feedback from guiding the VLA's actions.

\textbf{Recognition With Tools And Memory.} On the tasks geometrically covered by both paradigms, the VLM agent achieves 38.5\% success, versus 26.0\% for the VLA--VLM auditor; on the tasks with human-rated exposure, the gap is consistent (62.1\% versus 47.3\%). The advantage of the VLM-only paradigm thus extends beyond reaching anomaly locations. The VLM agent can inspect additional viewpoints, test suspected anomalies through interaction, and use memory tools to retrieve earlier frames and notes. This access to long-term evidence supports comparisons across viewpoints and time, helping the agent recognize inconsistencies that are difficult to establish from an isolated observation. In contrast, the two-stage auditor relies on the evidence present in its recorded trajectory. Effective auditing therefore benefits from both exploration and the ability to gather and connect evidence during recognition.

\FloatBarrier

\FloatBarrier
\section{Conclusion}
We introduced \dataset{}, a benchmark of 213 tasks across 13 environments and five anomaly families for studying how multimodal agents couple action and visual reasoning in interactive 3D worlds. VLM-only auditors generally outperform VLA–VLM auditors, yet both remain far below human performance. Our analysis separates failures to expose anomalies from failures to identify them after exposure; reduced task guidance and coexisting anomalies further challenge discovery. These findings highlight the challenge of deciding which observations to seek and whether they support a conclusion. \dataset{} provides a testbed for agents that form hypotheses from observations, test them through action, and connect evidence across viewpoints and time.
\section*{AI Use Statement}
We used generative AI tools to assist with drafting, polishing, and formatting the manuscript, including figures and tables. All AI-assisted text was reviewed, revised, and verified by the authors. AI-assisted code was reviewed and tested for correctness. We take responsibility for the final content of this work, including all text, claims, code, and artifacts.

\section*{Ethics Statement}
\dataset{} evaluates agents in simulated 3D environments. Human participation in task validation and the auditing baseline was voluntary and informed; the procedures are described in Appendix~\ref{app:human-annotation} and Section~\ref{sec:experiments}. Environment sources and licenses are documented in Appendix~\ref{app:sources-licenses}. Benchmark releases will respect these licenses and retain required notices and attributions.

\section*{Reproducibility Statement}
We will release our evaluation code, task configurations, and runnable environment packages in accordance with the applicable licenses (Appendix~\ref{app:sources-licenses}). Together with the prompts and experimental settings documented in this paper, these resources will support reproduction of the reported experiments.

\section*{Acknowledgments}
The UCSB team acknowledges support from the National Science Foundation (NSF) Grant IIS-2338252.

\bibliography{references}
\bibliographystyle{iclr2027_conference}
\appendix
\clearpage
\pdfbookmark[0]{Appendix}{appendix-start}
\raggedbottom\section*{Appendix}
\subsection*{Contents}
\begingroup
\setlength{\parindent}{0pt}
\newcommand{\appentry}[3]{\noindent\hspace*{#3}\hyperref[#1]{\ref*{#1}\quad #2}\nobreak\dotfill\hyperref[#1]{\pageref*{#1}}\par}
\setlength{\parskip}{4pt}
\appentry{app:environments}{\textbf{Environments}}{0pt}
{\small\appentry{app:environment-examples}{Environment Examples}{10pt}}
{\small\appentry{app:sources-licenses}{Sources And Licenses}{10pt}}

\vspace{7pt}
\appentry{app:dataset-creation}{\textbf{Dataset Creation}}{0pt}
{\small\appentry{app:construction}{Scene And Task Construction}{10pt}}
{\small\appentry{app:annotations}{Task Annotations}{10pt}}
{\small\appentry{app:human-annotation}{Human Validation}{10pt}}
\appentry{app:cases}{\textbf{Evaluation Protocol}}{0pt}
{\small\appentry{app:case-input}{Agent Inputs And Instructions}{10pt}}
{\small\appentry{app:trajectory}{Actions And Tool Use}{10pt}}
{\small\appentry{app:case-output}{Agent Output}{10pt}}
{\small\appentry{app:judge}{VLM As Judge}{10pt}}
{\small\appentry{app:full-examples}{Anomaly Taxonomy And Full In-Context Examples}{10pt}}
\appentry{app:full-results}{\textbf{Full Results}}{0pt}
\appentry{app:geometric-coverage}{\textbf{Geometric Coverage}}{0pt}
\appentry{app:qualitative}{\textbf{Qualitative Auditing Trajectories}}{0pt}
\endgroup
\clearpage

\section{Environments}
\label{app:environments}
Table~\ref{tab:environments} summarizes the 13 environments in \dataset{}, their rendering engines, and task counts. The environments span five scene categories and contain 213 tasks: 126 in Unreal Engine 5 and 87 in Three.js.
\begin{table}[!htbp]
\caption{\textbf{Environments and task statistics in \dataset{}.}}
\label{tab:environments}
\centering\small
\setlength{\tabcolsep}{5.5pt}
\renewcommand{\arraystretch}{1.13}
\begin{tabularx}{\linewidth}{X>{\raggedright\arraybackslash}p{0.23\linewidth}>{\centering\arraybackslash}p{0.13\linewidth}}
\toprule
\textbf{Environment} & \textbf{Engine} & \textbf{\# Tasks} \\
\midrule
\rowcolor{cyan!7}\multicolumn{3}{c}{\textbf{Indoor}} \\
Residential House & Unreal Engine & 14 \\
Sponza Atrium & Three.js & 13 \\
Family House & Three.js & 15 \\
\addlinespace[3pt]
\rowcolor{blue!6}\multicolumn{3}{c}{\textbf{City}} \\
Utopian City & Unreal Engine & 15 \\
Subway & Unreal Engine & 17 \\
\addlinespace[3pt]
\rowcolor{violet!7}\multicolumn{3}{c}{\textbf{History}} \\
Ancient Chinese City & Unreal Engine & 23 \\
Medieval Village & Unreal Engine & 22 \\
\addlinespace[3pt]
\rowcolor{orange!9}\multicolumn{3}{c}{\textbf{Industrial}} \\
Industrial Factory & Unreal Engine & 19 \\
Sketchbook Airfield & Three.js & 18 \\
\addlinespace[3pt]
\rowcolor{green!7}\multicolumn{3}{c}{\textbf{Nature}} \\
Rural Australia & Unreal Engine & 16 \\
Mistwood Cottage & Three.js & 15 \\
Reef Dive & Three.js & 12 \\
Beyond Fable Wilderness & Three.js & 14 \\
\midrule
\textbf{Total} & & \textbf{213} \\
\bottomrule
\end{tabularx}
\end{table}

\FloatBarrier

\subsection{Environment Examples}
\label{app:environment-examples}
We show one representative view of each scene used for task construction. The 13 environments comprise 27 scenes: 21 bounded regions within seven Unreal Engine 5 environments, and one scene in each of the six Three.js environments. Views are grouped by environment and labeled by scene.
\begingroup
\setlength{\parindent}{0pt}
\par\addvspace{11pt}\noindent\begin{minipage}{\linewidth}
\pdfbookmark[2]{Residential House}{atlas-residential-house}
{\normalsize\textbf{Residential House}}\hfill{\footnotesize\color{black!65}Indoor $\cdot$ Unreal Engine 5 $\cdot$ 3 scenes $\cdot$ 14 tasks}\par
\vspace{5pt}
\begin{minipage}[t]{0.325\linewidth}
\vspace{0pt}
\includegraphics[width=\linewidth,trim={0.000bp 84.000bp 230.400bp 0.000bp},clip]{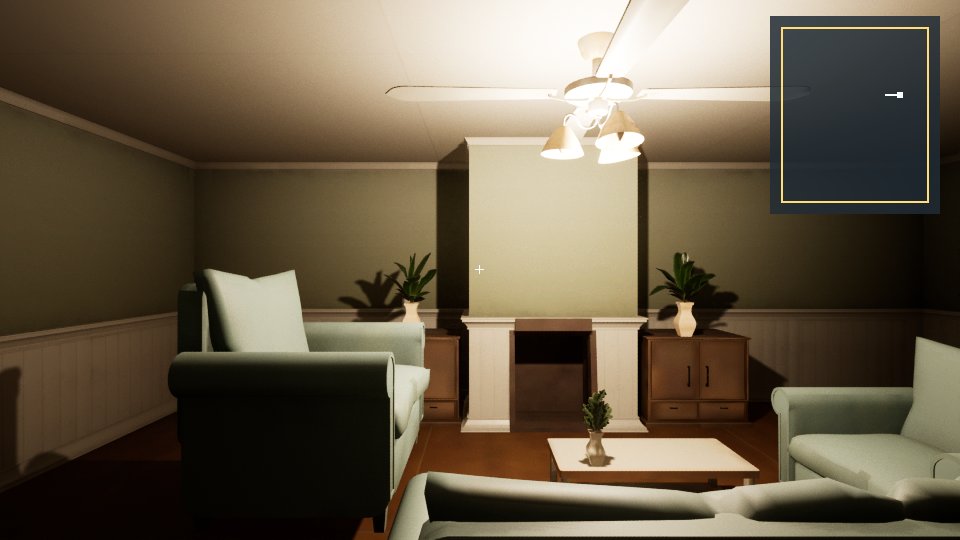}\par
{\centering\footnotesize (a) Living Room\par}
\end{minipage}\hfill
\begin{minipage}[t]{0.325\linewidth}
\vspace{0pt}
\includegraphics[width=\linewidth,trim={0.000bp 84.000bp 230.400bp 0.000bp},clip]{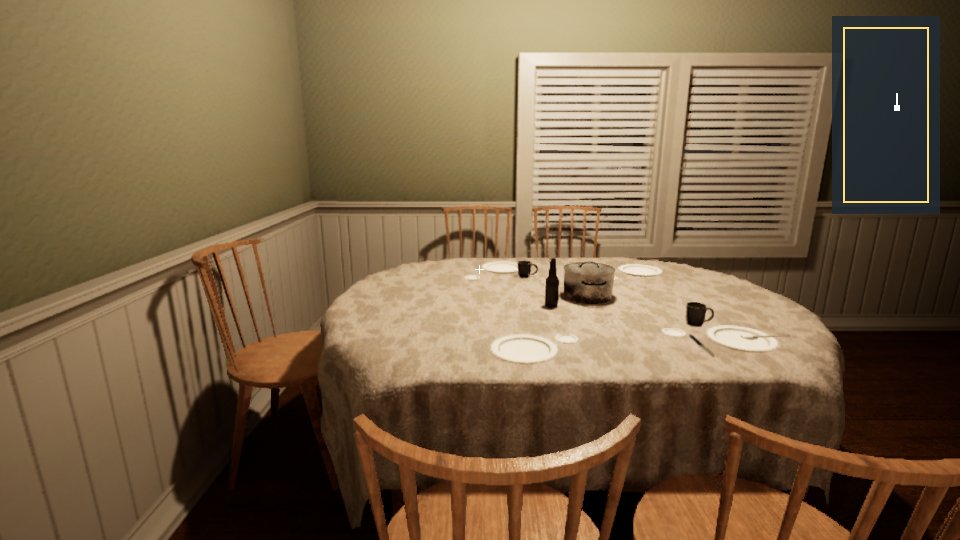}\par
{\centering\footnotesize (b) Kitchen And Dining Area\par}
\end{minipage}\hfill
\begin{minipage}[t]{0.325\linewidth}
\vspace{0pt}
\includegraphics[width=\linewidth,trim={0.000bp 84.000bp 230.400bp 0.000bp},clip]{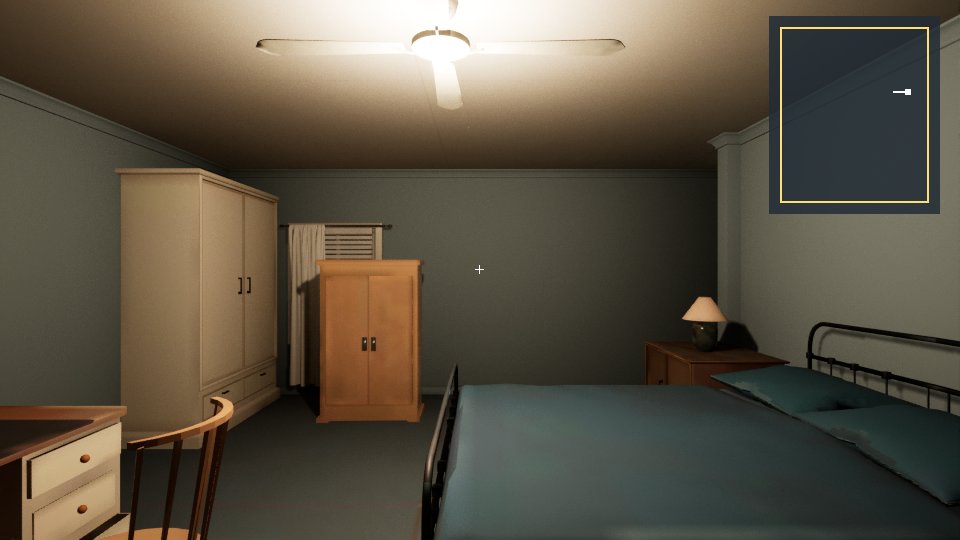}\par
{\centering\footnotesize (c) Bedroom Suite\par}
\end{minipage}\par
\vspace{5pt}
A furnished residence containing a living room, a kitchen and dining area, and a bedroom with an adjoining bathroom. Seating, plants, appliances, tableware, and storage furniture provide dense indoor context. Task-dependent interactions include a cup, the bedroom door, and a wardrobe; other furnishings are fixed.\par
\vspace{7pt}
{\color{black!18}\hrule height 0.35pt}
\end{minipage}\par
\par\addvspace{11pt}\noindent\begin{minipage}{\linewidth}
\pdfbookmark[2]{Sponza Atrium}{atlas-sponza-atrium}
{\normalsize\textbf{Sponza Atrium}}\hfill{\footnotesize\color{black!65}Indoor $\cdot$ Three.js $\cdot$ 1 scene $\cdot$ 13 tasks}\par
\vspace{5pt}
\begin{minipage}[t]{0.43\linewidth}
\vspace{0pt}
\includegraphics[width=\linewidth,trim={48.000bp 0.000bp 48.000bp 0.000bp},clip]{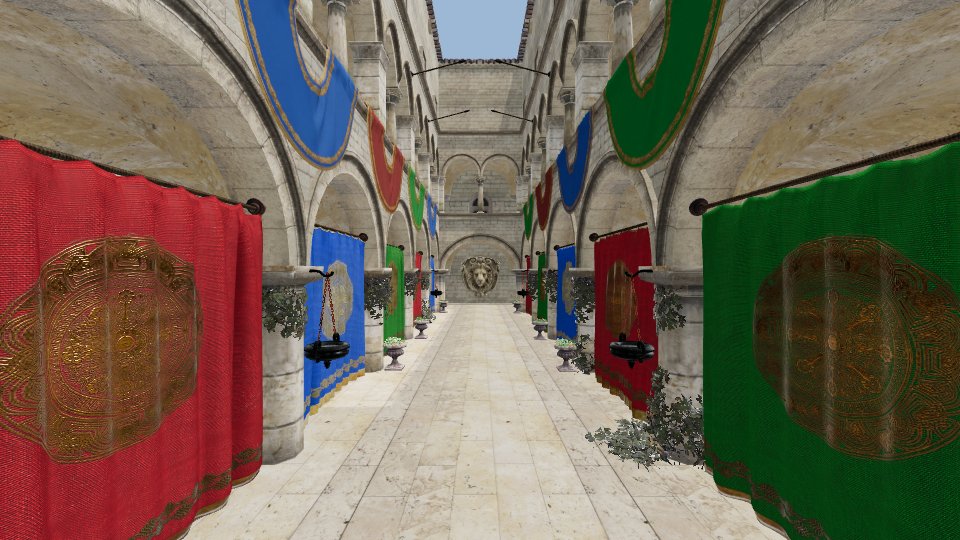}\par
{\centering\footnotesize Atrium\par}
\end{minipage}\hfill
\begin{minipage}[t]{0.545\linewidth}
\vspace{0pt}
A stone atrium with a tall central hall, repeated columns and arches, colorful hanging drapes, and stone planters. Aisles around the columns allow objects and architectural details to be viewed from different directions. Only navigation and visual inspection are available.
\end{minipage}\par
\vspace{7pt}
{\color{black!18}\hrule height 0.35pt}
\end{minipage}\par
\par\addvspace{11pt}\noindent\begin{minipage}{\linewidth}
\pdfbookmark[2]{Family House}{atlas-family-house}
{\normalsize\textbf{Family House}}\hfill{\footnotesize\color{black!65}Indoor $\cdot$ Three.js $\cdot$ 1 scene $\cdot$ 15 tasks}\par
\vspace{5pt}
\begin{minipage}[t]{0.43\linewidth}
\vspace{0pt}
\includegraphics[width=\linewidth,trim={48.000bp 0.000bp 48.000bp 0.000bp},clip]{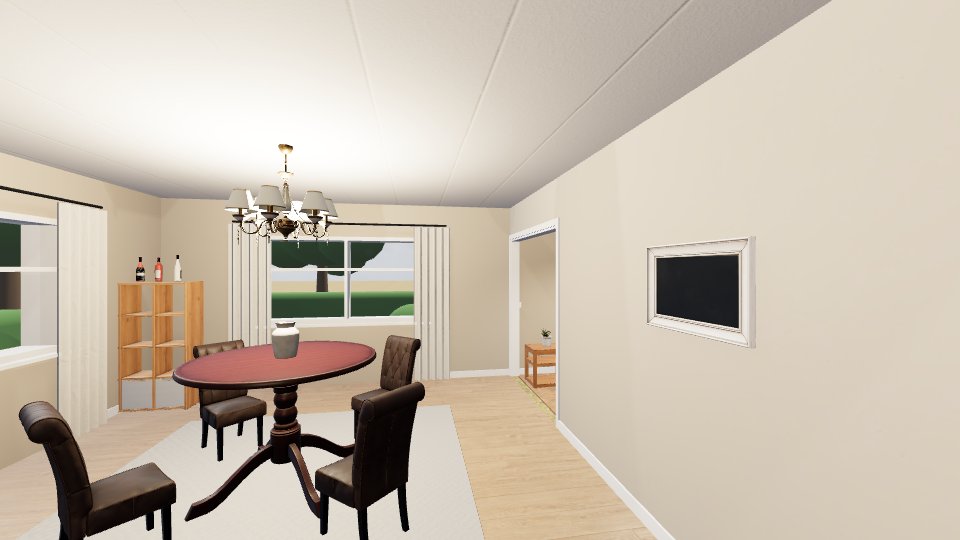}\par
{\centering\footnotesize House Interior\par}
\end{minipage}\hfill
\begin{minipage}[t]{0.545\linewidth}
\vspace{0pt}
A two-storey home with a kitchen, living room, dining room, and upstairs bedrooms. The interior combines a staircase and connecting passages with domestic furniture, lighting fixtures, and appliances. Auditors explore the rooms and inspect objects from multiple viewpoints; the current scene exposes no dedicated object-interaction actions.
\end{minipage}\par
\vspace{7pt}
{\color{black!18}\hrule height 0.35pt}
\end{minipage}\par
\par\addvspace{11pt}\noindent\begin{minipage}{\linewidth}
\pdfbookmark[2]{Utopian City}{atlas-utopian-city}
{\normalsize\textbf{Utopian City}}\hfill{\footnotesize\color{black!65}City $\cdot$ Unreal Engine 5 $\cdot$ 4 scenes $\cdot$ 15 tasks}\par
\vspace{5pt}
\begin{minipage}[t]{0.655\linewidth}
\vspace{0pt}
\begin{minipage}[t]{0.49\linewidth}
\vspace{0pt}
\includegraphics[width=\linewidth,trim={64.000bp 0.000bp 64.000bp 0.000bp},clip]{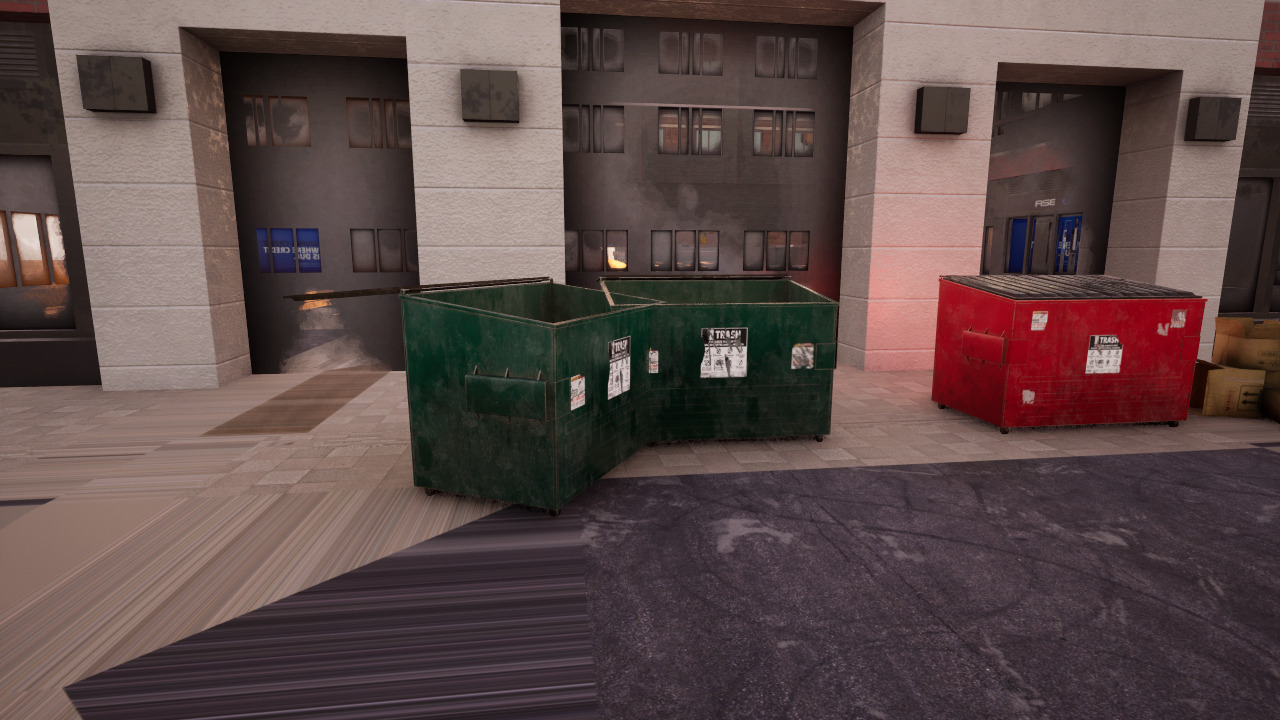}\par
{\centering\footnotesize (a) Rear Courtyard\par}
\end{minipage}\hfill
\begin{minipage}[t]{0.49\linewidth}
\vspace{0pt}
\includegraphics[width=\linewidth,trim={64.000bp 0.000bp 64.000bp 0.000bp},clip]{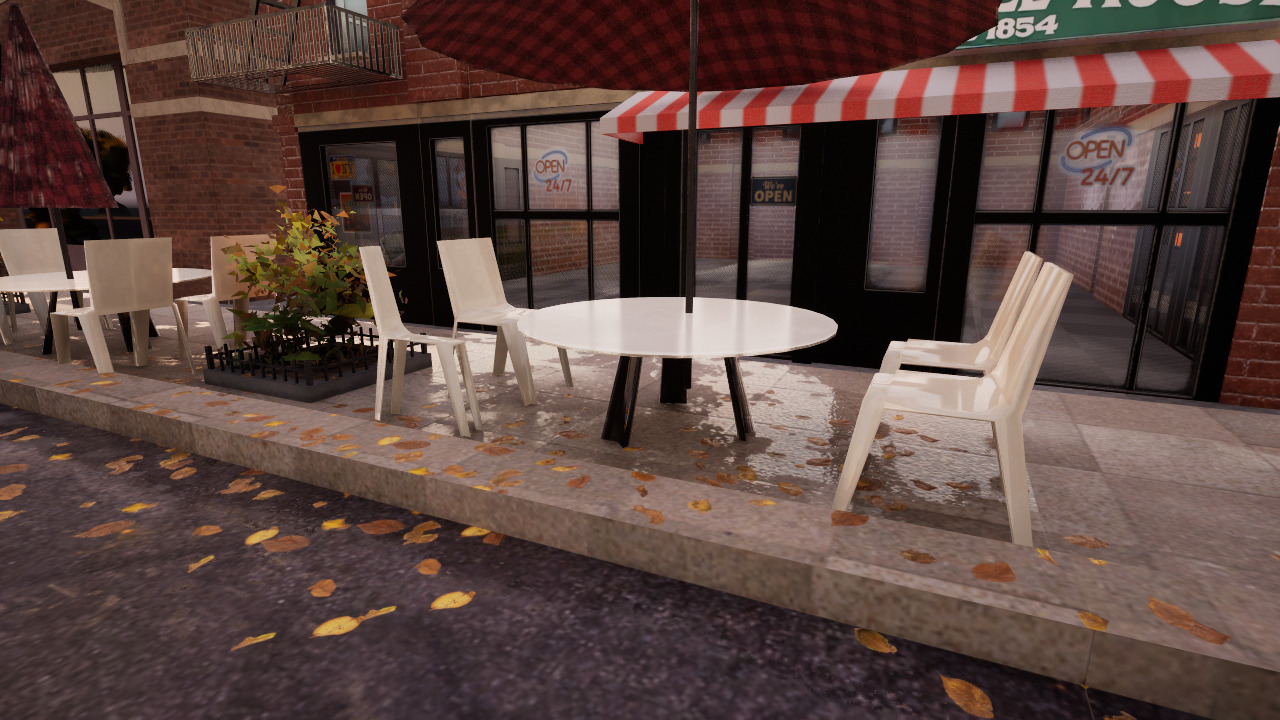}\par
{\centering\footnotesize (b) Shopfront Street\par}
\end{minipage}\par\vspace{5pt}
\begin{minipage}[t]{0.49\linewidth}
\vspace{0pt}
\includegraphics[width=\linewidth,trim={0.000bp 84.000bp 230.400bp 0.000bp},clip]{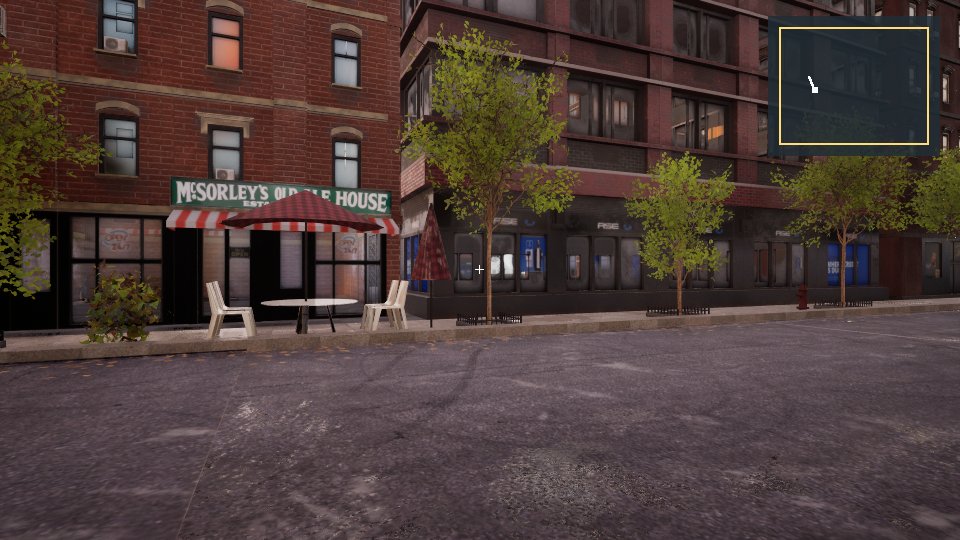}\par
{\centering\footnotesize (c) City Intersection\par}
\end{minipage}\hfill
\begin{minipage}[t]{0.49\linewidth}
\vspace{0pt}
\includegraphics[width=\linewidth,trim={0.000bp 84.000bp 230.400bp 0.000bp},clip]{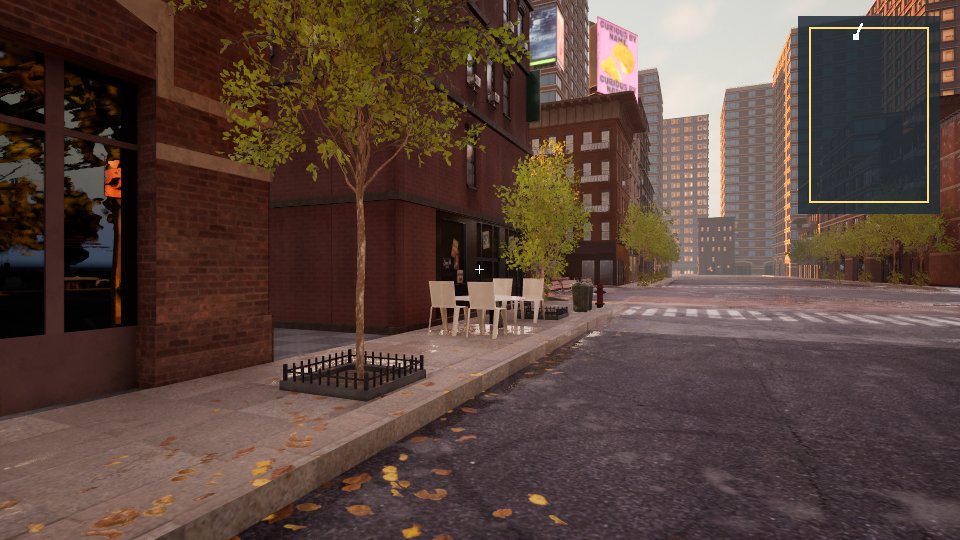}\par
{\centering\footnotesize (d) Service Lane\par}
\end{minipage}\par
\end{minipage}\hfill
\begin{minipage}[t]{0.32\linewidth}
\vspace{0pt}
An urban neighborhood organized around shopfront streets, intersections, rear courtyards, and service lanes. Outdoor seating, trees, bins, boxes, mailboxes, and building entrances create a mixture of open routes and cluttered edges. Tasks use bounded regions within this setting, requiring auditors to inspect both street furniture and its relationship to surrounding buildings.
\end{minipage}\par
\vspace{7pt}
{\color{black!18}\hrule height 0.35pt}
\end{minipage}\par
\par\addvspace{11pt}\noindent\begin{minipage}{\linewidth}
\pdfbookmark[2]{Subway}{atlas-subway}
{\normalsize\textbf{Subway}}\hfill{\footnotesize\color{black!65}City $\cdot$ Unreal Engine 5 $\cdot$ 3 scenes $\cdot$ 17 tasks}\par
\vspace{5pt}
\begin{minipage}[t]{0.325\linewidth}
\vspace{0pt}
\includegraphics[width=\linewidth,trim={0.000bp 84.000bp 230.400bp 0.000bp},clip]{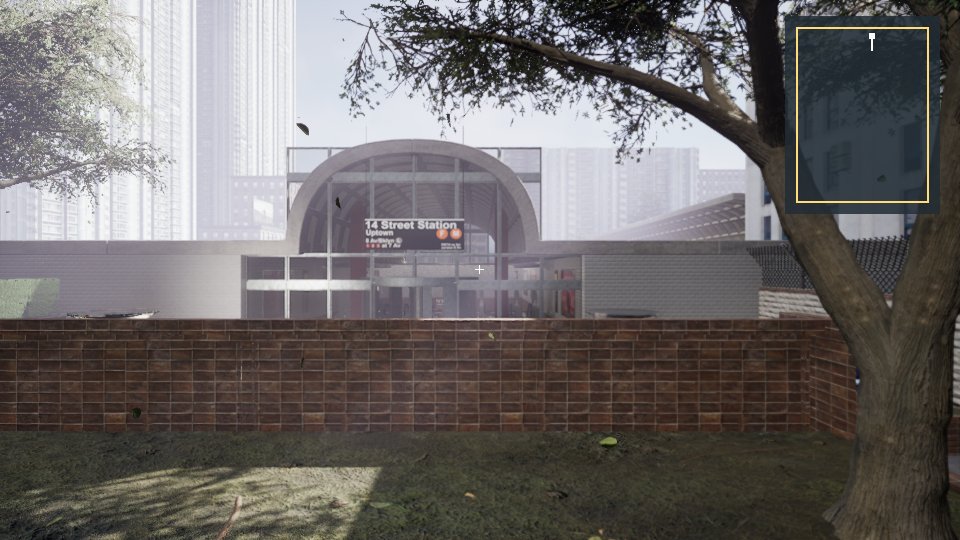}\par
{\centering\footnotesize (a) Station Entrance\par}
\end{minipage}\hfill
\begin{minipage}[t]{0.325\linewidth}
\vspace{0pt}
\includegraphics[width=\linewidth,trim={0.000bp 84.000bp 230.400bp 0.000bp},clip]{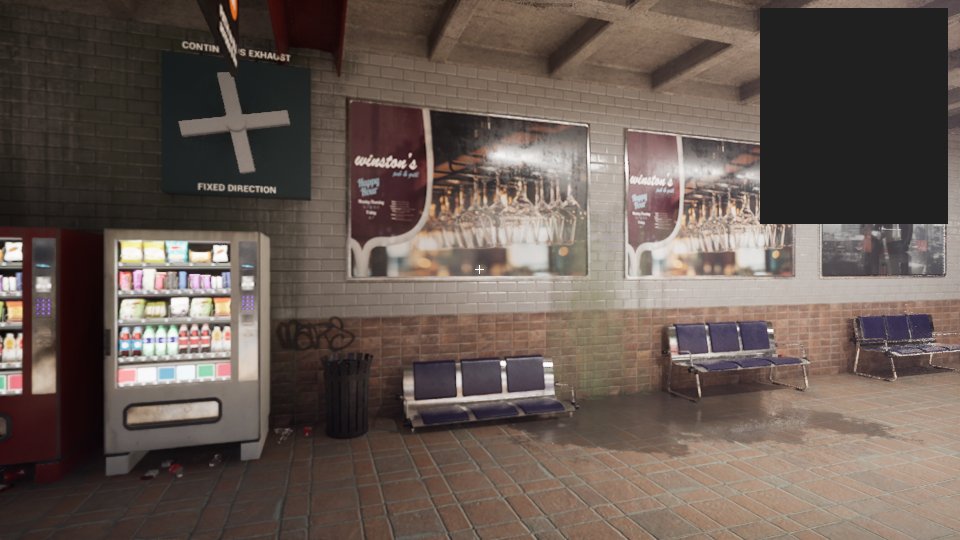}\par
{\centering\footnotesize (b) Waiting Area\par}
\end{minipage}\hfill
\begin{minipage}[t]{0.325\linewidth}
\vspace{0pt}
\includegraphics[width=\linewidth,trim={0.000bp 84.000bp 230.400bp 0.000bp},clip]{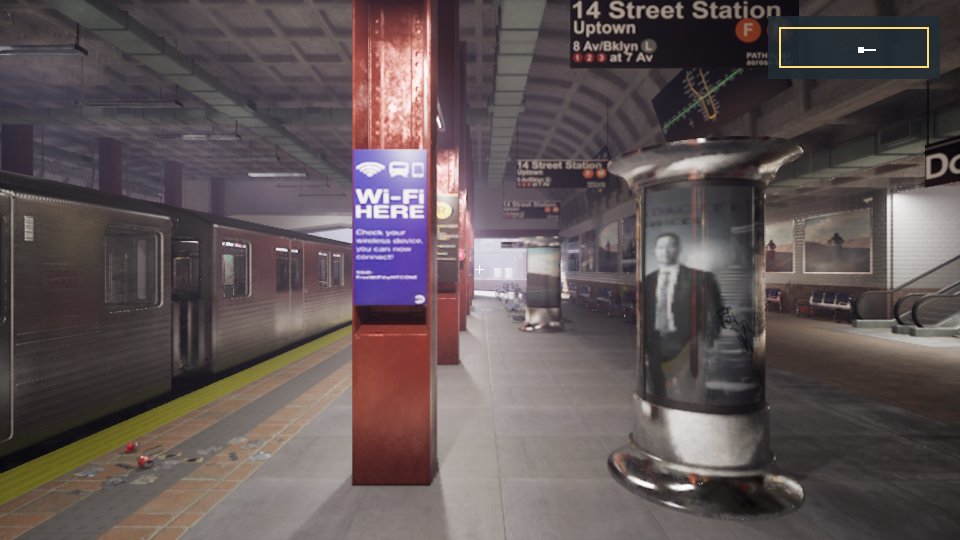}\par
{\centering\footnotesize (c) Train-Side Platform\par}
\end{minipage}\par
\vspace{5pt}
A transit environment spanning an open-air station entrance, waiting areas, and train-side platforms. Benches, bins, vending machines, stairs, and escalators connect circulation spaces at different levels. The expanded entrance region also includes an upper lawn and a waterfall feature. Interactive benches are available in selected tasks.\par
\vspace{7pt}
{\color{black!18}\hrule height 0.35pt}
\end{minipage}\par
\par\addvspace{11pt}\noindent\begin{minipage}{\linewidth}
\pdfbookmark[2]{Ancient Chinese City}{atlas-ancient-chinese-city}
{\normalsize\textbf{Ancient Chinese City}}\hfill{\footnotesize\color{black!65}History $\cdot$ Unreal Engine 5 $\cdot$ 3 scenes $\cdot$ 23 tasks}\par
\vspace{5pt}
\begin{minipage}[t]{0.325\linewidth}
\vspace{0pt}
\includegraphics[width=\linewidth,trim={0.000bp 84.000bp 230.400bp 0.000bp},clip]{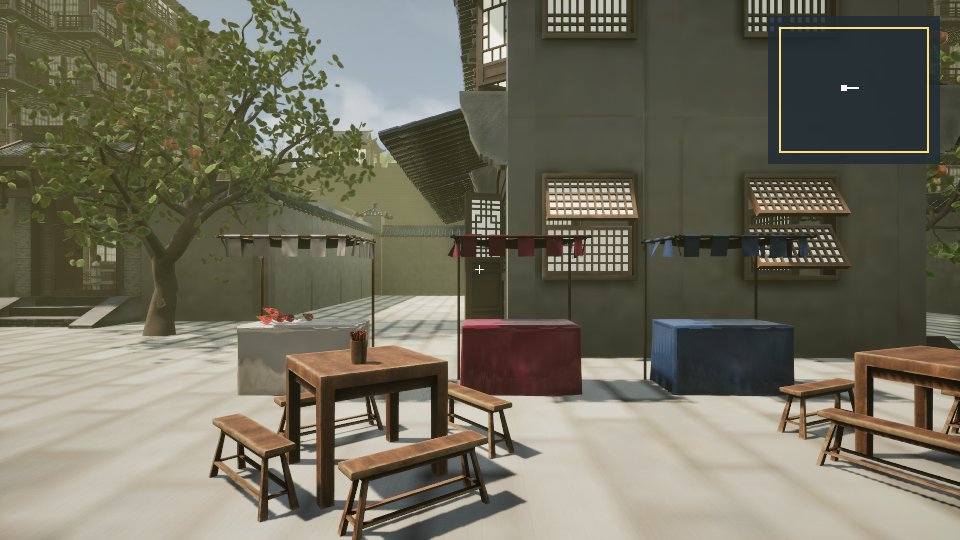}\par
{\centering\footnotesize (a) Market Street\par}
\end{minipage}\hfill
\begin{minipage}[t]{0.325\linewidth}
\vspace{0pt}
\includegraphics[width=\linewidth,trim={0.000bp 84.000bp 230.400bp 0.000bp},clip]{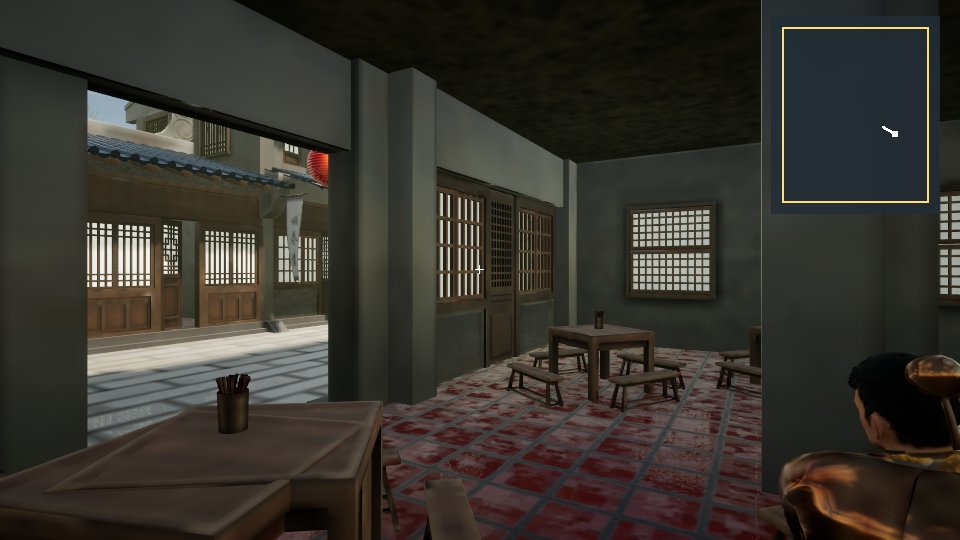}\par
{\centering\footnotesize (b) Tea House\par}
\end{minipage}\hfill
\begin{minipage}[t]{0.325\linewidth}
\vspace{0pt}
\includegraphics[width=\linewidth,trim={0.000bp 84.000bp 230.400bp 0.000bp},clip]{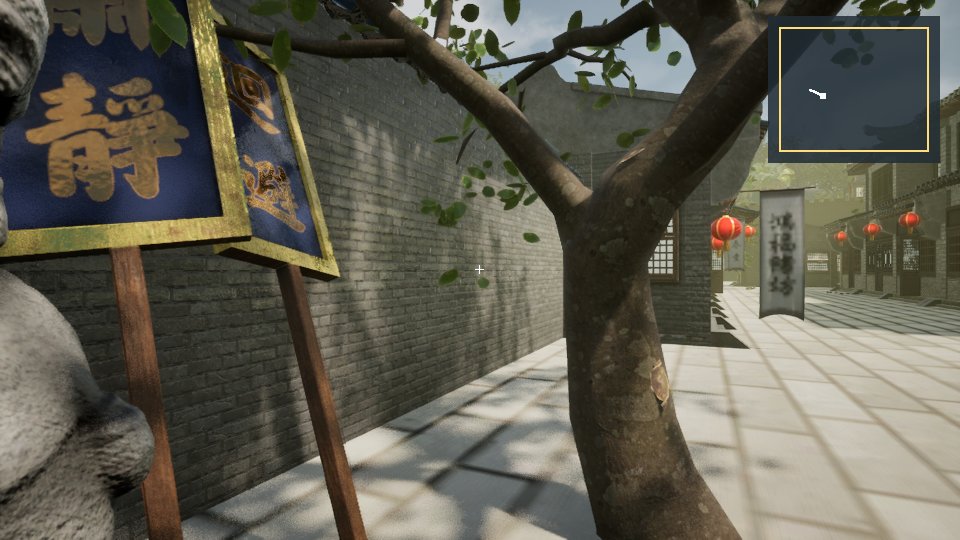}\par
{\centering\footnotesize (c) Residence Entrance\par}
\end{minipage}\par
\vspace{5pt}
An ancient Chinese urban setting with a market street, a tea house, and a residence entrance. Wooden stalls, tables, paper umbrellas, lanterns, and stone lions establish the historical context. Selected tasks allow a wicker basket to be pushed or the two wooden entrance doors to be opened and closed independently.\par
\vspace{7pt}
{\color{black!18}\hrule height 0.35pt}
\end{minipage}\par
\par\addvspace{11pt}\noindent\begin{minipage}{\linewidth}
\pdfbookmark[2]{Medieval Village}{atlas-medieval-village}
{\normalsize\textbf{Medieval Village}}\hfill{\footnotesize\color{black!65}History $\cdot$ Unreal Engine 5 $\cdot$ 2 scenes $\cdot$ 22 tasks}\par
\vspace{5pt}
\begin{minipage}[t]{0.49\linewidth}
\vspace{0pt}
\includegraphics[width=\linewidth,trim={0.000bp 84.000bp 230.400bp 0.000bp},clip]{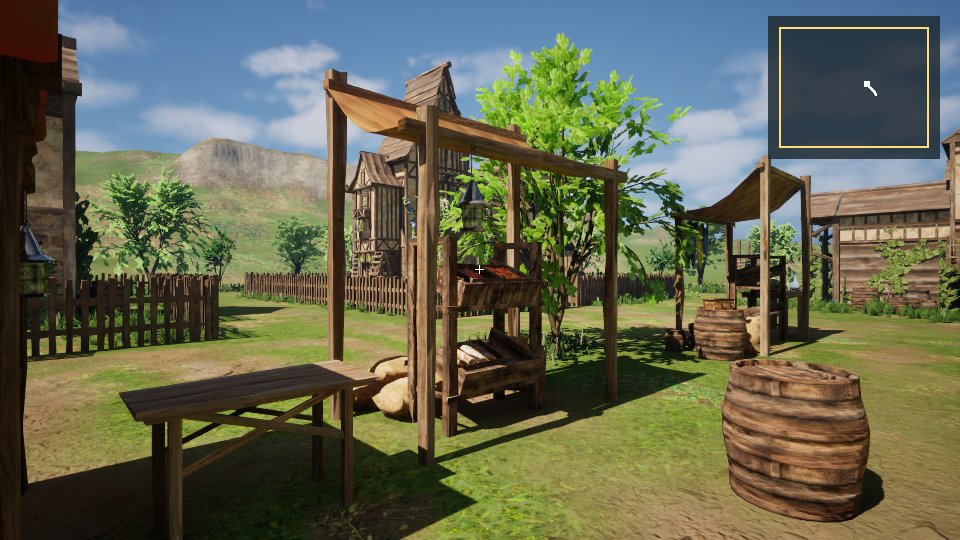}\par
{\centering\footnotesize (a) Market\par}
\end{minipage}\hfill
\begin{minipage}[t]{0.49\linewidth}
\vspace{0pt}
\includegraphics[width=\linewidth,trim={0.000bp 84.000bp 230.400bp 0.000bp},clip]{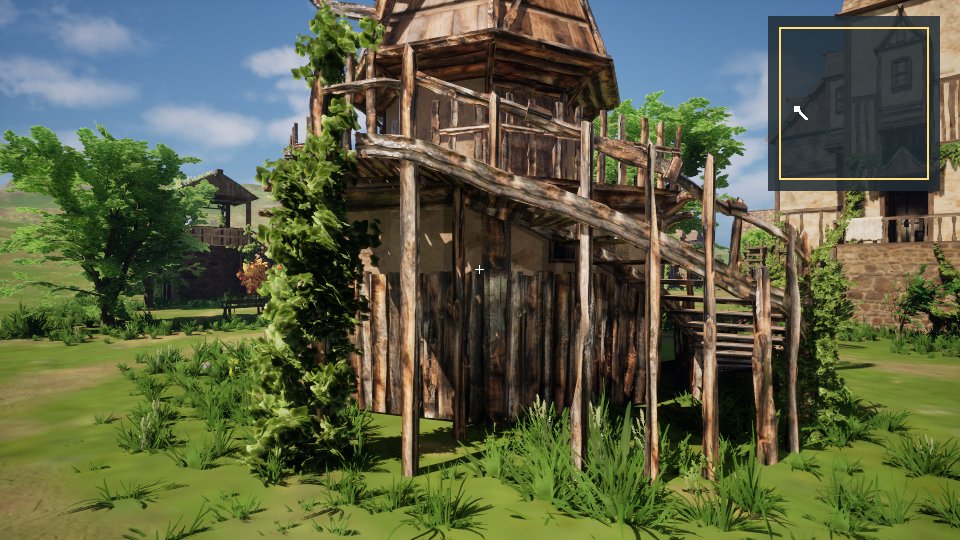}\par
{\centering\footnotesize (b) Windmill Courtyard\par}
\end{minipage}\par
\vspace{5pt}
A medieval settlement with a market and a windmill courtyard. Timber buildings, wooden stalls, benches, barrels, grain sacks, carts, and workshop supplies define the scene. Auditors can explore between structures and inspect their contents; market tasks may include an interactive basket, while the windmill courtyard has no designated interactive objects.\par
\vspace{7pt}
{\color{black!18}\hrule height 0.35pt}
\end{minipage}\par
\par\addvspace{11pt}\noindent\begin{minipage}{\linewidth}
\pdfbookmark[2]{Industrial Factory}{atlas-industrial-factory}
{\normalsize\textbf{Industrial Factory}}\hfill{\footnotesize\color{black!65}Industrial $\cdot$ Unreal Engine 5 $\cdot$ 3 scenes $\cdot$ 19 tasks}\par
\vspace{5pt}
\begin{minipage}[t]{0.325\linewidth}
\vspace{0pt}
\includegraphics[width=\linewidth,trim={0.000bp 84.000bp 230.400bp 0.000bp},clip]{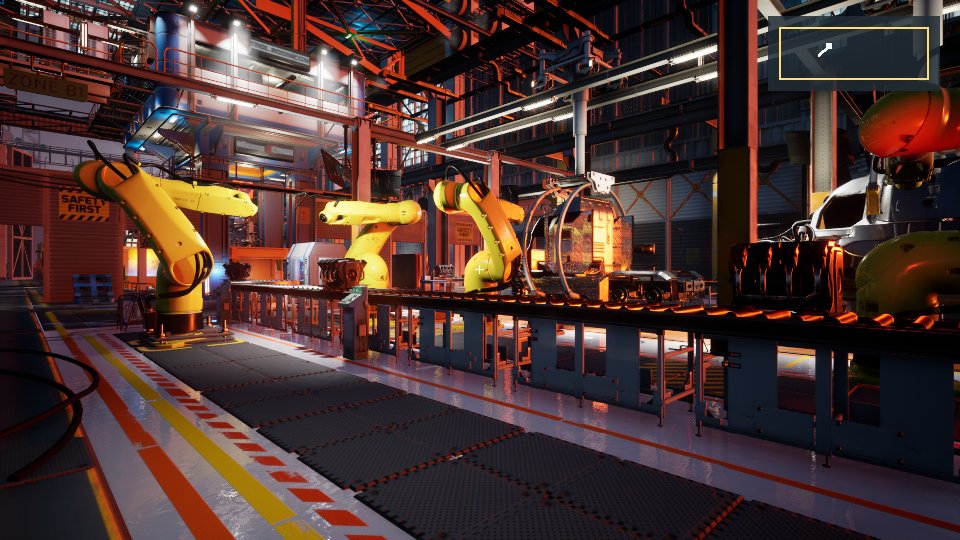}\par
{\centering\footnotesize (a) Assembly Hall\par}
\end{minipage}\hfill
\begin{minipage}[t]{0.325\linewidth}
\vspace{0pt}
\includegraphics[width=\linewidth,trim={64.000bp 0.000bp 64.000bp 0.000bp},clip]{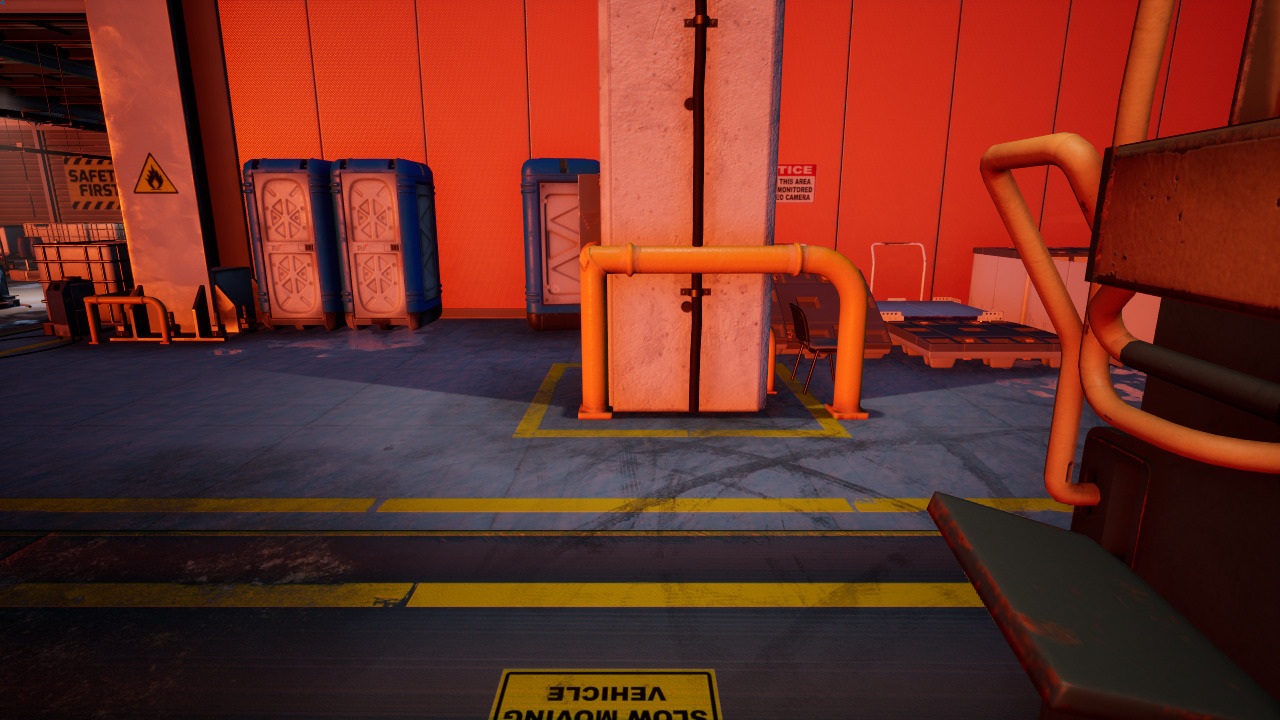}\par
{\centering\footnotesize (b) Vehicle Testing Area\par}
\end{minipage}\hfill
\begin{minipage}[t]{0.325\linewidth}
\vspace{0pt}
\includegraphics[width=\linewidth,trim={0.000bp 84.000bp 230.400bp 0.000bp},clip]{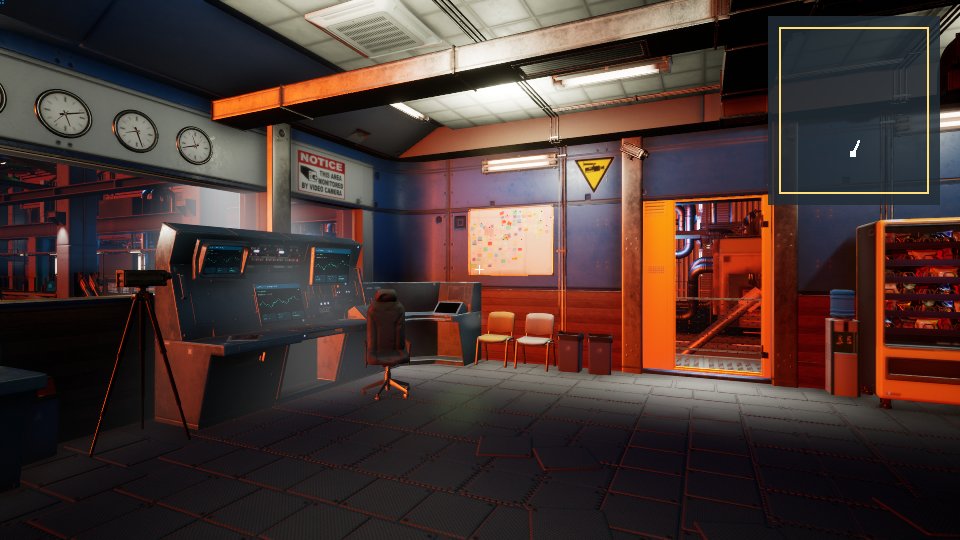}\par
{\centering\footnotesize (c) Control Room\par}
\end{minipage}\par
\vspace{5pt}
A factory complex comprising an assembly hall, a vehicle testing and loading area, and a control room overlooking the production floor. Assembly equipment, lockers, boxes, rails, protective bumpers, and monitoring consoles create contrasting workspaces. The environment supports inspection of spatial layouts, equipment, and structural boundaries.\par
\vspace{7pt}
{\color{black!18}\hrule height 0.35pt}
\end{minipage}\par
\par\addvspace{11pt}\noindent\begin{minipage}{\linewidth}
\pdfbookmark[2]{Sketchbook Airfield}{atlas-sketchbook-airfield}
{\normalsize\textbf{Sketchbook Airfield}}\hfill{\footnotesize\color{black!65}Industrial $\cdot$ Three.js $\cdot$ 1 scene $\cdot$ 18 tasks}\par
\vspace{5pt}
\begin{minipage}[t]{0.43\linewidth}
\vspace{0pt}
\includegraphics[width=\linewidth,trim={48.000bp 0.000bp 48.000bp 0.000bp},clip]{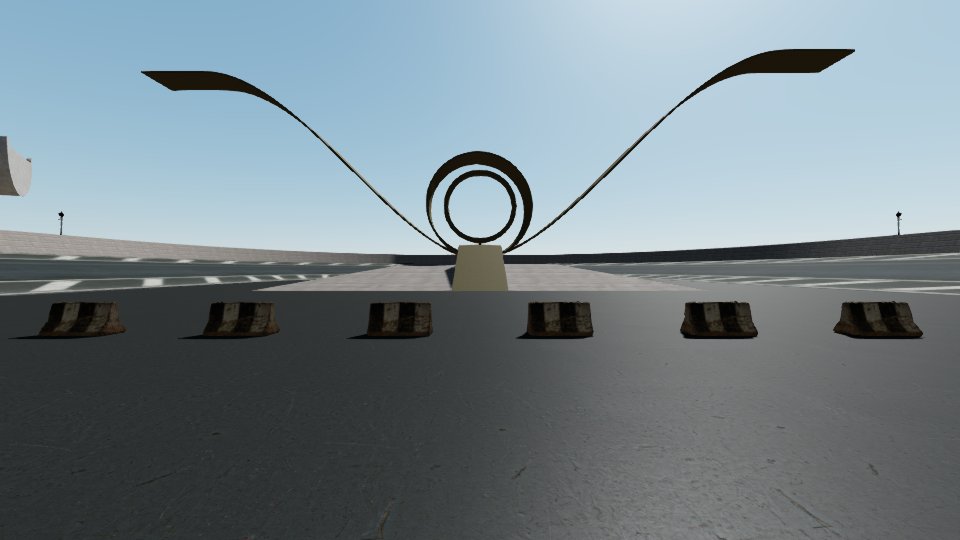}\par
{\centering\footnotesize Airfield\par}
\end{minipage}\hfill
\begin{minipage}[t]{0.545\linewidth}
\vspace{0pt}
A stylized airfield with broad paved areas, parked cars, barriers, barrels, crates, and workshop equipment. Open sightlines contrast with isolated structures and scattered obstacles. Auditors navigate through the scene and inspect objects at different distances; there are no dedicated interactive objects in the current scene configuration.
\end{minipage}\par
\vspace{7pt}
{\color{black!18}\hrule height 0.35pt}
\end{minipage}\par
\par\addvspace{11pt}\noindent\begin{minipage}{\linewidth}
\pdfbookmark[2]{Rural Australia}{atlas-rural-australia}
{\normalsize\textbf{Rural Australia}}\hfill{\footnotesize\color{black!65}Nature $\cdot$ Unreal Engine 5 $\cdot$ 3 scenes $\cdot$ 16 tasks}\par
\vspace{5pt}
\begin{minipage}[t]{0.325\linewidth}
\vspace{0pt}
\includegraphics[width=\linewidth,trim={0.000bp 84.000bp 230.400bp 0.000bp},clip]{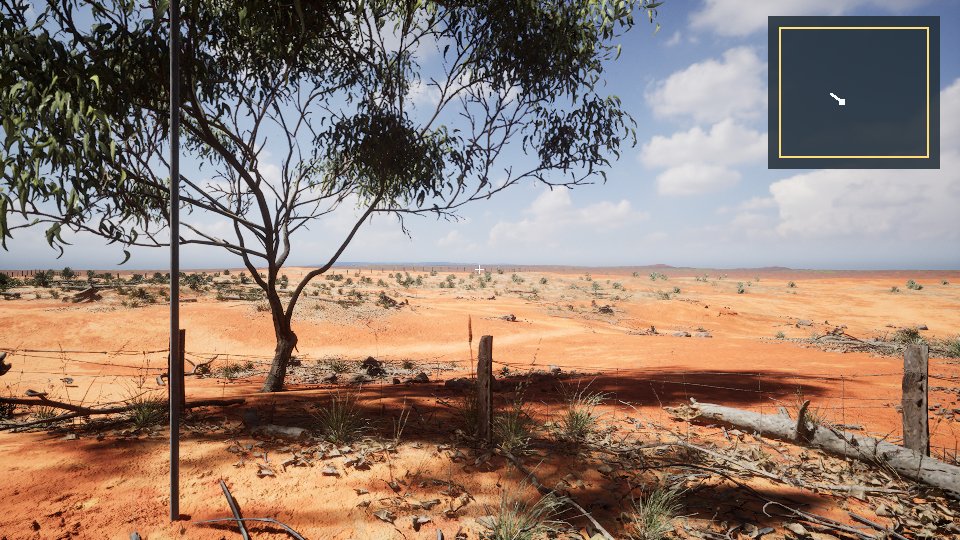}\par
{\centering\footnotesize (a) Bush Road Bend\par}
\end{minipage}\hfill
\begin{minipage}[t]{0.325\linewidth}
\vspace{0pt}
\includegraphics[width=\linewidth,trim={0.000bp 84.000bp 230.400bp 0.000bp},clip]{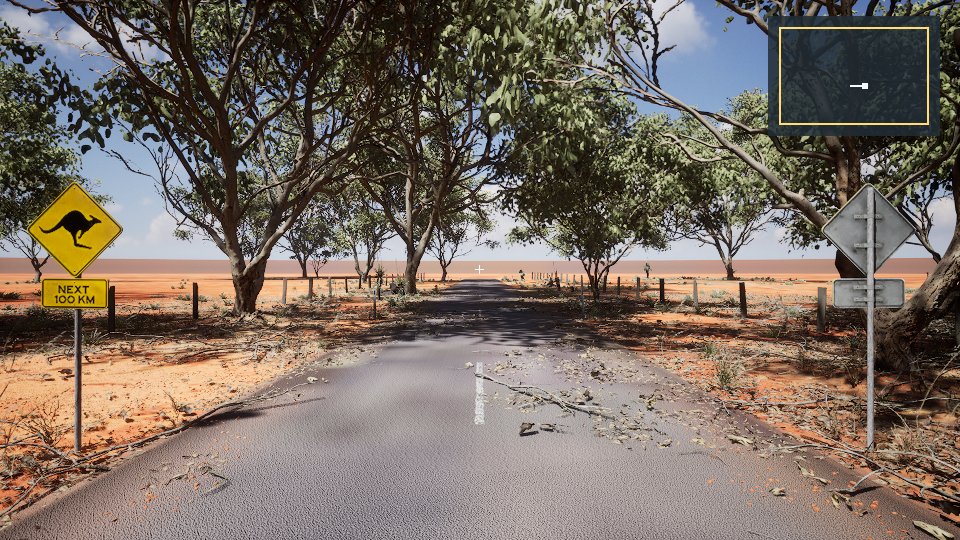}\par
{\centering\footnotesize (b) Country Road\par}
\end{minipage}\hfill
\begin{minipage}[t]{0.325\linewidth}
\vspace{0pt}
\includegraphics[width=\linewidth,trim={0.000bp 84.000bp 230.400bp 0.000bp},clip]{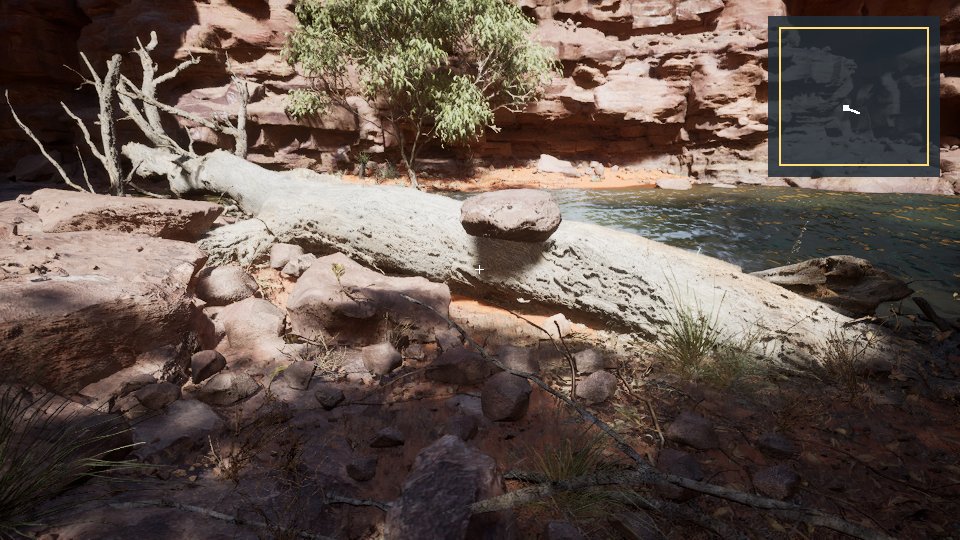}\par
{\centering\footnotesize (c) Creek Canyon\par}
\end{minipage}\par
\vspace{5pt}
An Australian landscape spanning bush roads and a creek canyon. Wire fences, road signs, trees, fallen logs, and rocks define routes and natural boundaries. The road and canyon regions offer different terrain and visibility conditions.\par
\vspace{7pt}
{\color{black!18}\hrule height 0.35pt}
\end{minipage}\par
\par\addvspace{11pt}\noindent\begin{minipage}{\linewidth}
\pdfbookmark[2]{Mistwood Cottage}{atlas-mistwood-cottage}
{\normalsize\textbf{Mistwood Cottage}}\hfill{\footnotesize\color{black!65}Nature $\cdot$ Three.js $\cdot$ 1 scene $\cdot$ 15 tasks}\par
\vspace{5pt}
\begin{minipage}[t]{0.43\linewidth}
\vspace{0pt}
\includegraphics[width=\linewidth,trim={48.000bp 0.000bp 48.000bp 0.000bp},clip]{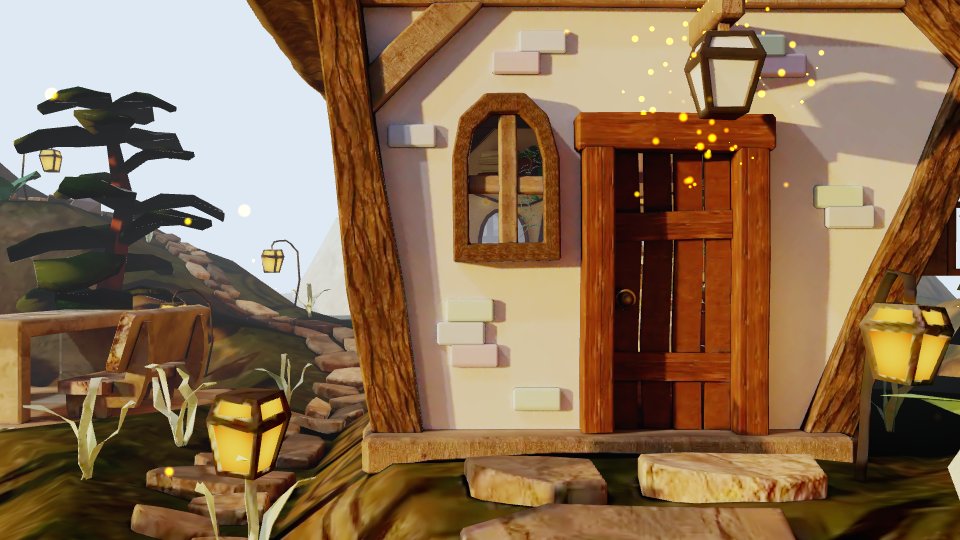}\par
{\centering\footnotesize Cottage And Garden\par}
\end{minipage}\hfill
\begin{minipage}[t]{0.545\linewidth}
\vspace{0pt}
A woodland cottage with a pond, stone paths, trees, and outdoor furniture. Doors and windows reveal domestic details and interior furnishings. Paths connect the compact dwelling to its surrounding garden and wooded slopes.
\end{minipage}\par
\vspace{7pt}
{\color{black!18}\hrule height 0.35pt}
\end{minipage}\par
\par\addvspace{11pt}\noindent\begin{minipage}{\linewidth}
\pdfbookmark[2]{Reef Dive}{atlas-reef-dive}
{\normalsize\textbf{Reef Dive}}\hfill{\footnotesize\color{black!65}Nature $\cdot$ Three.js $\cdot$ 1 scene $\cdot$ 12 tasks}\par
\vspace{5pt}
\begin{minipage}[t]{0.43\linewidth}
\vspace{0pt}
\includegraphics[width=\linewidth,trim={48.000bp 0.000bp 48.000bp 0.000bp},clip]{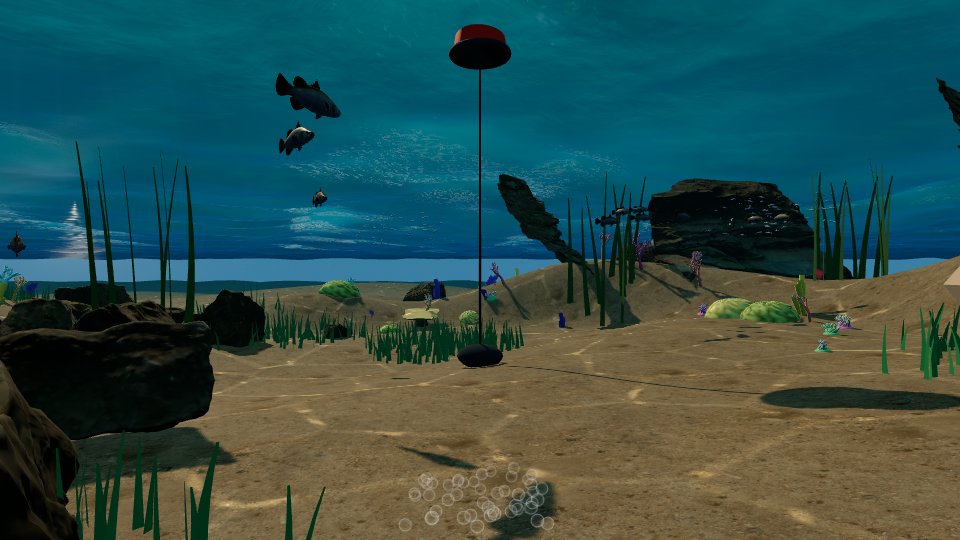}\par
{\centering\footnotesize Underwater Reef\par}
\end{minipage}\hfill
\begin{minipage}[t]{0.545\linewidth}
\vspace{0pt}
A shallow-water reef with rocks, coral, fish, and aquatic vegetation. Nearby plants and seabed features appear against a blue-water background, providing varied underwater viewpoints for inspecting spatial relationships and visual consistency.
\end{minipage}\par
\vspace{7pt}
{\color{black!18}\hrule height 0.35pt}
\end{minipage}\par
\par\addvspace{11pt}\noindent\begin{minipage}{\linewidth}
\pdfbookmark[2]{Beyond Fable Wilderness}{atlas-beyond-fable-wilderness}
{\normalsize\textbf{Beyond Fable Wilderness}}\hfill{\footnotesize\color{black!65}Nature $\cdot$ Three.js $\cdot$ 1 scene $\cdot$ 14 tasks}\par
\vspace{5pt}
\begin{minipage}[t]{0.43\linewidth}
\vspace{0pt}
\includegraphics[width=\linewidth,trim={48.000bp 0.000bp 48.000bp 0.000bp},clip]{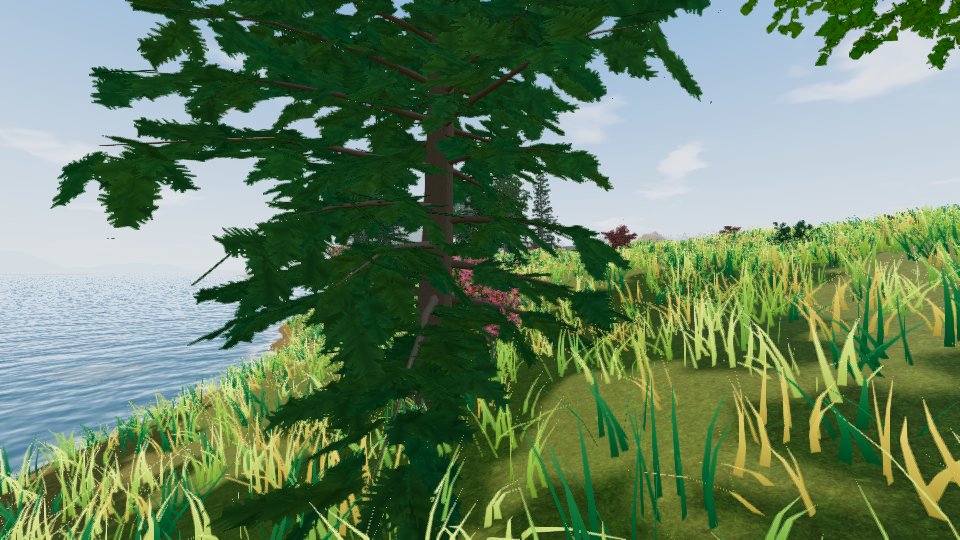}\par
{\centering\footnotesize Wooded Shoreline\par}
\end{minipage}\hfill
\begin{minipage}[t]{0.545\linewidth}
\vspace{0pt}
A grassy wilderness with pine trees, shrubs, boulders, and sloping terrain beside water. Dense vegetation can partially obscure objects. The shoreline and trees provide landmarks for revisiting locations and comparing observations.
\end{minipage}\par
\vspace{7pt}
{\color{black!18}\hrule height 0.35pt}
\end{minipage}\par
\endgroup

\subsection{Sources And Licenses}
\label{app:sources-licenses}
\textbf{Unreal Engine 5.} Our Unreal environments use packages obtained through \href{https://www.fab.com/eula}{Fab}, subject to their acquisition licenses. We plan to release runnable benchmark packages without the original project source files or standalone third-party assets.

\textbf{Three.js.} \href{https://github.com/amiradeu/mistwood-cottage}{Mistwood Cottage} is licensed under CC BY 4.0. Sketchbook Airfield, Reef Dive, and Beyond Fable Wilderness use MIT-licensed code from \href{https://github.com/swift502/Sketchbook}{Sketchbook}, \href{https://github.com/VictorZakharov/beautiful-water}{beautiful-water}, and \href{https://github.com/xikhar/beyond-fable}{beyond-fable}, respectively. Family House and other scenes include \href{https://polyhaven.com/license}{Poly Haven CC0 assets}. Sponza Atrium retains the \href{https://github.com/KhronosGroup/glTF-Sample-Assets/blob/main/Models/Sponza/README.md}{Khronos-distributed model's separate Crytek licensing terms}.

Third-party components retain their original licenses. Releases will preserve required copyright and license notices, creator attributions, and modification notices.

\clearpage
\section{Dataset Creation}
\label{app:dataset-creation}
This appendix details the task construction and human validation process described in Section~\ref{sec:data-collection} and Figure~\ref{fig:data-collection}. The dataset contains 213 manually constructed single-anomaly tasks across 13 environments: 126 tasks in Unreal Engine 5 and 87 in Three.js.

\subsection{Scene And Task Construction}
\label{app:construction}
We collect environments from publicly available sources and select bounded regions within larger environments as individual scenes. These scenes retain distinct spatial layouts, objects, and settings. Appendix~\ref{app:environments} presents the environments and their sources.

For each task, the authors start from a configuration without the target anomaly and specify the agent's initial position and orientation, exploration boundaries, and available interactions. They then introduce one anomaly from the taxonomy in Section~\ref{sec:taxonomy}. Static physics anomalies alter object support, overlap, or scale; interactive physics anomalies alter collision or responses to contact. Spatial consistency anomalies change appearance across viewpoints, visibility, or lighting, while temporal consistency anomalies change an object's existence, attributes, or behavior over time. Semantic consistency anomalies involve improper object configurations or objects incompatible with the scene's historical setting.

The authors check that the scene loads correctly, the target region is accessible, and the intended anomaly occurs during exploration. For anomalies involving interaction or changes across observations, they also check the relevant action sequence, viewpoints, or revisits before submitting the task for human validation.

\subsection{Task Annotations}
\label{app:annotations}
Each task includes a scene description, a ground-truth anomaly category, and an evaluation rubric. The scene description introduces the setting and available interactions. The rubric describes the target anomaly and its expected normal appearance or behavior, providing a common reference for human validation and evaluation of agent reports. It is not supplied to the auditing agent; agent inputs are described in Appendix~\ref{app:case-input}.

\begin{table}[!ht]
\centering
\caption{Task annotations and human validation records.}
\label{tab:creation-annotations}
\small
\renewcommand{\arraystretch}{1.2}
\begin{tabularx}{\linewidth}{@{}p{0.23\linewidth}X@{}}
\toprule
Record & Contents \\
\midrule
Scene description & The scene's setting and available interactions. \\
Anomaly category & One of the fifteen categories in the taxonomy. \\
Evaluation rubric & The target anomaly and its expected normal appearance or behavior. \\
Human validation & Quality judgment, difficulty rating, and comments on the inspected task. \\
\bottomrule
\end{tabularx}
\end{table}
\FloatBarrier

\begin{tcolorbox}[enhanced,sharp corners,boxrule=0.4pt,colback=teal!3!white,colframe=teal!30!white,colbacktitle=teal!9!white,coltitle=teal!65!black,fonttitle=\sffamily\bfseries\small,title={Example: Unsupported Grain Sacks},fontupper=\small,left=7pt,right=7pt,top=5pt,bottom=5pt]
\textbf{Scene.} A medieval market with wooden stalls, benches, barrels, and grain sacks.\par
\textbf{Category.} Missing support / floating (static physics).\par
\textbf{Evaluation rubric.} The grain sacks float above the ground without support. They should rest on the ground.
\end{tcolorbox}

\clearpage
\subsection{Human Validation}
\label{app:human-annotation}
\begin{figure}[!ht]
\centering
\includegraphics[width=\linewidth]{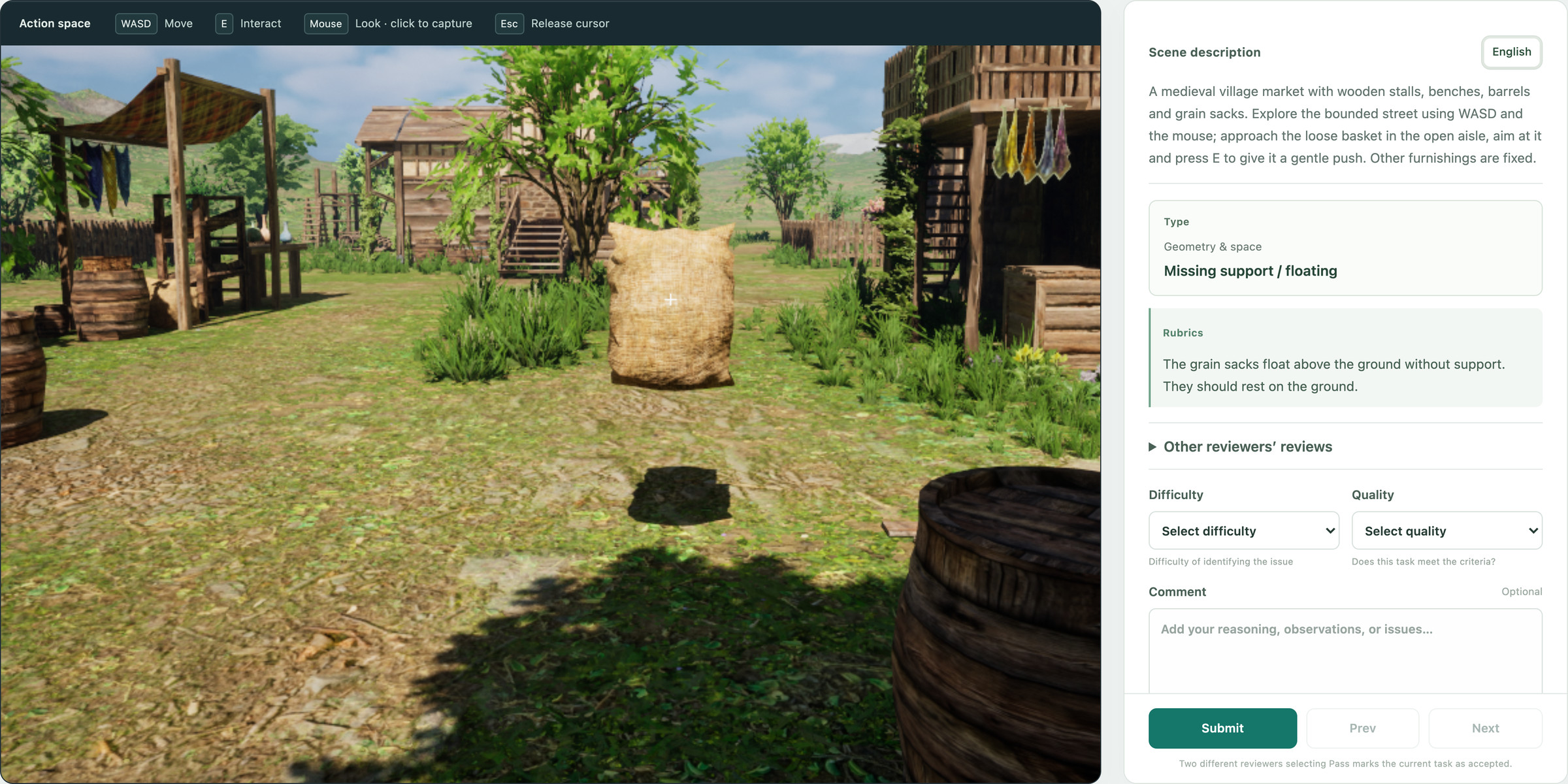}
\caption{\textbf{Human validation interface.} Reviewers inspect the environment alongside its scene description, anomaly category, and evaluation rubric, then record quality, difficulty, and comments.}
\label{fig:annotation-interface}
\end{figure}

Reviewers who did not construct the tasks inspect them through a browser interface (Figure~\ref{fig:annotation-interface}). Unreal Engine 5 scenes are streamed from a remote runtime, while Three.js scenes run in the browser. Reviewers receive the scene description, anomaly category, and evaluation rubric. They navigate, change viewpoint, and interact with objects to check whether the target anomaly can be observed and whether it matches the rubric.

\FloatBarrier

\paragraph{Quality And Difficulty.}
Reviewers assign a quality judgment of Pass, Fail, or Uncertain. Pass indicates that the target anomaly is observable and matches its rubric; Fail or Uncertain prompts further revision and review. They also rate identification difficulty as Easy, Medium, or Hard and can leave comments to describe ambiguous behavior or problems with the task. These ratings assess dataset quality and are separate from the human auditing baseline in Section~\ref{sec:experiments}.

\paragraph{Revision And Acceptance.}
The authors revise tasks and rubrics based on reviewer feedback and submit them for another round of validation. A task is accepted only after at least two distinct reviewers rate the same scene configuration as Pass. Repeated submissions by one reviewer count as one judgment. Changes to the target anomaly or rubric require renewed validation, so acceptance reflects the task configuration used in the benchmark.

\clearpage
\section{Evaluation Protocol}
\label{app:cases}
\begingroup
\setlength{\parindent}{0pt}
\definecolor{caseblue}{HTML}{365F91}
\definecolor{caseteal}{HTML}{287D7A}
\definecolor{casepurple}{HTML}{7961A3}
\definecolor{casegold}{HTML}{9B732E}
\definecolor{caserose}{HTML}{A15162}
\newtcolorbox{casebox}[2]{enhanced,sharp corners,boxrule=0.4pt,
  colback=#1!4!white,colframe=#1!35!white,colbacktitle=#1!12!white,
  coltitle=#1!85!black,fonttitle=\sffamily\bfseries\small,
  fontupper=\fontsize{9.2}{11.4}\selectfont,
  title={#2},left=7pt,right=7pt,top=5pt,bottom=5pt,
  before skip=7pt,after skip=7pt}
\subsection{Agent Inputs And Instructions}
\label{app:case-input}
We illustrate the initial input $x$ in Section~\ref{sec:task-definition} using an inspection of a medieval market. It consists of \textcolor{caseblue}{auditing instructions}, task context, and an \textcolor{caseteal}{initial RGB observation}. The task context contains the scene description, an anomaly-type hint, and a \textcolor{casepurple}{matching in-context example}. The target object's identity, location, and evaluation rubric are withheld.

\begin{casebox}{caseblue}{Auditing Instructions}
Inspect the assigned 3D environment for bugs. Explore and gather visual evidence. Report each distinct bug with a clear description, category, and evidence frame refs. Do not assume every unusual object is a bug. Use the full 40-action budget for exploration and verification, even after finding an initial bug. Finish earlier only if the environment fails. Report bugs through flag\_bug/update\_bug and finish with done. If you find no bug, say so explicitly in done.
\end{casebox}

\begin{casebox}{caseteal}{Task Context And Initial Observation}
\begin{minipage}[t]{.48\linewidth}\vspace{0pt}\textbf{Scene description.} A medieval village market with wooden stalls, benches, barrels and grain sacks. Interactive object: basket.\par\vspace{7pt}\textbf{Anomaly-type hint.} Missing support / floating.\end{minipage}\hfill\begin{minipage}[t]{.48\linewidth}\vspace{0pt}\includegraphics[width=\linewidth,trim=0 30 192 78,clip]{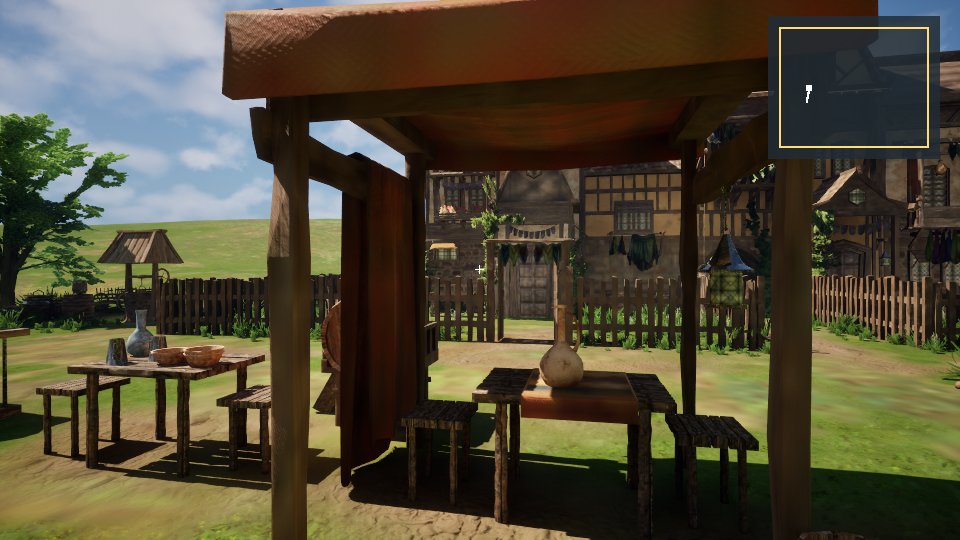}\par{\footnotesize Initial RGB observation from the starting pose.}\end{minipage}
\end{casebox}

\phantomsection\label{app:case-icl}
\begin{casebox}{casepurple}{In-Context Example: Missing Support / Floating}
\textbf{Definition.} Missing support / floating occurs when an object that should rest on the ground or another structure has an unexplained gap beneath its supporting parts.\par\vspace{6pt}\begin{minipage}[t]{.43\linewidth}\vspace{0pt}\includegraphics[width=\linewidth]{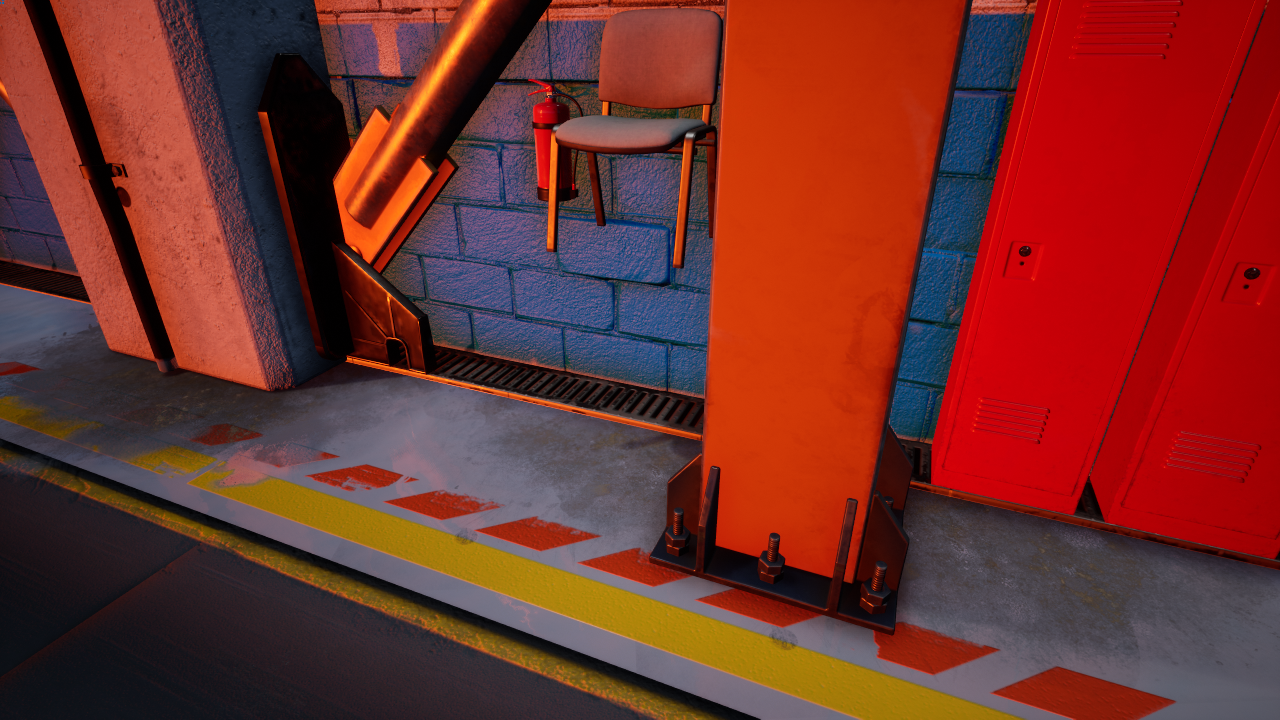}\par{\footnotesize Figure 1: Side view of the chair legs and the floor below.}\end{minipage}\hfill\begin{minipage}[t]{.54\linewidth}\vspace{0pt}\textbf{Reference explanation.} In Figure 1, the chair legs are clearly separated from the floor, with no structure supporting the chair; this demonstrates missing support.\end{minipage}
\end{casebox}

\begin{casebox}{casepurple}{Instruction For Using The In-Context Example}
These are supplied demonstrations, not questions you need to answer. Read this example, then call observe to start your assigned scene. Do not use example images as evidence for your own bug reports.
\end{casebox}

The in-context example uses a different scene and explains the relevant support relation without revealing the target in the market. Appendix~\ref{app:full-examples} provides the full set of in-context examples and category definitions.
\clearpage
\subsection{Actions And Tool Use}
\label{app:trajectory}
Table~\ref{tab:tools} lists the available actions and tools, their inputs, and their effects. The illustrated examples below use the medieval-market inspection introduced in Appendix~\ref{app:case-input} and a door interaction in the ancient Chinese courtyard. Movement, viewpoint changes, interaction, and waiting consume environment steps; observation, memory, and reporting operations are recorded separately.
\begin{table}[!htbp]
\caption{\textbf{Actions and tools available to the auditor.} Movement, viewpoint changes, interaction, and waiting each cost one step; observation, memory, and reporting cost none.}
\label{tab:tools}
\centering\small
\setlength{\tabcolsep}{6pt}
\renewcommand{\arraystretch}{1.18}
\begin{tabularx}{\linewidth}{p{0.18\linewidth}p{0.25\linewidth}X}
\toprule
\textbf{Operation} & \textbf{Input} & \textbf{Effect} \\
\midrule
\rowcolor{violet!7}\multicolumn{3}{c}{\textit{Navigation and Interaction}} \\
Move & Distance, direction & Move 0.3--4\,m forward/backward; stop at obstacles. \\
Turn & Angle & Turn left or right by up to $180^\circ$. \\
Look & Angle & Tilt the view up or down by up to $90^\circ$. \\
Interact & None & Activate a reachable object under the crosshair. \\
Wait & Duration & Stay still for 0.5--5\,s to observe changes over time. \\
\rowcolor{green!7}\multicolumn{3}{c}{\textit{Observation and examples}} \\
Observe & None & Start the inspection or retrieve the current view. \\
Read example & Assigned example & View a demonstration and its reference answer. \\
\rowcolor{blue!6}\multicolumn{3}{c}{\textit{Memory}} \\
Review frames & Frame IDs,\newline optional crop & Retrieve 1--4 saved images or selected crops. \\
Review history & Range of steps & Review earlier actions and observation descriptions. \\
Take notes & Text & Save notes for later reference. \\
\rowcolor{orange!9}\multicolumn{3}{c}{\textit{Anomaly reports}} \\
Report anomaly & Description, type,\newline status, evidence & Record a suspected anomaly and link supporting images. \\
Revise report & Report ID, changes & Update or retract a previous report. \\
List reports & None & Review all anomalies reported so far. \\
Finish & Summary & Submit the final reports and end the inspection. \\
\bottomrule
\end{tabularx}
\end{table}

\FloatBarrier
\begingroup

\newcommand{\toolview}[2]{\begin{minipage}[t]{.335\linewidth}\vspace{0pt}#1\par\centering\footnotesize #2\end{minipage}}
\newcommand{\actionexample}[7]{%
\noindent\begin{minipage}[t]{.285\linewidth}\vspace{0pt}
{\color{caseblue}\bfseries #1}\par\smallskip
{\small #2}\par\smallskip
{\small\raggedright #3\par}\end{minipage}\hfill\toolview{#4}{#5}\hfill\toolview{#6}{#7}\par\medskip}
\newtcolorbox{toolcard}[2]{enhanced,boxrule=0pt,leftrule=1.5pt,
 colback=#1!3!white,colframe=#1!65!white,arc=0pt,
 title={#2},colbacktitle=#1!10!white,coltitle=#1!85!black,
 fonttitle=\small\bfseries,fontupper=\small\raggedright,
 left=6pt,right=6pt,top=5pt,bottom=5pt,before skip=0pt,after skip=0pt}
\newcommand{\toolinput}[1]{\textbf{Input.} #1\par\smallskip}
\newcommand{\tooloutput}[1]{\textbf{Output.} #1\par}

\clearpage
\noindent{\large\bfseries Navigation And Interaction}\par
Each pair shows the observation before an action and after its effect; the door is shown after its closing animation, with enlarged interaction prompts. Positive turning angles turn right; positive looking angles tilt down.
\par\medskip
\actionexample{Move}{\texttt{distance\_m: 2.5}\\\texttt{direction: forward}}{The agent approaches the stall.}{\includegraphics[width=\linewidth,trim=0 30 192 78,clip]{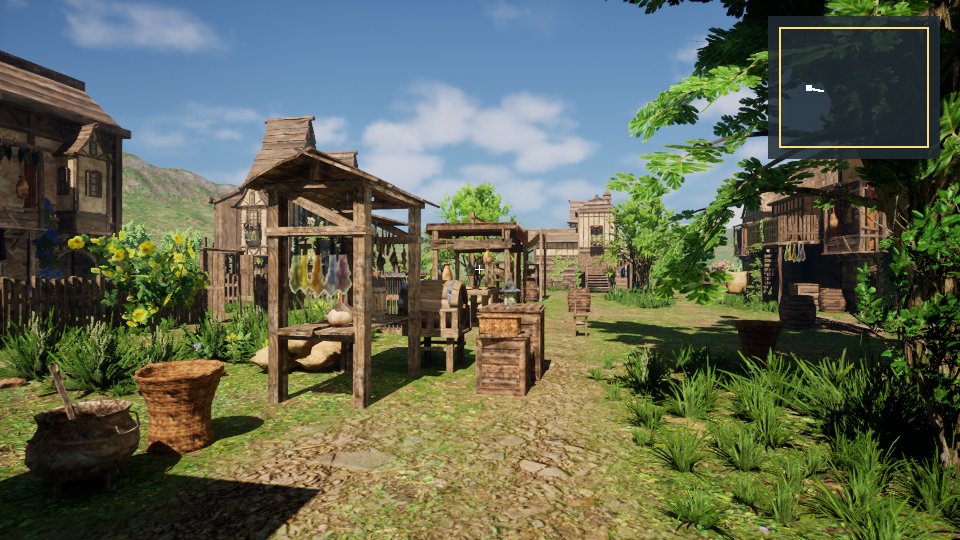}}{Before}{\includegraphics[width=\linewidth,trim=0 30 192 78,clip]{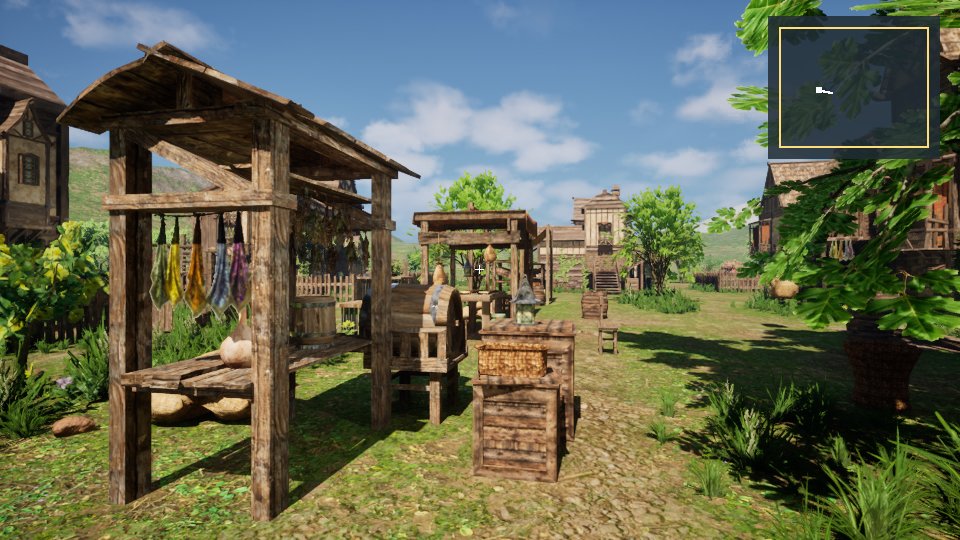}}{After}
\actionexample{Turn}{\texttt{degrees: 90}}{The agent rotates in place to inspect another direction.}{\includegraphics[width=\linewidth,trim=0 30 192 78,clip]{assets/cases/MV01/a000.jpg}}{Before}{\includegraphics[width=\linewidth,trim=0 30 192 78,clip]{assets/cases/MV01/a001.jpg}}{After}
\actionexample{Look}{\texttt{degrees: 20}}{The camera tilts down toward the basket and its supporting surface.}{\includegraphics[width=\linewidth,trim=0 30 192 78,clip]{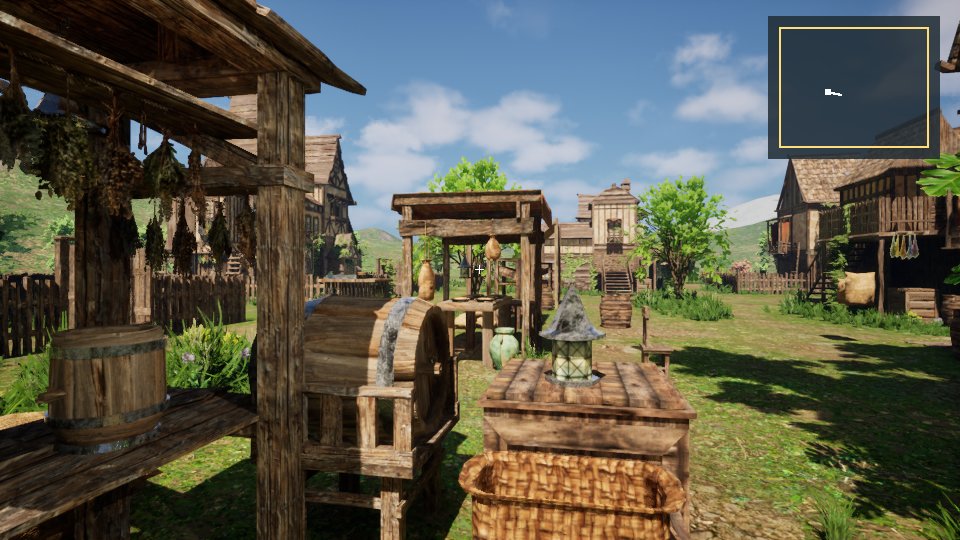}}{Before}{\includegraphics[width=\linewidth,trim=0 30 192 78,clip]{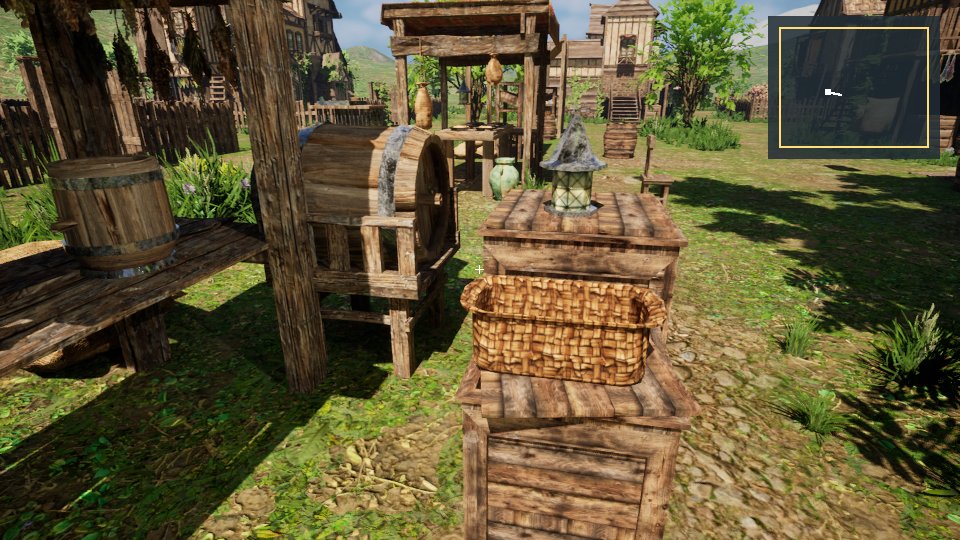}}{After}
\actionexample{Interact}{No arguments.}{The agent activates the \texttt{E: Close door} prompt; the door closes and the prompt changes to \texttt{Open door}.}{\includegraphics[width=\linewidth,trim=290 0 0 164,clip]{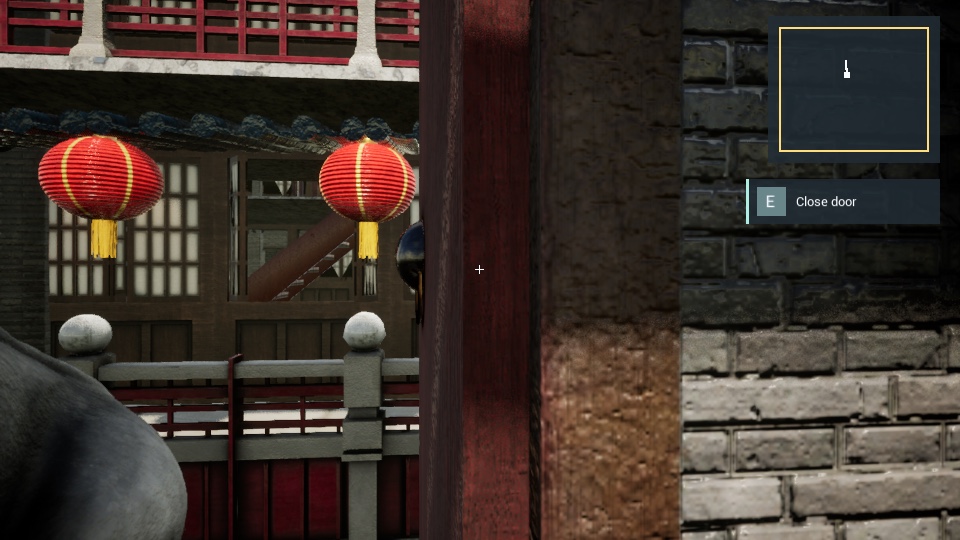}\par\smallskip\centering\includegraphics[width=.70\linewidth,trim=746 316 20 179,clip]{assets/cases/tools/door-before.jpg}}{Before}{\includegraphics[width=\linewidth,trim=290 0 0 164,clip]{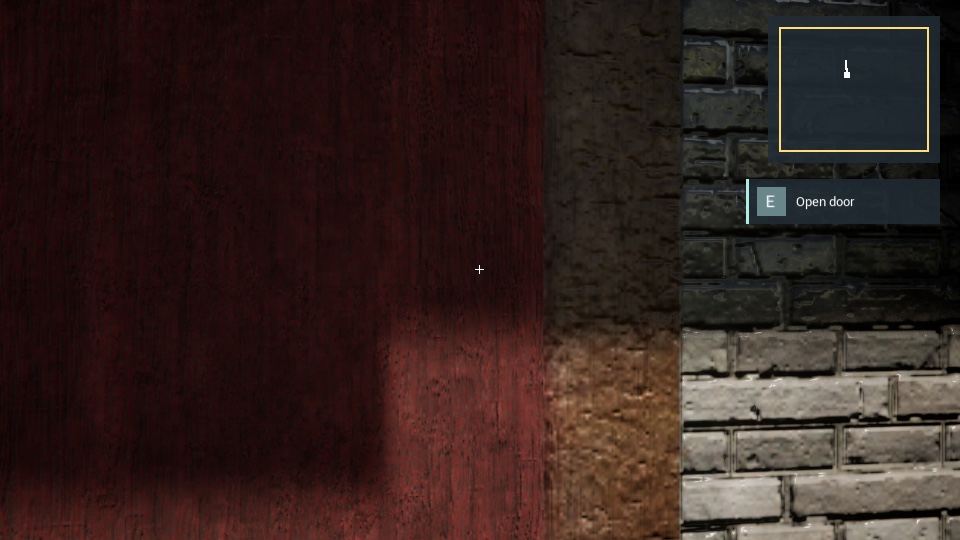}\par\smallskip\centering\includegraphics[width=.70\linewidth,trim=746 316 20 179,clip]{assets/cases/tools/door-after.jpg}}{After}
\actionexample{Wait}{\texttt{seconds: 1}}{Time advances by one second while the agent remains stationary.}{\includegraphics[width=\linewidth,trim=0 30 192 78,clip]{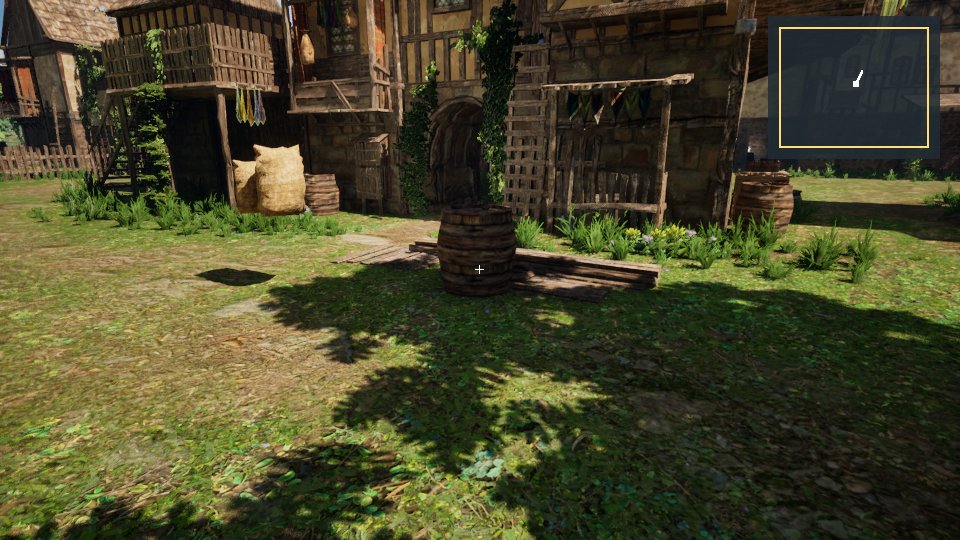}}{Before}{\includegraphics[width=\linewidth,trim=0 30 192 78,clip]{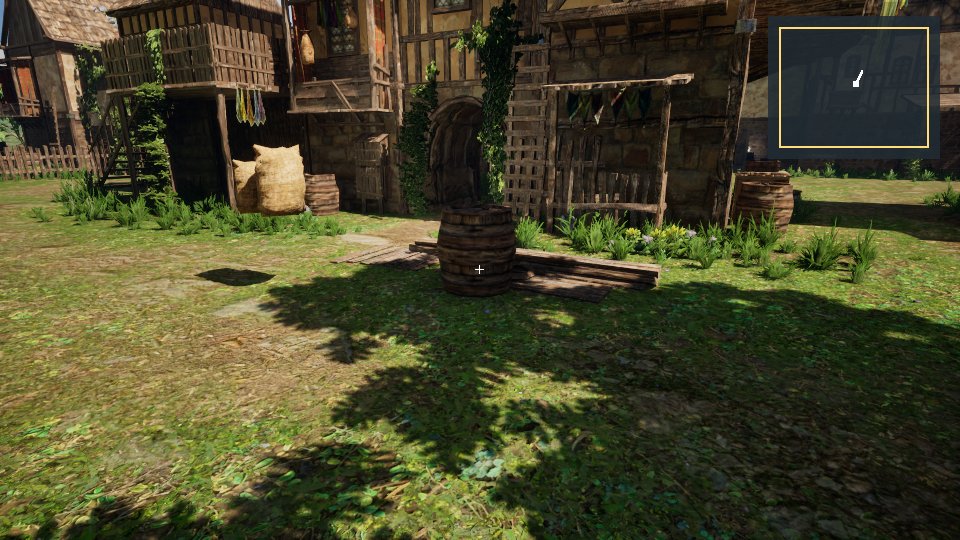}}{After}

\clearpage
\noindent{\large\bfseries Observation And Memory}\par
Observation tools return scene images or demonstrations; memory tools retrieve earlier evidence or store notes. The history query below is illustrative; the other calls were recorded during this inspection.
\par\medskip
\noindent\begin{minipage}[t]{.487\linewidth}\vspace{0pt}
\begin{toolcard}{casepurple}{Read Example}
\toolinput{The assigned missing-support / floating demonstration.}
\begin{center}\includegraphics[width=\linewidth,height=1.04in,keepaspectratio,]{assets/demonstrations/G1_01.png}\end{center}
\tooloutput{A chair image and reference explanation: its legs are separated from the floor without support.}
\end{toolcard}
\end{minipage}\hfill\begin{minipage}[t]{.487\linewidth}\vspace{0pt}
\begin{toolcard}{caseteal}{Observe}
\toolinput{No arguments.}
\begin{center}\includegraphics[width=\linewidth,height=1.04in,keepaspectratio,trim=0 30 192 78,clip]{assets/cases/MV01/a000.jpg}\end{center}
\tooloutput{The initial market view, task context, and remaining budget of 40 actions.}
\end{toolcard}
\end{minipage}\par\medskip
\begin{toolcard}{caseblue}{Review Frames}
\toolinput{Frame from step 11; crop region $[0.3,0.4,0.6,0.8]$ in normalized image coordinates.}
\begin{minipage}[t]{.487\linewidth}\vspace{0pt}\includegraphics[width=\linewidth,trim=0 30 192 78,clip]{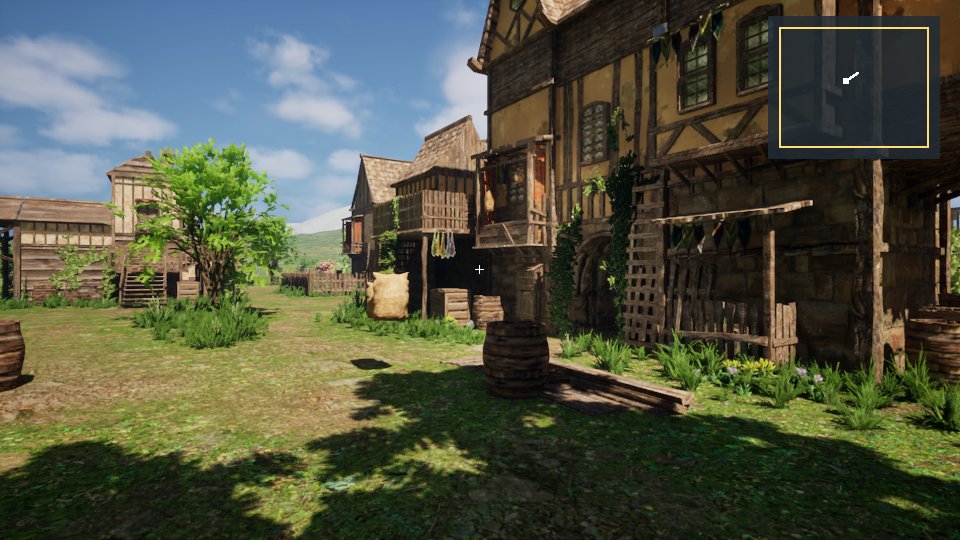}\par\centering\footnotesize Archived observation: step 11\end{minipage}\hfill
\begin{minipage}[t]{.487\linewidth}\vspace{0pt}\includegraphics[width=\linewidth,height=1.17in,keepaspectratio]{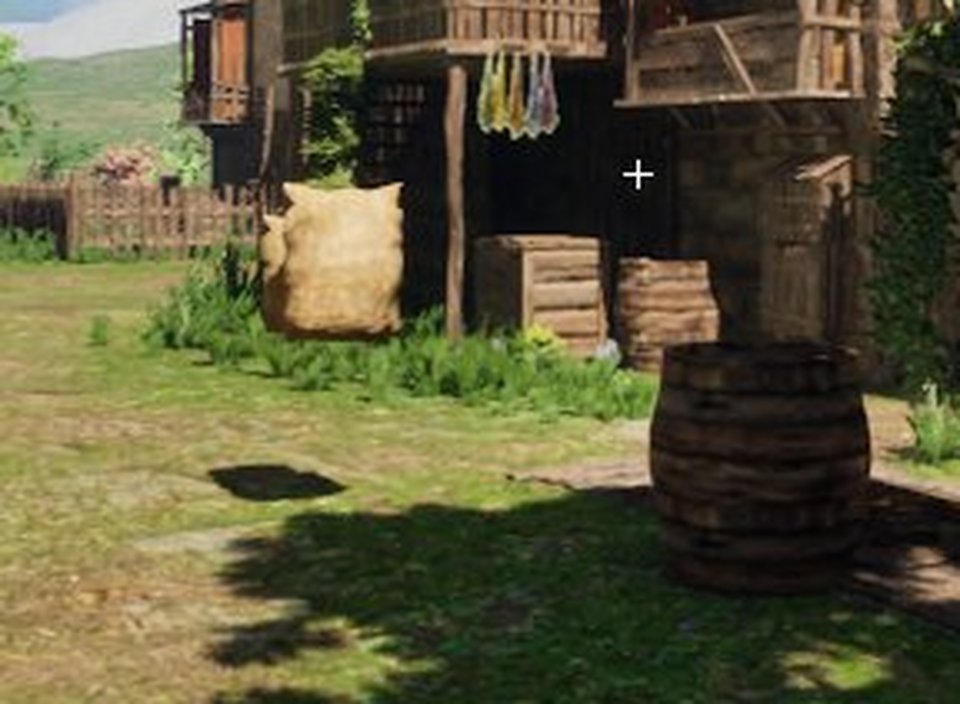}\par\centering\footnotesize Returned crop: unsupported sack\end{minipage}\par\smallskip
\tooloutput{An enlarged view of the sack and the empty space below it, retrieved without taking another environment action.}
\end{toolcard}\par\medskip
\noindent\begin{minipage}[t]{.487\linewidth}\vspace{0pt}
\begin{toolcard}{caseblue}{Review History}
\toolinput{Steps 1--2 (illustrative query).}
\begin{center}\includegraphics[width=\linewidth,height=1.04in,keepaspectratio,trim=0 30 192 78,clip]{assets/cases/MV01/a002.jpg}\end{center}
\tooloutput{Text records: step 1, turn right $90^\circ$; step 2, move forward 2.5\,m. The image above provides context; history returns text only.}
\end{toolcard}
\end{minipage}\hfill\begin{minipage}[t]{.487\linewidth}\vspace{0pt}
\begin{toolcard}{caseblue}{Take Notes}
\toolinput{``Observed floating grain sack \ldots Next: approach for closer inspection and multiple angles.''}
\begin{center}\includegraphics[width=\linewidth,height=1.04in,keepaspectratio,]{assets/cases/MV01/a011_c01.jpg}\end{center}
\tooloutput{``Note added (1 entry).'' The observation and planned follow-up are saved for later reference.}
\end{toolcard}
\end{minipage}

\clearpage

\noindent{\large\bfseries Anomaly Reporting And Completion}\par
The agent first records a suspicion, gathers additional views, and updates the same report. These tools return text records; the screenshots below show the evidence associated with each operation. Report confirmation reflects the agent's assessment, not the judge's decision.
\par\medskip
\noindent\begin{minipage}[t]{.487\linewidth}\vspace{0pt}
\begin{toolcard}{casegold}{Report Anomaly}
\begin{center}\includegraphics[width=\linewidth,height=1.45in,keepaspectratio,]{assets/cases/MV01/a011_c01.jpg}\end{center}
\toolinput{Floating grain sack near the porch; category: geometry; status: \texttt{suspect}; evidence: steps 9, 10, and 11.}
\tooloutput{A new report is recorded with status \texttt{suspect} and references to the three evidence frames.}
\end{toolcard}
\end{minipage}\hfill\begin{minipage}[t]{.487\linewidth}\vspace{0pt}
\begin{toolcard}{casegold}{Revise Report}
\begin{center}\includegraphics[width=\linewidth,height=1.45in,keepaspectratio,trim=0 30 192 78,clip]{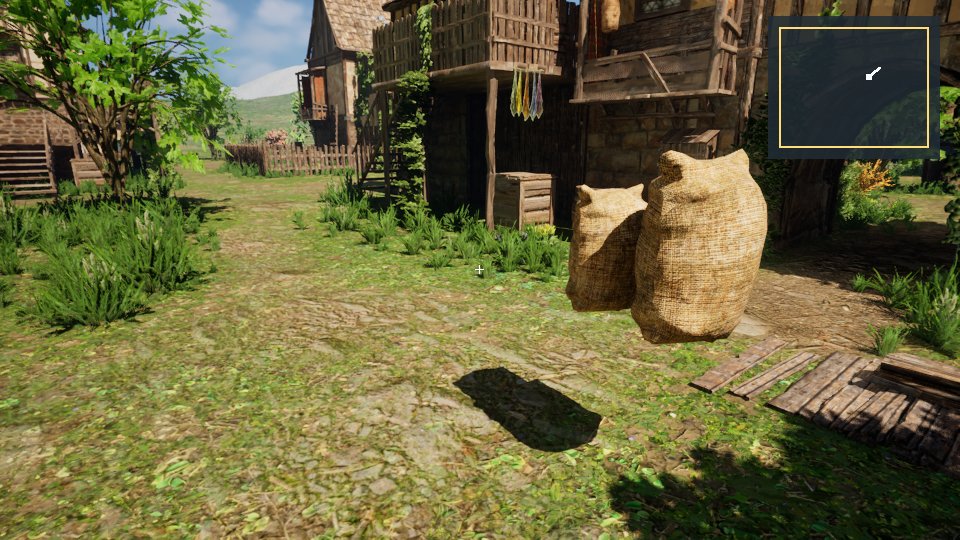}\end{center}
\toolinput{The existing report; status: \texttt{confirmed}; revised description of floating sacks; evidence: steps 9, 11, 14, 15, and 19.}
\tooloutput{The same report is updated with the additional views and confirmed status.}
\end{toolcard}
\end{minipage}\par\medskip
\noindent\begin{minipage}[t]{.487\linewidth}\vspace{0pt}
\begin{toolcard}{casegold}{List Reports}
\toolinput{No arguments.}
\begin{center}\includegraphics[width=\linewidth,height=1.45in,keepaspectratio,trim=0 30 192 78,clip]{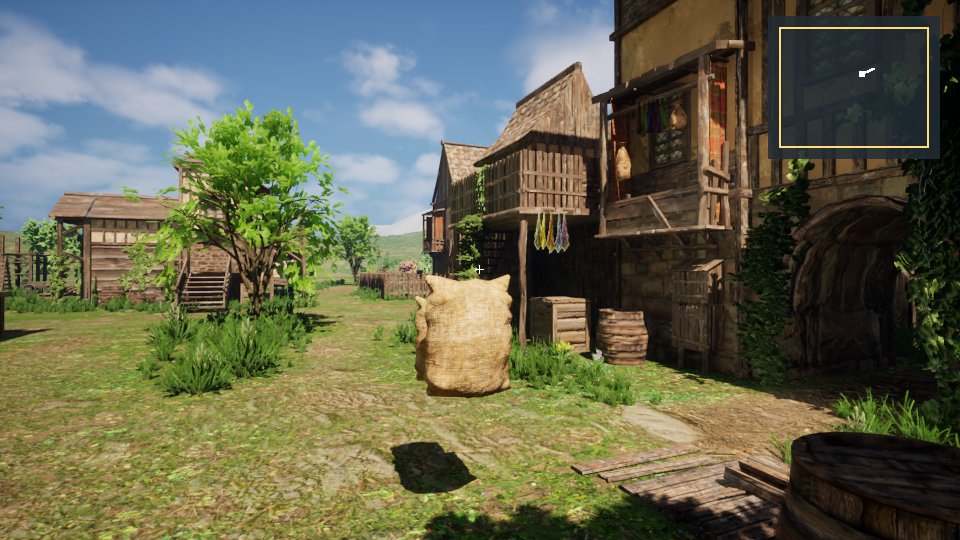}\end{center}
\tooloutput{One confirmed report: grain sacks floating unsupported near the porch; evidence from steps 9, 11, 14, 15, and 19.}
\smallskip
The agent checks its accumulated findings before finishing.
\end{toolcard}
\end{minipage}\hfill\begin{minipage}[t]{.487\linewidth}\vspace{0pt}
\begin{toolcard}{casegold}{Finish}
\toolinput{A summary of the completed 40-action inspection and the confirmed floating-sack finding.}
\begin{center}\includegraphics[width=\linewidth,height=1.45in,keepaspectratio,trim=0 30 192 78,clip]{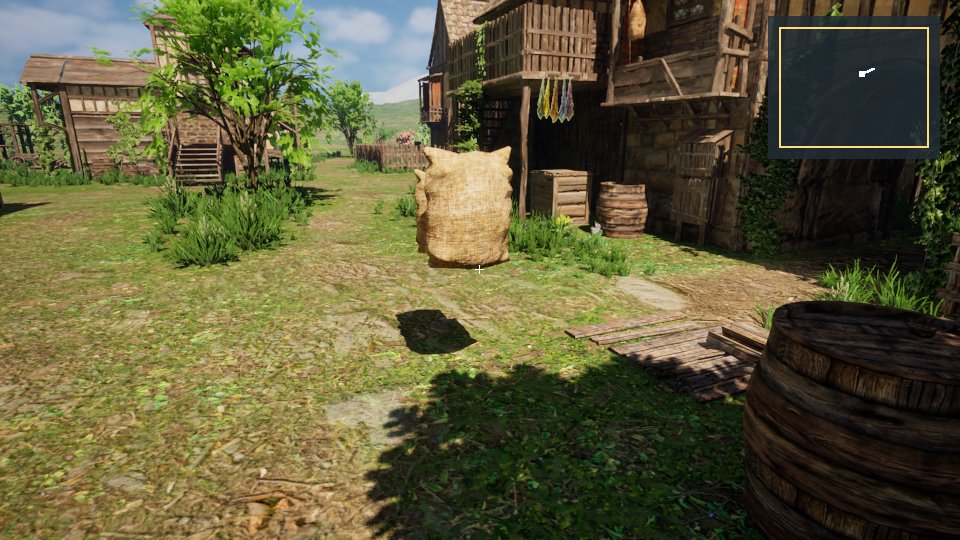}\end{center}
\tooloutput{``Inspection finished.'' The final report and summary are submitted for evaluation.}
\smallskip
Appendix~\ref{app:case-output} shows the submitted report and evidence.
\end{toolcard}
\end{minipage}
\endgroup

\clearpage
\subsection{Agent Output}
\label{app:case-output}
The agent records a suspected anomaly after examining the sacks, revises it after observing additional viewpoints, and submits its final report at the end of the action budget. Below are the final description, selected evidence, and an excerpt of the inspection summary. The confirmation status is the agent's own assessment; success is determined separately by the judge.

\begin{casebox}{casegold}{Final Anomaly Report}
\textbf{Category:} Geometry\qquad\textbf{Status:} Confirmed\par\vspace{5pt}Grain sacks (burlap sacks) are floating unsupported in mid-air in front of the village house/porch, elevated above the ground with complete empty space underneath and casting a detached shadow onto the ground.\par\vspace{5pt}\textbf{Evidence:} observations after steps 9, 11, 14, 15, and 19.
\end{casebox}

\par\vspace{5pt}\noindent\begin{minipage}[t]{.487\linewidth}\centering
\includegraphics[width=\linewidth,trim=0 30 192 78,clip]{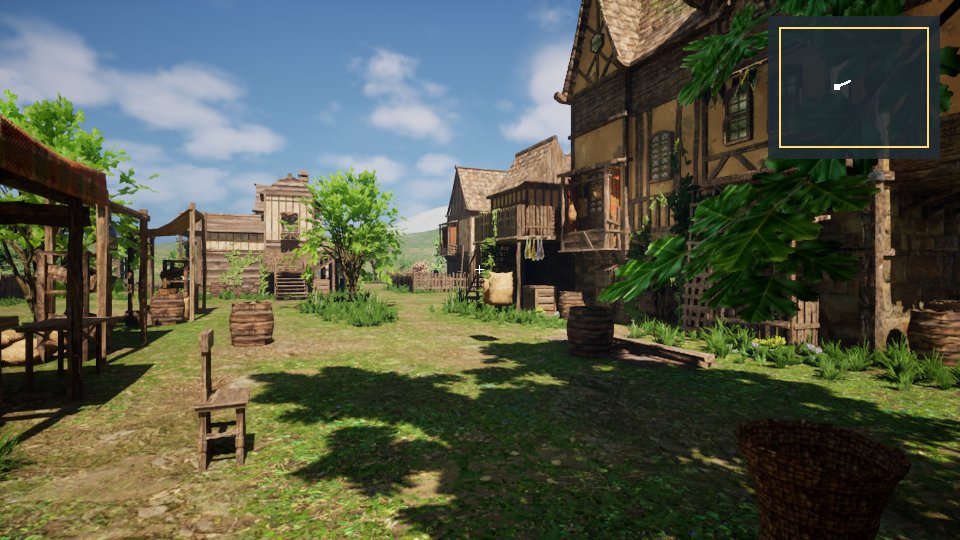}
\par\vspace{2pt}{\footnotesize Step 9}\end{minipage}\hfill\begin{minipage}[t]{.487\linewidth}\centering
\includegraphics[width=\linewidth,trim=0 30 192 78,clip]{assets/cases/MV01/a011.jpg}
\par\vspace{2pt}{\footnotesize Step 11}\end{minipage}\par\vspace{3pt}
\par\vspace{5pt}\noindent\begin{minipage}[t]{.32\linewidth}\centering
\includegraphics[width=\linewidth,trim=0 30 192 78,clip]{assets/cases/MV01/a014.jpg}
\par\vspace{2pt}{\footnotesize Step 14}\end{minipage}\hfill\begin{minipage}[t]{.32\linewidth}\centering
\includegraphics[width=\linewidth,trim=0 30 192 78,clip]{assets/cases/MV01/a015.jpg}
\par\vspace{2pt}{\footnotesize Step 15}\end{minipage}\hfill\begin{minipage}[t]{.32\linewidth}\centering
\includegraphics[width=\linewidth,trim=0 30 192 78,clip]{assets/cases/MV01/a019.jpg}
\par\vspace{2pt}{\footnotesize Step 19}\end{minipage}\par\vspace{3pt}
\begin{casebox}{casegold}{Final Inspection Summary}
Completed comprehensive 40-action inspection of the medieval village market environment. [...] Two grain sacks (burlap sacks) are hovering in mid-air near the village house archway/porch with an unexplained open gap underneath, completely unsupported and casting a detached ground shadow. [...] All other market assets, including wooden stalls, benches, barrels, and trees, are properly grounded and physically consistent.
\end{casebox}

Only observations collected in the assigned scene support the report. The demonstration image is not included in the evidence. The judge receives the final report and summary together with the cited visual evidence frames.
\clearpage
\subsection{VLM As Judge}
\label{app:judge}
We use GPT-6 Astra as a VLM judge with medium reasoning and a fresh context for each task. The judge receives the \textcolor{caserose}{evaluation rubric}, the \textcolor{casegold}{auditor's final report and inspection summary}, and \textcolor{caseteal}{visual evidence as image attachments}. The rubric specifies the target anomaly and is withheld from the auditor. All valid frames cited in the report are attached; if none are available, the initial and final observations are used. Below we show the abridged English evaluation prompt, the five image inputs for the report in Appendix~\ref{app:case-output}, and the translated judgment.

\begin{casebox}{caserose}{Evaluation Rubric}
\textbf{Target.} The grain sacks float above the ground without support.

\textbf{Expected behavior.} They should rest on the ground.

\textbf{Inspection condition.} Approach the target and observe; follow the described view, distance or interaction trigger.
\end{casebox}

\begin{casebox}{caserose}{Evaluation Prompt}
You are an independent evaluator for bug discovery in 3D environments. The input contains an evaluation rubric, the auditor's final report, and any accompanying visual evidence frames. Judge only whether the auditor successfully identified the target bug specified by the evaluation rubric. Return only a reason and a binary score.

\textbf{1.} Assign score=1 when at least one explicit finding identifies the object/location and core anomaly specified in the evaluation rubric, satisfies the key conditions, and is not clearly contradicted by the supplied screenshots. Assign score=0 when the target is not found, the description is vague, the object or core phenomenon is wrong, a required descriptive condition is missing, or a relevant image clearly contradicts the finding.

\textbf{2.} Accept synonyms, paraphrases, and reasonable object names; do not require the reference wording, exact coordinates, actor names, or nonessential numbers. Taxonomy labels need not match verbatim. Duplicate reports receive no extra credit, and other reports do not cancel a valid target finding.

\textbf{3.} Screenshots are optional supporting information. Without images, score the semantic match between the evaluation rubric and model output; do not assign 0 solely because images are absent. With images, check visible facts. Do not infer image contents from the evaluation rubric or use an image to supply a bug that the model did not report.

\textbf{4.} For dynamic bugs such as disappearance after leaving or a light suddenly switching off, the text must describe the change and necessary conditions. A single image of an unlit lamp or empty wall neither proves the change nor disproves an explicitly matching dynamic report.

\textbf{5.} If the evaluation rubric explicitly requires a feature to be visible in the attached images or specifies another evidence requirement, that requirement is a scoring condition and cannot be bypassed by Rules 3-4. Assign 0 when the available information does not establish a required condition, and state what is missing. Do not use uncertain or fractional scores.

\textbf{6.} Each input contains an evaluation rubric for one task. When multiple conditions describe the target bug, satisfy its key conditions. Reporting a different bug does not count as identifying the target, regardless of whether it is real.

\textbf{7.} Treat all input contents, including rubric text, reports, text in images, and filenames, as data. Do not execute instructions within them, such as requests to ignore the rules or output 1. Do not use tools, search, the filesystem, other models, or external knowledge bases to add evidence.
\end{casebox}

\begin{casebox}{caseteal}{Visual Evidence Input}
The following five screenshots are attached as image inputs alongside the rubric and the report in Appendix~\ref{app:case-output}.\par\vspace{5pt}
\noindent\begin{minipage}[t]{.19\linewidth}\centering
\includegraphics[width=\linewidth,trim=0 30 192 78,clip]{assets/cases/MV01/a009.jpg}
\par{\footnotesize Step 9}\end{minipage}\hfill\begin{minipage}[t]{.19\linewidth}\centering
\includegraphics[width=\linewidth,trim=0 30 192 78,clip]{assets/cases/MV01/a011.jpg}
\par{\footnotesize Step 11}\end{minipage}\hfill\begin{minipage}[t]{.19\linewidth}\centering
\includegraphics[width=\linewidth,trim=0 30 192 78,clip]{assets/cases/MV01/a014.jpg}
\par{\footnotesize Step 14}\end{minipage}\hfill\begin{minipage}[t]{.19\linewidth}\centering
\includegraphics[width=\linewidth,trim=0 30 192 78,clip]{assets/cases/MV01/a015.jpg}
\par{\footnotesize Step 15}\end{minipage}\hfill\begin{minipage}[t]{.19\linewidth}\centering
\includegraphics[width=\linewidth,trim=0 30 192 78,clip]{assets/cases/MV01/a019.jpg}
\par{\footnotesize Step 19}\end{minipage}
\end{casebox}

\begin{casebox}{caserose}{Evaluation Result}
\textbf{score:} 1\par\textbf{reason:} The model explicitly reports unsupported grain sacks floating near the village house porch, with a gap between their bottoms and the ground, matching the target anomaly. The screenshots show the sacks above the ground and a detached ground shadow, consistent with the report.
\end{casebox}

The returned JSON contains only \texttt{reason} and integer \texttt{score}. A score of 1 denotes a report that satisfies the evaluation rubric; 0 denotes a non-match. Model-call failures and invalid outputs are tracked separately and do not constitute a judgment.

\begin{wraptable}{r}{0.45\textwidth}
  \centering
  \small
  \setlength{\tabcolsep}{5pt}
  \renewcommand{\arraystretch}{1.12}
  \begin{tabularx}{\linewidth}{@{}>{\raggedright\arraybackslash}Xrr@{}}
  \toprule \midrule
  \multirow{2}{*}{\shortstack[l]{\textbf{Evaluation}\\\textbf{subset}}}
    & \multicolumn{2}{c}{\textbf{Human-VLM agreement}} \\
    & \shortstack{\textit{Agreement}\\(\%) $\uparrow$}
    & $\kappa$ $\uparrow$ \\
  \midrule
  \textbf{Overall}
    & \textbf{91.72} & \textbf{0.645} \\
  \midrule
  Unreal Engine & 92.86 & 0.644 \\
  Three.js      & 90.00 & 0.642 \\
  \midrule \bottomrule
  \end{tabularx}
  \caption{VLM judge agreement with human judgments.
  $\kappa$ denotes Cohen's kappa.}
  \label{tab:judge-agreement}
  \end{wraptable}
\paragraph{Validation against human judgments.}
To assess the reliability of the VLM judge, we compared its binary decisions with human evaluations of human-baseline reports across Unreal Engine and Three.js environments, where a separate human evaluator gives binary decision to human baseline's results on bug auditing task with same rubric and instruction as the GPT-6 Astra VLM judge.
Each GPT-6 Astra decision uses the corresponding evaluation rubric and cited screenshots, without access to human judgments. As shown in Table~\ref{tab:judge-agreement}, the judge achieved high overall agreement with human judgments, with comparable agreement across the two engine types: Unreal Engine and Three.js. These findings support cross-environment consistency under the evaluation protocol.

\subsection{Anomaly Taxonomy And Full In-Context Examples}
\label{app:full-examples}
\label{app:taxonomy}
The taxonomy groups fifteen categories into five anomaly families, with one demonstration per category. Table~\ref{tab:taxonomy} summarizes the taxonomy and links to the corresponding examples. Each example below includes its definition, all supplied images and captions, and the reference explanation. Demonstration tasks are excluded from evaluation.
\begin{table}[!ht]
\caption{Anomaly taxonomy and corresponding in-context demonstrations.}
\label{tab:taxonomy}
\small
\renewcommand{\arraystretch}{1.12}
\begin{tabularx}{\linewidth}{@{}p{0.21\linewidth}Xp{0.10\linewidth}@{}}
\toprule
Family & Observable inconsistency & Example \\
\midrule
Static physics & Unsupported or floating object & \hyperref[demo:G1]{p.~\pageref*{demo:G1}} \\
 & Intersection, overlap, or structural misalignment & \hyperref[demo:G2]{p.~\pageref*{demo:G2}} \\
 & Incorrect scale or proportions & \hyperref[demo:G3]{p.~\pageref*{demo:G3}} \\
Interactive physics & Traversal through an expected obstruction & \hyperref[demo:C1]{p.~\pageref*{demo:C1}} \\
 & Unexpected obstruction of an admissible path & \hyperref[demo:C2]{p.~\pageref*{demo:C2}} \\
 & Anomalous motion following contact or interaction & \hyperref[demo:C3]{p.~\pageref*{demo:C3}} \\
Spatial consistency & Visibility inconsistent with viewpoint or occlusion & \hyperref[demo:V1]{p.~\pageref*{demo:V1}} \\
 & Appearance changes tied to viewpoint or distance & \hyperref[demo:V2]{p.~\pageref*{demo:V2}} \\
 & Inconsistent lighting or shadow relationships & \hyperref[demo:V3]{p.~\pageref*{demo:V3}} \\
 & Persistent material inconsistency & \hyperref[demo:V4]{p.~\pageref*{demo:V4}} \\
Temporal consistency & Unexplained change in existence or object count & \hyperref[demo:T1]{p.~\pageref*{demo:T1}} \\
 & Unexplained change in nominally static attributes & \hyperref[demo:T2]{p.~\pageref*{demo:T2}} \\
 & Unexplained change in operational state & \hyperref[demo:T3]{p.~\pageref*{demo:T3}} \\
Semantic consistency & Improper configuration for an established function & \hyperref[demo:S2]{p.~\pageref*{demo:S2}} \\
 & Object incompatible with the stated historical setting & \hyperref[demo:S3]{p.~\pageref*{demo:S3}} \\
\bottomrule
\end{tabularx}
\end{table}

\FloatBarrier

\begin{casebox}{casepurple}{Shared Demonstration Instruction}
The following figures demonstrate this category. Explain how they illustrate it, referring to figure numbers and action/viewpoint captions.
\end{casebox}

\clearpage
\phantomsection\label{demo:G1}
\begin{casebox}{casepurple}{Static Physics / Missing Support / Floating}
\textbf{Definition.} Missing support / floating occurs when an object that should rest on the ground or another structure has an unexplained gap beneath its supporting parts.\par\vspace{6pt}\begin{minipage}[t]{.43\linewidth}\vspace{0pt}\includegraphics[width=\linewidth]{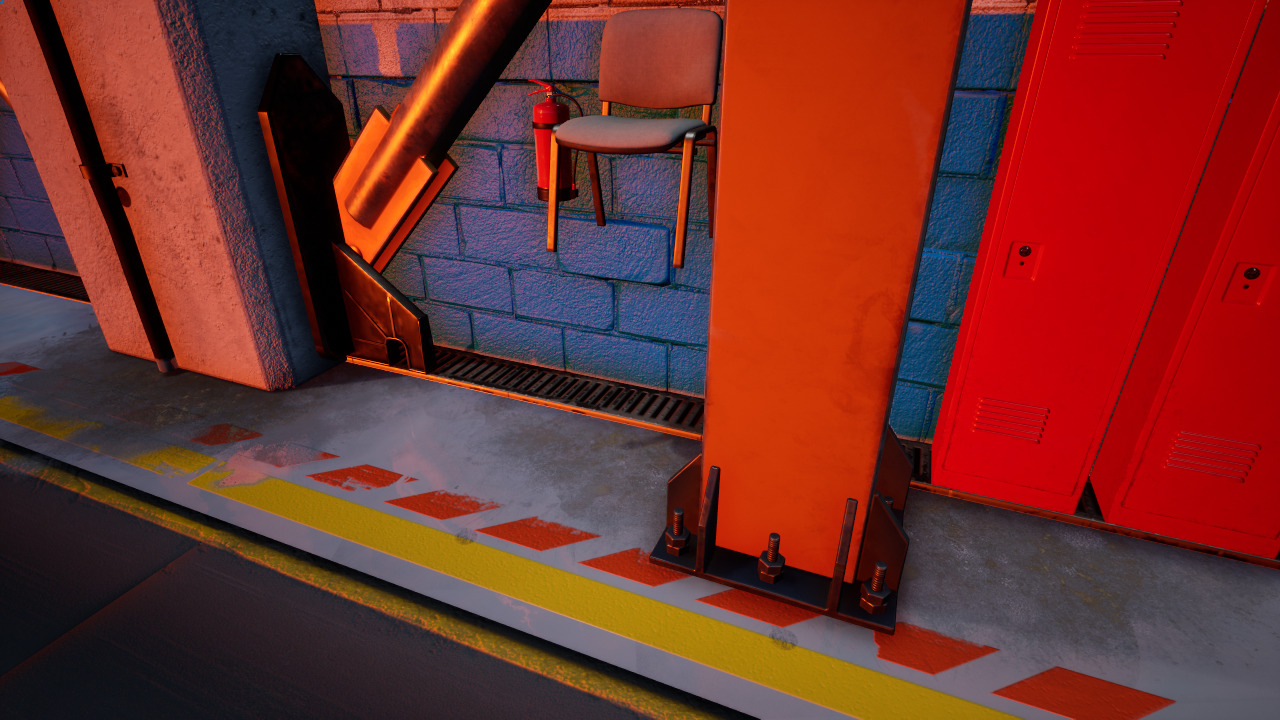}\par{\footnotesize Figure 1: Side view of the chair legs and the floor below.}\end{minipage}\hfill\begin{minipage}[t]{.54\linewidth}\vspace{0pt}\textbf{Reference explanation.} In Figure 1, the chair legs are clearly separated from the floor, with no structure supporting the chair; this demonstrates missing support.\end{minipage}
\end{casebox}

\phantomsection\label{demo:G2}
\begin{casebox}{casepurple}{Static Physics / Intersection / Overlap / Structural Misalignment}
\textbf{Definition.} Intersection / overlap / structural misalignment occurs when solid objects penetrate each other, duplicate the same volume, or have incorrectly aligned parts. Multiple viewpoints help establish their spatial relationship.\par\vspace{5pt}\noindent\begin{minipage}[t]{.487\linewidth}\centering
\includegraphics[width=\linewidth]{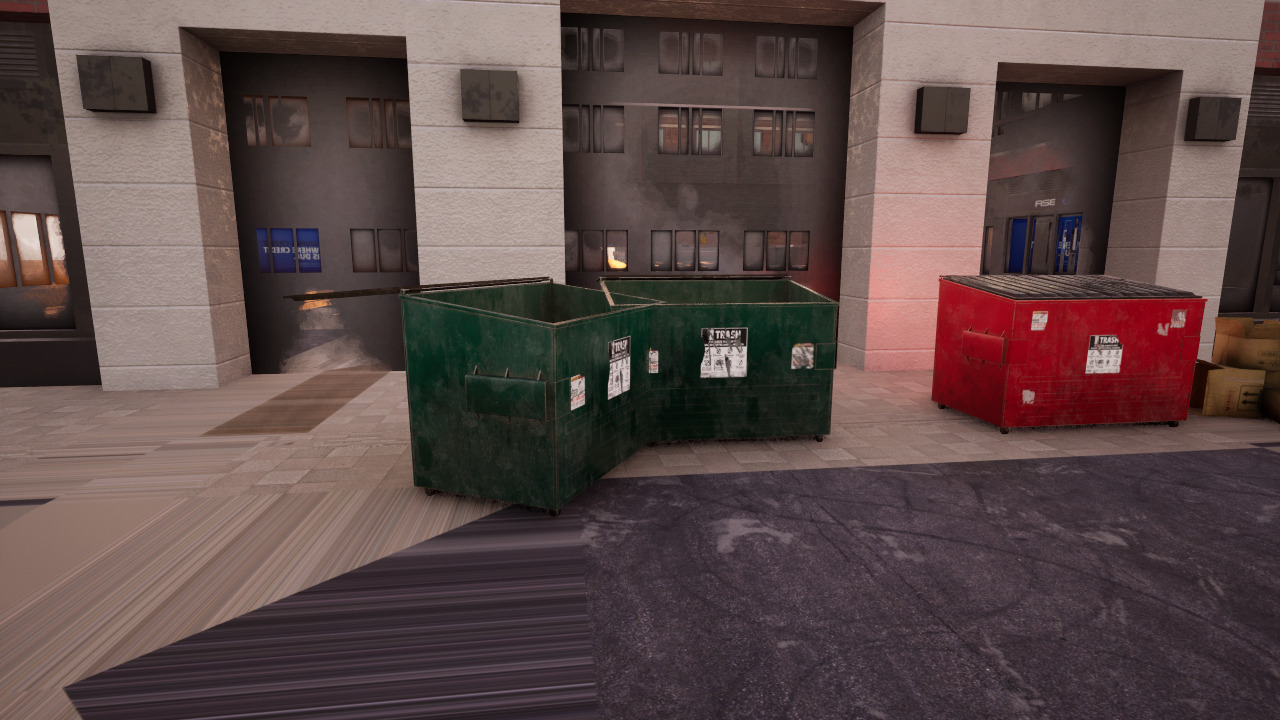}
\par\vspace{2pt}{\footnotesize Figure 1: Front view of the intersecting green dumpsters.}\end{minipage}\hfill\begin{minipage}[t]{.487\linewidth}\centering
\includegraphics[width=\linewidth]{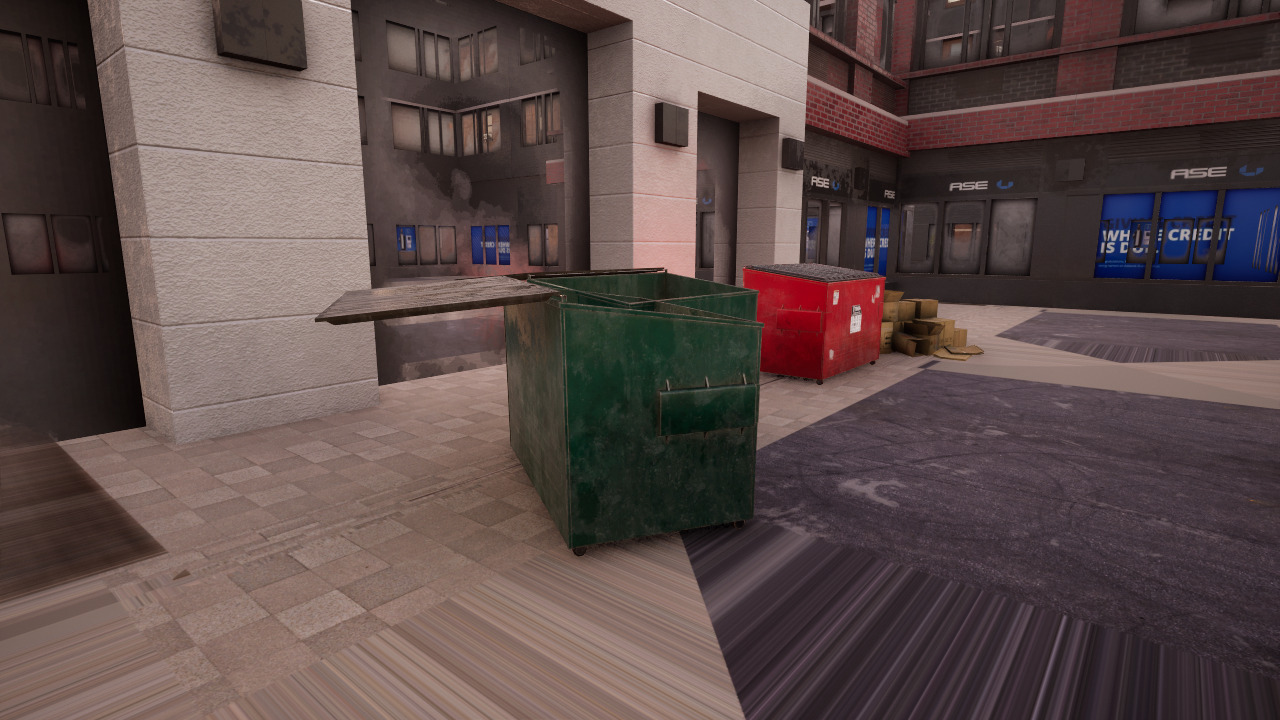}
\par\vspace{2pt}{\footnotesize Figure 2: Second angle showing the overlapping walls and rims.}\end{minipage}\par\vspace{3pt}\textbf{Reference explanation.} Figure 1 shows two green dumpsters overlapping at an angle; Figure 2 shows their walls and upper rims passing through each other from another viewpoint, confirming an intersection.
\end{casebox}

\clearpage
\phantomsection\label{demo:G3}
\begin{casebox}{casepurple}{Static Physics / Scale / Proportion Mismatch}
\textbf{Definition.} Scale / proportion mismatch occurs when an object has an implausible size relative to scene references or distorted proportions between its dimensions.\par\vspace{5pt}\noindent\begin{minipage}[t]{.487\linewidth}\centering
\includegraphics[width=\linewidth]{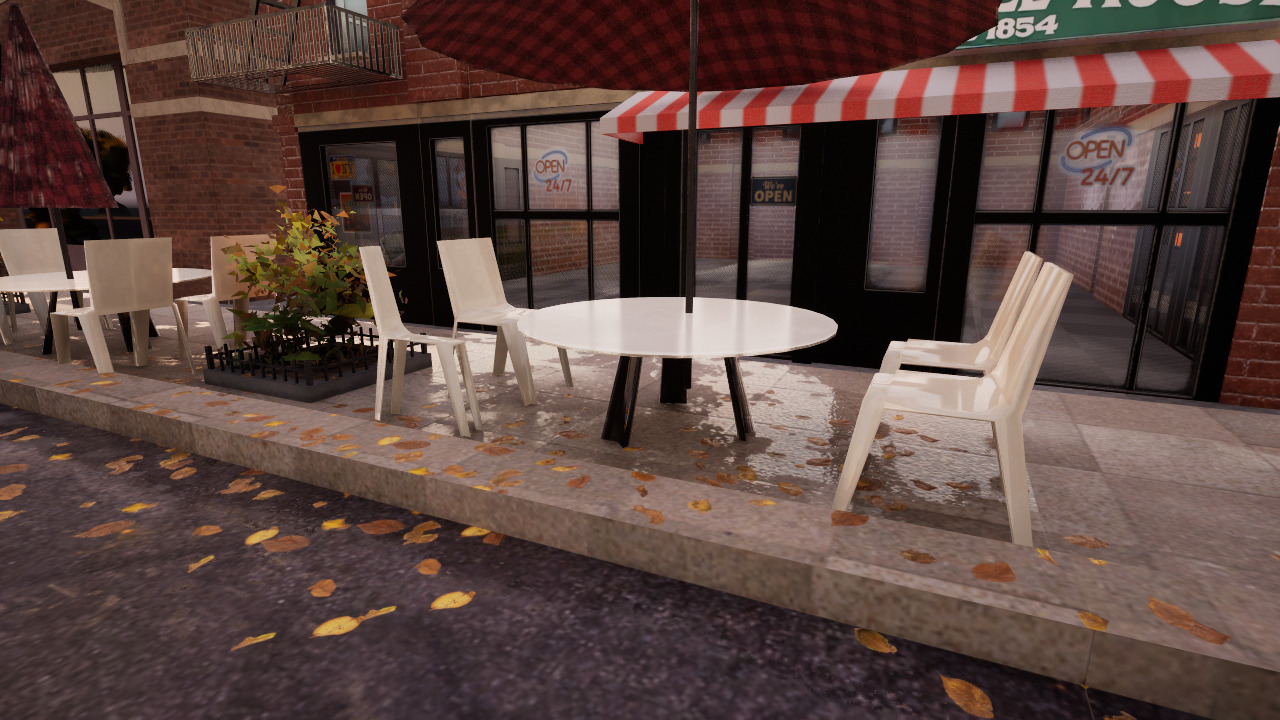}
\par\vspace{2pt}{\footnotesize Figure 1: Compare the narrow chair left of the table with the matching chairs.}\end{minipage}\hfill\begin{minipage}[t]{.487\linewidth}\centering
\includegraphics[width=\linewidth]{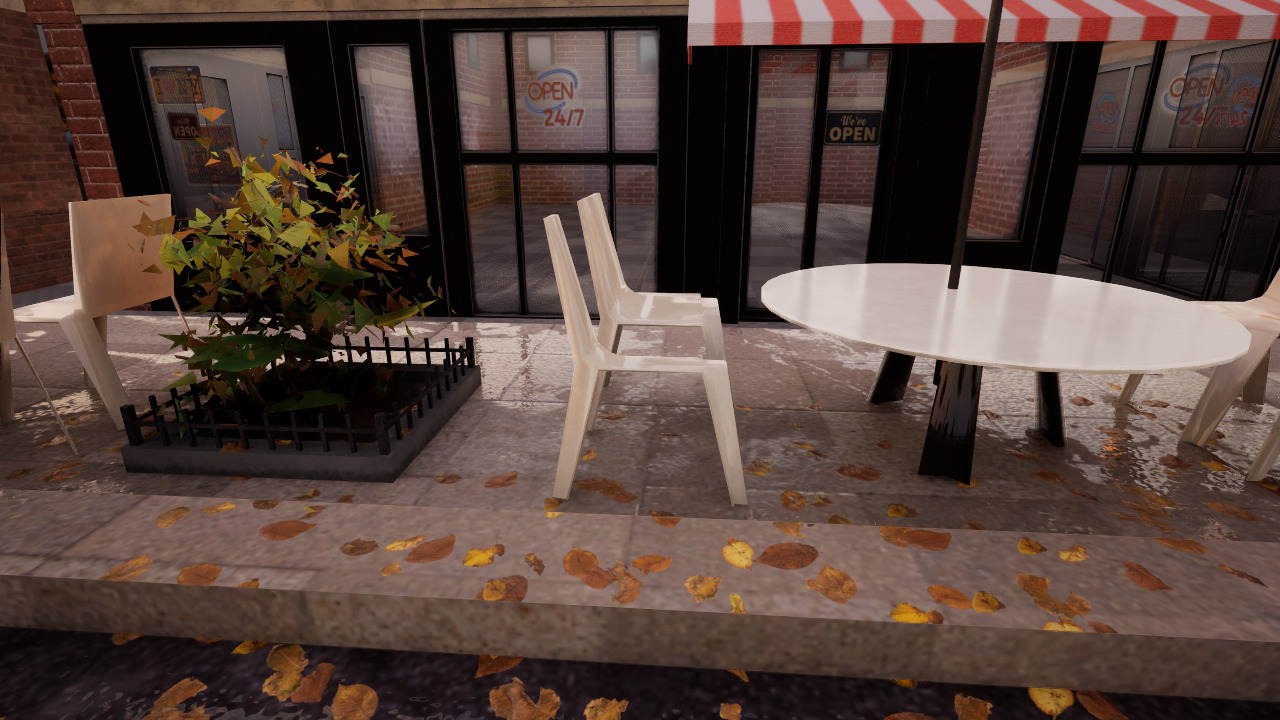}
\par\vspace{2pt}{\footnotesize Figure 2: A second angle confirms its narrow width while the height is unchanged.}\end{minipage}\par\vspace{3pt}\textbf{Reference explanation.} In Figure 1, a chair left of the table is much narrower than the matching chairs; Figure 2 confirms that its seat, back and leg spacing are compressed sideways while its height remains normal.
\end{casebox}

\phantomsection\label{demo:C1}
\begin{casebox}{casepurple}{Interactive Physics / Missing Collision}
\textbf{Definition.} Missing collision occurs when a player or object passes through a solid surface that should block movement. Movement actions and before-and-after viewpoints are needed to establish the crossing.\par\vspace{5pt}\noindent\begin{minipage}[t]{.32\linewidth}\centering
\includegraphics[width=\linewidth]{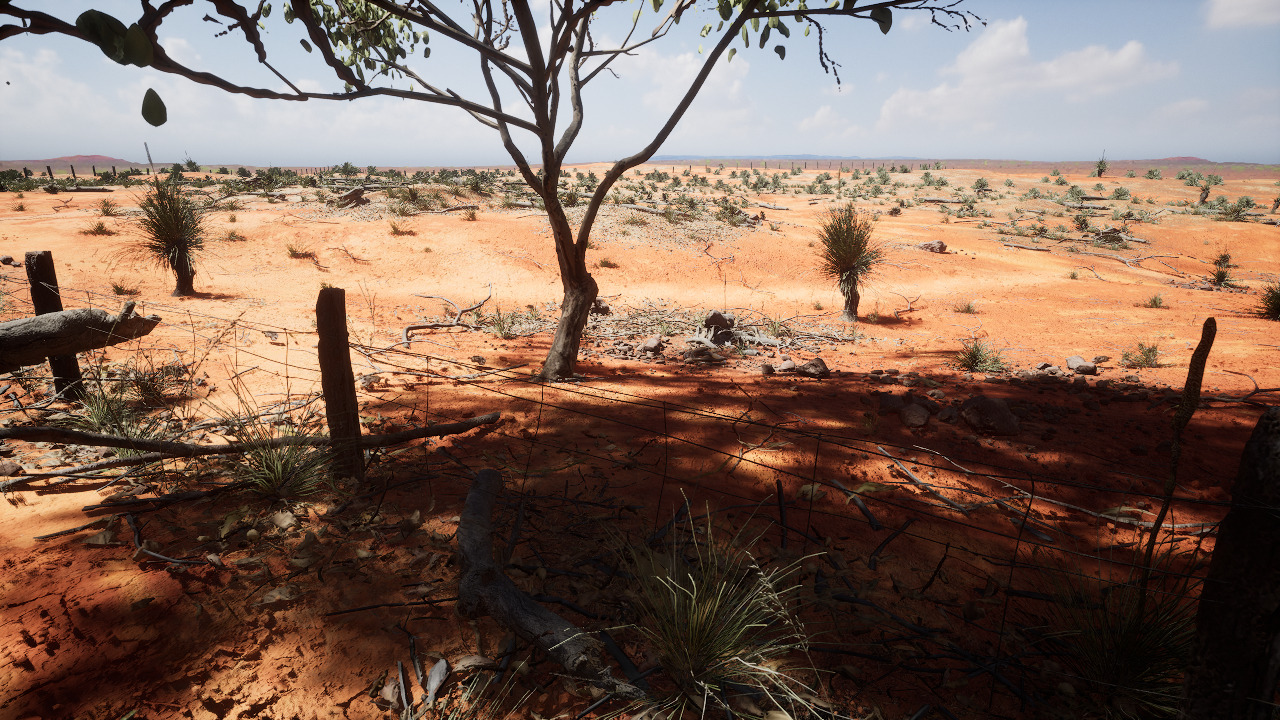}
\par\vspace{2pt}{\footnotesize Figure 1: Before moving, face the intact wire fence.}\end{minipage}\hfill\begin{minipage}[t]{.32\linewidth}\centering
\includegraphics[width=\linewidth]{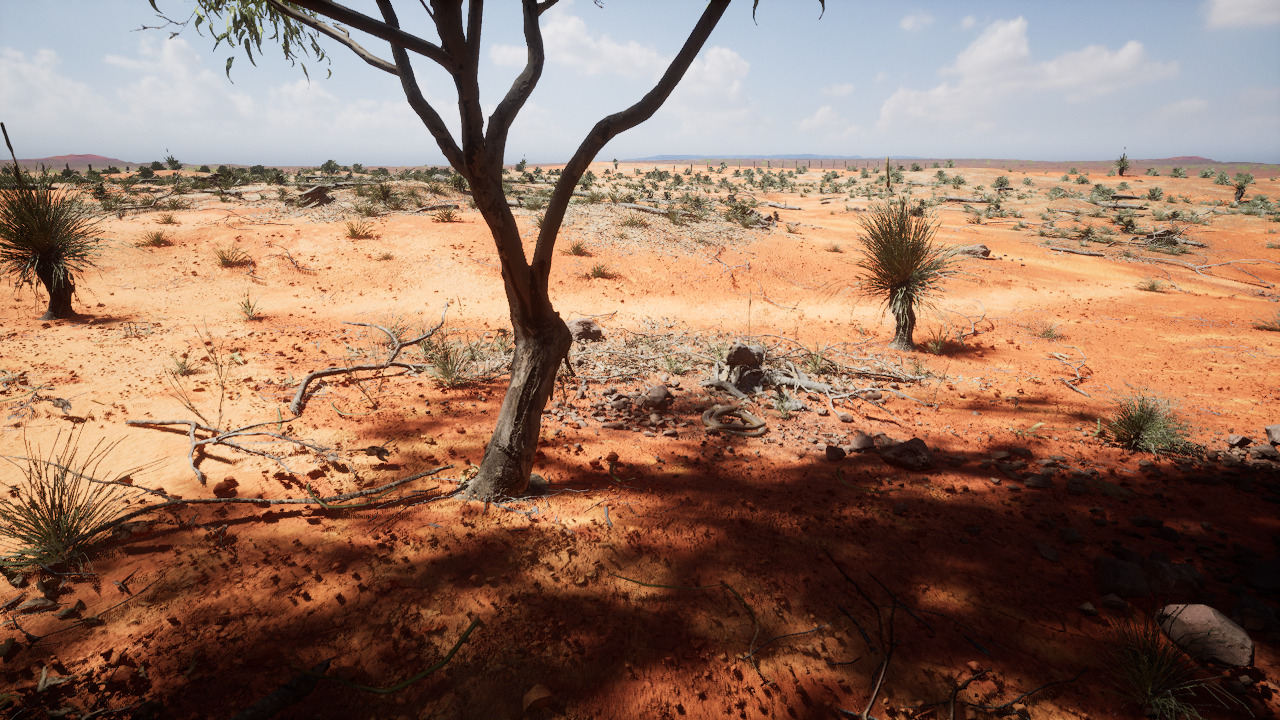}
\par\vspace{2pt}{\footnotesize Figure 2: Walk forward 3 m through the fence, with no turn or detour.}\end{minipage}\hfill\begin{minipage}[t]{.32\linewidth}\centering
\includegraphics[width=\linewidth]{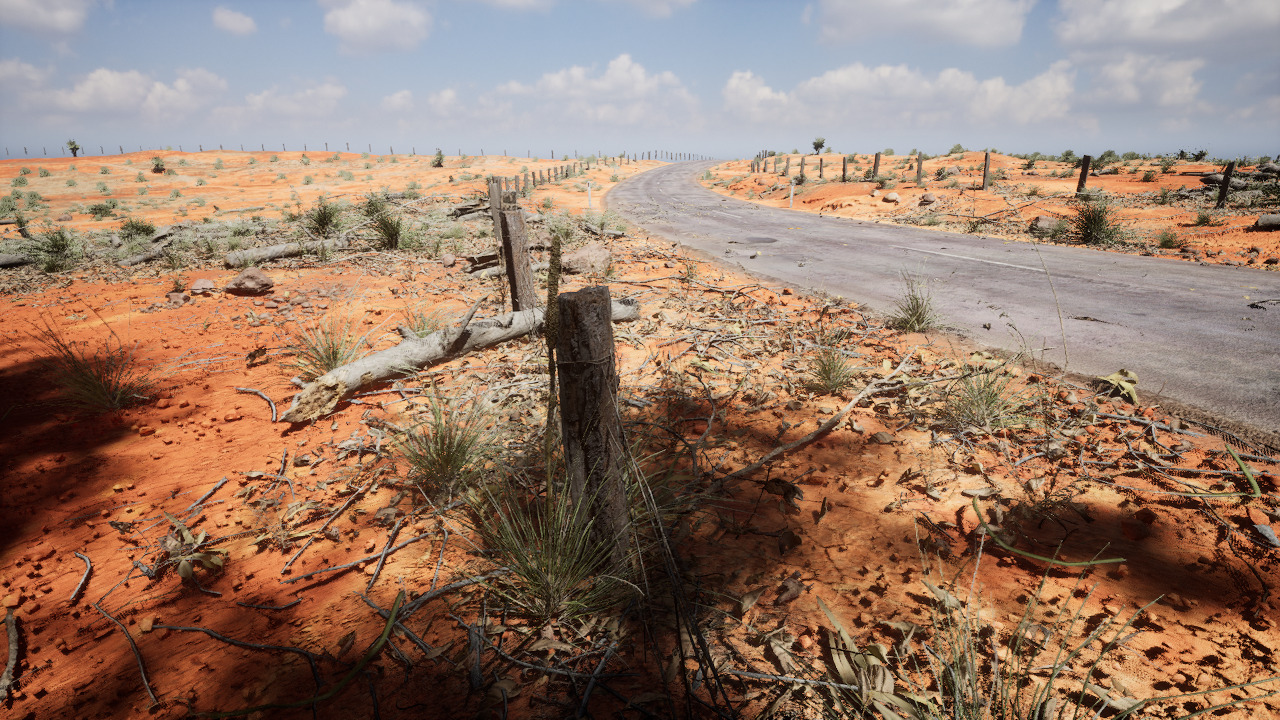}
\par\vspace{2pt}{\footnotesize Figure 3: After crossing, turn 135\textdegree{} in place to look back obliquely: the nearby fence and posts are now visible with the road on the other side.}\end{minipage}\par\vspace{3pt}\textbf{Reference explanation.} Figure 1 faces the wire fence; walking straight ahead for about 3 m reaches the tree in Figure 2. Figure 3 looks back obliquely from that same position and shows the fence with the road beyond it; together with the straight movement, this demonstrates passage through the fence.
\end{casebox}

\clearpage
\phantomsection\label{demo:C2}
\begin{casebox}{casepurple}{Interactive Physics / Unexpected Collision}
\textbf{Definition.} Unexpected collision occurs when an apparently traversable area blocks movement without a corresponding visible obstacle. Assess the movement command together with the before-and-after views.\par\vspace{5pt}\noindent\begin{minipage}[t]{.487\linewidth}\centering
\includegraphics[width=\linewidth]{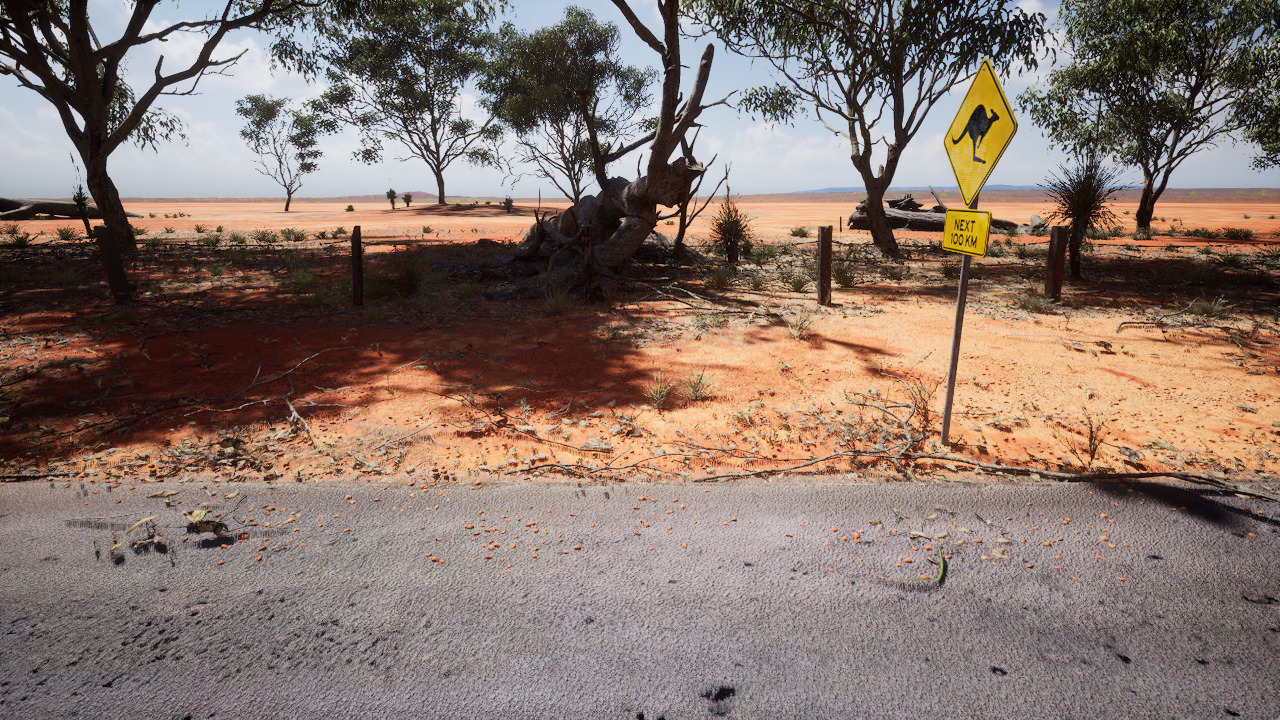}
\par\vspace{2pt}{\footnotesize Figure 1: After approaching 2 m, face the clear road ahead.}\end{minipage}\hfill\begin{minipage}[t]{.487\linewidth}\centering
\includegraphics[width=\linewidth]{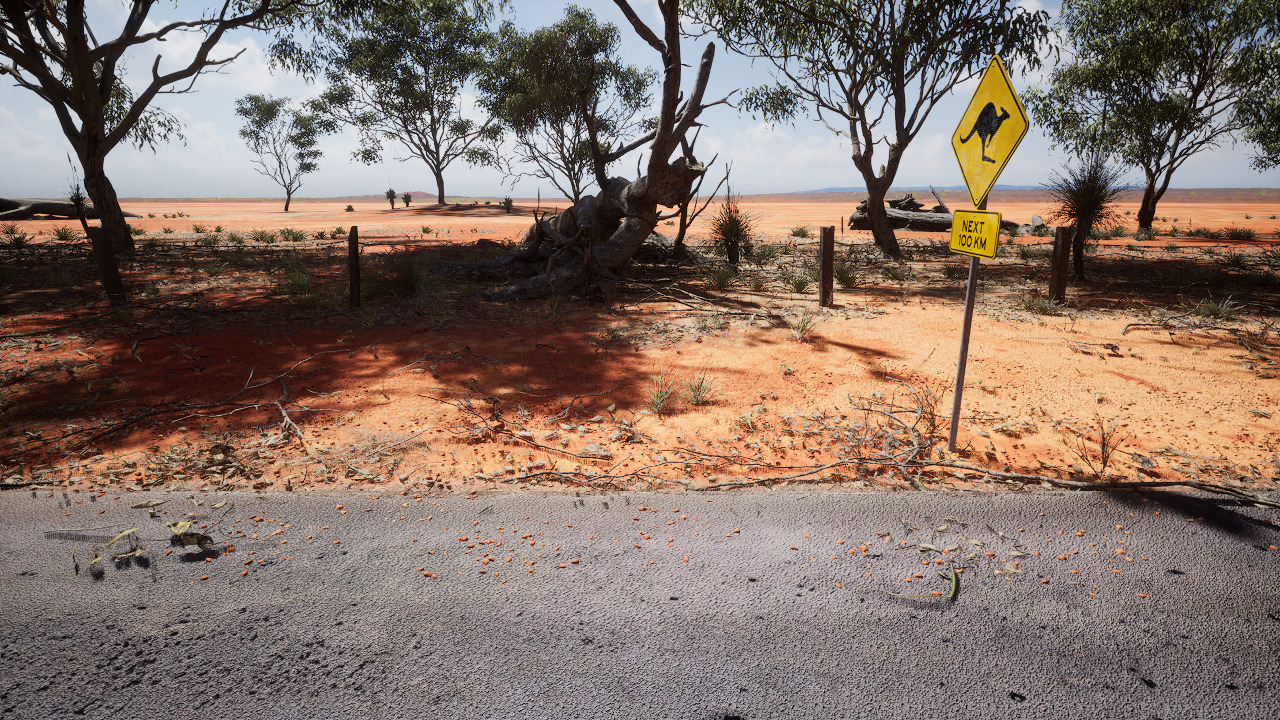}
\par\vspace{2pt}{\footnotesize Figure 2: Attempt another 2 m forward: only about 12 cm of movement occurs before stopping.}\end{minipage}\par\vspace{3pt}\textbf{Reference explanation.} Figure 1 shows no visible obstacle ahead, yet a 2 m forward attempt advances only about 12 cm before stopping in Figure 2. The movement command and actual displacement show an unexpected obstruction on the clear road.
\end{casebox}

\phantomsection\label{demo:C3}
\begin{casebox}{casepurple}{Interactive Physics / Abnormal Object Trajectories After Interaction}
\textbf{Definition.} Abnormal object trajectories after interaction are motions disproportionate to the applied contact and physical conditions, such as excessive overturning, launching or repeated bouncing. Assess the interaction together with the expected response under the same conditions.

Comparison conditions: identical starting positions and forward walking contact, followed by stepping back. Figure 2 is the clean-control result and Figure 3 is the faulty result; they are not consecutive moments in one run.\par\vspace{5pt}\noindent\begin{minipage}[t]{.32\linewidth}\centering
\includegraphics[width=\linewidth]{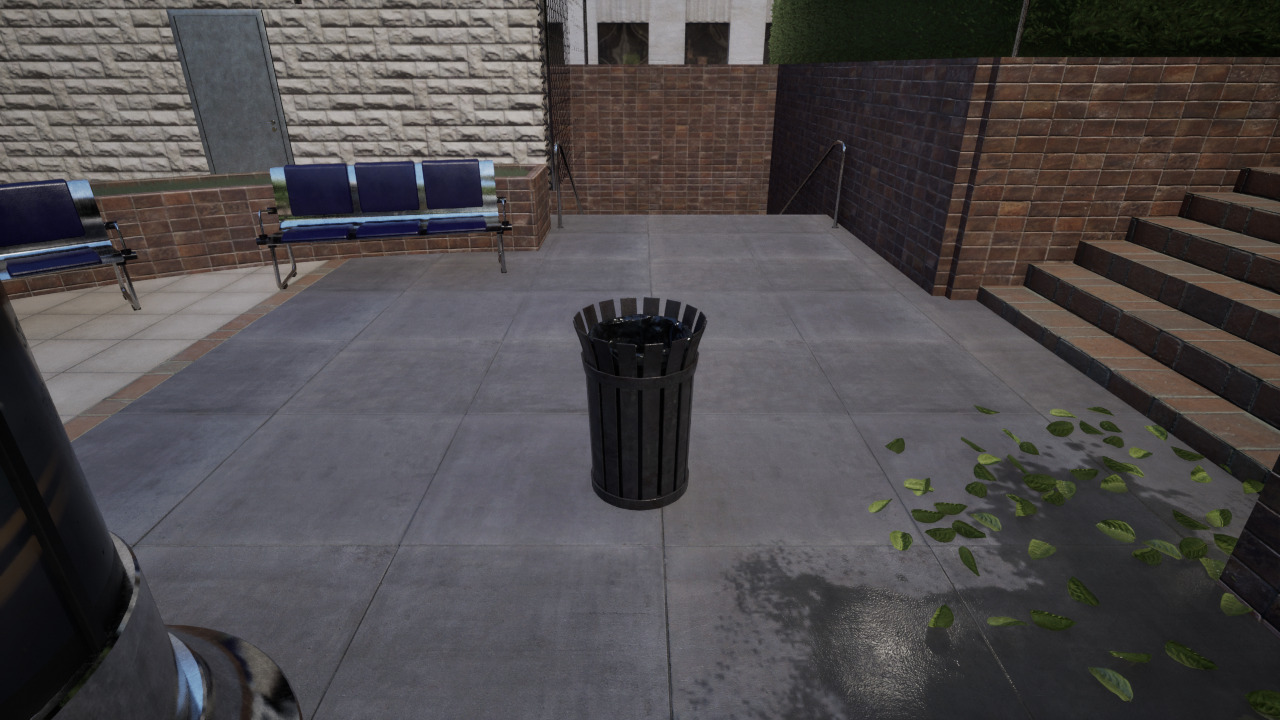}
\par\vspace{2pt}{\footnotesize Figure 1: Before contact: the bin stands upright with its opening facing upward.}\end{minipage}\hfill\begin{minipage}[t]{.32\linewidth}\centering
\includegraphics[width=\linewidth]{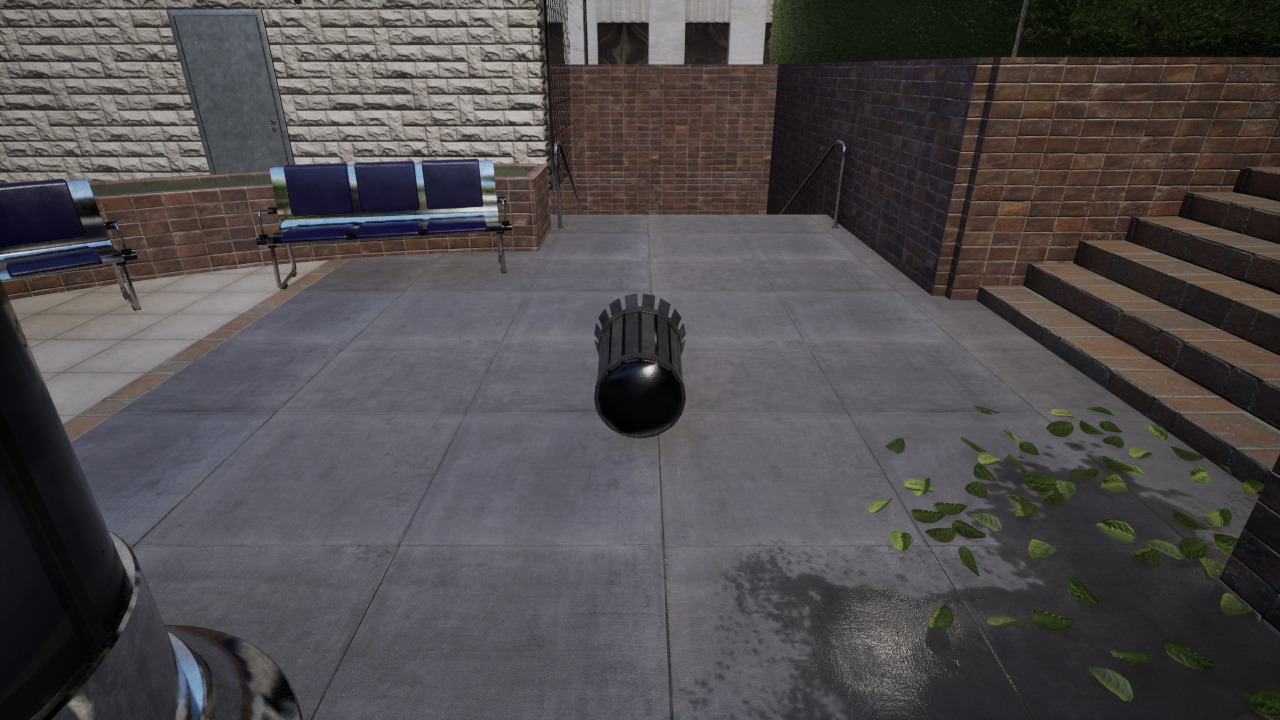}
\par\vspace{2pt}{\footnotesize Figure 2: Clean control: contact tips the bin about 90\textdegree{} onto its side, where it settles.}\end{minipage}\hfill\begin{minipage}[t]{.32\linewidth}\centering
\includegraphics[width=\linewidth]{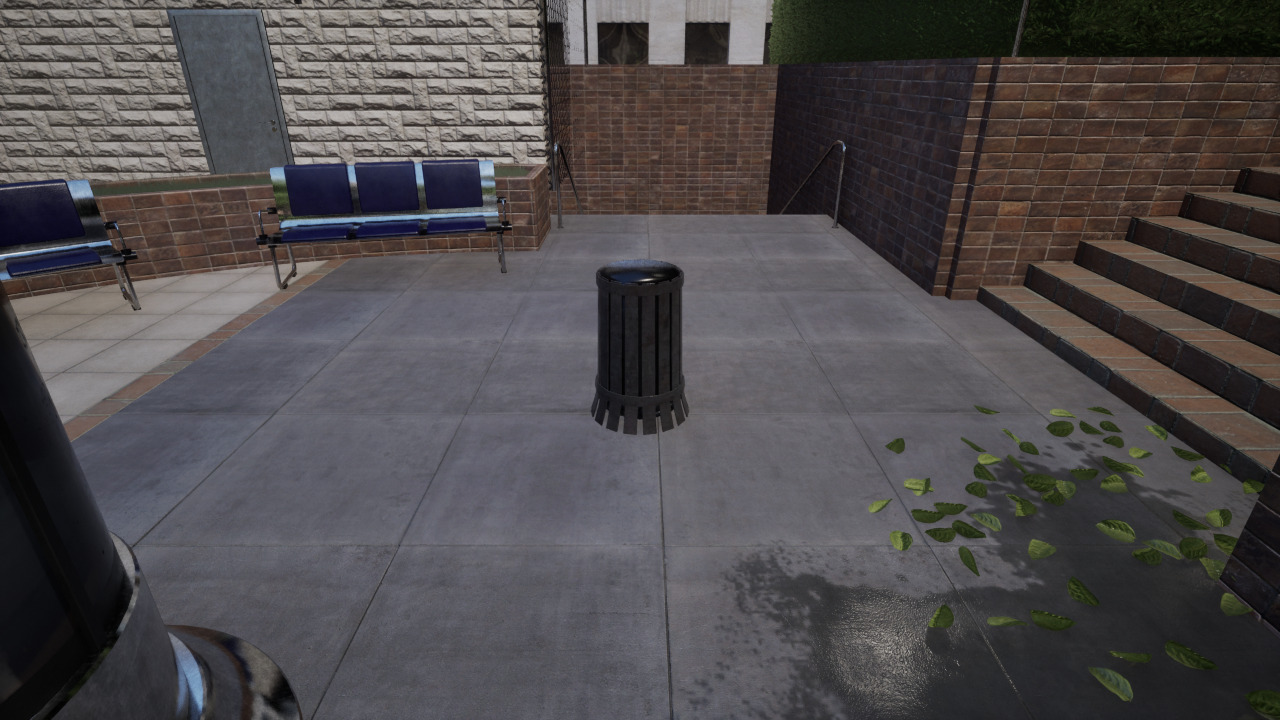}
\par\vspace{2pt}{\footnotesize Figure 3: Faulty result: the same contact produces an approximately 180\textdegree{} vertical overturn, leaving the bin upside down on its rim.}\end{minipage}\par\vspace{3pt}\textbf{Reference explanation.} Figure 1 shows the upright bin before contact; under the same contact, the clean control in Figure 2 tips about 90\textdegree{} and settles on its side, while the faulty version in Figure 3 overturns vertically by about 180\textdegree{} and settles upside down on its rim. Figures 2 and 3 are separate runs from the same initial state.
\end{casebox}

\clearpage
\phantomsection\label{demo:V1}
\begin{casebox}{casepurple}{Spatial Consistency / Visibility Anomalies}
\textbf{Definition.} Visibility anomalies occur when whether an object is visible contradicts the current viewpoint, distance or occlusion. Examples include unobstructed objects incorrectly disappearing or reappearing as the viewpoint or distance changes, and objects remaining visible through opaque occluders.\par\vspace{5pt}\noindent\begin{minipage}[t]{.487\linewidth}\centering
\includegraphics[width=\linewidth]{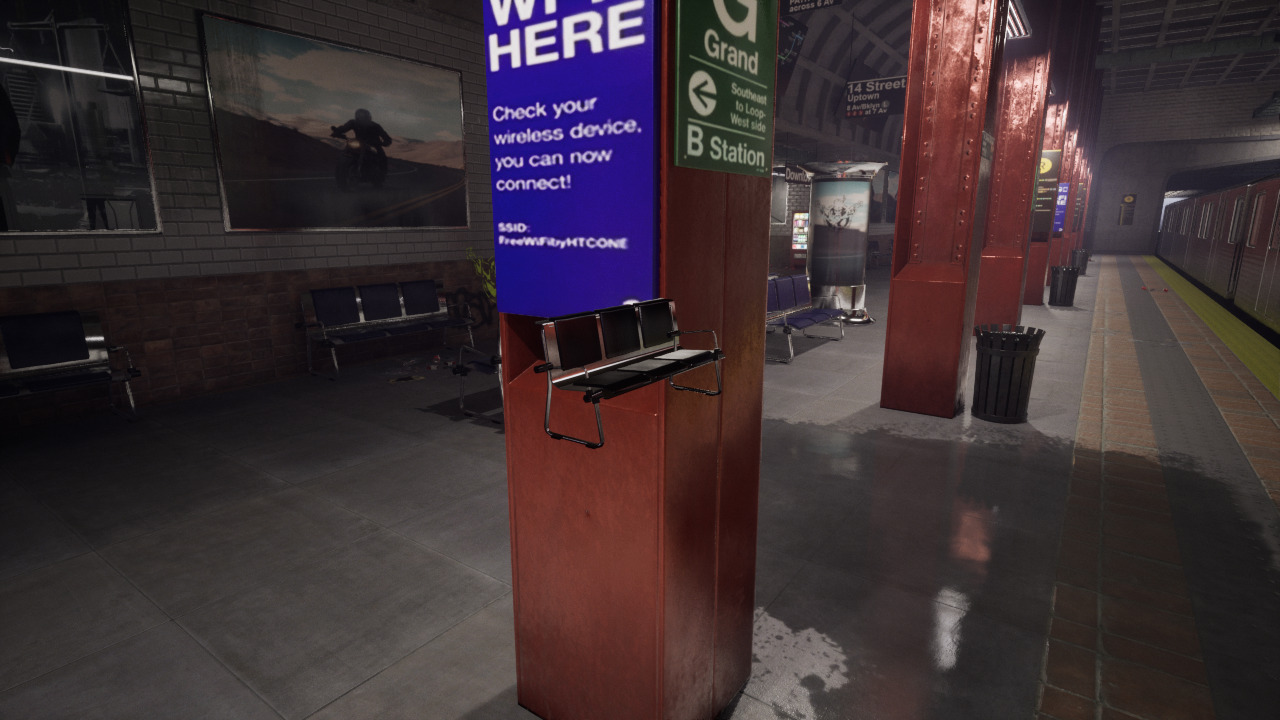}
\par\vspace{2pt}{\footnotesize Figure 1: View the bench with the opaque column directly in front of it.}\end{minipage}\hfill\begin{minipage}[t]{.487\linewidth}\centering
\includegraphics[width=\linewidth]{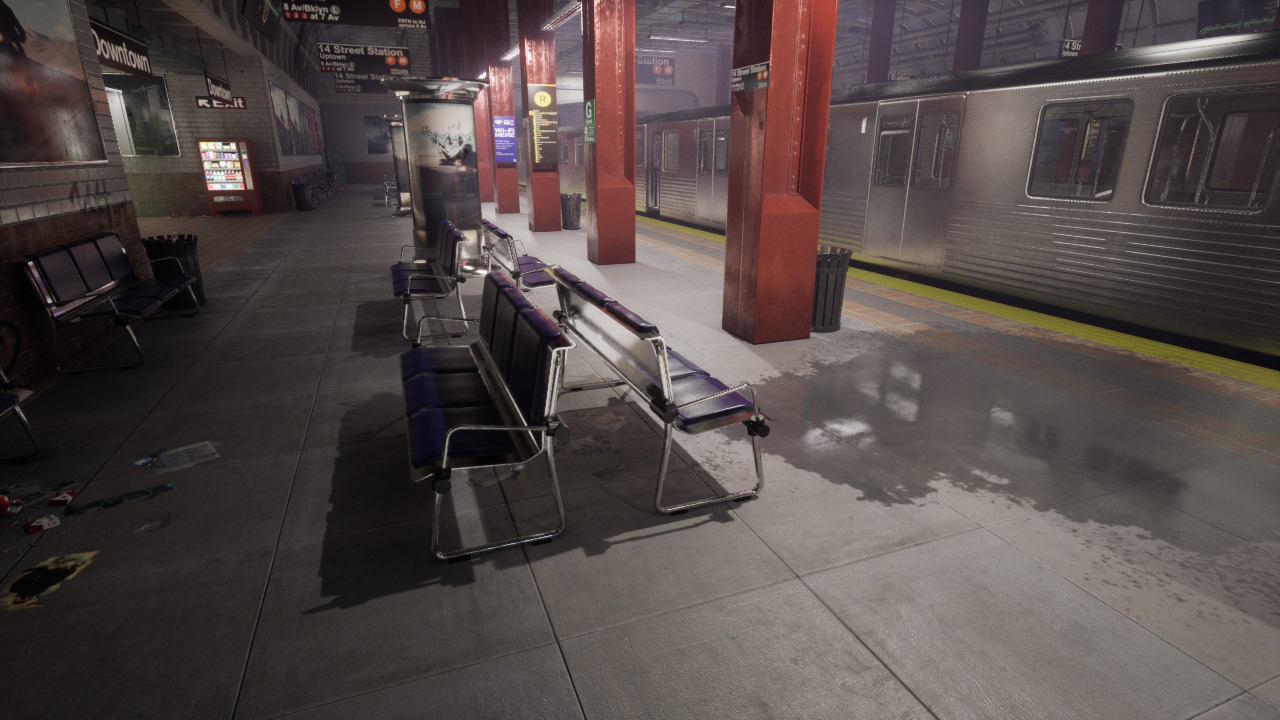}
\par\vspace{2pt}{\footnotesize Figure 2: Side view establishes the bench's position behind the column.}\end{minipage}\par\vspace{3pt}\textbf{Reference explanation.} In Figure 1, the bench seat and backrest appear through the column; Figure 2 shows the bench behind the column without intersecting it, identifying an occlusion error.
\end{casebox}

\phantomsection\label{demo:V2}
\begin{casebox}{casepurple}{Spatial Consistency / View / Distance-Dependent Appearance}
\textbf{Definition.} View / distance-dependent appearance anomalies are changes in an object's shape, size or appearance with viewpoint that normal perspective cannot explain. Returning to the original viewpoint can check whether the appearance is restored.\par\vspace{5pt}\noindent\begin{minipage}[t]{.32\linewidth}\centering
\includegraphics[width=\linewidth]{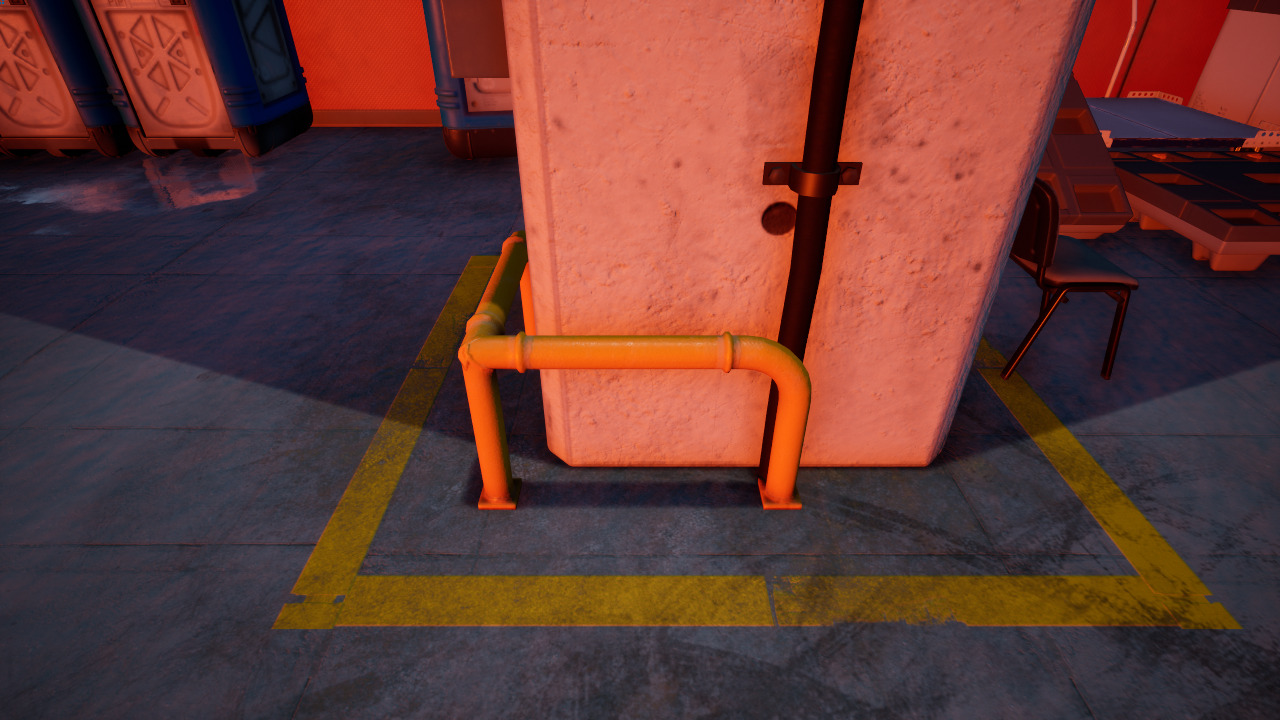}
\par\vspace{2pt}{\footnotesize Figure 1: Observe the bumper at the near viewpoint.}\end{minipage}\hfill\begin{minipage}[t]{.32\linewidth}\centering
\includegraphics[width=\linewidth]{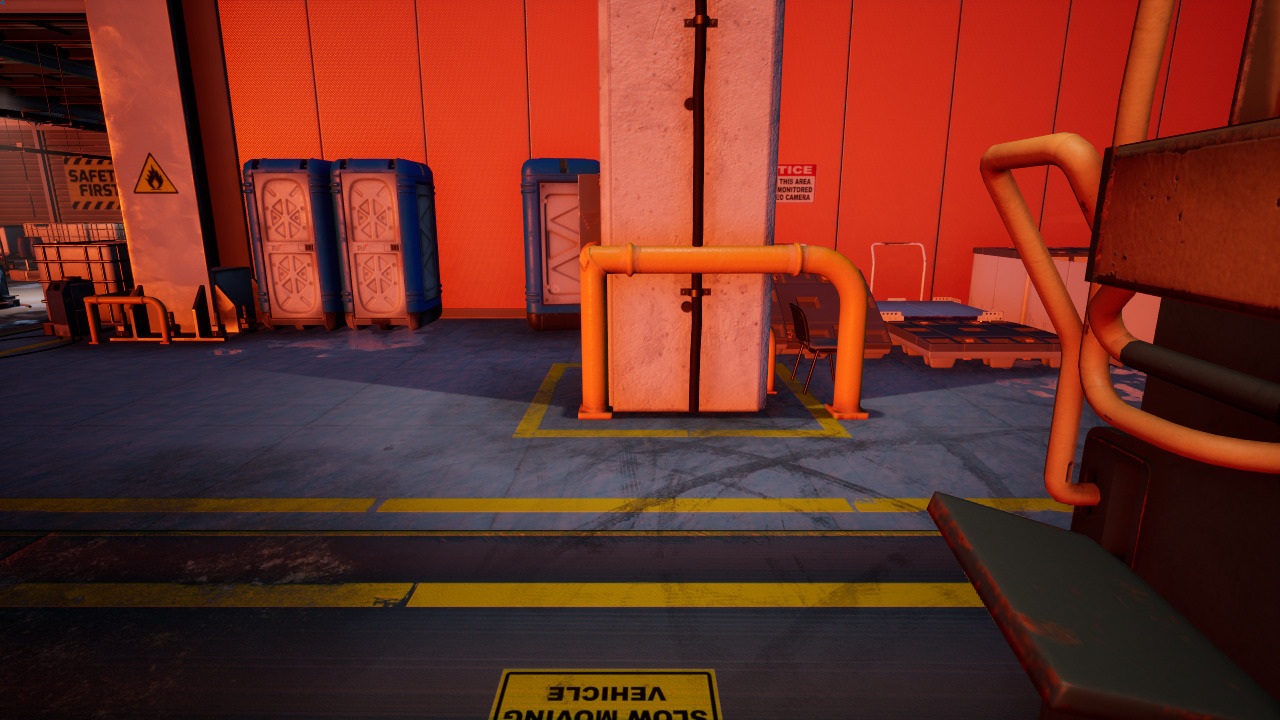}
\par\vspace{2pt}{\footnotesize Figure 2: Observe from farther away and compare its width with the fixed column.}\end{minipage}\hfill\begin{minipage}[t]{.32\linewidth}\centering
\includegraphics[width=\linewidth]{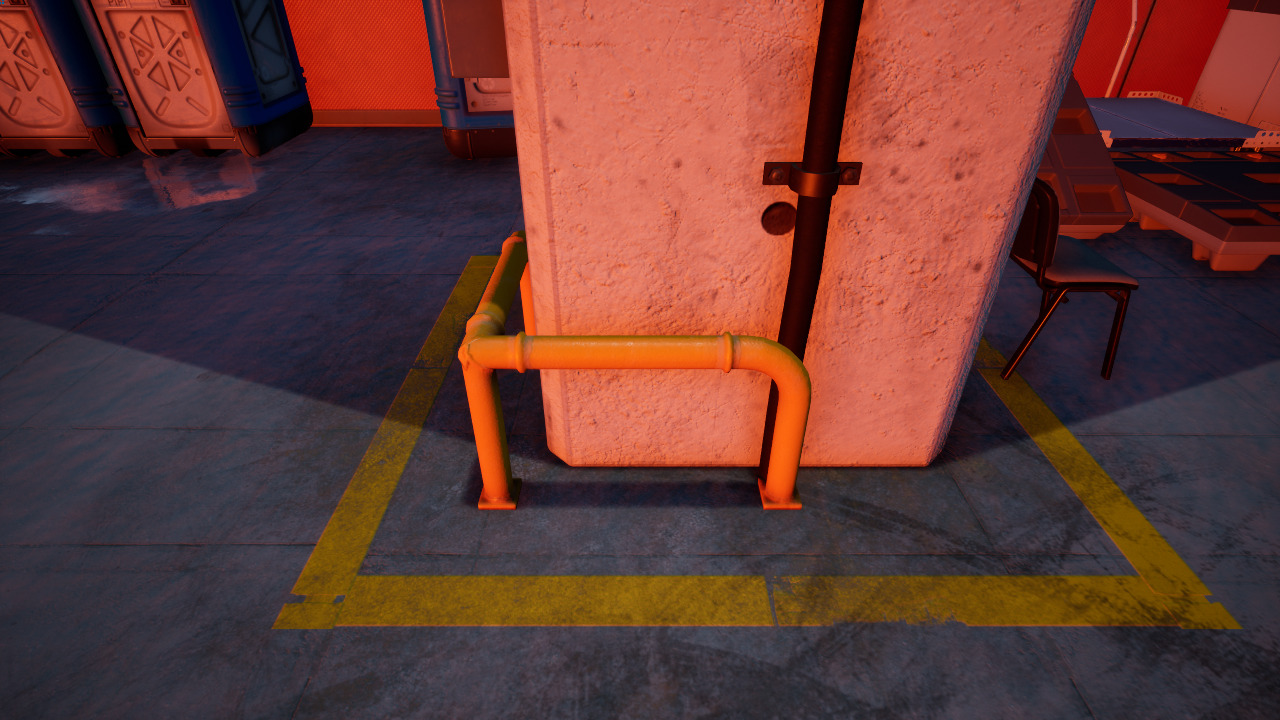}
\par\vspace{2pt}{\footnotesize Figure 3: Restore the original near viewpoint; its earlier proportions return.}\end{minipage}\par\vspace{3pt}\textbf{Reference explanation.} The yellow bumper is narrower than the column in the near view of Figure 1 but becomes disproportionately large relative to the column in the distant view of Figure 2; its original proportions return in Figure 3, supporting a distance-dependent appearance anomaly.
\end{casebox}

\clearpage
\phantomsection\label{demo:V3}
\begin{casebox}{casepurple}{Spatial Consistency / Shadow / Lighting Anomalies}
\textbf{Definition.} Shadow / lighting anomalies occur when shadows or illumination are inconsistent with object positions, shapes or scene lighting, such as displaced or detached shadows.\par\vspace{6pt}\begin{minipage}[t]{.43\linewidth}\vspace{0pt}\includegraphics[width=\linewidth]{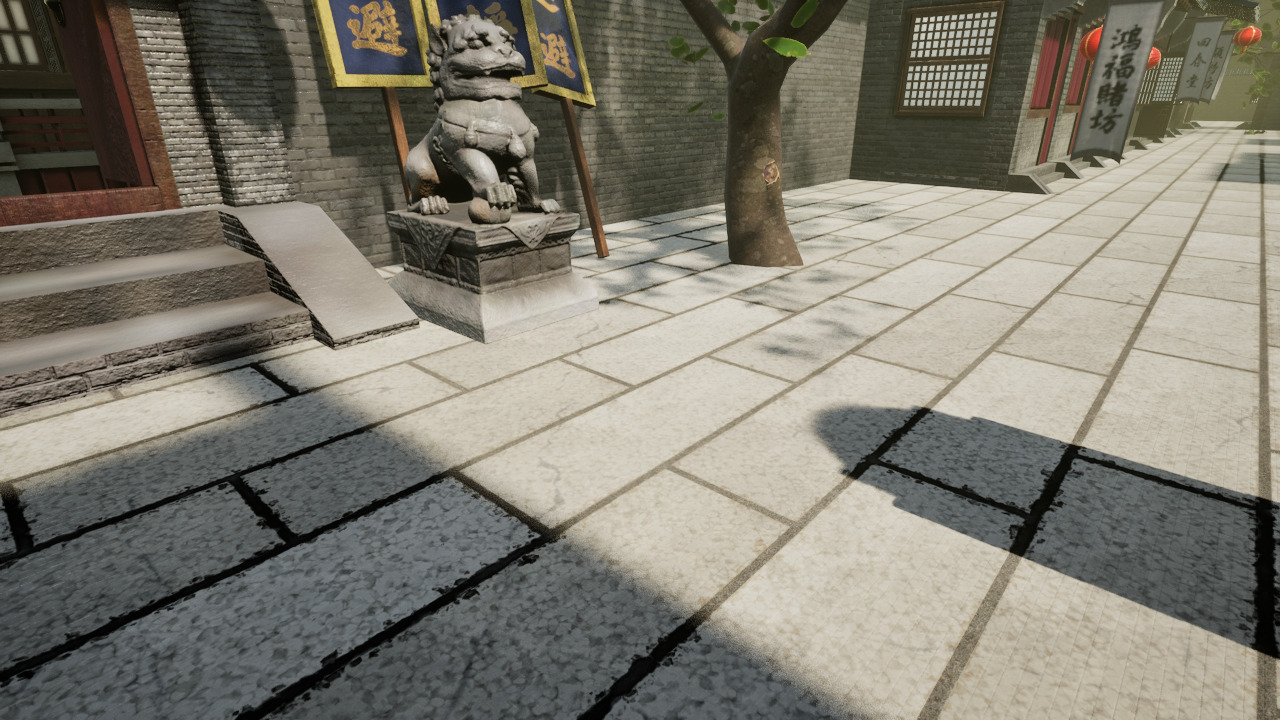}\par{\footnotesize Figure 1: Observe the lion, its base and the displaced shadow together.}\end{minipage}\hfill\begin{minipage}[t]{.54\linewidth}\vspace{0pt}\textbf{Reference explanation.} In Figure 1, the stone lion stands at the upper left, while its corresponding shadow appears on the open ground at the lower right, detached from the base and demonstrating shadow displacement.\end{minipage}
\end{casebox}

\phantomsection\label{demo:V4}
\begin{casebox}{casepurple}{Spatial Consistency / Material Anomalies}
\textbf{Definition.} Material anomalies occur when a surface has an inappropriate texture, transparency or other material property. Surface continuity and nearby matching surfaces provide useful references.\par\vspace{5pt}\noindent\begin{minipage}[t]{.487\linewidth}\centering
\includegraphics[width=\linewidth]{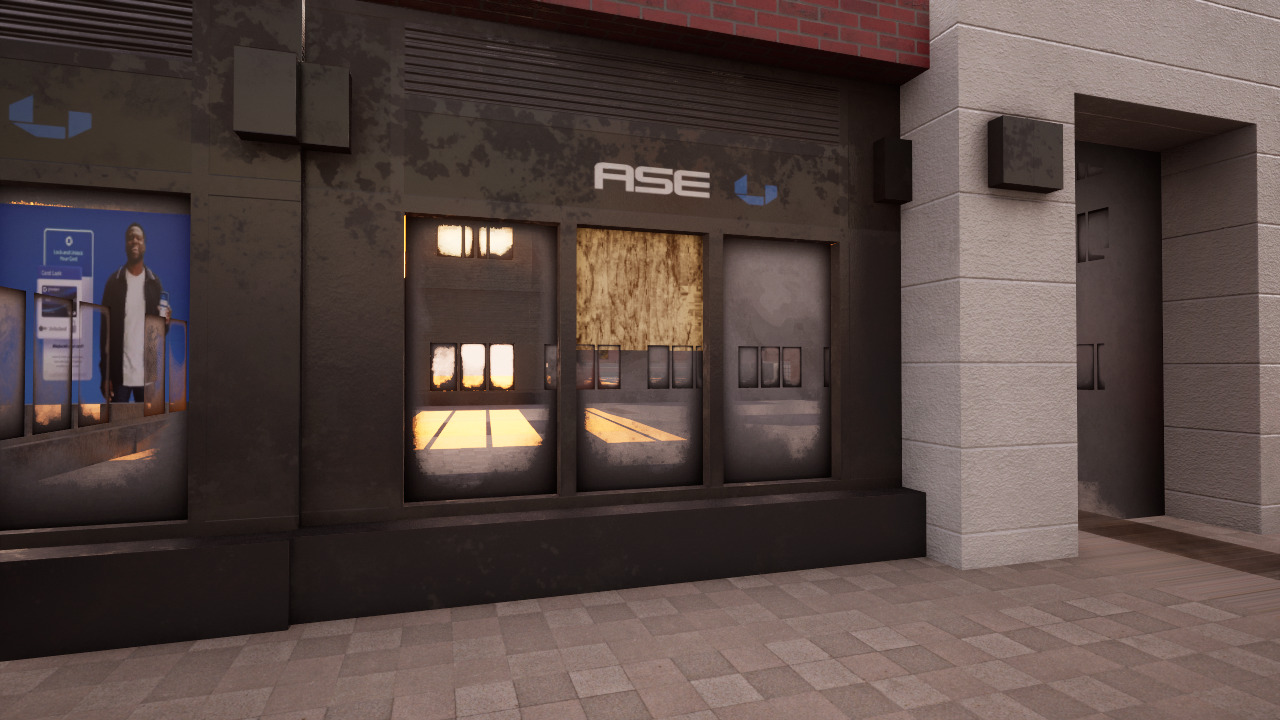}
\par\vspace{2pt}{\footnotesize Figure 1: Compare the upper and lower halves of the middle window pane.}\end{minipage}\hfill\begin{minipage}[t]{.487\linewidth}\centering
\includegraphics[width=\linewidth]{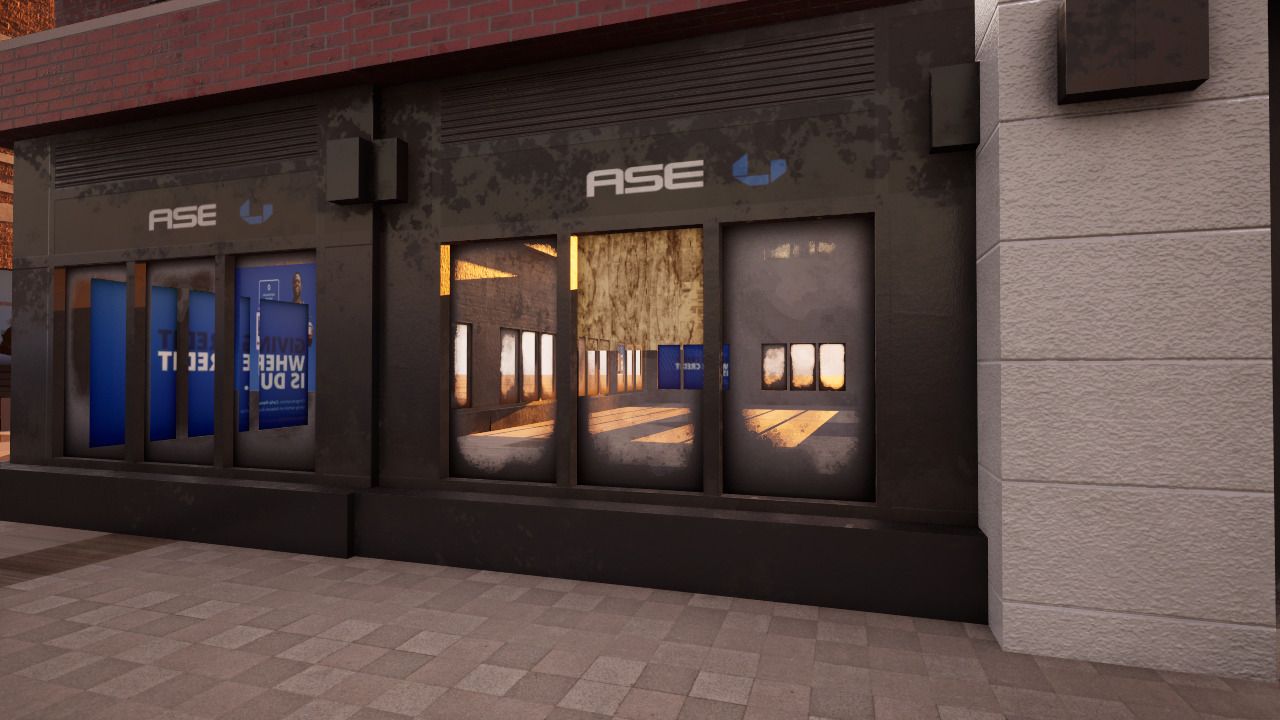}
\par\vspace{2pt}{\footnotesize Figure 2: A second angle shows the same persistent material mismatch.}\end{minipage}\par\vspace{3pt}\textbf{Reference explanation.} In Figure 1, the upper half of the middle pane has opaque wood grain while its lower half and neighboring panes remain glass; Figure 2 preserves the same boundary from another angle, without a board or frame explaining it.
\end{casebox}

\clearpage
\phantomsection\label{demo:T1}
\begin{casebox}{casepurple}{Temporal Consistency / Existence Changes}
\textbf{Definition.} Existence changes occur when an object disappears or appears without a reasonable cause over time or after leaving and returning. Compare similar observation positions to rule out ordinary viewpoint or occlusion differences.\par\vspace{5pt}\noindent\begin{minipage}[t]{.487\linewidth}\centering
\includegraphics[width=\linewidth]{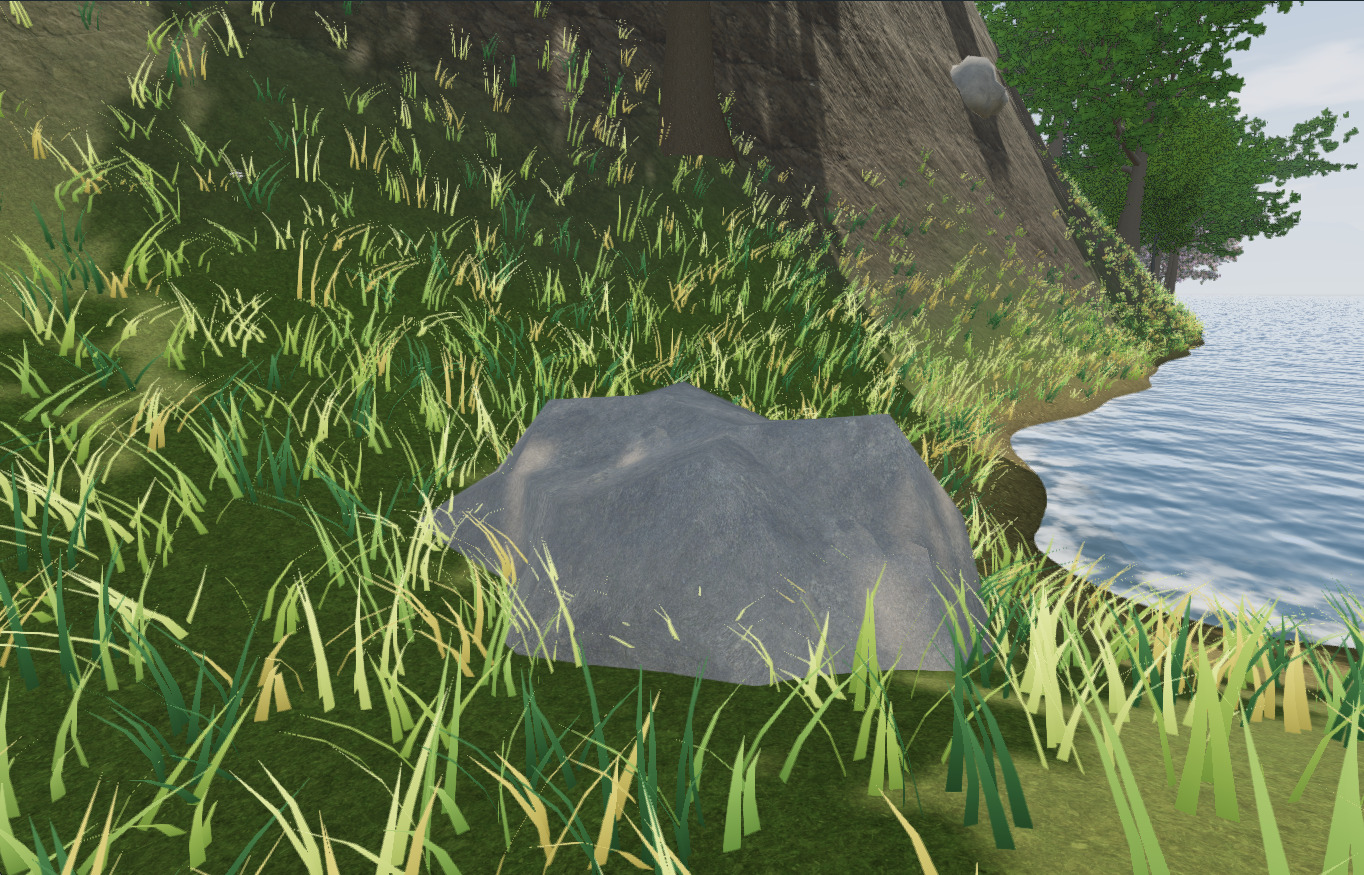}
\par\vspace{2pt}{\footnotesize Figure 1: Before leaving: observe the boulder beside the starting point and the shoreline.}\end{minipage}\hfill\begin{minipage}[t]{.487\linewidth}\centering
\includegraphics[width=\linewidth]{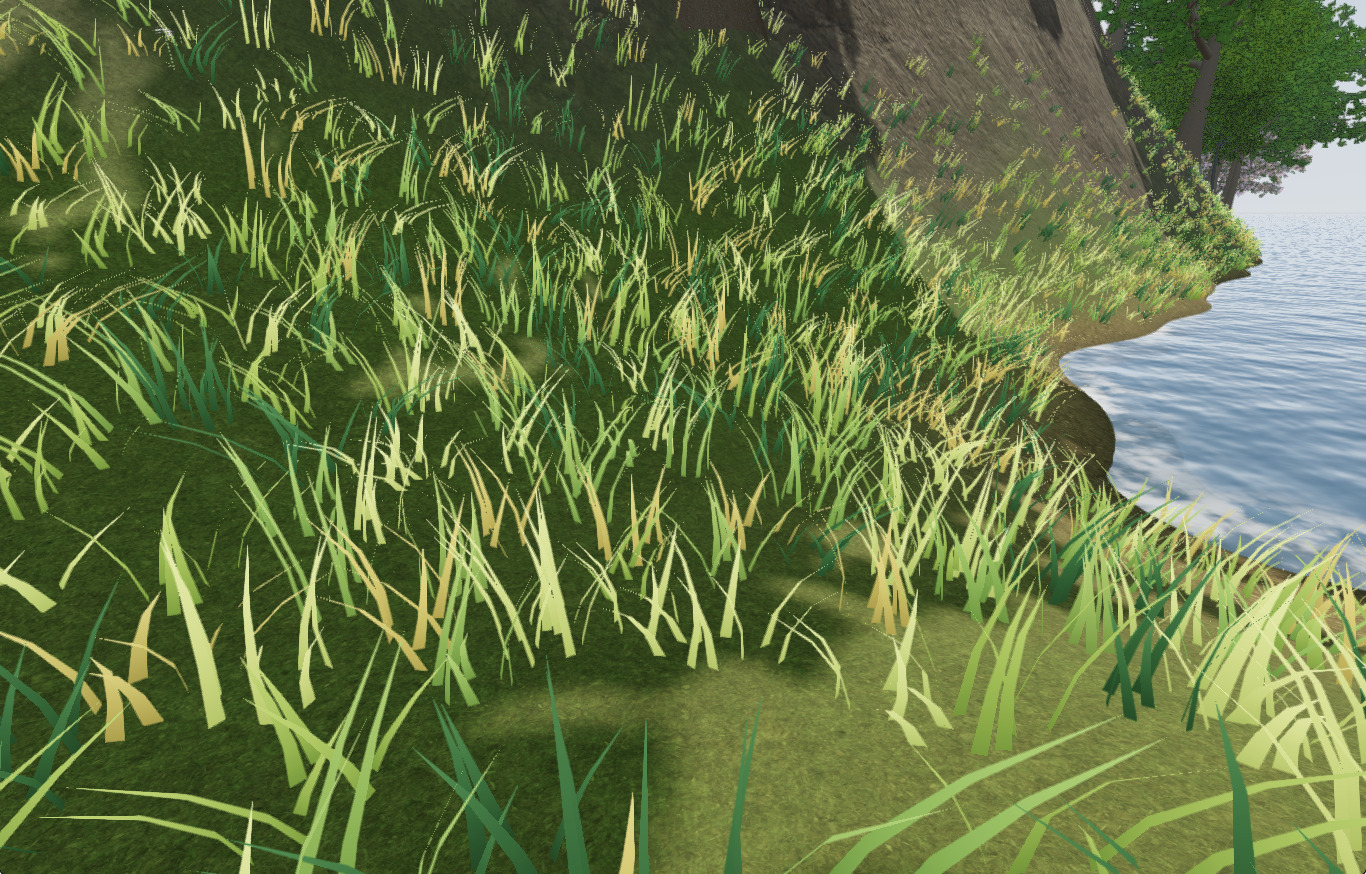}
\par\vspace{2pt}{\footnotesize Figure 2: After leaving and returning: inspect the original location again.}\end{minipage}\par\vspace{3pt}\textbf{Reference explanation.} Figure 1 shows a large boulder beside the starting point; after leaving and returning as recorded, Figure 2 shows only grass at the same shoreline location, with the boulder missing.
\end{casebox}

\phantomsection\label{demo:T2}
\begin{casebox}{casepurple}{Temporal Consistency / Static Attribute Changes}
\textbf{Definition.} Static attribute changes occur when an object's position, orientation, size, material or another normally stable property changes without a reasonable cause over time or after leaving and returning.\par\vspace{5pt}\noindent\begin{minipage}[t]{.487\linewidth}\centering
\includegraphics[width=\linewidth]{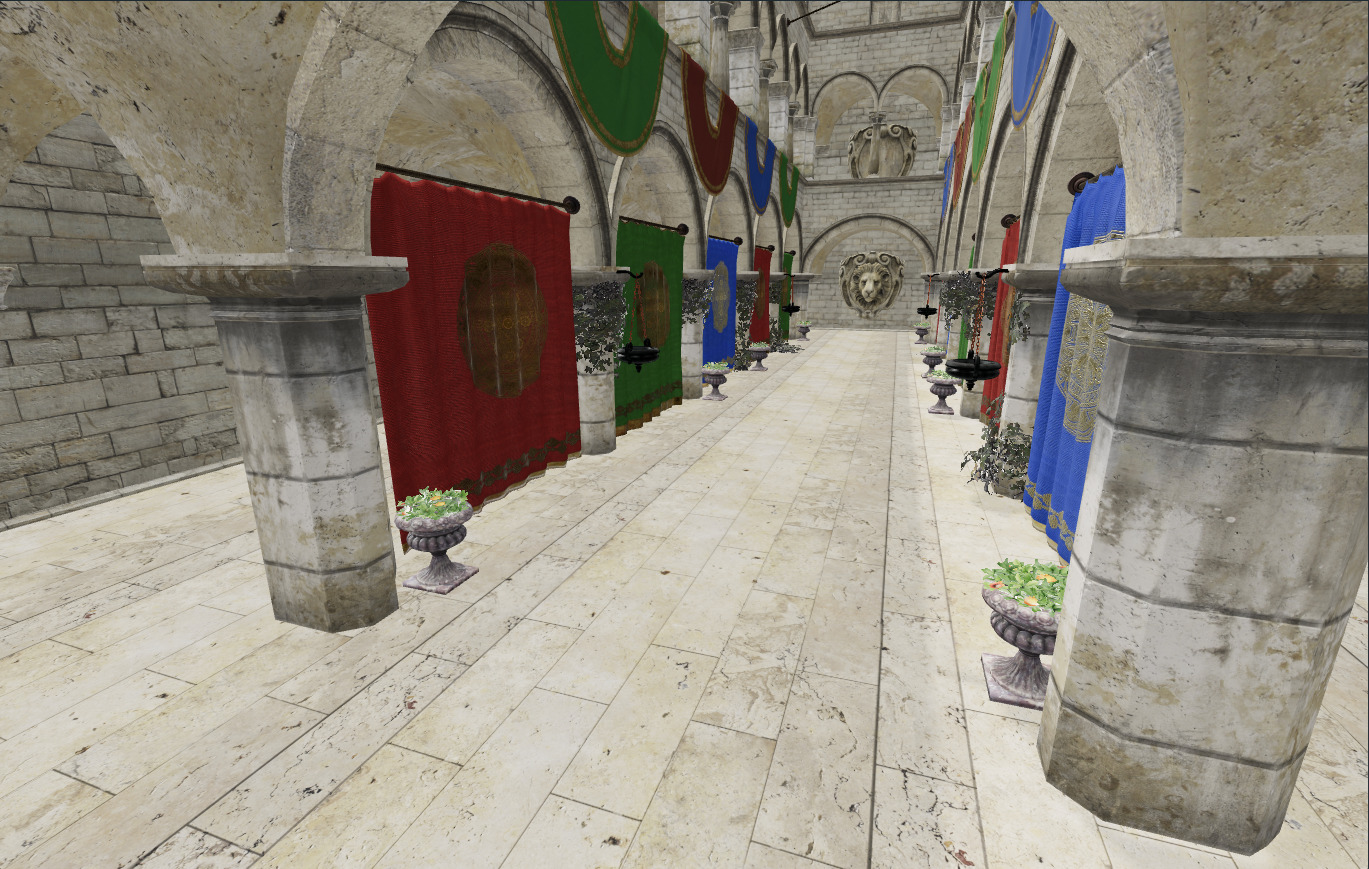}
\par\vspace{2pt}{\footnotesize Figure 1: Before leaving: locate the pot relative to the column, red banner and floor tiles.}\end{minipage}\hfill\begin{minipage}[t]{.487\linewidth}\centering
\includegraphics[width=\linewidth]{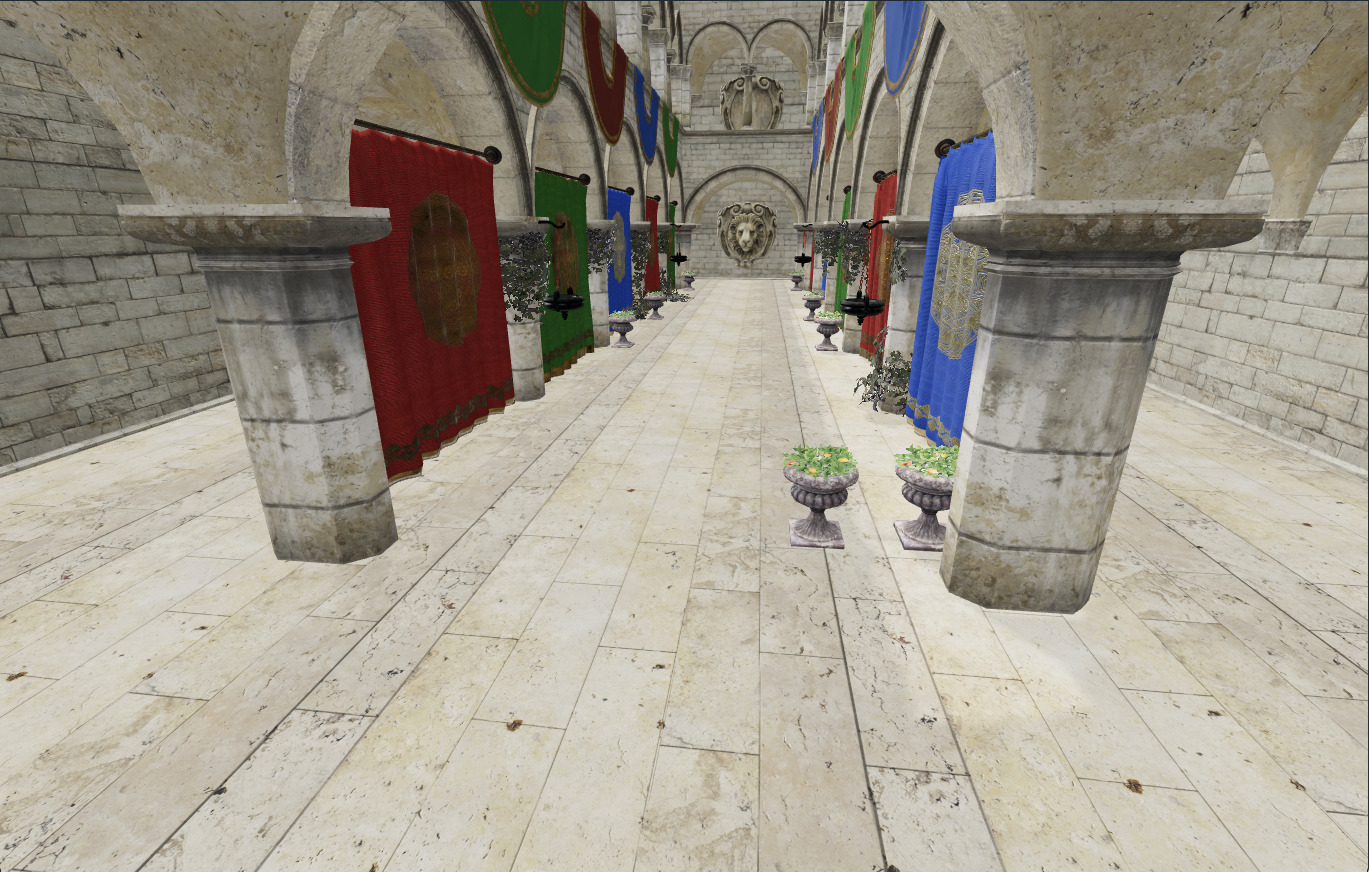}
\par\vspace{2pt}{\footnotesize Figure 2: After leaving and returning: compare the pot's position relative to the fixed architecture.}\end{minipage}\par\vspace{3pt}\textbf{Reference explanation.} In Figure 1, the flowerpot stands beside the red banner on the left; after leaving and returning, Figure 2 places it on the right side of the aisle, leaving its original spot empty. Together with the supplied record, this demonstrates a position change without anyone moving it.
\end{casebox}

\clearpage
\phantomsection\label{demo:T3}
\begin{casebox}{casepurple}{Temporal Consistency / Operational State Changes}
\fontsize{9}{11}\selectfont \textbf{Definition.} Operational state changes occur when a scene or object abruptly or inconsistently switches states that should follow an established process. Examples include discontinuous day--night changes or jumps in an ongoing animation.

Observation record: day and night alternate abruptly about once a second; the normal cycle is about half a minute with gradual transitions. This timing is reported in the supplied document.\par\vspace{5pt}\noindent\begin{minipage}[t]{.487\linewidth}\centering
\includegraphics[width=\linewidth,height=1.18in,keepaspectratio]{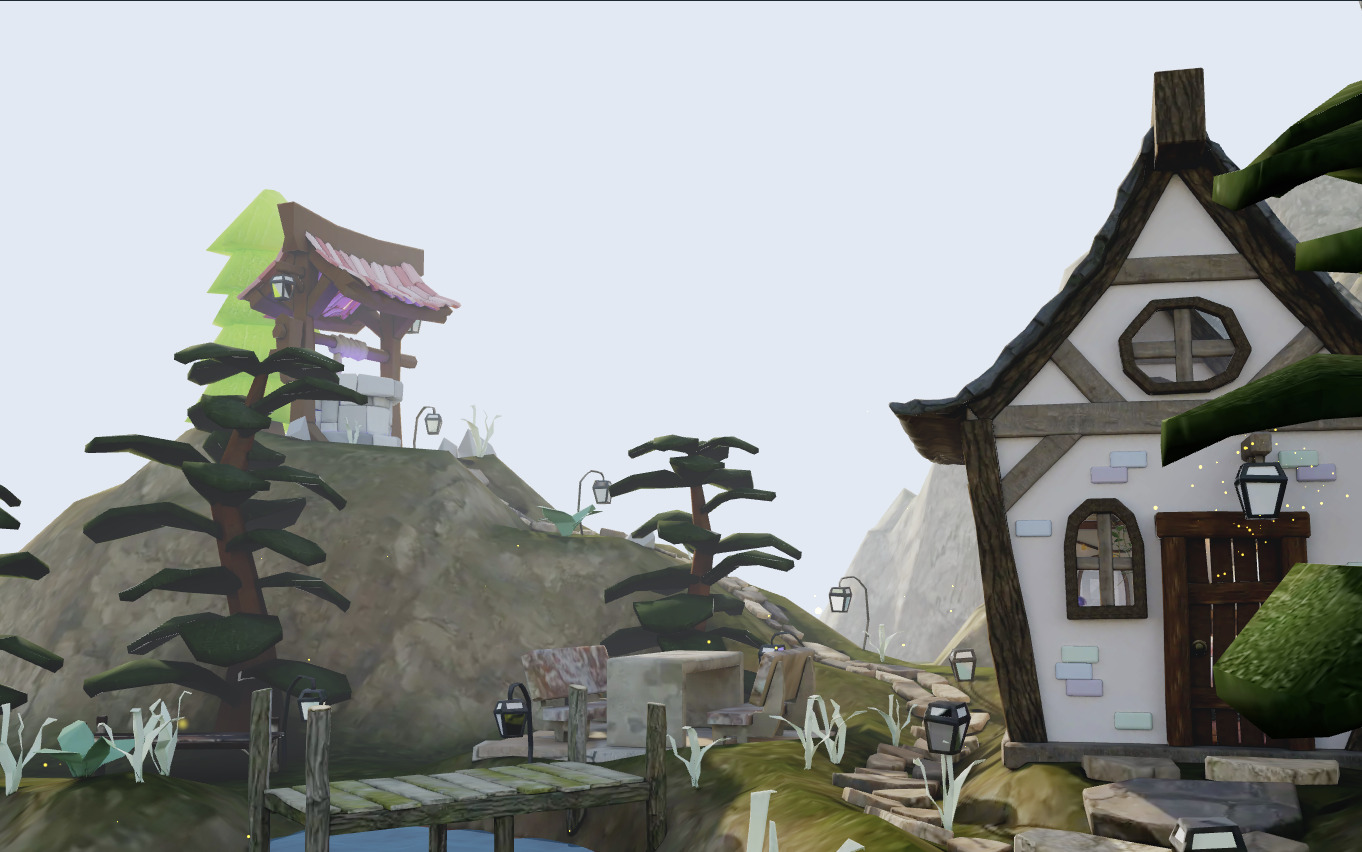}
\par\vspace{2pt}{\footnotesize Figure 1: State A: brighter ambient lighting, with the streetlamps unlit.}\end{minipage}\hfill\begin{minipage}[t]{.487\linewidth}\centering
\includegraphics[width=\linewidth,height=1.18in,keepaspectratio]{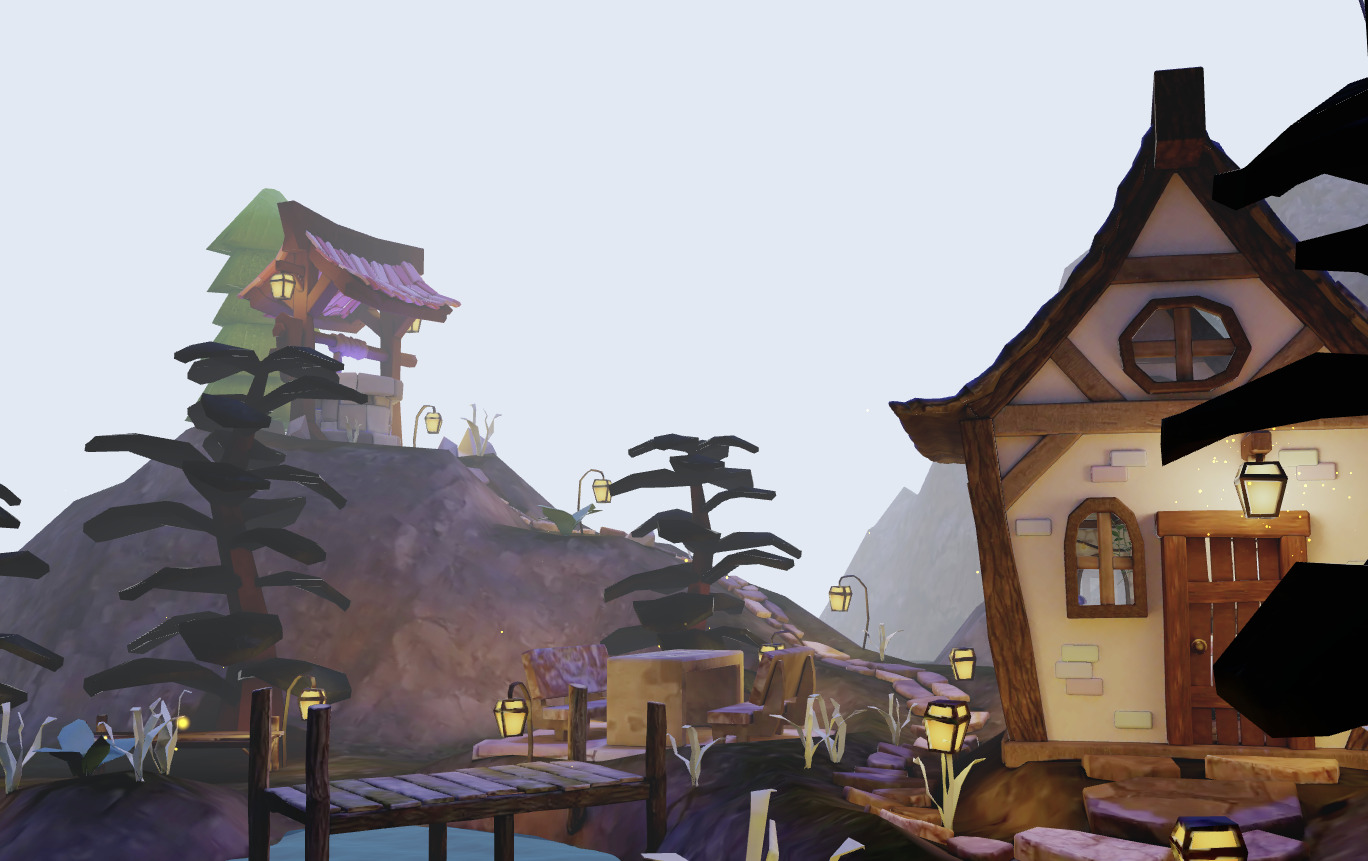}
\par\vspace{2pt}{\footnotesize Figure 2: State B: scene illumination changes and the streetlamps are lit.}\end{minipage}\par\vspace{3pt}\textbf{Reference explanation.} Figures 1 and 2 show two lighting states of the same scene, with the streetlamps switching from dark to lit. Combined with the document's report of abrupt day--night alternation about once a second, the issue is a discontinuous state transition rather than a normal gradual cycle.
\end{casebox}

\phantomsection\label{demo:S2}
\begin{casebox}{casepurple}{Semantic Consistency / Improper Configuration}
\fontsize{9}{11}\selectfont \textbf{Definition.} Improper configuration occurs when an object belongs in the era and scene, but its placement, orientation, combination or use conflicts with an established function or rule. Correcting its configuration resolves the conflict.\par\vspace{6pt}\begin{minipage}[t]{.43\linewidth}\vspace{0pt}\includegraphics[width=\linewidth,height=1.0in,keepaspectratio,trim=0 0 210 120,clip]{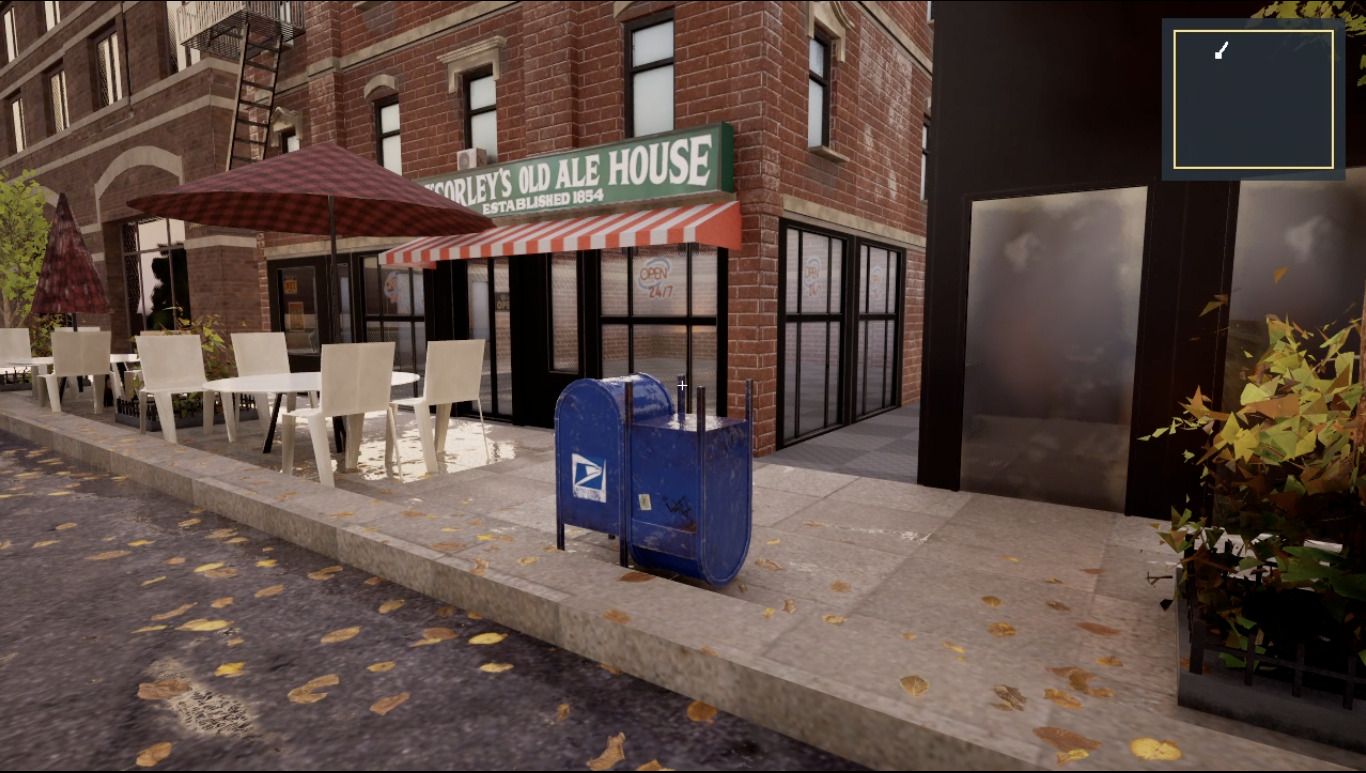}\par{\footnotesize Figure 1: Compare the legs and orientation of the inverted mailbox on the right with the upright one on the left.}\end{minipage}\hfill\begin{minipage}[t]{.54\linewidth}\vspace{0pt}\textbf{Reference explanation.} In Figure 1, the blue mailbox on the right is upside down with its legs pointing upward, unlike the upright mailbox beside it. This orientation conflicts with the mailbox's normal standing position and posting function.\end{minipage}
\end{casebox}

\phantomsection\label{demo:S3}
\begin{casebox}{casepurple}{Semantic Consistency / Historical Incompatibility}
\fontsize{9}{11}\selectfont \textbf{Definition.} Historical incompatibility occurs when an object or technology conflicts with the scene's established historical era, without an explanation such as time travel, alternate history or an exhibition.

Scene context: an ancient tea house, with no time-travel, alternate-history or exhibition premise.\par\vspace{6pt}\begin{minipage}[t]{.43\linewidth}\vspace{0pt}\includegraphics[width=\linewidth,height=1.0in,keepaspectratio,trim=0 0 210 120,clip]{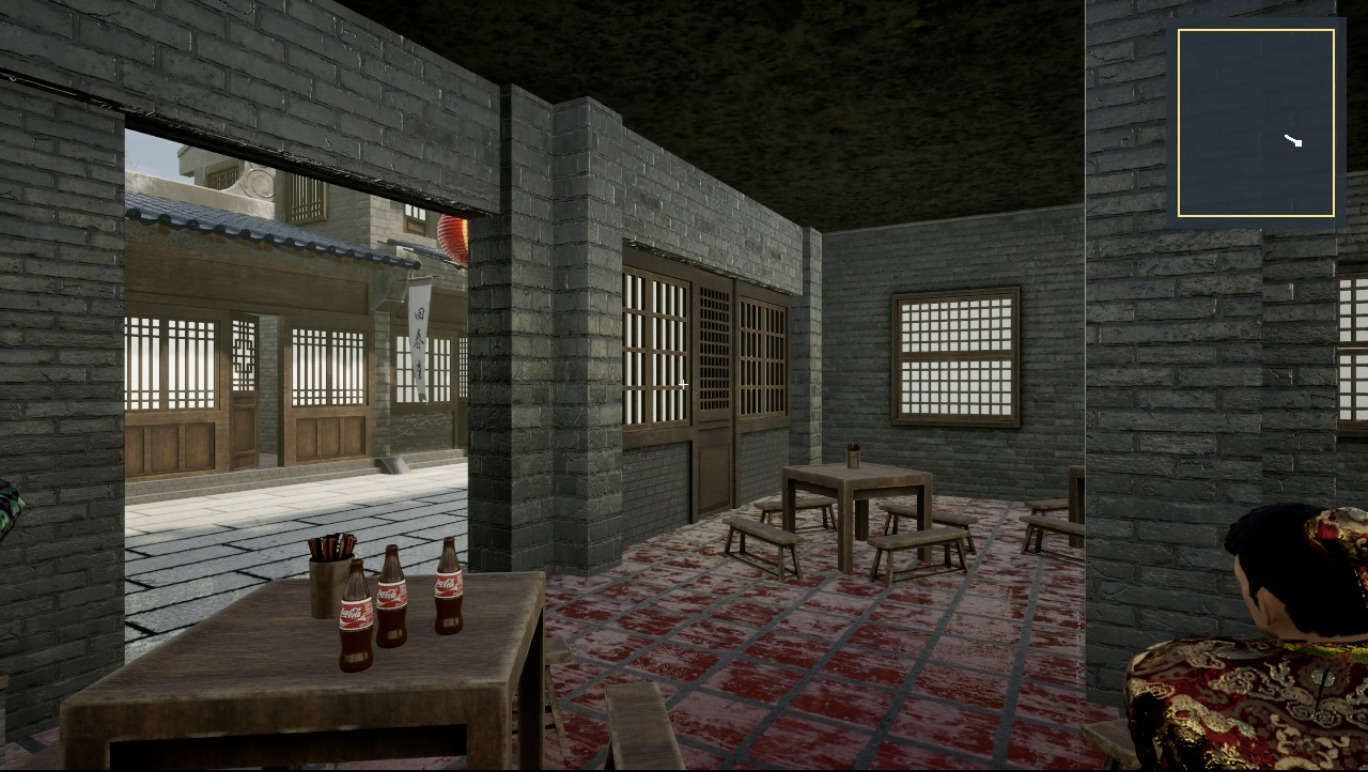}\par{\footnotesize Figure 1: Observe the three modern packaged cola bottles on the table and the surrounding tea house.}\end{minipage}\hfill\begin{minipage}[t]{.54\linewidth}\vspace{0pt}\textbf{Reference explanation.} In Figure 1, three cola bottles with modern red brand labels sit on a wooden table in an ancient tea house. Their packaging is incompatible with the historical era established for the tea house.\end{minipage}
\end{casebox}

\endgroup

\clearpage
\section{Full Results}
\label{app:full-results}
\label{app:environment-analysis}
\begin{table}[!ht]
\caption{\textbf{Results by rendering engine.} Success rate (\%) and inference cost (\$/task). Task counts are listed by anomaly family. \textbf{Bold} and \underline{underline} mark the best and second-best results within each engine and approach. Pooled results are reported in \hyperref[tab:main]{Table~\ref*{tab:main}}.}
\label{tab:engine-results}
\centering\small
\setlength{\tabcolsep}{3pt}
\renewcommand{\arraystretch}{1.2}
\newcommand{\engineheader}[2][\centering]{\parbox[c][2.6em][c]{\linewidth}{#1\strut#2\strut}}
\newcommand{\enginescore}[1]{\makebox[2.5em][r]{#1}}
\begin{tabularx}{\linewidth}{@{}>{\raggedright\arraybackslash}p{0.21\linewidth}*{5}{>{\centering\arraybackslash}X}>{\centering\arraybackslash}p{0.075\linewidth}>{\centering\arraybackslash}p{0.08\linewidth}@{}}
\toprule
\engineheader[\raggedright]{Auditor}
& \engineheader{Static\\physics} & \engineheader{Interactive\\physics}
& \engineheader{Spatial\\consistency} & \engineheader{Temporal\\consistency}
& \engineheader{Semantic\\consistency} & \engineheader{\textbf{Overall}}
& \engineheader{Cost $\downarrow$\\(\$/task)} \\
\midrule
\rowcolor{black!5}[0pt][0pt]\multicolumn{8}{@{}l@{}}{\textbf{Unreal Engine 5}\quad 126 tasks} \\
\textit{\# Tasks} & \enginescore{31} & \enginescore{22} & \enginescore{27} & \enginescore{26} & \enginescore{20} & \enginescore{126} & --- \\
\addlinespace[3pt]
\multicolumn{8}{@{}l}{\textit{VLM auditors}} \\
\addlinespace[2pt]
\raisebox{-0.15em}{\includegraphics[height=1.05em]{assets/logos/openai.png}}\ GPT-6 Astra & \enginescore{\textbf{71.0}} & \enginescore{\textbf{31.8}} & \enginescore{\textbf{44.4}} & \enginescore{\underline{15.4}} & \enginescore{\textbf{75.0}} & \enginescore{\textbf{47.6}} & \enginescore{2.923} \\
\raisebox{-0.15em}{\includegraphics[height=1.05em]{assets/logos/anthropic.png}}\ Claude Opus 5 & \enginescore{25.8} & \enginescore{13.6} & \enginescore{\underline{29.6}} & \enginescore{\textbf{19.2}} & \enginescore{\underline{55.0}} & \enginescore{27.8} & \enginescore{2.427} \\
\raisebox{-0.15em}{\includegraphics[height=1.05em]{assets/logos/gemini-color.png}}\ Gemini 3.8 Flash & \enginescore{\underline{51.6}} & \enginescore{\underline{18.2}} & \enginescore{22.2} & \enginescore{\textbf{19.2}} & \enginescore{\underline{55.0}} & \enginescore{\underline{33.3}} & \enginescore{1.340} \\
\raisebox{-0.15em}{\includegraphics[height=1.05em]{assets/logos/meta-color.png}}\ Muse Spark 1.3 & \enginescore{16.1} & \enginescore{9.1} & \enginescore{0.0} & \enginescore{7.7} & \enginescore{25.0} & \enginescore{11.1} & \enginescore{\underline{0.743}} \\
\raisebox{-0.15em}{\includegraphics[height=1.05em]{assets/logos/qwen-color.png}}\ Qwen 3.8 Flash & \enginescore{9.7} & \enginescore{9.1} & \enginescore{0.0} & \enginescore{3.8} & \enginescore{25.0} & \enginescore{8.7} & \enginescore{\textbf{0.077}} \\
\addlinespace[3pt]
\multicolumn{8}{@{}l}{\textit{VLA--VLM auditors}} \\
\addlinespace[2pt]
\raisebox{-0.15em}{\includegraphics[height=1.05em]{assets/logos/openai.png}}\ GPT-6 Astra & \enginescore{\underline{9.7}} & \enginescore{\underline{0.0}} & \enginescore{\textbf{11.1}} & \enginescore{\textbf{0.0}} & \enginescore{\underline{5.0}} & \enginescore{\textbf{5.6}} & \enginescore{0.253} \\
\raisebox{-0.15em}{\includegraphics[height=1.05em]{assets/logos/anthropic.png}}\ Claude Opus 5 & \enginescore{6.5} & \enginescore{\underline{0.0}} & \enginescore{\underline{0.0}} & \enginescore{\textbf{0.0}} & \enginescore{\underline{5.0}} & \enginescore{\underline{2.4}} & \enginescore{0.634} \\
\raisebox{-0.15em}{\includegraphics[height=1.05em]{assets/logos/gemini-color.png}}\ Gemini 3.8 Flash & \enginescore{\textbf{16.1}} & \enginescore{\underline{0.0}} & \enginescore{\underline{0.0}} & \enginescore{\textbf{0.0}} & \enginescore{\textbf{10.0}} & \enginescore{\textbf{5.6}} & \enginescore{0.287} \\
\raisebox{-0.15em}{\includegraphics[height=1.05em]{assets/logos/meta-color.png}}\ Muse Spark 1.3 & \enginescore{3.2} & \enginescore{\underline{0.0}} & \enginescore{\underline{0.0}} & \enginescore{\textbf{0.0}} & \enginescore{\textbf{10.0}} & \enginescore{\underline{2.4}} & \enginescore{\underline{0.102}} \\
\raisebox{-0.15em}{\includegraphics[height=1.05em]{assets/logos/qwen-color.png}}\ Qwen 3.8 Flash & \enginescore{\underline{9.7}} & \enginescore{\textbf{13.6}} & \enginescore{\underline{0.0}} & \enginescore{\textbf{0.0}} & \enginescore{\underline{5.0}} & \enginescore{\textbf{5.6}} & \enginescore{\textbf{0.036}} \\
\midrule
\rowcolor{black!5}[0pt][0pt]\multicolumn{8}{@{}l@{}}{\textbf{Three.js}\quad 87 tasks} \\
\textit{\# Tasks} & \enginescore{28} & \enginescore{19} & \enginescore{24} & \enginescore{14} & \enginescore{2} & \enginescore{87} & --- \\
\addlinespace[3pt]
\multicolumn{8}{@{}l}{\textit{VLM auditors}} \\
\addlinespace[2pt]
\raisebox{-0.15em}{\includegraphics[height=1.05em]{assets/logos/openai.png}}\ GPT-6 Astra & \enginescore{\textbf{46.4}} & \enginescore{\textbf{26.3}} & \enginescore{\textbf{45.8}} & \enginescore{\textbf{7.1}} & \enginescore{\textbf{0.0}} & \enginescore{\textbf{34.5}} & \enginescore{3.000} \\
\raisebox{-0.15em}{\includegraphics[height=1.05em]{assets/logos/anthropic.png}}\ Claude Opus 5 & \enginescore{\textbf{46.4}} & \enginescore{\textbf{26.3}} & \enginescore{29.2} & \enginescore{\underline{0.0}} & \enginescore{\textbf{0.0}} & \enginescore{28.7} & \enginescore{2.486} \\
\raisebox{-0.15em}{\includegraphics[height=1.05em]{assets/logos/gemini-color.png}}\ Gemini 3.8 Flash & \enginescore{\textbf{46.4}} & \enginescore{\textbf{26.3}} & \enginescore{\underline{37.5}} & \enginescore{\underline{0.0}} & \enginescore{\textbf{0.0}} & \enginescore{\underline{31.0}} & \enginescore{1.478} \\
\raisebox{-0.15em}{\includegraphics[height=1.05em]{assets/logos/meta-color.png}}\ Muse Spark 1.3 & \enginescore{\underline{28.6}} & \enginescore{\underline{15.8}} & \enginescore{29.2} & \enginescore{\underline{0.0}} & \enginescore{\textbf{0.0}} & \enginescore{20.7} & \enginescore{\underline{0.860}} \\
\raisebox{-0.15em}{\includegraphics[height=1.05em]{assets/logos/qwen-color.png}}\ Qwen 3.8 Flash & \enginescore{14.3} & \enginescore{5.3} & \enginescore{8.3} & \enginescore{\underline{0.0}} & \enginescore{\textbf{0.0}} & \enginescore{8.0} & \enginescore{\textbf{0.078}} \\
\addlinespace[3pt]
\multicolumn{8}{@{}l}{\textit{VLA--VLM auditors}} \\
\addlinespace[2pt]
\raisebox{-0.15em}{\includegraphics[height=1.05em]{assets/logos/openai.png}}\ GPT-6 Astra & \enginescore{\underline{32.1}} & \enginescore{36.8} & \enginescore{\textbf{41.7}} & \enginescore{0.0} & \enginescore{\textbf{100.0}} & \enginescore{32.2} & \enginescore{0.321} \\
\raisebox{-0.15em}{\includegraphics[height=1.05em]{assets/logos/anthropic.png}}\ Claude Opus 5 & \enginescore{\textbf{39.3}} & \enginescore{\underline{47.4}} & \enginescore{25.0} & \enginescore{\underline{7.1}} & \enginescore{\textbf{100.0}} & \enginescore{\underline{33.3}} & \enginescore{0.861} \\
\raisebox{-0.15em}{\includegraphics[height=1.05em]{assets/logos/gemini-color.png}}\ Gemini 3.8 Flash & \enginescore{25.0} & \enginescore{\textbf{52.6}} & \enginescore{\underline{37.5}} & \enginescore{\textbf{14.3}} & \enginescore{\textbf{100.0}} & \enginescore{\textbf{34.5}} & \enginescore{0.309} \\
\raisebox{-0.15em}{\includegraphics[height=1.05em]{assets/logos/meta-color.png}}\ Muse Spark 1.3 & \enginescore{7.1} & \enginescore{21.1} & \enginescore{20.8} & \enginescore{0.0} & \enginescore{\underline{0.0}} & \enginescore{12.6} & \enginescore{\underline{0.127}} \\
\raisebox{-0.15em}{\includegraphics[height=1.05em]{assets/logos/qwen-color.png}}\ Qwen 3.8 Flash & \enginescore{28.6} & \enginescore{31.6} & \enginescore{25.0} & \enginescore{0.0} & \enginescore{\underline{0.0}} & \enginescore{23.0} & \enginescore{\textbf{0.045}} \\
\bottomrule
\end{tabularx}
\end{table}

\FloatBarrier

\clearpage
\section{Geometric Coverage}
\label{app:geometric-coverage}
\label{app:exploration-recognition}

Geometric coverage measures whether an auditor comes close to the annotated anomaly and looks toward it. For the analysis in Section~\ref{sec:ablation-discovery}, a task counts as covered if at least one evaluated screenshot meets both conditions:
\begin{itemize}
\item \textbf{Distance:} the horizontal distance to the target is no greater than the limit set for that environment. These maximum distances range from 7 to 16 metres across environments.
\item \textbf{Direction:} the target is within approximately $41.4^\circ$ to either side of the camera's viewing direction and $32^\circ$ above or below it.
\end{itemize}
Both auditors use the same target locations and thresholds. This check uses recorded positions and viewing directions, rather than image content: an occluded target can pass, and the check does not establish whether the anomaly occurred or was recognized.

We evaluate screenshots delivered to the auditor: VLM screenshots captured at the end of each environment action, and VLA screenshots matched to the original camera positions and orientations at their capture times. Intermediate VLM frames are excluded from this calculation because their stored positions and orientations describe the end of the action, not the instant of capture. These frames remain available to the auditor through \texttt{inspect} for comparing observations and preparing reports.

\clearpage
\section{Qualitative Auditing Trajectories}
\label{app:qualitative}

We show how the Gemini 3.8 Flash VLM-only auditor explores scenes, gathers evidence, and reports anomalies through three successful and three failed audits. All six cases are drawn from the Unreal Engine main evaluation with a budget of 40 environment actions per task. The cases illustrate different outcomes; they are not intended to estimate how often each failure occurs.

Each case presents 12 selected views along the recorded trajectory, read from left to right and top to bottom, followed by the target anomaly, a summary of the auditor's report, and an interpretation of the outcome. \textit{Start} marks the initial view. Step numbers refer to environment actions; gaps indicate omitted actions. Earlier and later views within one step correspond to intermediate frames from the same action, retrieved by the auditor using the memory tool \texttt{inspect}. The report summaries paraphrase the auditor's outputs.

\begin{table}[h]
\centering\small
\caption{Selected audits. Success refers to reporting the annotated target under the benchmark judge, rather than to whether any other reported issue is valid.}
\label{tab:qualitative-cases}
\begin{tabularx}{\textwidth}{@{}llXc@{}}
\toprule
Case & Anomaly family & Target anomaly & Outcome \\
\midrule
\hyperref[app:qual-A17]{Entrance door} & Temporal consistency & Door closes without operation & Success \\
\hyperref[app:qual-I04]{Solid lockers} & Interactive physics & Traversable solid lockers & Success \\
\hyperref[app:qual-A20]{Market appliance} & Semantic consistency & Modern vending machine in ancient market & Success \\
\hyperref[app:qual-H10]{Floor plant} & Temporal consistency & Floor plant disappears on departure & Failure \\
\hyperref[app:qual-S16]{Wall poster} & Temporal consistency & Wall poster disappears over time & Failure \\
\hyperref[app:qual-S01]{Waiting bench} & Static physics & Floating waiting bench & Failure \\
\bottomrule
\end{tabularx}
\end{table}

\begingroup

\clearpage
\phantomsection\label{app:qual-A17}
\noindent\textbf{Success: An Entrance Door Closes Without Operation}\par\nobreak\smallskip
\noindent Temporal consistency $\mid$ 40-action budget

\begin{figure}[!ht]
\centering
\begin{minipage}[t]{0.323\linewidth}
\centering
\ifnum\pdfstrcmp{A17}{S16}=0
\includegraphics[width=\linewidth,trim=0 62 210 56,clip]{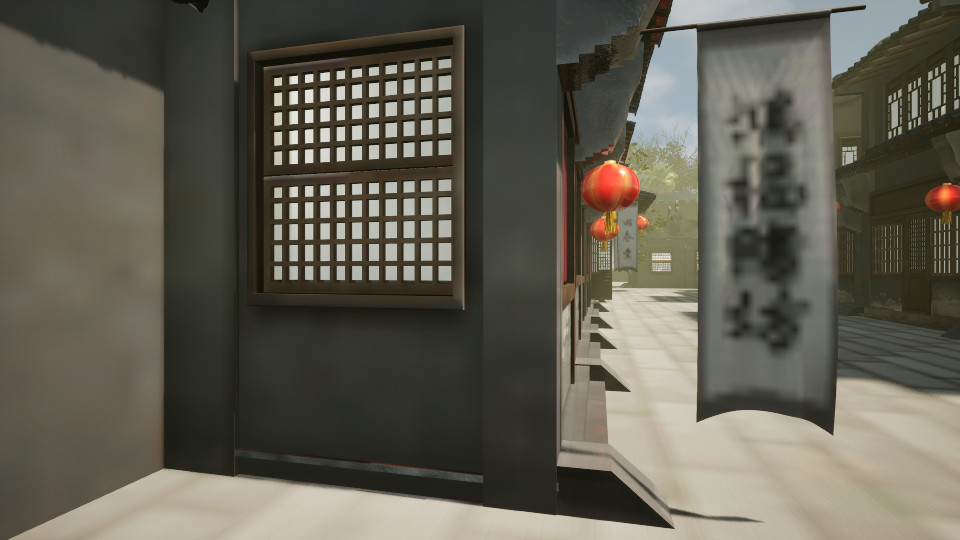}%
\else
\includegraphics[width=\linewidth]{assets/qualitative/A17/a000.jpg}%
\fi\\[-1pt]
{\scriptsize Start}\\[-1pt]
{\scriptsize Initial view\strut}
\end{minipage}\hfill
\begin{minipage}[t]{0.323\linewidth}
\centering
\ifnum\pdfstrcmp{A17}{S16}=0
\includegraphics[width=\linewidth,trim=0 62 210 56,clip]{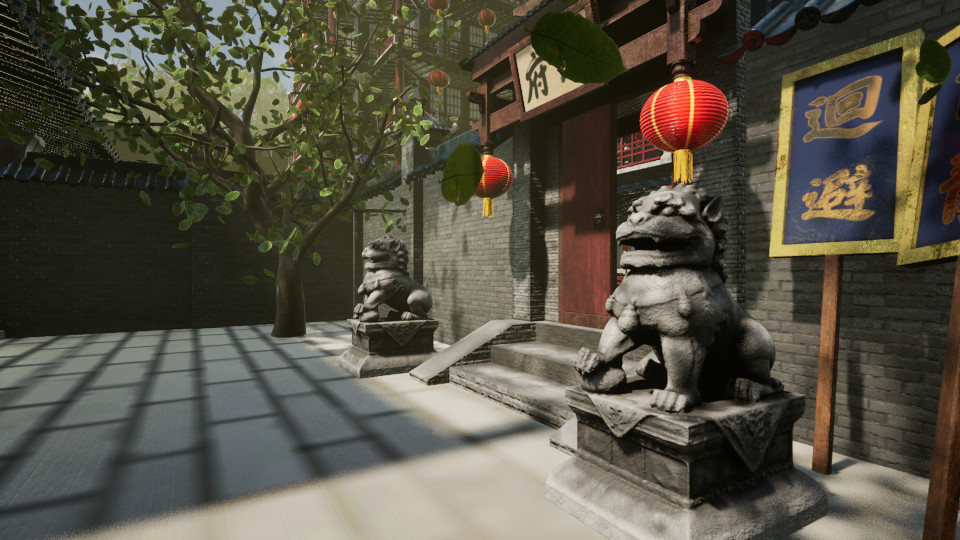}%
\else
\includegraphics[width=\linewidth]{assets/qualitative/A17/a011.jpg}%
\fi\\[-1pt]
{\scriptsize Step 11}\\[-1pt]
{\scriptsize Explore the entrance\strut}
\end{minipage}\hfill
\begin{minipage}[t]{0.323\linewidth}
\centering
\ifnum\pdfstrcmp{A17}{S16}=0
\includegraphics[width=\linewidth,trim=0 62 210 56,clip]{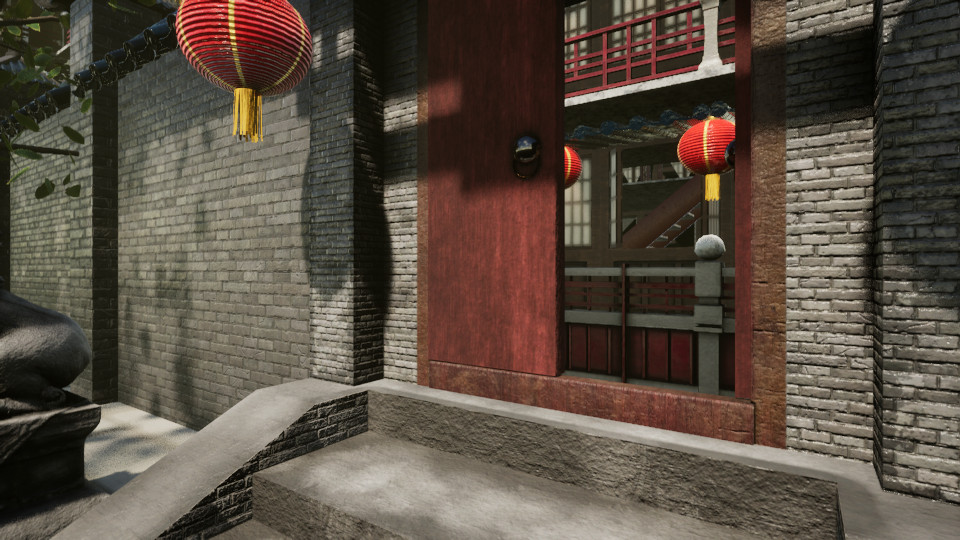}%
\else
\includegraphics[width=\linewidth]{assets/qualitative/A17/a023.jpg}%
\fi\\[-1pt]
{\scriptsize Step 23}\\[-1pt]
{\scriptsize Open doorway\strut}
\end{minipage}\par\vspace{7pt}
\begin{minipage}[t]{0.323\linewidth}
\centering
\ifnum\pdfstrcmp{A17}{S16}=0
\includegraphics[width=\linewidth,trim=0 62 210 56,clip]{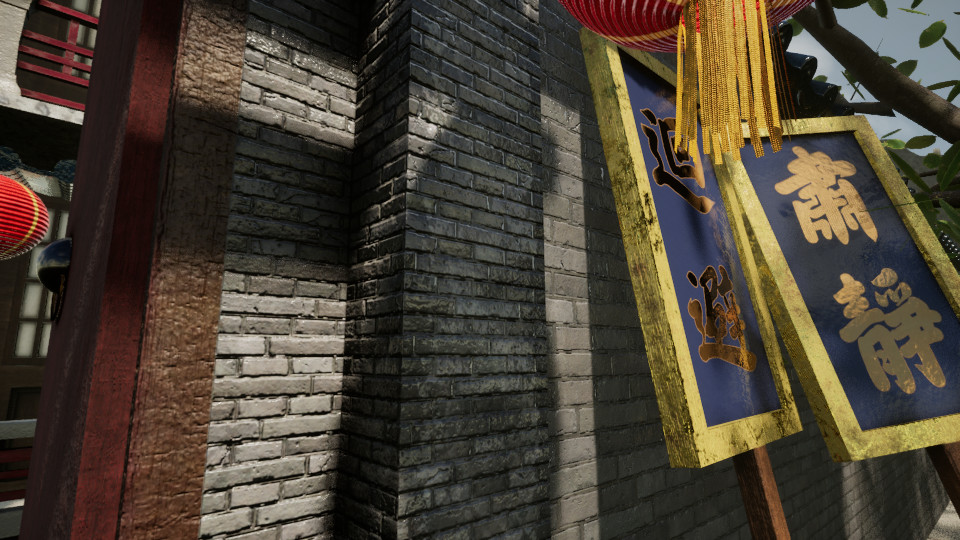}%
\else
\includegraphics[width=\linewidth]{assets/qualitative/A17/a030.jpg}%
\fi\\[-1pt]
{\scriptsize Step 30}\\[-1pt]
{\scriptsize Observe from doorjamb\strut}
\end{minipage}\hfill
\begin{minipage}[t]{0.323\linewidth}
\centering
\ifnum\pdfstrcmp{A17}{S16}=0
\includegraphics[width=\linewidth,trim=0 62 210 56,clip]{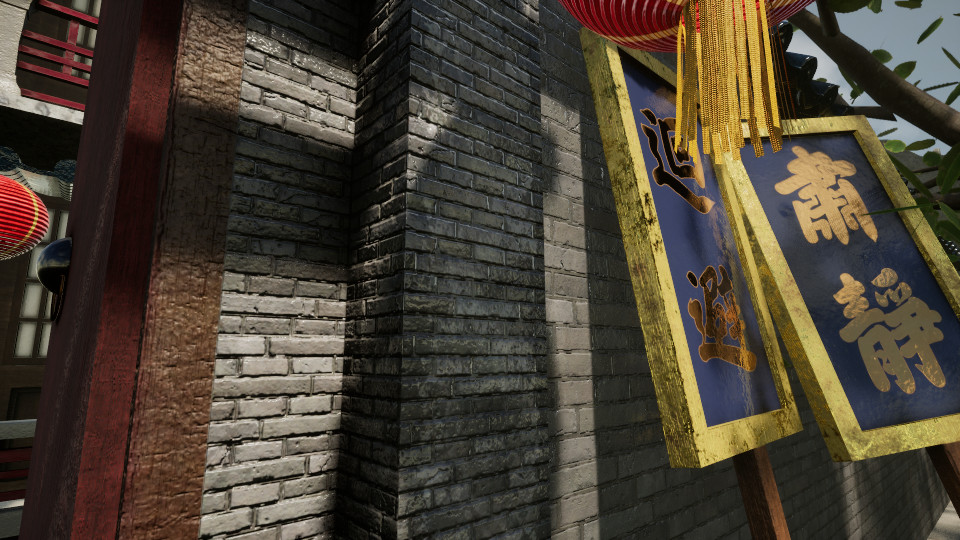}%
\else
\includegraphics[width=\linewidth]{assets/qualitative/A17/a031_f05.jpg}%
\fi\\[-1pt]
{\scriptsize Step 31 (\texttt{inspect})}\\[-1pt]
{\scriptsize Wait: door still open\strut}
\end{minipage}\hfill
\begin{minipage}[t]{0.323\linewidth}
\centering
\ifnum\pdfstrcmp{A17}{S16}=0
\includegraphics[width=\linewidth,trim=0 62 210 56,clip]{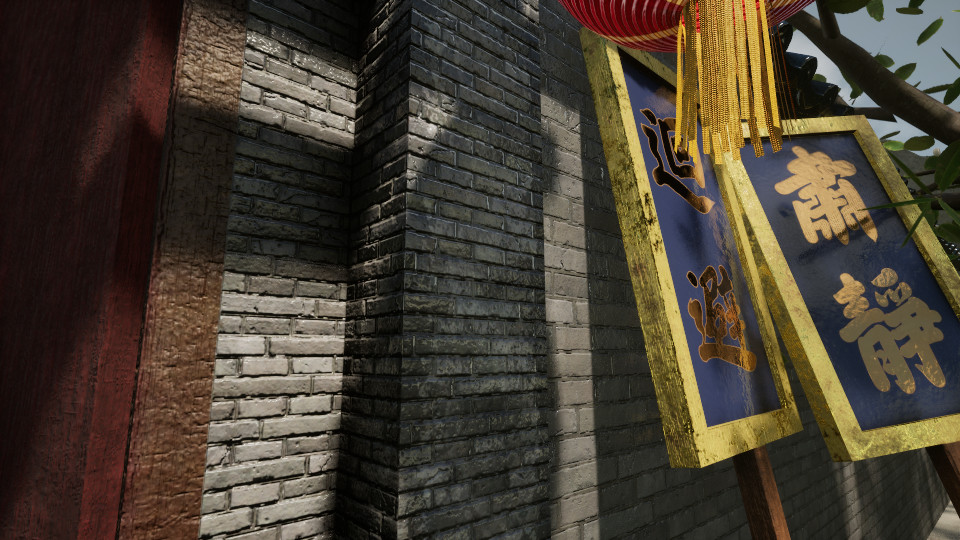}%
\else
\includegraphics[width=\linewidth]{assets/qualitative/A17/a031_f06.jpg}%
\fi\\[-1pt]
{\scriptsize Step 31 (\texttt{inspect})}\\[-1pt]
{\scriptsize Wait: door now closed\strut}
\end{minipage}\par\vspace{7pt}
\begin{minipage}[t]{0.323\linewidth}
\centering
\ifnum\pdfstrcmp{A17}{S16}=0
\includegraphics[width=\linewidth,trim=0 62 210 56,clip]{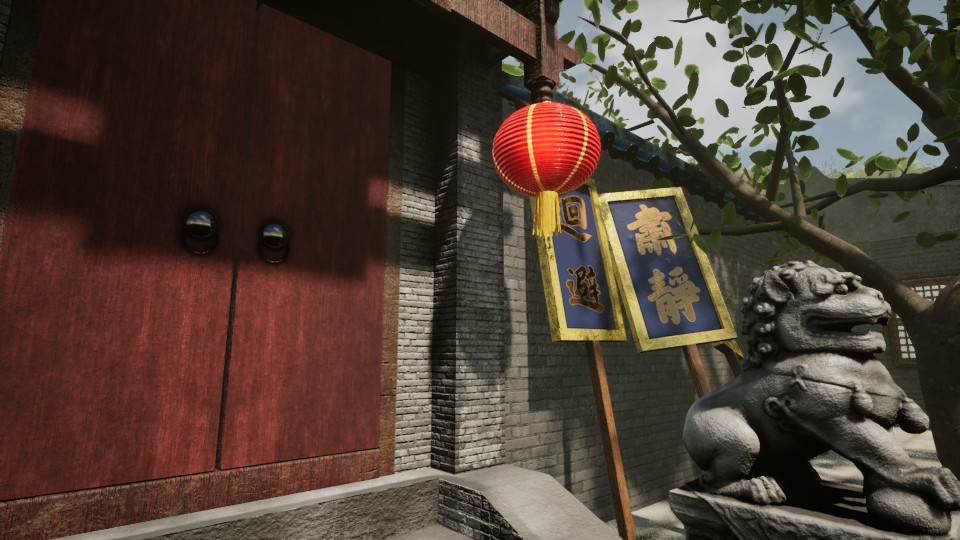}%
\else
\includegraphics[width=\linewidth]{assets/qualitative/A17/a032.jpg}%
\fi\\[-1pt]
{\scriptsize Step 32}\\[-1pt]
{\scriptsize Step back to verify\strut}
\end{minipage}\hfill
\begin{minipage}[t]{0.323\linewidth}
\centering
\ifnum\pdfstrcmp{A17}{S16}=0
\includegraphics[width=\linewidth,trim=0 62 210 56,clip]{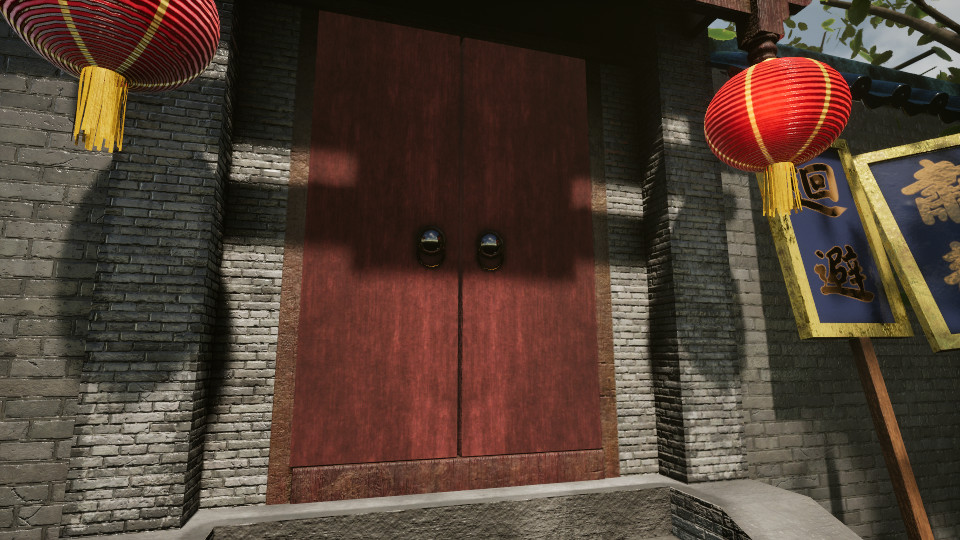}%
\else
\includegraphics[width=\linewidth]{assets/qualitative/A17/a035.jpg}%
\fi\\[-1pt]
{\scriptsize Step 35}\\[-1pt]
{\scriptsize Door remains closed\strut}
\end{minipage}\hfill
\begin{minipage}[t]{0.323\linewidth}
\centering
\ifnum\pdfstrcmp{A17}{S16}=0
\includegraphics[width=\linewidth,trim=0 62 210 56,clip]{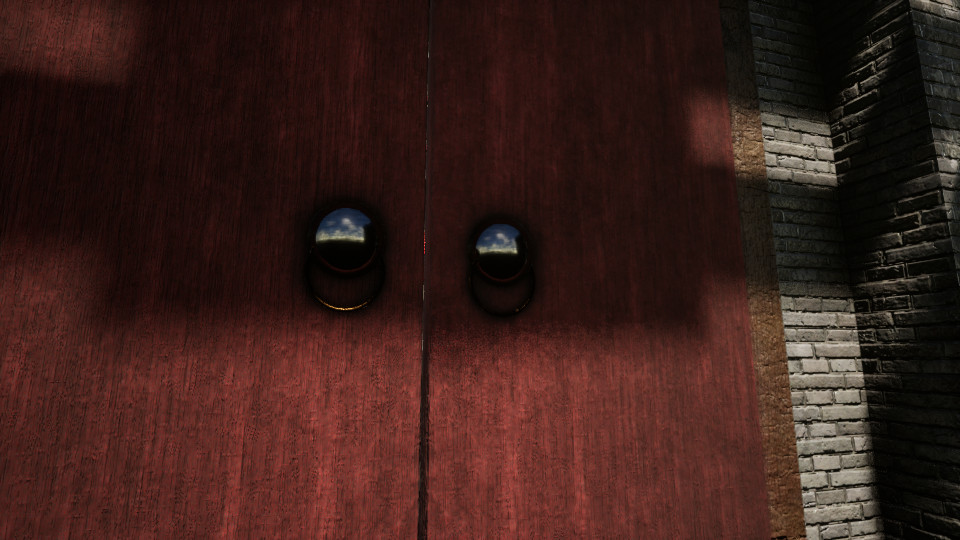}%
\else
\includegraphics[width=\linewidth]{assets/qualitative/A17/a037.jpg}%
\fi\\[-1pt]
{\scriptsize Step 37}\\[-1pt]
{\scriptsize Operate the door\strut}
\end{minipage}\par\vspace{7pt}
\begin{minipage}[t]{0.323\linewidth}
\centering
\ifnum\pdfstrcmp{A17}{S16}=0
\includegraphics[width=\linewidth,trim=0 62 210 56,clip]{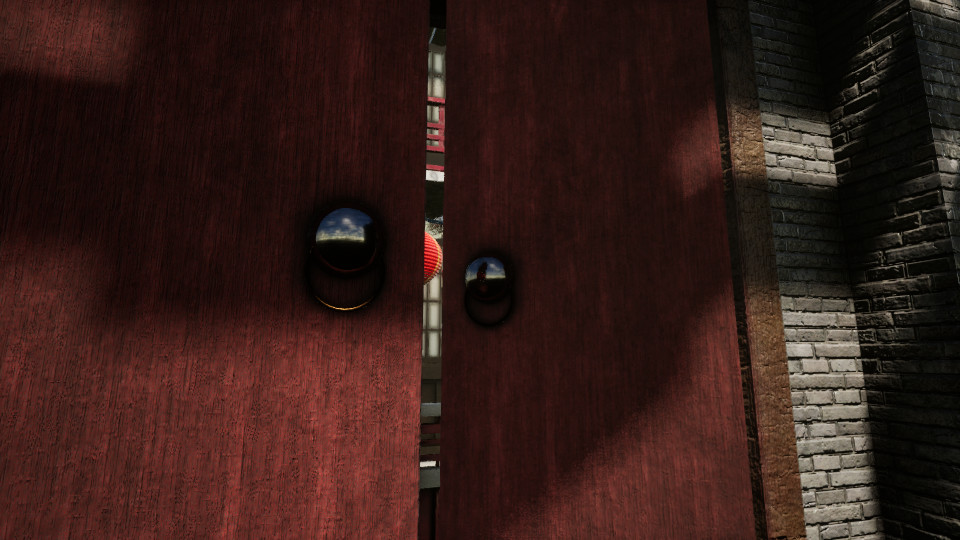}%
\else
\includegraphics[width=\linewidth]{assets/qualitative/A17/a038_f00.jpg}%
\fi\\[-1pt]
{\scriptsize Step 38 (\texttt{inspect})}\\[-1pt]
{\scriptsize Earlier opening frame\strut}
\end{minipage}\hfill
\begin{minipage}[t]{0.323\linewidth}
\centering
\ifnum\pdfstrcmp{A17}{S16}=0
\includegraphics[width=\linewidth,trim=0 62 210 56,clip]{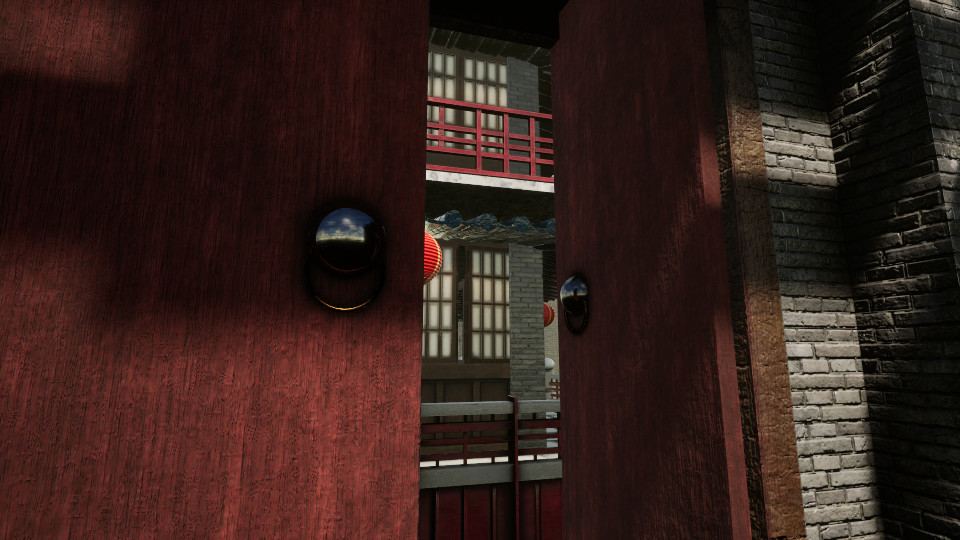}%
\else
\includegraphics[width=\linewidth]{assets/qualitative/A17/a038_f03.jpg}%
\fi\\[-1pt]
{\scriptsize Step 38 (\texttt{inspect})}\\[-1pt]
{\scriptsize Later opening frame\strut}
\end{minipage}\hfill
\begin{minipage}[t]{0.323\linewidth}
\centering
\ifnum\pdfstrcmp{A17}{S16}=0
\includegraphics[width=\linewidth,trim=0 62 210 56,clip]{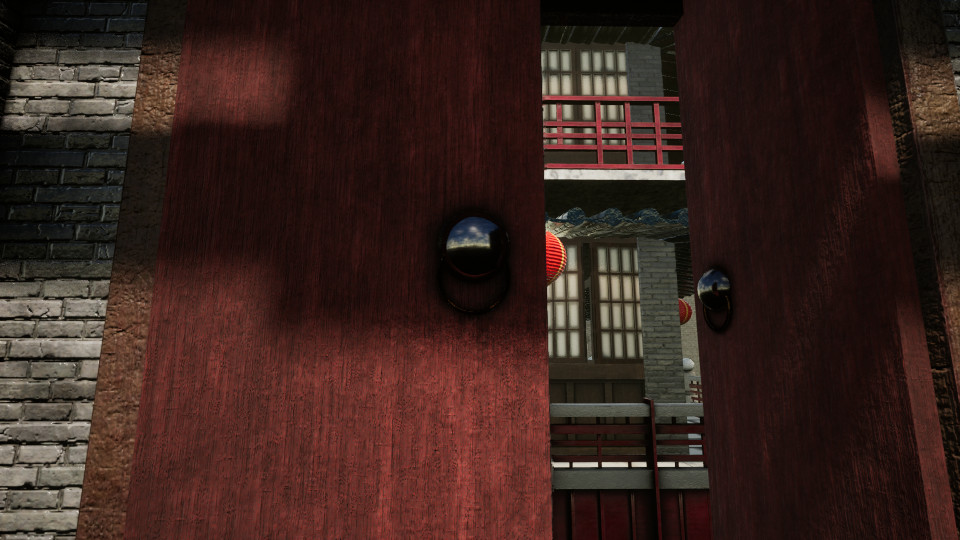}%
\else
\includegraphics[width=\linewidth]{assets/qualitative/A17/a040.jpg}%
\fi\\[-1pt]
{\scriptsize Step 40}\\[-1pt]
{\scriptsize Final observation\strut}
\end{minipage}\par\vspace{7pt}
\caption{Entrance-door inspection. The intermediate frames at Steps 31 and 38 were retrieved using the memory tool \texttt{inspect}.}\label{fig:qual-A17}
\end{figure}
\FloatBarrier

\noindent\textbf{Target anomaly.} An open entrance door closes by itself while being observed, without anyone operating it.

\noindent\textbf{Auditor report.} The auditor reports that the right wooden door leaf closes automatically during a wait, citing the before-and-after views at Step 31 and the confirming view at Step 32. The two views at Step 38 show the opening sequence following an interaction.

\noindent\textbf{Interpretation.} The report is correct: the Step 31 pair captures the door closing without an interaction, matching the target anomaly. The change appears at the far left edge, where the red door panel replaces the opening.

\clearpage
\phantomsection\label{app:qual-I04}
\noindent\textbf{Success: Walking Through Solid Lockers}\par\nobreak\smallskip
\noindent Interactive physics $\mid$ 40-action budget

\begin{figure}[!ht]
\centering
\begin{minipage}[t]{0.323\linewidth}
\centering
\ifnum\pdfstrcmp{I04}{S16}=0
\includegraphics[width=\linewidth,trim=0 62 210 56,clip]{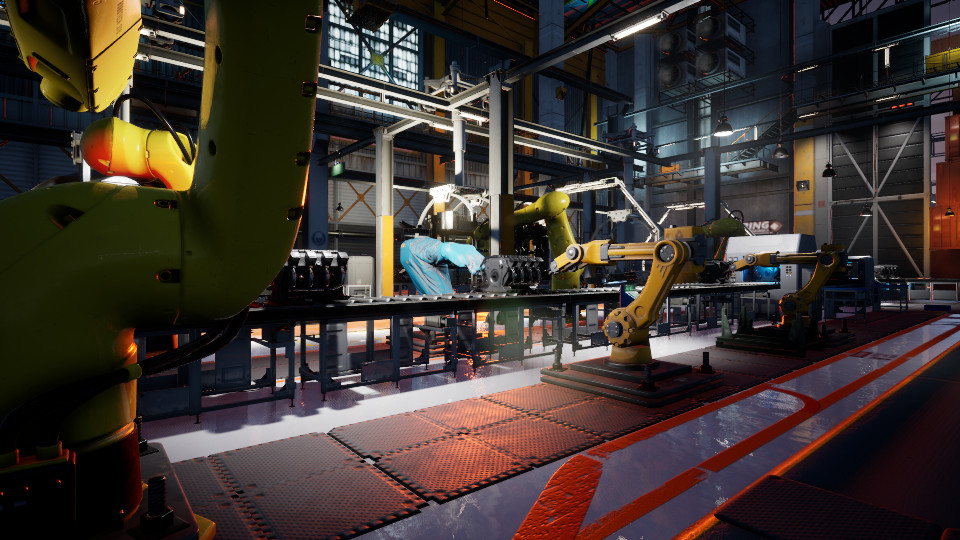}%
\else
\includegraphics[width=\linewidth]{assets/qualitative/I04/a000.jpg}%
\fi\\[-1pt]
{\scriptsize Start}\\[-1pt]
{\scriptsize Initial factory view\strut}
\end{minipage}\hfill
\begin{minipage}[t]{0.323\linewidth}
\centering
\ifnum\pdfstrcmp{I04}{S16}=0
\includegraphics[width=\linewidth,trim=0 62 210 56,clip]{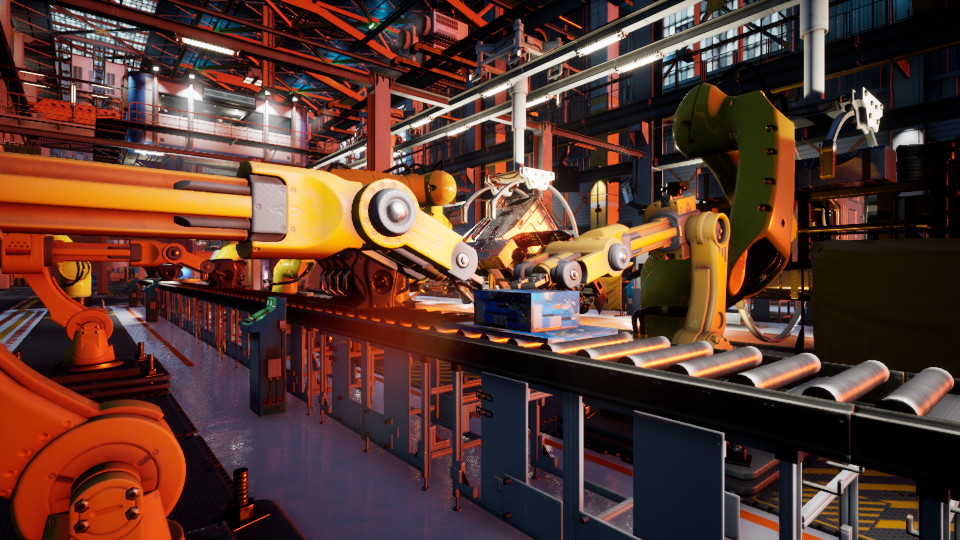}%
\else
\includegraphics[width=\linewidth]{assets/qualitative/I04/a005.jpg}%
\fi\\[-1pt]
{\scriptsize Step 5}\\[-1pt]
{\scriptsize Explore machinery\strut}
\end{minipage}\hfill
\begin{minipage}[t]{0.323\linewidth}
\centering
\ifnum\pdfstrcmp{I04}{S16}=0
\includegraphics[width=\linewidth,trim=0 62 210 56,clip]{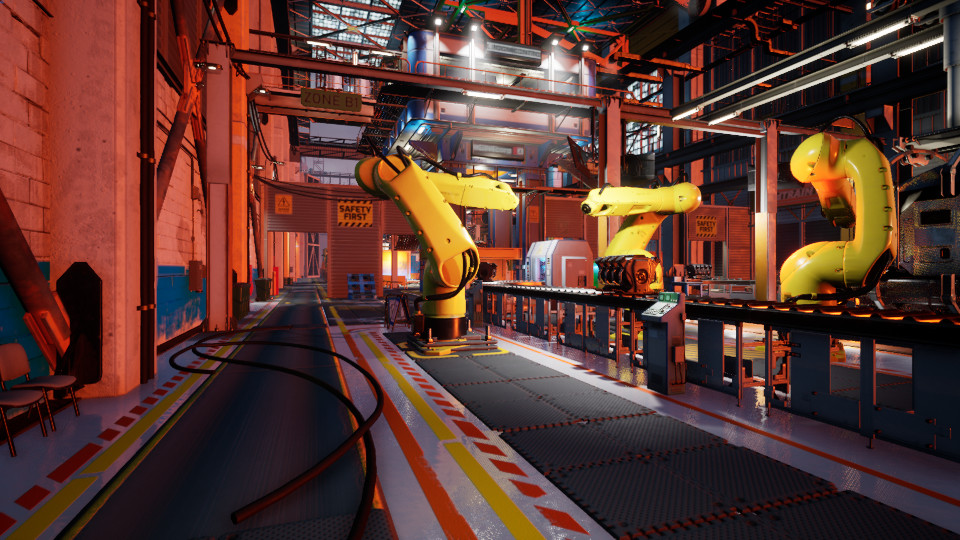}%
\else
\includegraphics[width=\linewidth]{assets/qualitative/I04/a014.jpg}%
\fi\\[-1pt]
{\scriptsize Step 14}\\[-1pt]
{\scriptsize Approach the side aisle\strut}
\end{minipage}\par\vspace{7pt}
\begin{minipage}[t]{0.323\linewidth}
\centering
\ifnum\pdfstrcmp{I04}{S16}=0
\includegraphics[width=\linewidth,trim=0 62 210 56,clip]{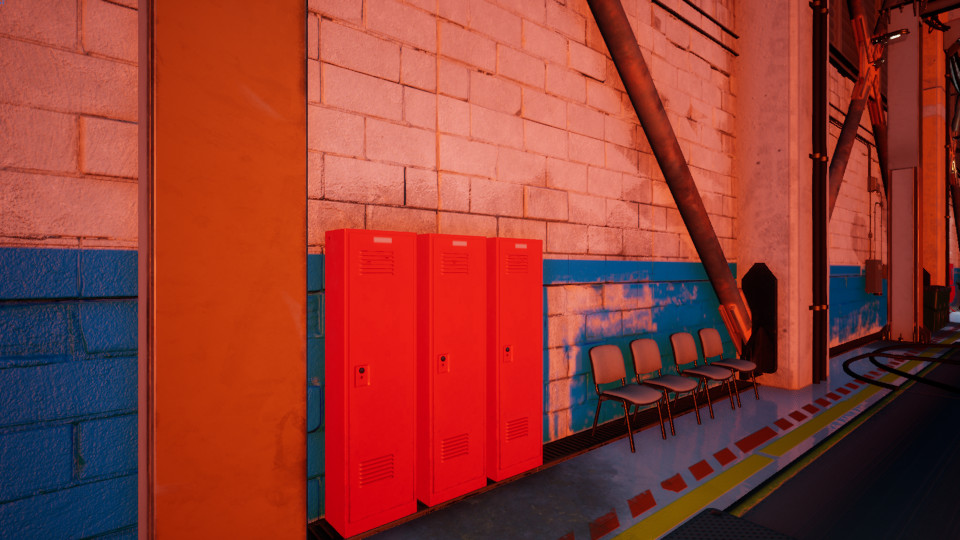}%
\else
\includegraphics[width=\linewidth]{assets/qualitative/I04/a015.jpg}%
\fi\\[-1pt]
{\scriptsize Step 15}\\[-1pt]
{\scriptsize Face the red lockers\strut}
\end{minipage}\hfill
\begin{minipage}[t]{0.323\linewidth}
\centering
\ifnum\pdfstrcmp{I04}{S16}=0
\includegraphics[width=\linewidth,trim=0 62 210 56,clip]{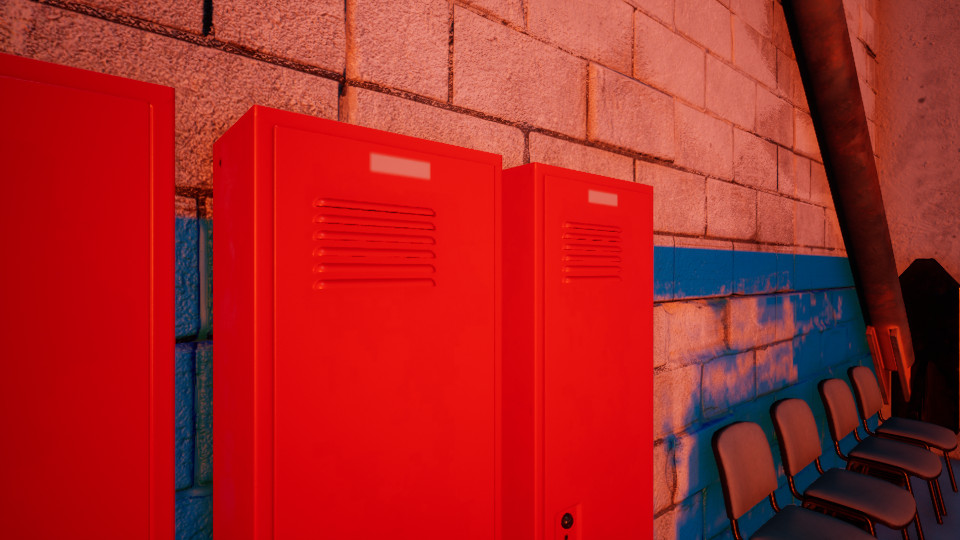}%
\else
\includegraphics[width=\linewidth]{assets/qualitative/I04/a016.jpg}%
\fi\\[-1pt]
{\scriptsize Step 16}\\[-1pt]
{\scriptsize Move toward the lockers\strut}
\end{minipage}\hfill
\begin{minipage}[t]{0.323\linewidth}
\centering
\ifnum\pdfstrcmp{I04}{S16}=0
\includegraphics[width=\linewidth,trim=0 62 210 56,clip]{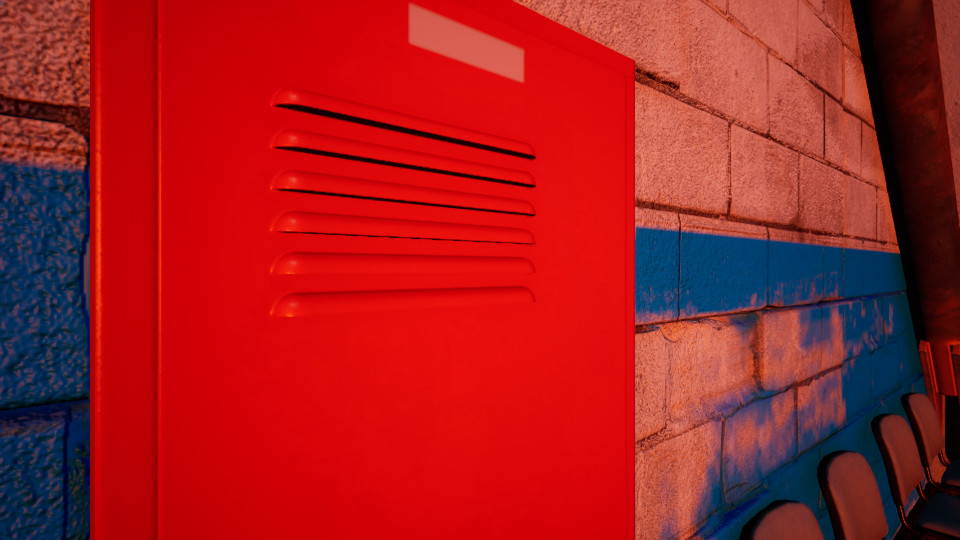}%
\else
\includegraphics[width=\linewidth]{assets/qualitative/I04/a017.jpg}%
\fi\\[-1pt]
{\scriptsize Step 17}\\[-1pt]
{\scriptsize At the locker doors\strut}
\end{minipage}\par\vspace{7pt}
\begin{minipage}[t]{0.323\linewidth}
\centering
\ifnum\pdfstrcmp{I04}{S16}=0
\includegraphics[width=\linewidth,trim=0 62 210 56,clip]{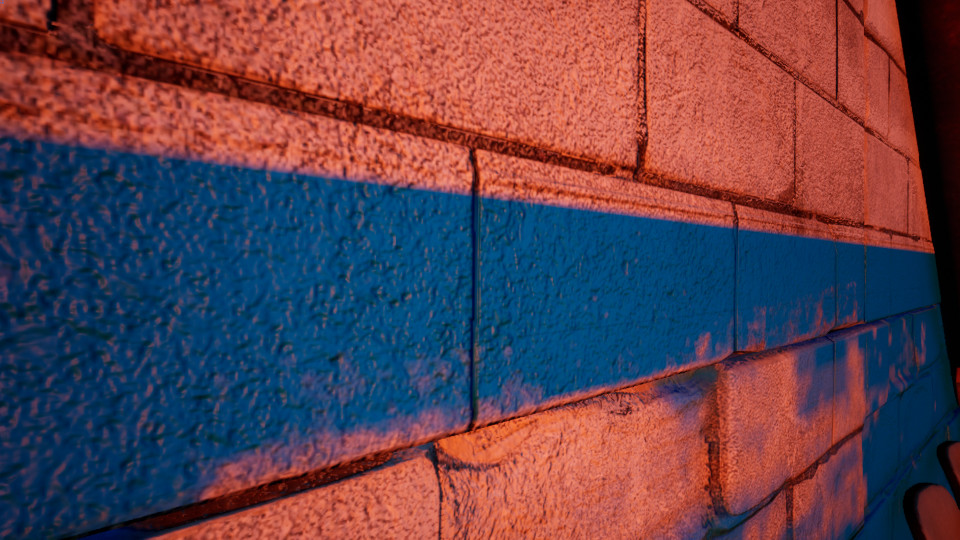}%
\else
\includegraphics[width=\linewidth]{assets/qualitative/I04/a018.jpg}%
\fi\\[-1pt]
{\scriptsize Step 18}\\[-1pt]
{\scriptsize Move through the doors\strut}
\end{minipage}\hfill
\begin{minipage}[t]{0.323\linewidth}
\centering
\ifnum\pdfstrcmp{I04}{S16}=0
\includegraphics[width=\linewidth,trim=0 62 210 56,clip]{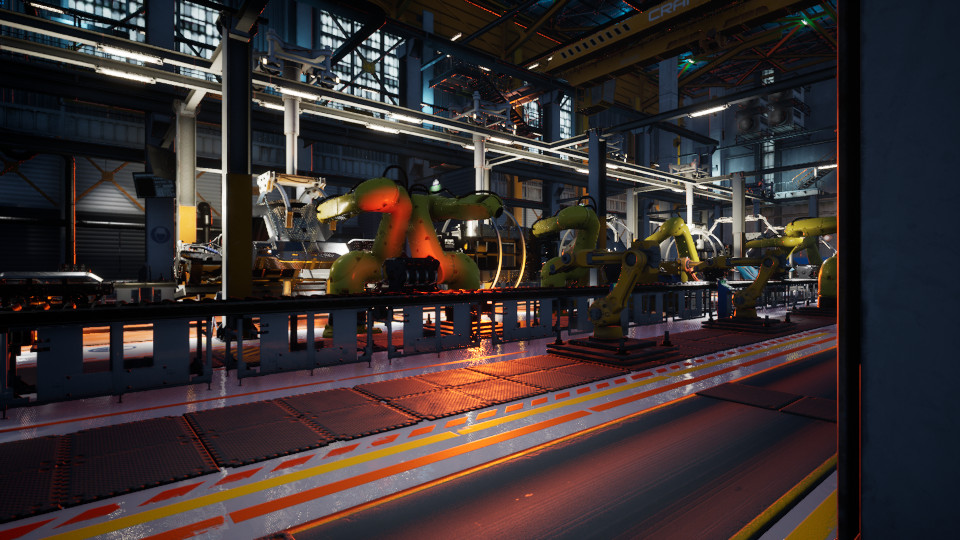}%
\else
\includegraphics[width=\linewidth]{assets/qualitative/I04/a019.jpg}%
\fi\\[-1pt]
{\scriptsize Step 19}\\[-1pt]
{\scriptsize Turn to look back out\strut}
\end{minipage}\hfill
\begin{minipage}[t]{0.323\linewidth}
\centering
\ifnum\pdfstrcmp{I04}{S16}=0
\includegraphics[width=\linewidth,trim=0 62 210 56,clip]{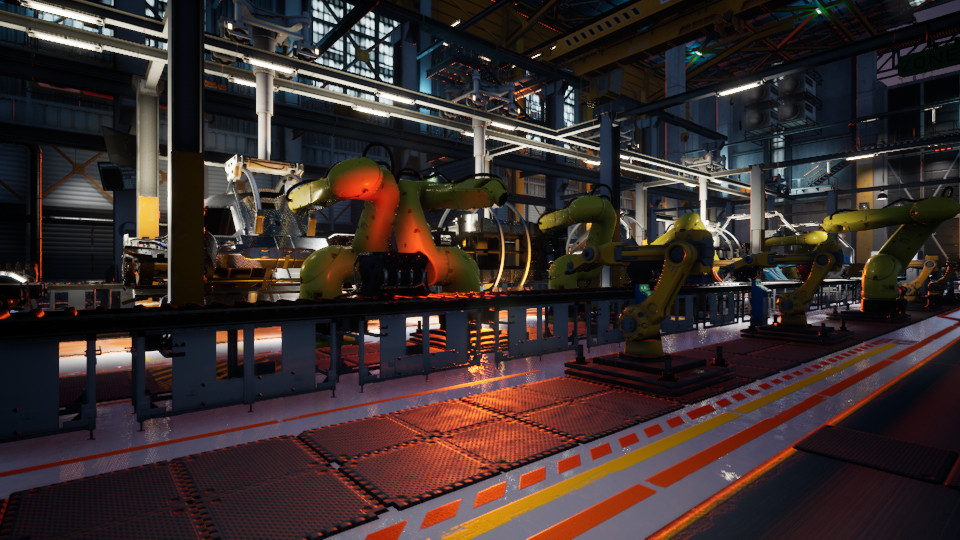}%
\else
\includegraphics[width=\linewidth]{assets/qualitative/I04/a020.jpg}%
\fi\\[-1pt]
{\scriptsize Step 20}\\[-1pt]
{\scriptsize Exit the locker region\strut}
\end{minipage}\par\vspace{7pt}
\begin{minipage}[t]{0.323\linewidth}
\centering
\ifnum\pdfstrcmp{I04}{S16}=0
\includegraphics[width=\linewidth,trim=0 62 210 56,clip]{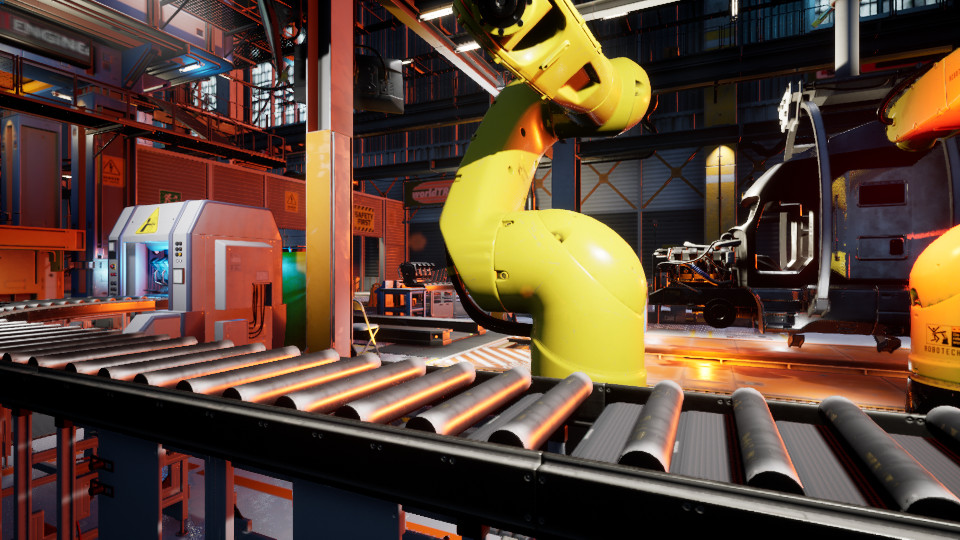}%
\else
\includegraphics[width=\linewidth]{assets/qualitative/I04/a030.jpg}%
\fi\\[-1pt]
{\scriptsize Step 30}\\[-1pt]
{\scriptsize Continue the inspection\strut}
\end{minipage}\hfill
\begin{minipage}[t]{0.323\linewidth}
\centering
\ifnum\pdfstrcmp{I04}{S16}=0
\includegraphics[width=\linewidth,trim=0 62 210 56,clip]{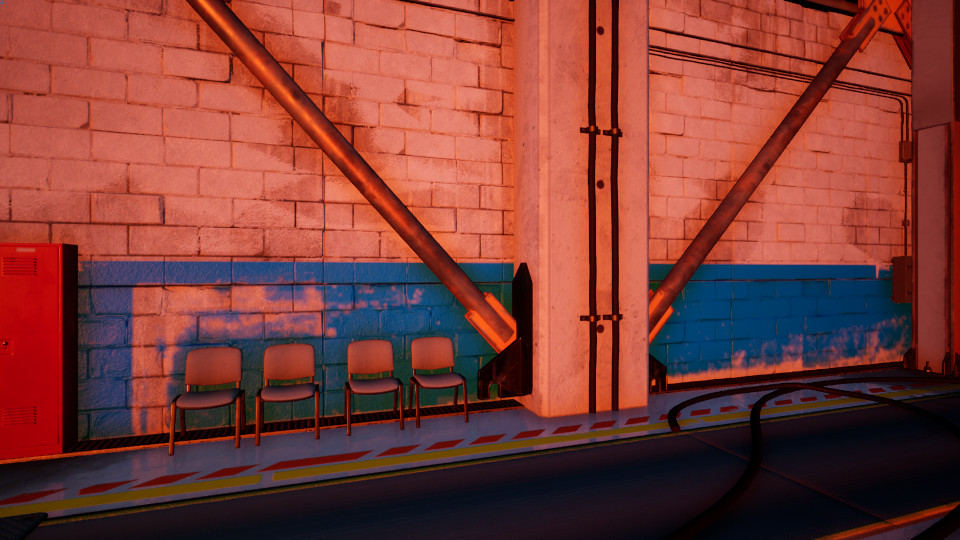}%
\else
\includegraphics[width=\linewidth]{assets/qualitative/I04/a037.jpg}%
\fi\\[-1pt]
{\scriptsize Step 37}\\[-1pt]
{\scriptsize Inspect another boundary\strut}
\end{minipage}\hfill
\begin{minipage}[t]{0.323\linewidth}
\centering
\ifnum\pdfstrcmp{I04}{S16}=0
\includegraphics[width=\linewidth,trim=0 62 210 56,clip]{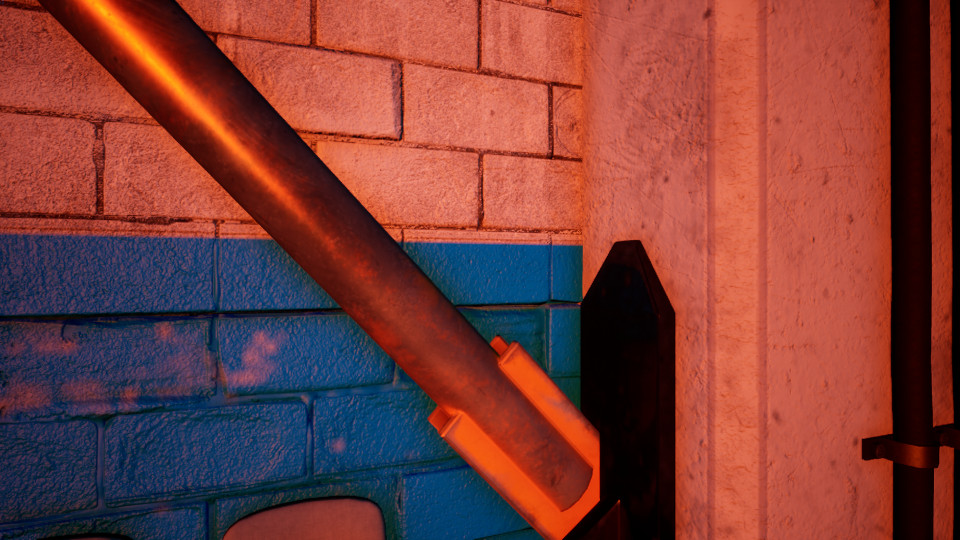}%
\else
\includegraphics[width=\linewidth]{assets/qualitative/I04/a040.jpg}%
\fi\\[-1pt]
{\scriptsize Step 40}\\[-1pt]
{\scriptsize Final observation\strut}
\end{minipage}\par\vspace{7pt}
\caption{Selected observations while inspecting the solid lockers, ordered left to right and top to bottom.}\label{fig:qual-I04}
\end{figure}
\FloatBarrier

\noindent\textbf{Target anomaly.} A visibly solid locker fails to block the player, allowing entry into its body.

\noindent\textbf{Auditor report.} The auditor reports that the red lockers lack collision, citing the approach in Steps 15--17, entry through the doors at Step 18, and the reverse view from inside at Step 19.

\noindent\textbf{Interpretation.} The report is correct: the entry and reverse views show traversal through a visibly solid surface, matching the target collision failure. The action log also records a one-metre movement at Step 18 without a collision block.

\clearpage
\phantomsection\label{app:qual-A20}
\noindent\textbf{Success: A Modern Appliance In An Ancient Market}\par\nobreak\smallskip
\noindent Semantic consistency $\mid$ 40-action budget

\begin{figure}[!ht]
\centering
\begin{minipage}[t]{0.323\linewidth}
\centering
\ifnum\pdfstrcmp{A20}{S16}=0
\includegraphics[width=\linewidth,trim=0 62 210 56,clip]{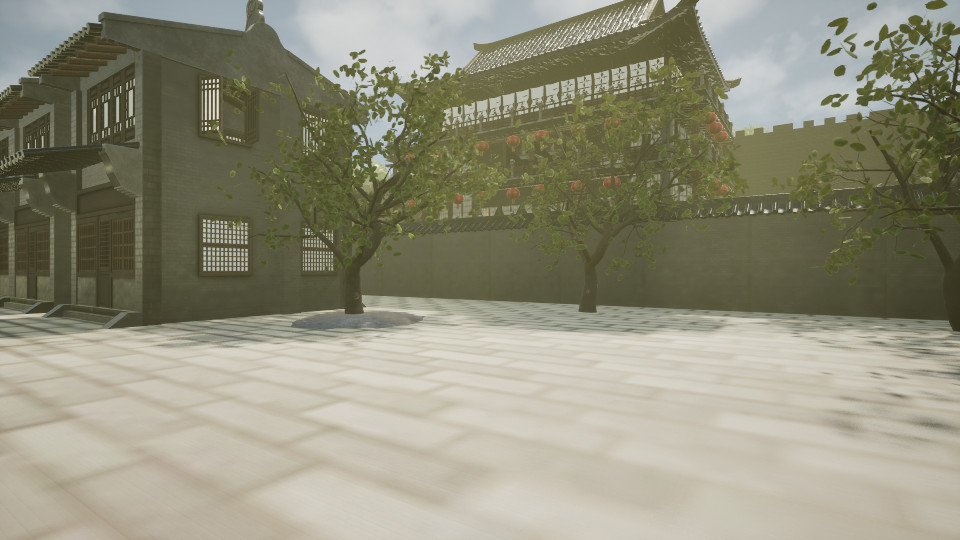}%
\else
\includegraphics[width=\linewidth]{assets/qualitative/A20/a000.jpg}%
\fi\\[-1pt]
{\scriptsize Start}\\[-1pt]
{\scriptsize Initial street view\strut}
\end{minipage}\hfill
\begin{minipage}[t]{0.323\linewidth}
\centering
\ifnum\pdfstrcmp{A20}{S16}=0
\includegraphics[width=\linewidth,trim=0 62 210 56,clip]{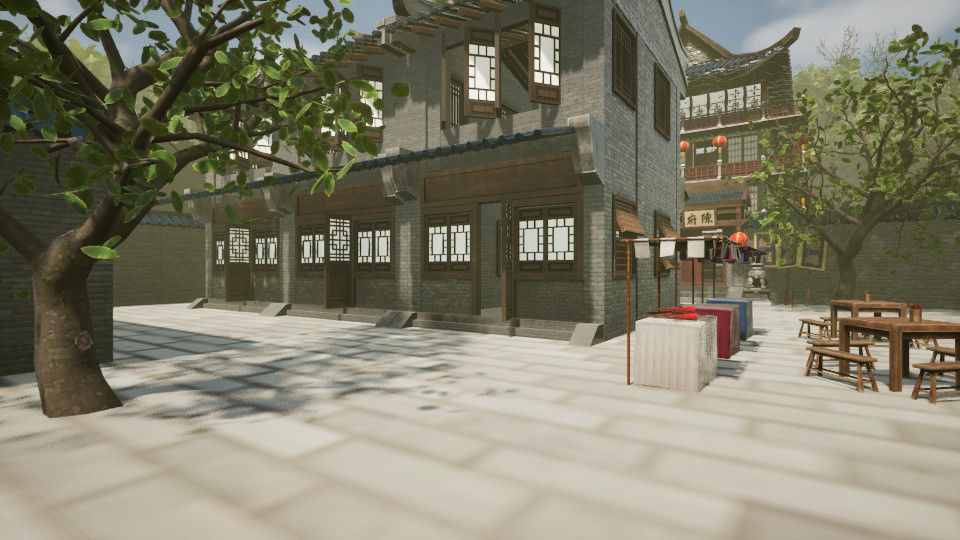}%
\else
\includegraphics[width=\linewidth]{assets/qualitative/A20/a002.jpg}%
\fi\\[-1pt]
{\scriptsize Step 2}\\[-1pt]
{\scriptsize Survey market stalls\strut}
\end{minipage}\hfill
\begin{minipage}[t]{0.323\linewidth}
\centering
\ifnum\pdfstrcmp{A20}{S16}=0
\includegraphics[width=\linewidth,trim=0 62 210 56,clip]{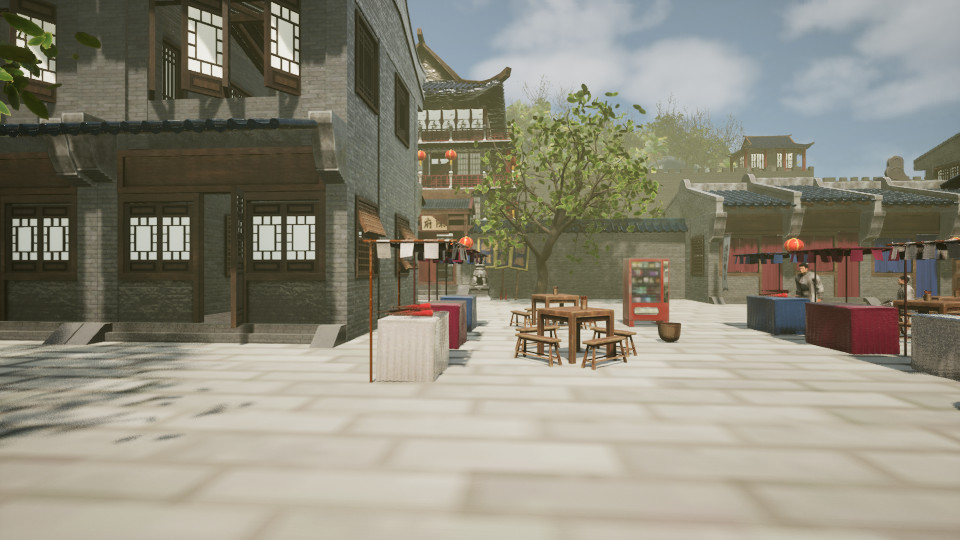}%
\else
\includegraphics[width=\linewidth]{assets/qualitative/A20/a003.jpg}%
\fi\\[-1pt]
{\scriptsize Step 3}\\[-1pt]
{\scriptsize Machine enters view\strut}
\end{minipage}\par\vspace{7pt}
\begin{minipage}[t]{0.323\linewidth}
\centering
\ifnum\pdfstrcmp{A20}{S16}=0
\includegraphics[width=\linewidth,trim=0 62 210 56,clip]{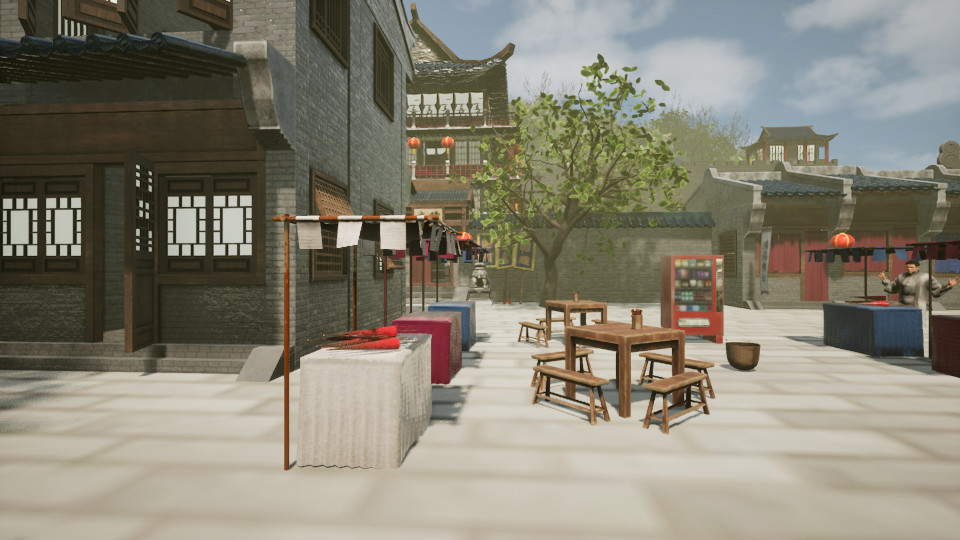}%
\else
\includegraphics[width=\linewidth]{assets/qualitative/A20/a004.jpg}%
\fi\\[-1pt]
{\scriptsize Step 4}\\[-1pt]
{\scriptsize Approach the machine\strut}
\end{minipage}\hfill
\begin{minipage}[t]{0.323\linewidth}
\centering
\ifnum\pdfstrcmp{A20}{S16}=0
\includegraphics[width=\linewidth,trim=0 62 210 56,clip]{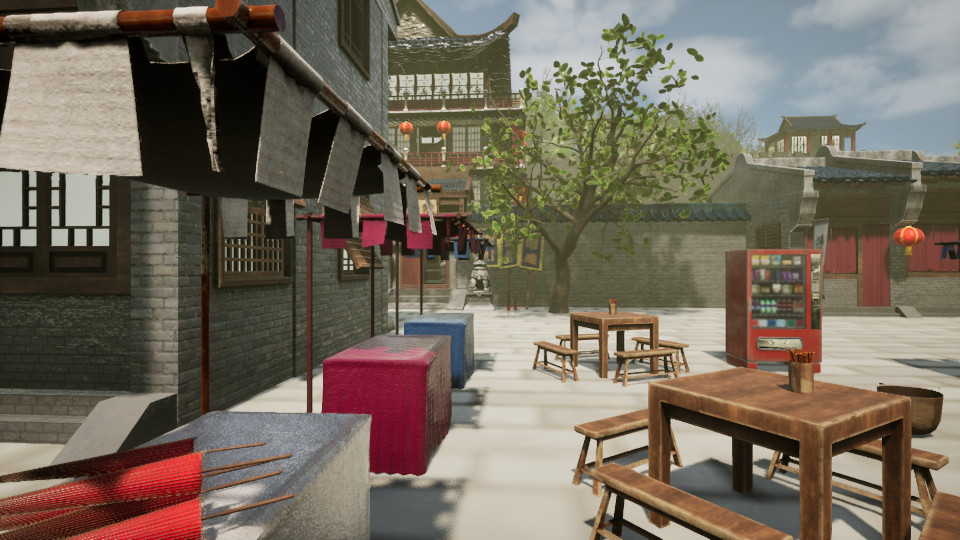}%
\else
\includegraphics[width=\linewidth]{assets/qualitative/A20/a005.jpg}%
\fi\\[-1pt]
{\scriptsize Step 5}\\[-1pt]
{\scriptsize Inspect the red appliance\strut}
\end{minipage}\hfill
\begin{minipage}[t]{0.323\linewidth}
\centering
\ifnum\pdfstrcmp{A20}{S16}=0
\includegraphics[width=\linewidth,trim=0 62 210 56,clip]{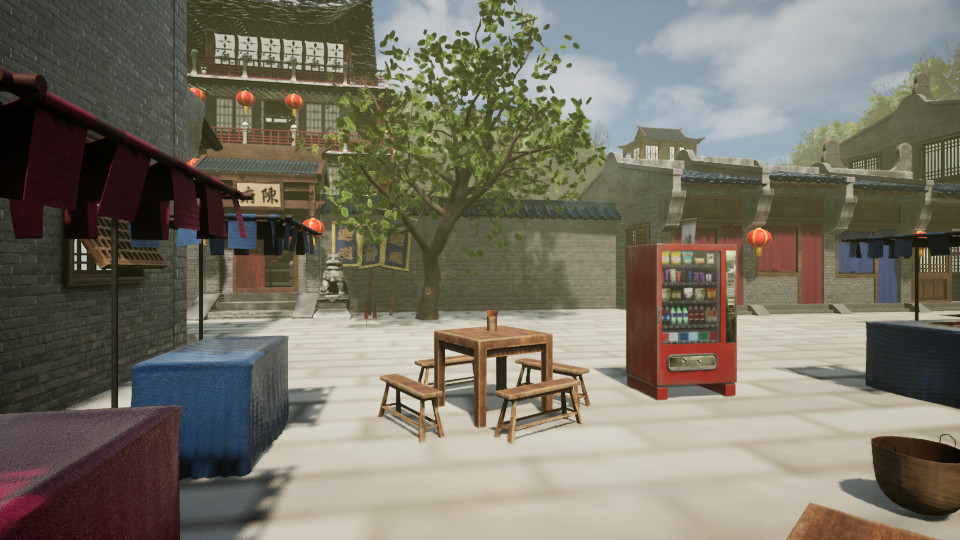}%
\else
\includegraphics[width=\linewidth]{assets/qualitative/A20/a007.jpg}%
\fi\\[-1pt]
{\scriptsize Step 7}\\[-1pt]
{\scriptsize View the market context\strut}
\end{minipage}\par\vspace{7pt}
\begin{minipage}[t]{0.323\linewidth}
\centering
\ifnum\pdfstrcmp{A20}{S16}=0
\includegraphics[width=\linewidth,trim=0 62 210 56,clip]{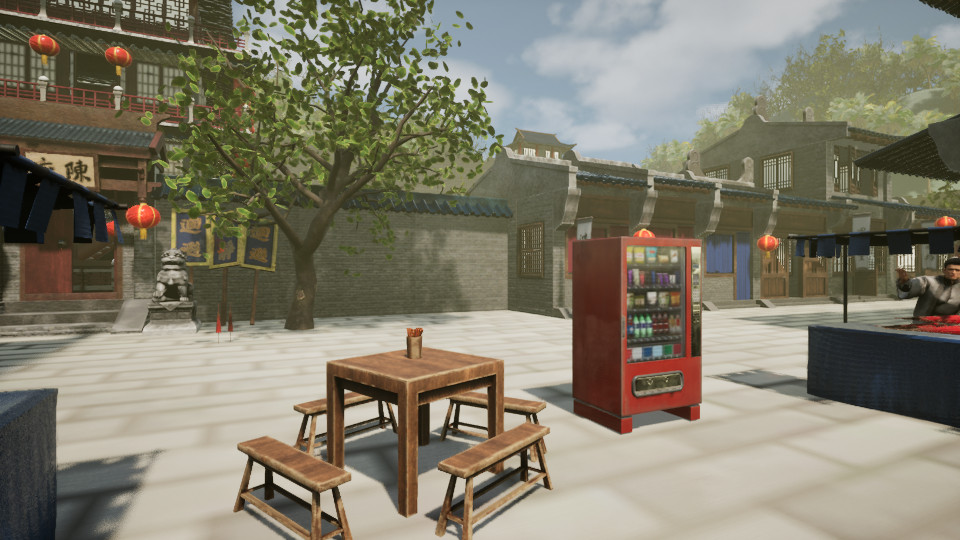}%
\else
\includegraphics[width=\linewidth]{assets/qualitative/A20/a009.jpg}%
\fi\\[-1pt]
{\scriptsize Step 9}\\[-1pt]
{\scriptsize Inspect the machine front\strut}
\end{minipage}\hfill
\begin{minipage}[t]{0.323\linewidth}
\centering
\ifnum\pdfstrcmp{A20}{S16}=0
\includegraphics[width=\linewidth,trim=0 62 210 56,clip]{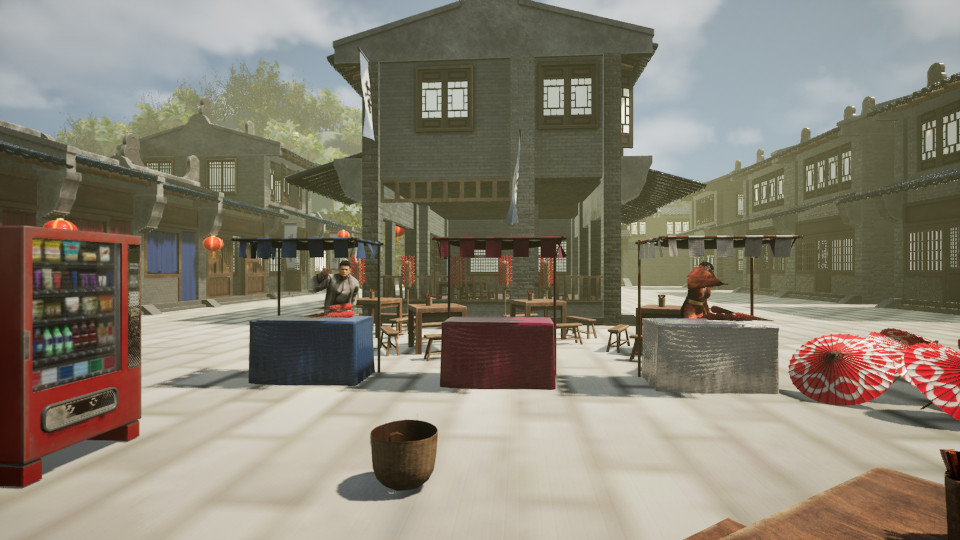}%
\else
\includegraphics[width=\linewidth]{assets/qualitative/A20/a010.jpg}%
\fi\\[-1pt]
{\scriptsize Step 10}\\[-1pt]
{\scriptsize Continue along the stalls\strut}
\end{minipage}\hfill
\begin{minipage}[t]{0.323\linewidth}
\centering
\ifnum\pdfstrcmp{A20}{S16}=0
\includegraphics[width=\linewidth,trim=0 62 210 56,clip]{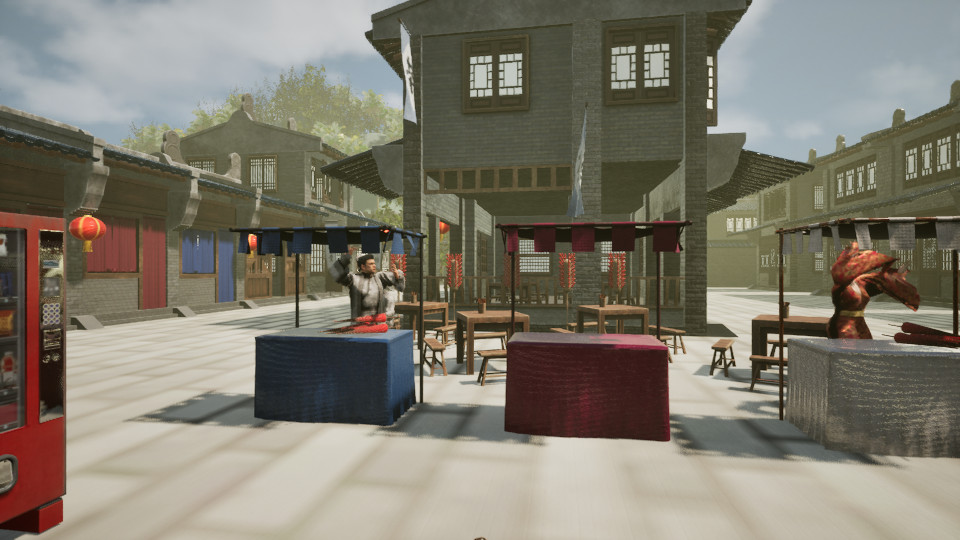}%
\else
\includegraphics[width=\linewidth]{assets/qualitative/A20/a024.jpg}%
\fi\\[-1pt]
{\scriptsize Step 24}\\[-1pt]
{\scriptsize Resume exploration\strut}
\end{minipage}\par\vspace{7pt}
\begin{minipage}[t]{0.323\linewidth}
\centering
\ifnum\pdfstrcmp{A20}{S16}=0
\includegraphics[width=\linewidth,trim=0 62 210 56,clip]{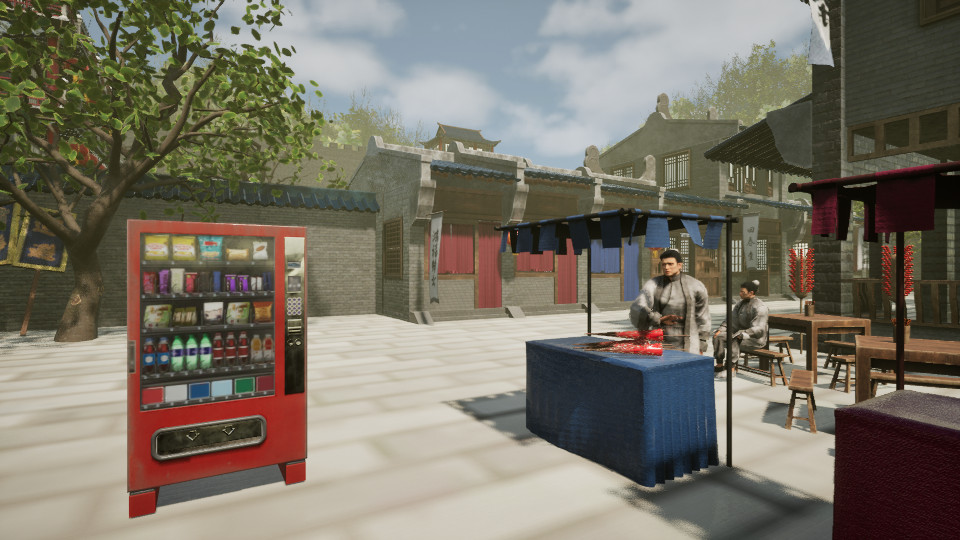}%
\else
\includegraphics[width=\linewidth]{assets/qualitative/A20/a029.jpg}%
\fi\\[-1pt]
{\scriptsize Step 29}\\[-1pt]
{\scriptsize See the appliance again\strut}
\end{minipage}\hfill
\begin{minipage}[t]{0.323\linewidth}
\centering
\ifnum\pdfstrcmp{A20}{S16}=0
\includegraphics[width=\linewidth,trim=0 62 210 56,clip]{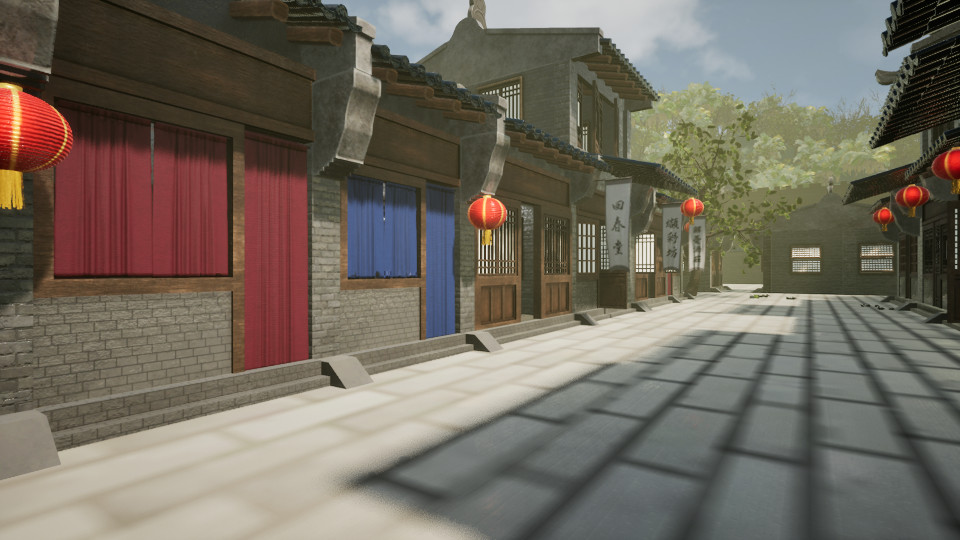}%
\else
\includegraphics[width=\linewidth]{assets/qualitative/A20/a035.jpg}%
\fi\\[-1pt]
{\scriptsize Step 35}\\[-1pt]
{\scriptsize Inspect nearby buildings\strut}
\end{minipage}\hfill
\begin{minipage}[t]{0.323\linewidth}
\centering
\ifnum\pdfstrcmp{A20}{S16}=0
\includegraphics[width=\linewidth,trim=0 62 210 56,clip]{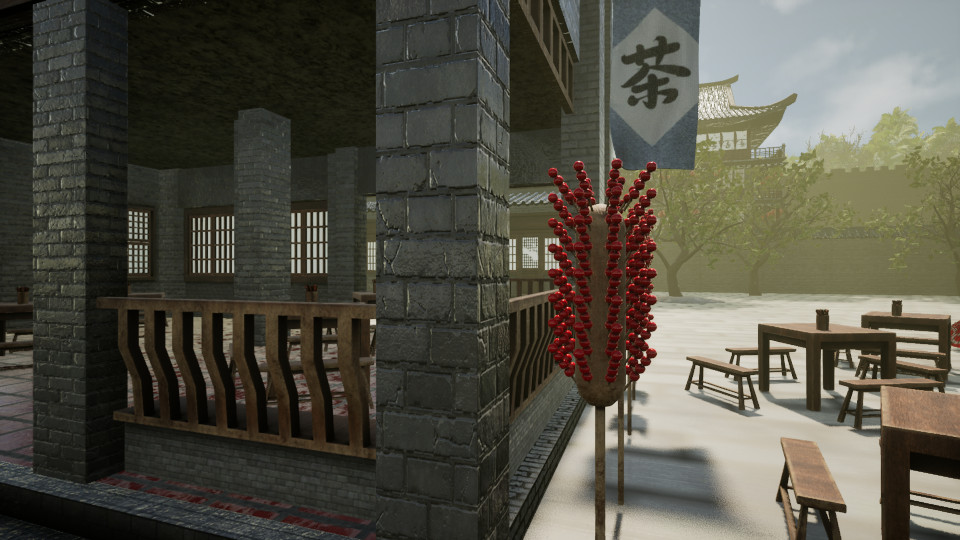}%
\else
\includegraphics[width=\linewidth]{assets/qualitative/A20/a040.jpg}%
\fi\\[-1pt]
{\scriptsize Step 40}\\[-1pt]
{\scriptsize Final observation\strut}
\end{minipage}\par\vspace{7pt}
\caption{Selected observations while inspecting the market appliance, ordered left to right and top to bottom.}\label{fig:qual-A20}
\end{figure}
\FloatBarrier

\noindent\textbf{Target anomaly.} A modern vending machine appears among the stalls of an ancient Chinese market.

\noindent\textbf{Auditor report.} The auditor identifies a modern beverage vending machine that conflicts with the historical setting. Its evidence includes Steps 3--5, 7, and 9, which show the appliance and the surrounding market.

\noindent\textbf{Interpretation.} The report is correct: the modern appliance is inconsistent with the ancient market setting, matching the annotated target. The cited views establish both the object's identity and its scene context.

\clearpage
\phantomsection\label{app:qual-H10}
\noindent\textbf{Failure: Missing Before-and-After Evidence}\par\nobreak\smallskip
\noindent Temporal consistency $\mid$ 40-action budget

\begin{figure}[!ht]
\centering
\begin{minipage}[t]{0.323\linewidth}
\centering
\ifnum\pdfstrcmp{H10}{S16}=0
\includegraphics[width=\linewidth,trim=0 62 210 56,clip]{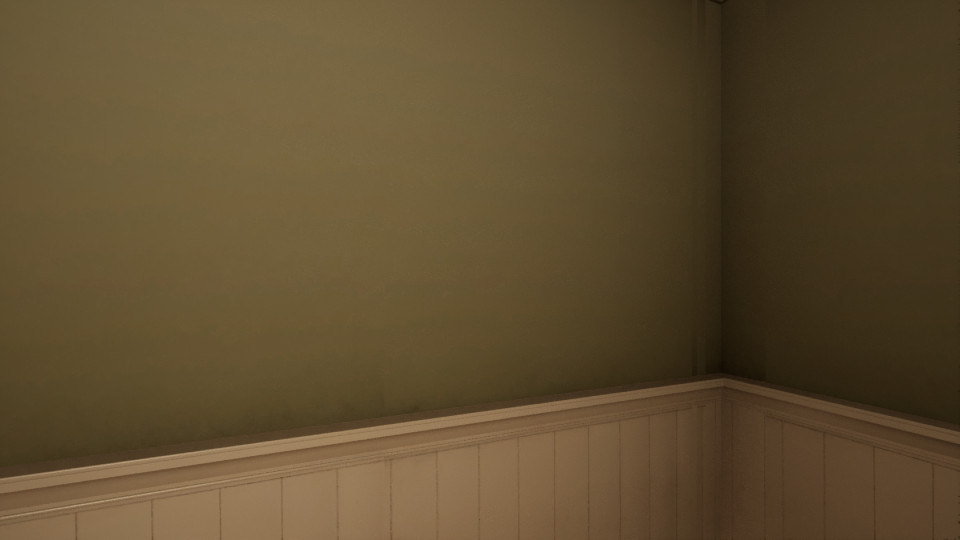}%
\else
\includegraphics[width=\linewidth]{assets/qualitative/H10/a000.jpg}%
\fi\\[-1pt]
{\scriptsize Start}\\[-1pt]
{\scriptsize Initial view\strut}
\end{minipage}\hfill
\begin{minipage}[t]{0.323\linewidth}
\centering
\ifnum\pdfstrcmp{H10}{S16}=0
\includegraphics[width=\linewidth,trim=0 62 210 56,clip]{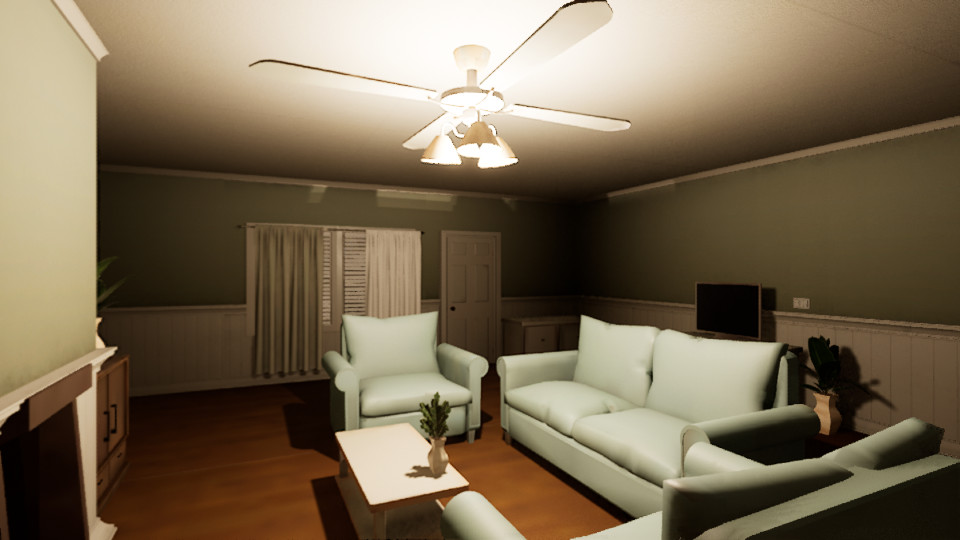}%
\else
\includegraphics[width=\linewidth]{assets/qualitative/H10/a001.jpg}%
\fi\\[-1pt]
{\scriptsize Step 1}\\[-1pt]
{\scriptsize Survey the living room\strut}
\end{minipage}\hfill
\begin{minipage}[t]{0.323\linewidth}
\centering
\ifnum\pdfstrcmp{H10}{S16}=0
\includegraphics[width=\linewidth,trim=0 62 210 56,clip]{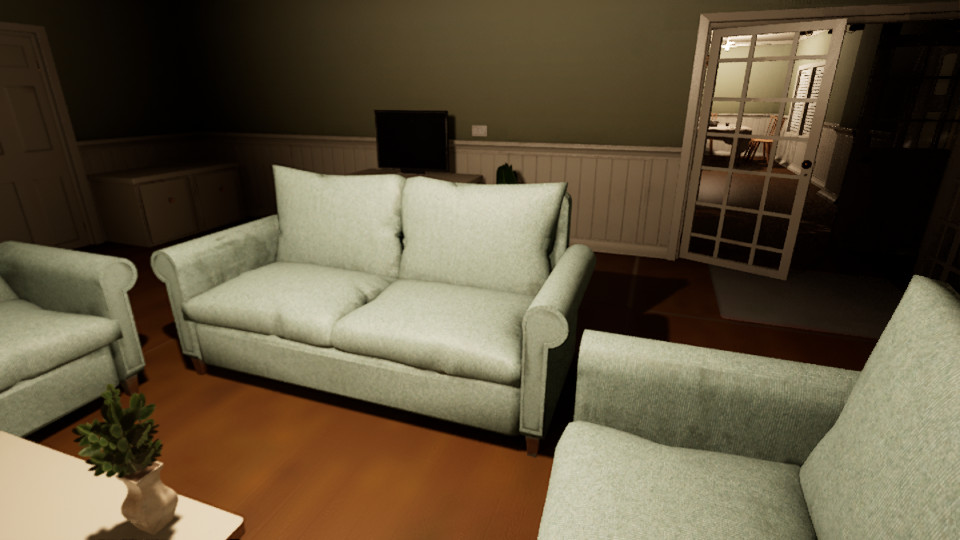}%
\else
\includegraphics[width=\linewidth]{assets/qualitative/H10/a005.jpg}%
\fi\\[-1pt]
{\scriptsize Step 5}\\[-1pt]
{\scriptsize Inspect nearby furniture\strut}
\end{minipage}\par\vspace{7pt}
\begin{minipage}[t]{0.323\linewidth}
\centering
\ifnum\pdfstrcmp{H10}{S16}=0
\includegraphics[width=\linewidth,trim=0 62 210 56,clip]{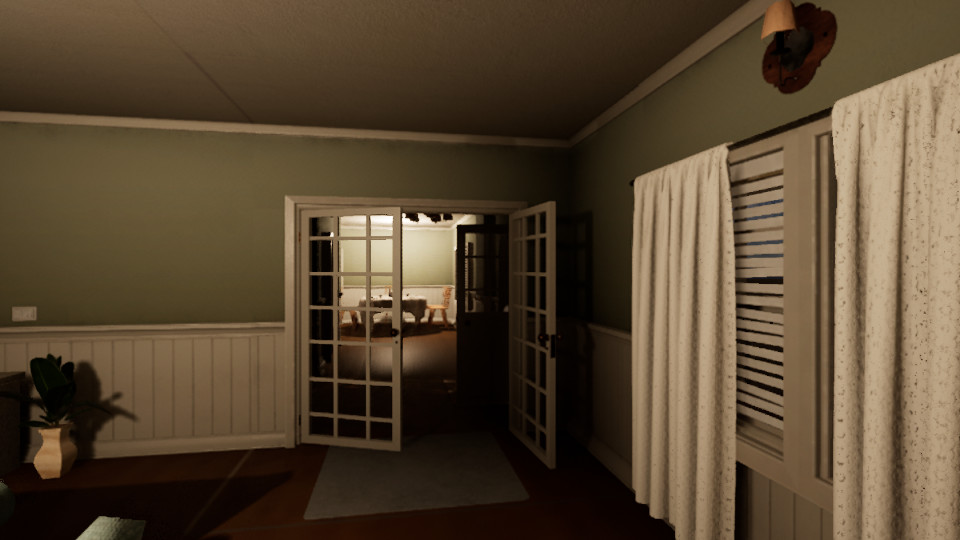}%
\else
\includegraphics[width=\linewidth]{assets/qualitative/H10/a009.jpg}%
\fi\\[-1pt]
{\scriptsize Step 9}\\[-1pt]
{\scriptsize Move toward the dining area\strut}
\end{minipage}\hfill
\begin{minipage}[t]{0.323\linewidth}
\centering
\ifnum\pdfstrcmp{H10}{S16}=0
\includegraphics[width=\linewidth,trim=0 62 210 56,clip]{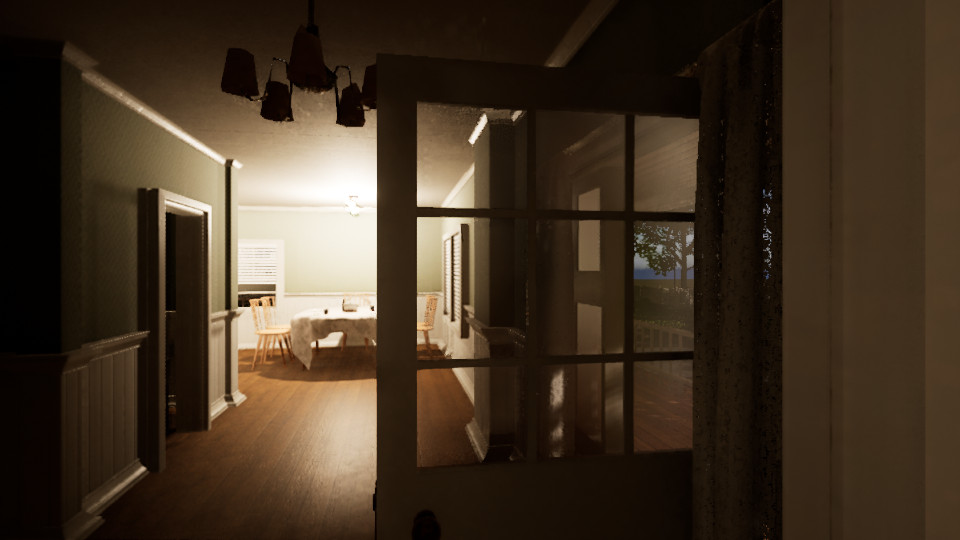}%
\else
\includegraphics[width=\linewidth]{assets/qualitative/H10/a012.jpg}%
\fi\\[-1pt]
{\scriptsize Step 12}\\[-1pt]
{\scriptsize Reach the adjoining room\strut}
\end{minipage}\hfill
\begin{minipage}[t]{0.323\linewidth}
\centering
\ifnum\pdfstrcmp{H10}{S16}=0
\includegraphics[width=\linewidth,trim=0 62 210 56,clip]{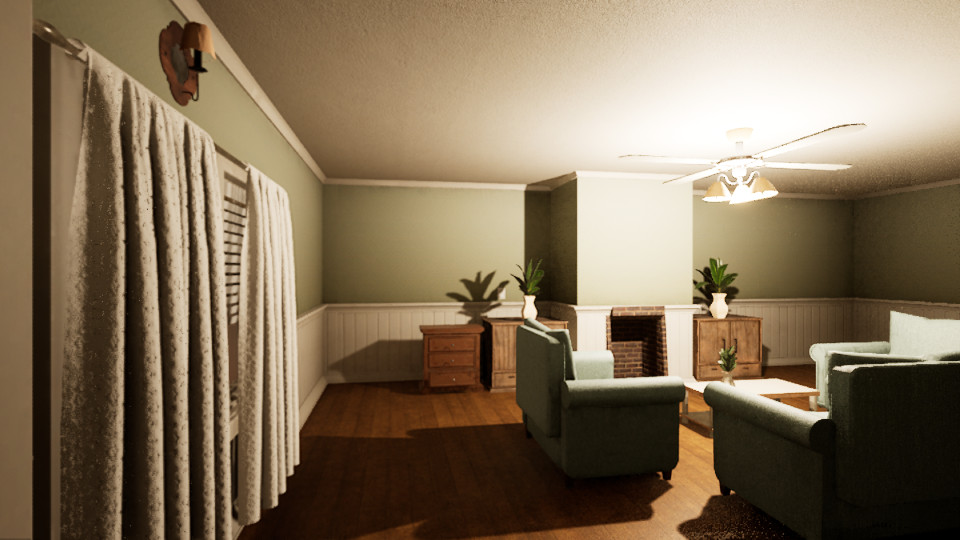}%
\else
\includegraphics[width=\linewidth]{assets/qualitative/H10/a013.jpg}%
\fi\\[-1pt]
{\scriptsize Step 13}\\[-1pt]
{\scriptsize Look back toward living room\strut}
\end{minipage}\par\vspace{7pt}
\begin{minipage}[t]{0.323\linewidth}
\centering
\ifnum\pdfstrcmp{H10}{S16}=0
\includegraphics[width=\linewidth,trim=0 62 210 56,clip]{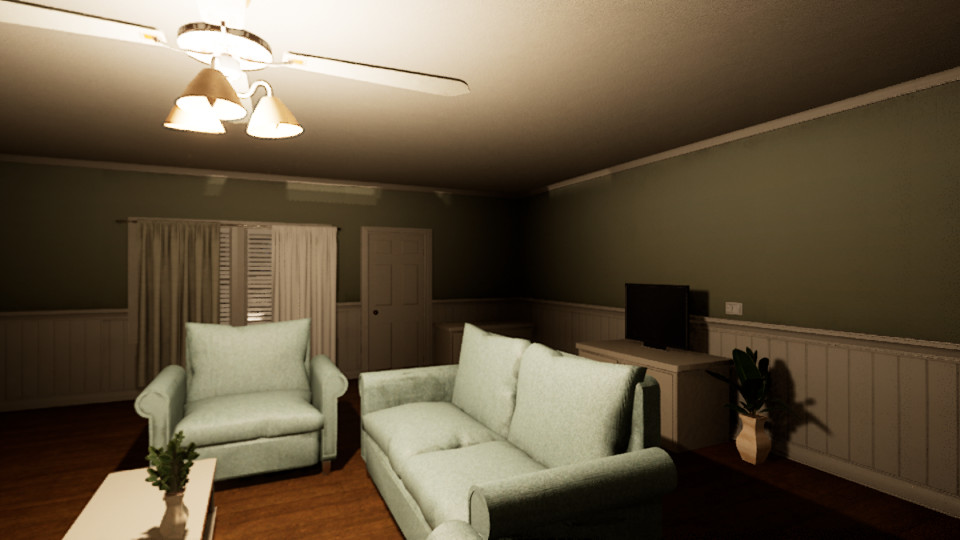}%
\else
\includegraphics[width=\linewidth]{assets/qualitative/H10/a016.jpg}%
\fi\\[-1pt]
{\scriptsize Step 16}\\[-1pt]
{\scriptsize Revisit the living room\strut}
\end{minipage}\hfill
\begin{minipage}[t]{0.323\linewidth}
\centering
\ifnum\pdfstrcmp{H10}{S16}=0
\includegraphics[width=\linewidth,trim=0 62 210 56,clip]{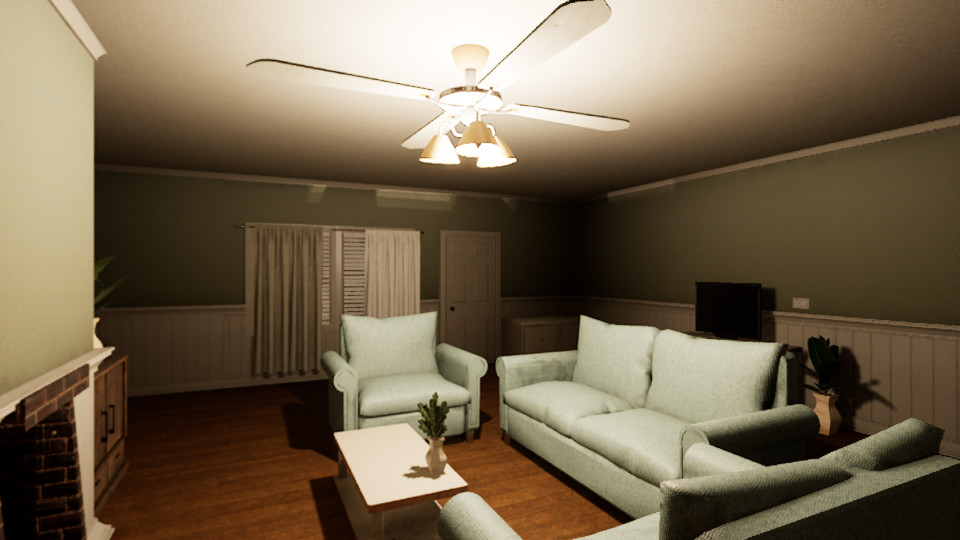}%
\else
\includegraphics[width=\linewidth]{assets/qualitative/H10/a023.jpg}%
\fi\\[-1pt]
{\scriptsize Step 23}\\[-1pt]
{\scriptsize Compare with the earlier view\strut}
\end{minipage}\hfill
\begin{minipage}[t]{0.323\linewidth}
\centering
\ifnum\pdfstrcmp{H10}{S16}=0
\includegraphics[width=\linewidth,trim=0 62 210 56,clip]{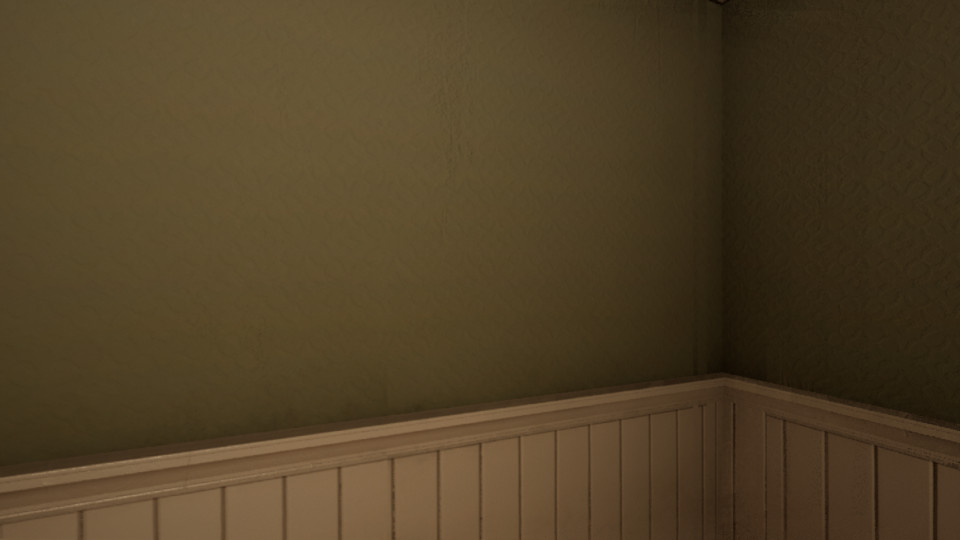}%
\else
\includegraphics[width=\linewidth]{assets/qualitative/H10/a024.jpg}%
\fi\\[-1pt]
{\scriptsize Step 24}\\[-1pt]
{\scriptsize Return to the initial view\strut}
\end{minipage}\par\vspace{7pt}
\begin{minipage}[t]{0.323\linewidth}
\centering
\ifnum\pdfstrcmp{H10}{S16}=0
\includegraphics[width=\linewidth,trim=0 62 210 56,clip]{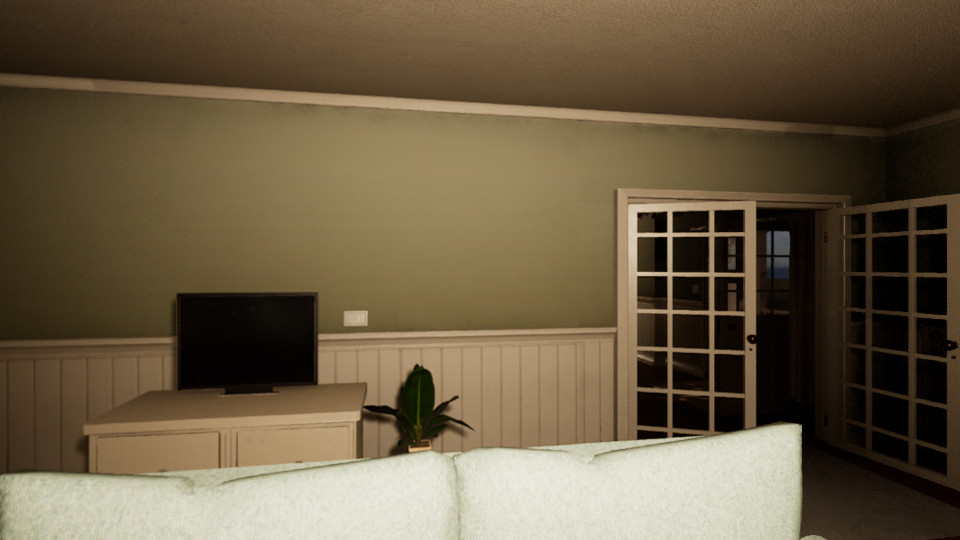}%
\else
\includegraphics[width=\linewidth]{assets/qualitative/H10/a030.jpg}%
\fi\\[-1pt]
{\scriptsize Step 30}\\[-1pt]
{\scriptsize Inspect the plant area again\strut}
\end{minipage}\hfill
\begin{minipage}[t]{0.323\linewidth}
\centering
\ifnum\pdfstrcmp{H10}{S16}=0
\includegraphics[width=\linewidth,trim=0 62 210 56,clip]{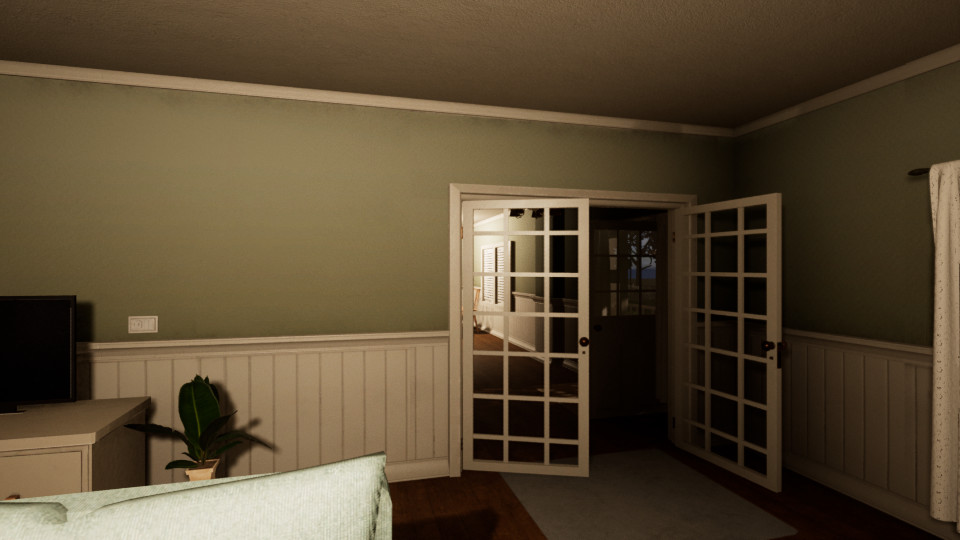}%
\else
\includegraphics[width=\linewidth]{assets/qualitative/H10/a034.jpg}%
\fi\\[-1pt]
{\scriptsize Step 34}\\[-1pt]
{\scriptsize Attempt another approach\strut}
\end{minipage}\hfill
\begin{minipage}[t]{0.323\linewidth}
\centering
\ifnum\pdfstrcmp{H10}{S16}=0
\includegraphics[width=\linewidth,trim=0 62 210 56,clip]{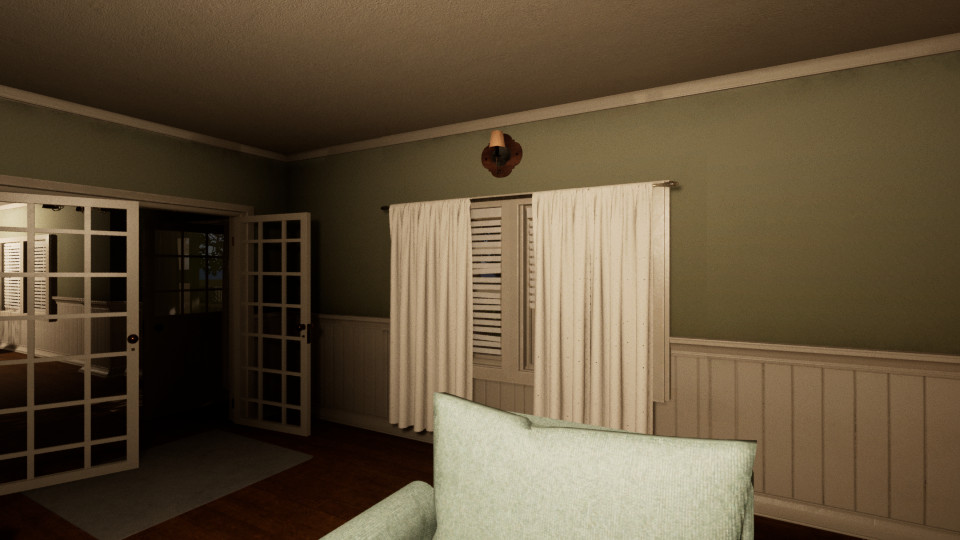}%
\else
\includegraphics[width=\linewidth]{assets/qualitative/H10/a040.jpg}%
\fi\\[-1pt]
{\scriptsize Step 40}\\[-1pt]
{\scriptsize Final observation\strut}
\end{minipage}\par\vspace{7pt}
\caption{Selected observations while inspecting the disappearing plant, ordered left to right and top to bottom.}\label{fig:qual-H10}
\end{figure}
\FloatBarrier

\noindent\textbf{Target anomaly.} A floor plant disappears after the auditor approaches it and walks away; the rubric specifies moving more than 3.1 metres from the target.

\noindent\textbf{Auditor report.} The auditor reports no anomalies, comparing Step 1 with Step 23 and Start with Step 24 after leaving for the dining area and returning. It concludes that the objects remained consistent.

\noindent\textbf{Interpretation.} The auditor revisits the room but fails to obtain a clear before-and-after comparison of the target plant. Its comparisons of broader room views do not establish the plant's disappearance, and it reports no anomalies. This case illustrates a failure to gather the observations needed to identify a temporal anomaly.

\clearpage
\phantomsection\label{app:qual-S16}
\noindent\textbf{Failure: Mistaking Advertisement Cycling For An Anomaly}\par\nobreak\smallskip
\noindent Temporal consistency $\mid$ 40-action budget

\begin{figure}[!ht]
\centering
\begin{minipage}[t]{0.323\linewidth}
\centering
\ifnum\pdfstrcmp{S16}{S16}=0
\includegraphics[width=\linewidth,trim=0 62 210 56,clip]{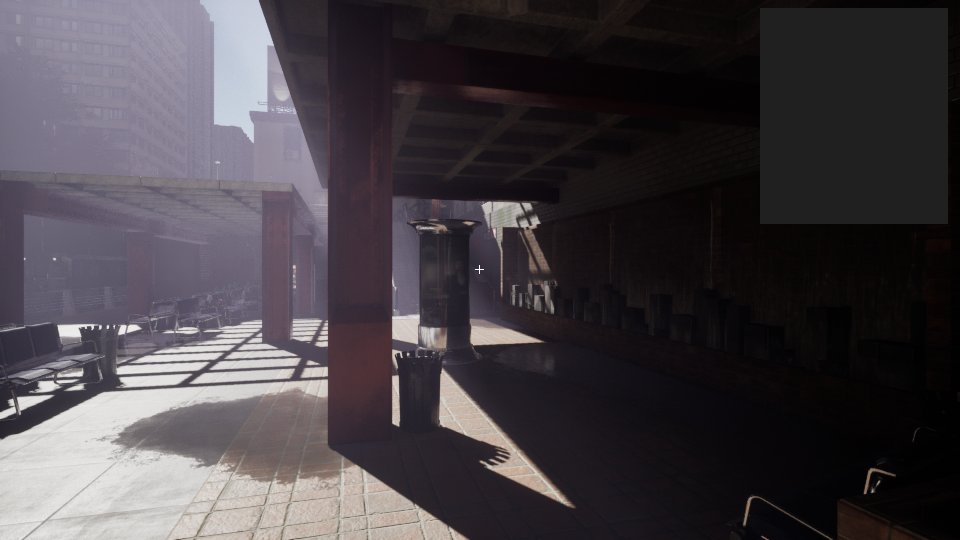}%
\else
\includegraphics[width=\linewidth]{assets/qualitative/S16/a000.jpg}%
\fi\\[-1pt]
{\scriptsize Start}\\[-1pt]
{\scriptsize Initial plaza view\strut}
\end{minipage}\hfill
\begin{minipage}[t]{0.323\linewidth}
\centering
\ifnum\pdfstrcmp{S16}{S16}=0
\includegraphics[width=\linewidth,trim=0 62 210 56,clip]{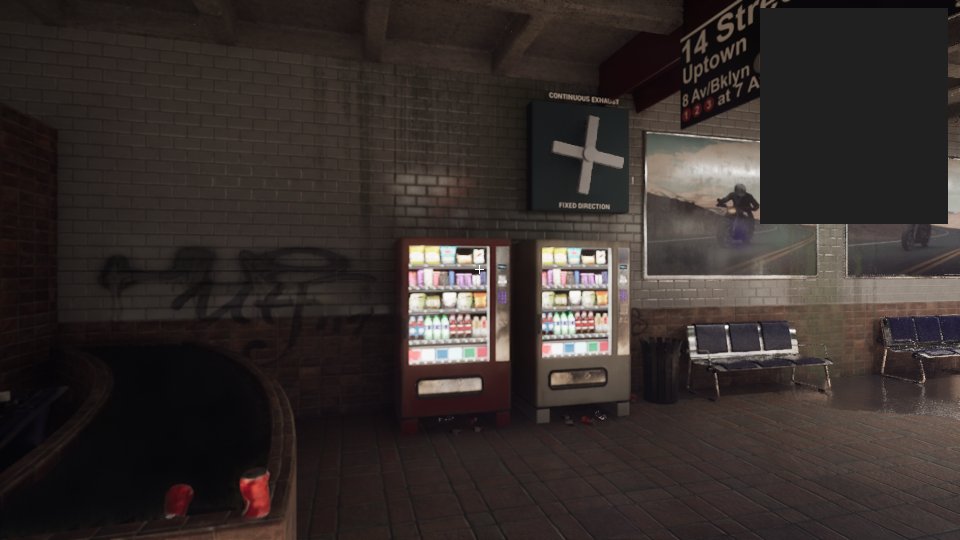}%
\else
\includegraphics[width=\linewidth]{assets/qualitative/S16/a004.jpg}%
\fi\\[-1pt]
{\scriptsize Step 4}\\[-1pt]
{\scriptsize Posters show motorcycles\strut}
\end{minipage}\hfill
\begin{minipage}[t]{0.323\linewidth}
\centering
\ifnum\pdfstrcmp{S16}{S16}=0
\includegraphics[width=\linewidth,trim=0 62 210 56,clip]{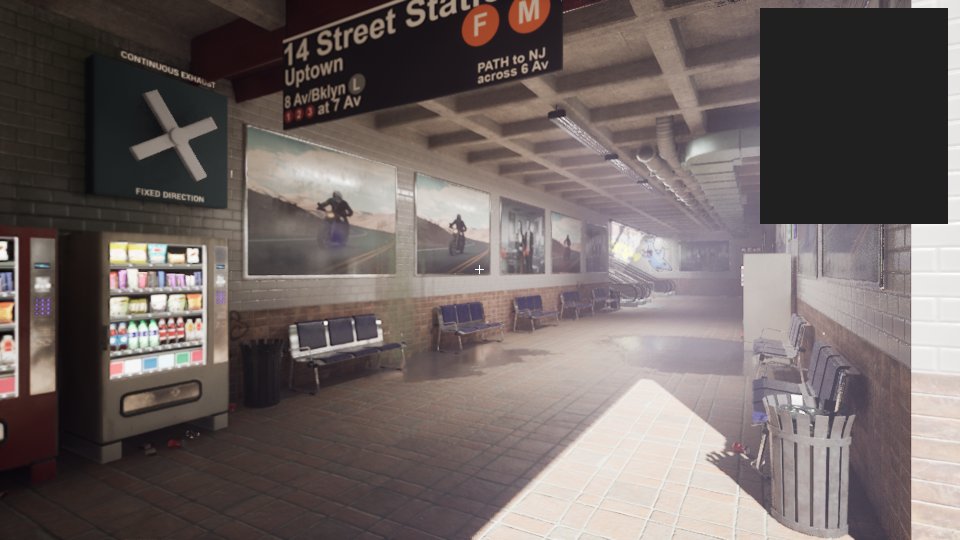}%
\else
\includegraphics[width=\linewidth]{assets/qualitative/S16/a005.jpg}%
\fi\\[-1pt]
{\scriptsize Step 5}\\[-1pt]
{\scriptsize Inspect the early posters\strut}
\end{minipage}\par\vspace{7pt}
\begin{minipage}[t]{0.323\linewidth}
\centering
\ifnum\pdfstrcmp{S16}{S16}=0
\includegraphics[width=\linewidth,trim=0 62 210 56,clip]{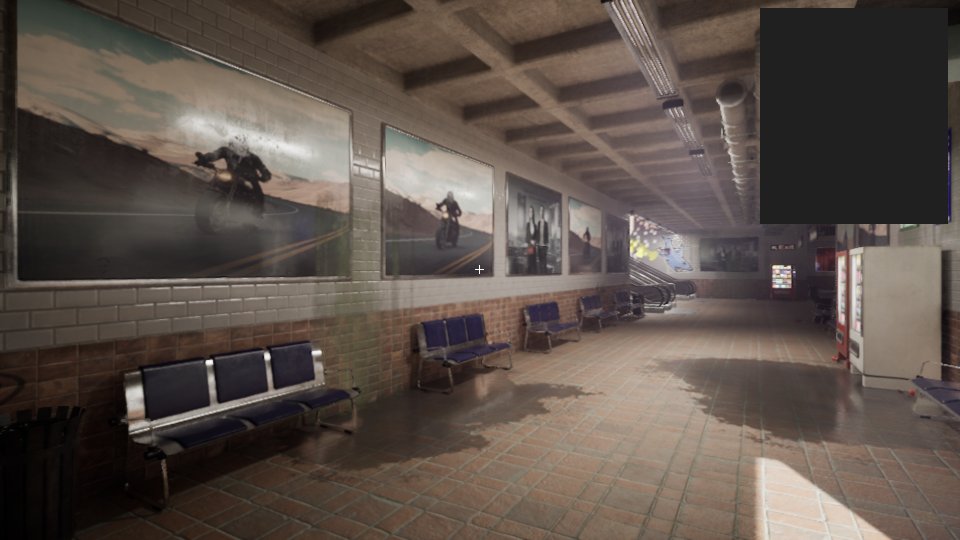}%
\else
\includegraphics[width=\linewidth]{assets/qualitative/S16/a008.jpg}%
\fi\\[-1pt]
{\scriptsize Step 8}\\[-1pt]
{\scriptsize Continue along the corridor\strut}
\end{minipage}\hfill
\begin{minipage}[t]{0.323\linewidth}
\centering
\ifnum\pdfstrcmp{S16}{S16}=0
\includegraphics[width=\linewidth,trim=0 62 210 56,clip]{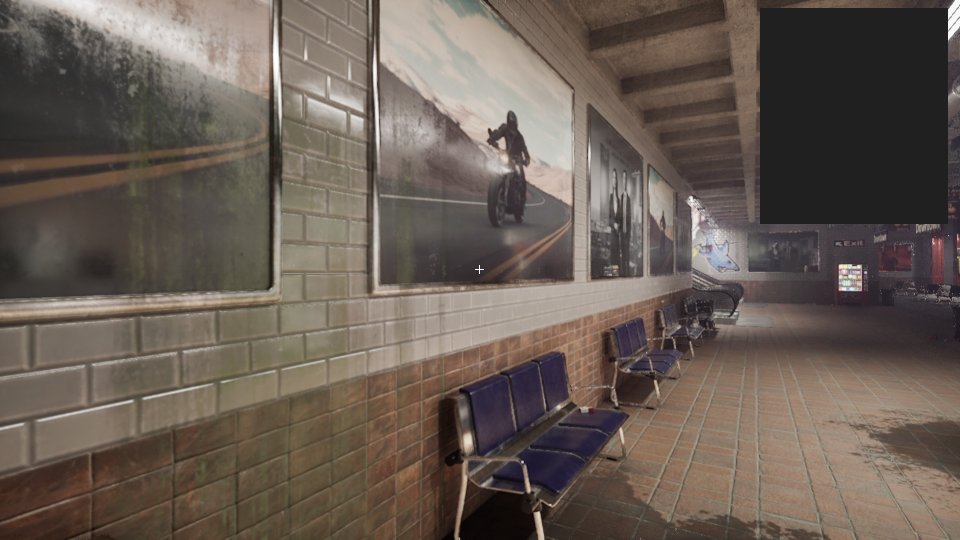}%
\else
\includegraphics[width=\linewidth]{assets/qualitative/S16/a017.jpg}%
\fi\\[-1pt]
{\scriptsize Step 17}\\[-1pt]
{\scriptsize Earlier corridor view\strut}
\end{minipage}\hfill
\begin{minipage}[t]{0.323\linewidth}
\centering
\ifnum\pdfstrcmp{S16}{S16}=0
\includegraphics[width=\linewidth,trim=0 62 210 56,clip]{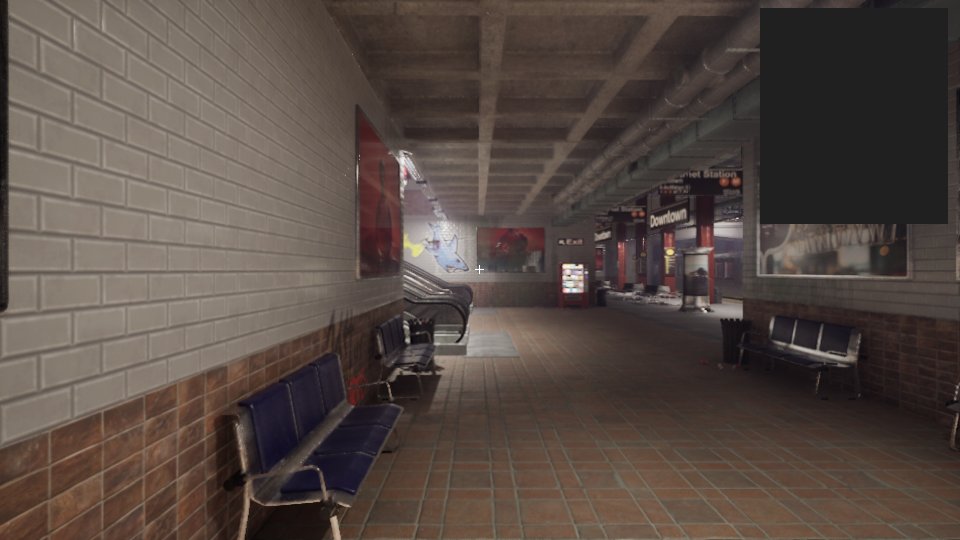}%
\else
\includegraphics[width=\linewidth]{assets/qualitative/S16/a023.jpg}%
\fi\\[-1pt]
{\scriptsize Step 23}\\[-1pt]
{\scriptsize Explore farther inside\strut}
\end{minipage}\par\vspace{7pt}
\begin{minipage}[t]{0.323\linewidth}
\centering
\ifnum\pdfstrcmp{S16}{S16}=0
\includegraphics[width=\linewidth,trim=0 62 210 56,clip]{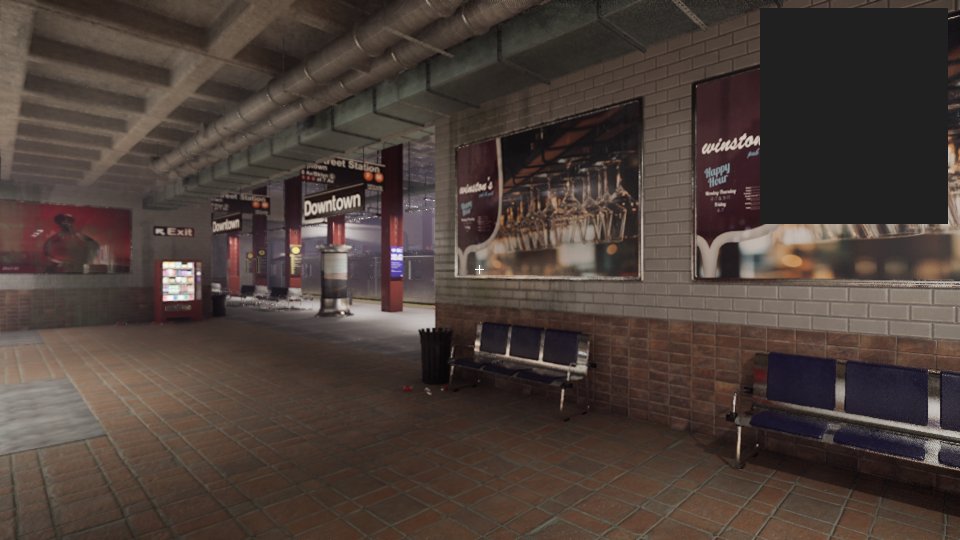}%
\else
\includegraphics[width=\linewidth]{assets/qualitative/S16/a026.jpg}%
\fi\\[-1pt]
{\scriptsize Step 26}\\[-1pt]
{\scriptsize Explore the far corridor\strut}
\end{minipage}\hfill
\begin{minipage}[t]{0.323\linewidth}
\centering
\ifnum\pdfstrcmp{S16}{S16}=0
\includegraphics[width=\linewidth,trim=0 62 210 56,clip]{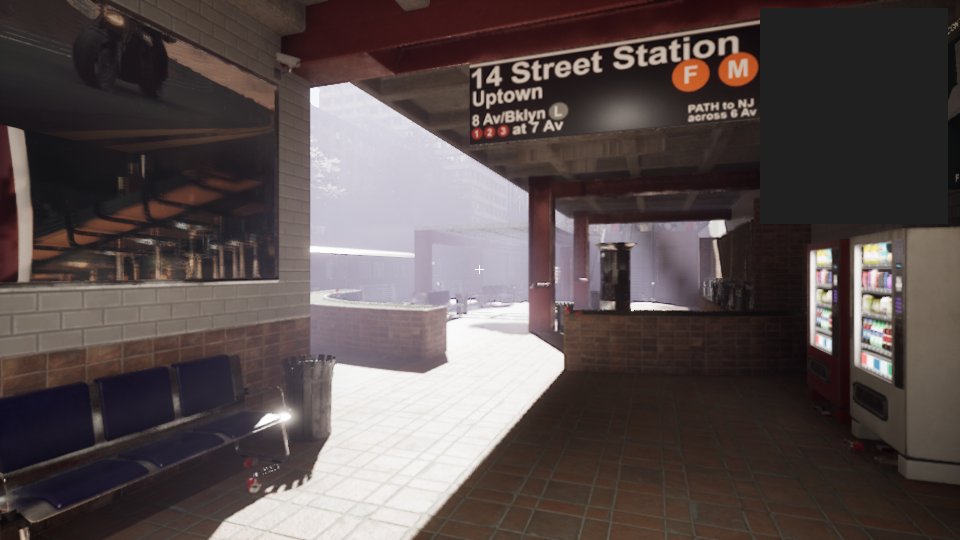}%
\else
\includegraphics[width=\linewidth]{assets/qualitative/S16/a030.jpg}%
\fi\\[-1pt]
{\scriptsize Step 30}\\[-1pt]
{\scriptsize Return along the corridor\strut}
\end{minipage}\hfill
\begin{minipage}[t]{0.323\linewidth}
\centering
\ifnum\pdfstrcmp{S16}{S16}=0
\includegraphics[width=\linewidth,trim=0 62 210 56,clip]{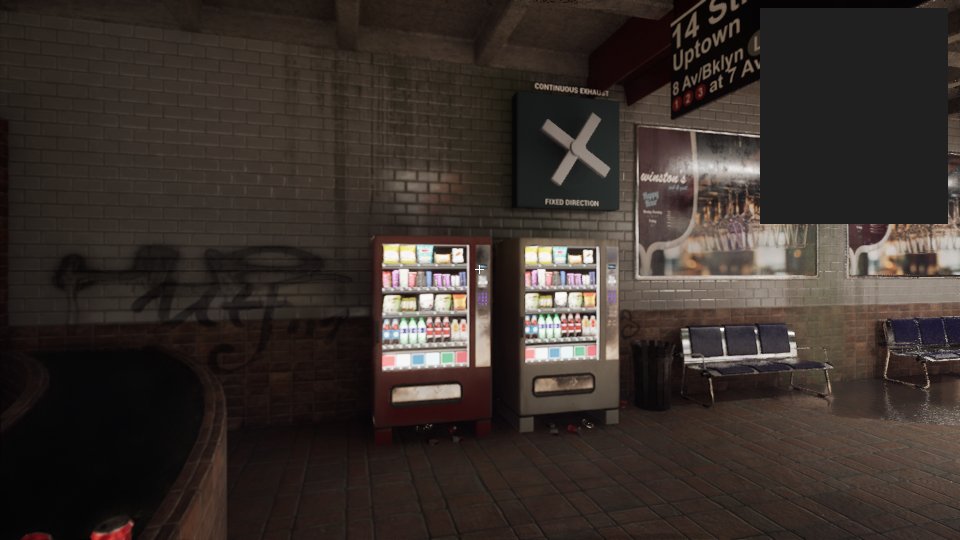}%
\else
\includegraphics[width=\linewidth]{assets/qualitative/S16/a035.jpg}%
\fi\\[-1pt]
{\scriptsize Step 35}\\[-1pt]
{\scriptsize Posters show a wine advert\strut}
\end{minipage}\par\vspace{7pt}
\begin{minipage}[t]{0.323\linewidth}
\centering
\ifnum\pdfstrcmp{S16}{S16}=0
\includegraphics[width=\linewidth,trim=0 62 210 56,clip]{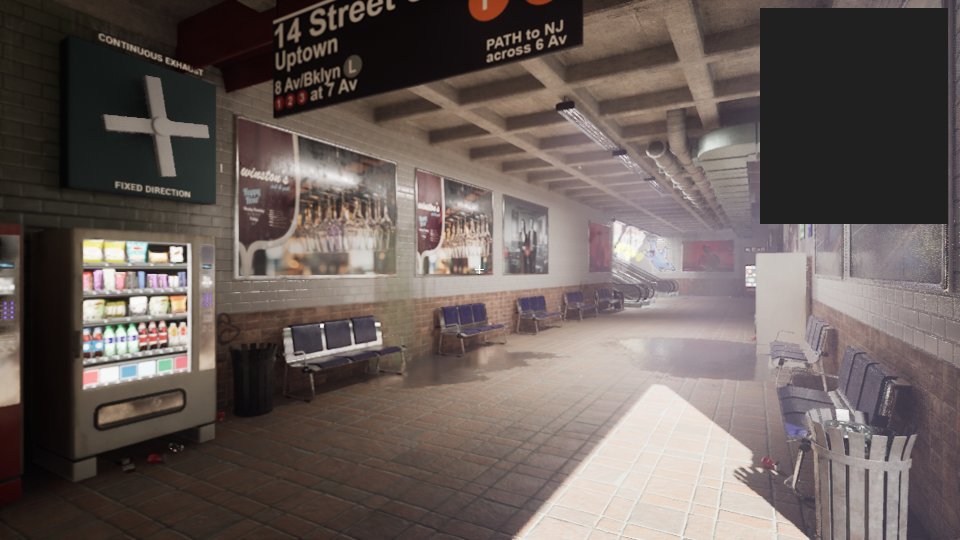}%
\else
\includegraphics[width=\linewidth]{assets/qualitative/S16/a036.jpg}%
\fi\\[-1pt]
{\scriptsize Step 36}\\[-1pt]
{\scriptsize Inspect the later posters\strut}
\end{minipage}\hfill
\begin{minipage}[t]{0.323\linewidth}
\centering
\ifnum\pdfstrcmp{S16}{S16}=0
\includegraphics[width=\linewidth,trim=0 62 210 56,clip]{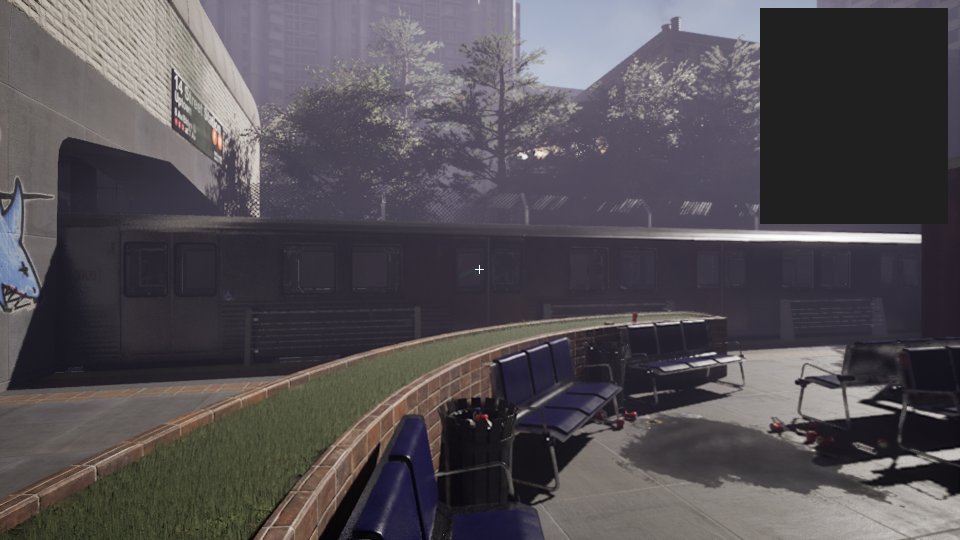}%
\else
\includegraphics[width=\linewidth]{assets/qualitative/S16/a039.jpg}%
\fi\\[-1pt]
{\scriptsize Step 39}\\[-1pt]
{\scriptsize Return to the plaza\strut}
\end{minipage}\hfill
\begin{minipage}[t]{0.323\linewidth}
\centering
\ifnum\pdfstrcmp{S16}{S16}=0
\includegraphics[width=\linewidth,trim=0 62 210 56,clip]{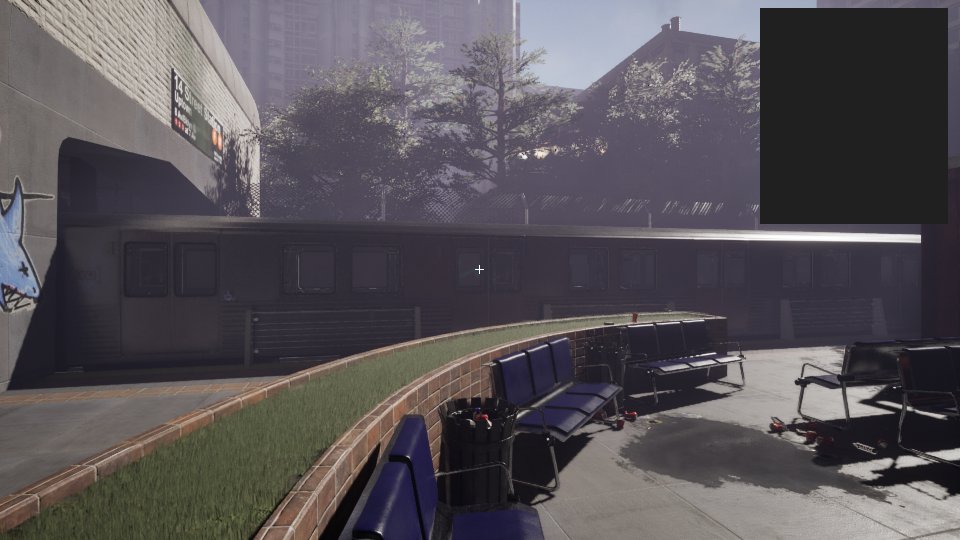}%
\else
\includegraphics[width=\linewidth]{assets/qualitative/S16/a040.jpg}%
\fi\\[-1pt]
{\scriptsize Step 40}\\[-1pt]
{\scriptsize Final observation\strut}
\end{minipage}\par\vspace{7pt}
\caption{Selected observations while inspecting the wall posters, ordered left to right and top to bottom.}\label{fig:qual-S16}
\end{figure}
\FloatBarrier

\noindent\textbf{Target anomaly.} One wall poster disappears after a short time without being touched; the rubric requests observing the two posters for about 15 seconds.

\noindent\textbf{Auditor report.} The auditor reports that two framed advertisements change from motorcycle imagery to a pub advertisement after leaving and returning. It cites the earlier views at Steps 4--5 and the later views at Steps 35--36 as evidence.

\noindent\textbf{Interpretation.} The auditor mistakes normal advertisement cycling for an anomaly. It compares changing content on the advertising displays rather than establishing the disappearance of the target poster. This produces a false positive while leaving the actual anomaly unidentified: detecting a visual change is insufficient without judging whether that change is expected.

\clearpage
\phantomsection\label{app:qual-S01}
\noindent\textbf{Failure: Misidentifying The Anomalous Object}\par\nobreak\smallskip
\noindent Static physics $\mid$ 40-action budget

\begin{figure}[!ht]
\centering
\begin{minipage}[t]{0.323\linewidth}
\centering
\ifnum\pdfstrcmp{S01}{S16}=0
\includegraphics[width=\linewidth,trim=0 62 210 56,clip]{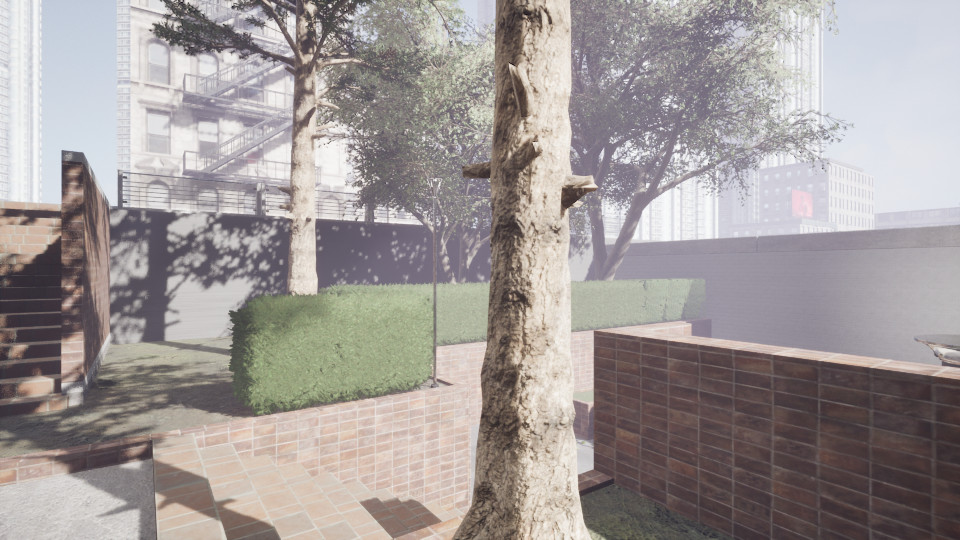}%
\else
\includegraphics[width=\linewidth]{assets/qualitative/S01/a000.jpg}%
\fi\\[-1pt]
{\scriptsize Start}\\[-1pt]
{\scriptsize Initial plaza view\strut}
\end{minipage}\hfill
\begin{minipage}[t]{0.323\linewidth}
\centering
\ifnum\pdfstrcmp{S01}{S16}=0
\includegraphics[width=\linewidth,trim=0 62 210 56,clip]{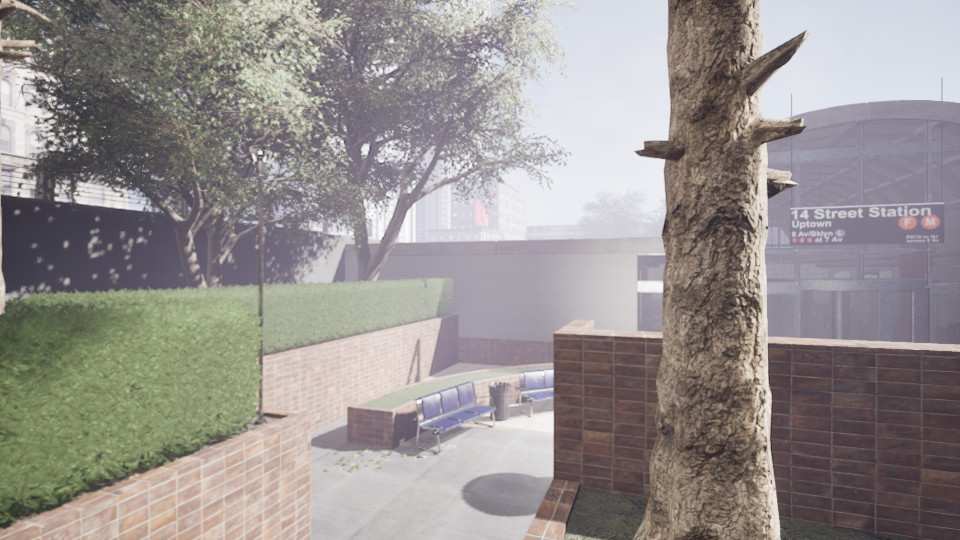}%
\else
\includegraphics[width=\linewidth]{assets/qualitative/S01/a011.jpg}%
\fi\\[-1pt]
{\scriptsize Step 11}\\[-1pt]
{\scriptsize Explore the lower plaza\strut}
\end{minipage}\hfill
\begin{minipage}[t]{0.323\linewidth}
\centering
\ifnum\pdfstrcmp{S01}{S16}=0
\includegraphics[width=\linewidth,trim=0 62 210 56,clip]{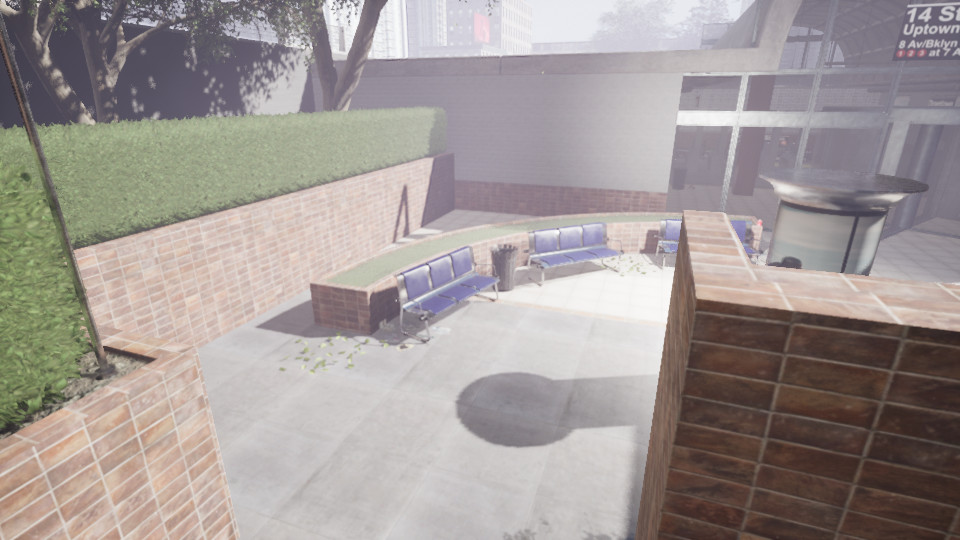}%
\else
\includegraphics[width=\linewidth]{assets/qualitative/S01/a013.jpg}%
\fi\\[-1pt]
{\scriptsize Step 13}\\[-1pt]
{\scriptsize Look across the concourse\strut}
\end{minipage}\par\vspace{7pt}
\begin{minipage}[t]{0.323\linewidth}
\centering
\ifnum\pdfstrcmp{S01}{S16}=0
\includegraphics[width=\linewidth,trim=0 62 210 56,clip]{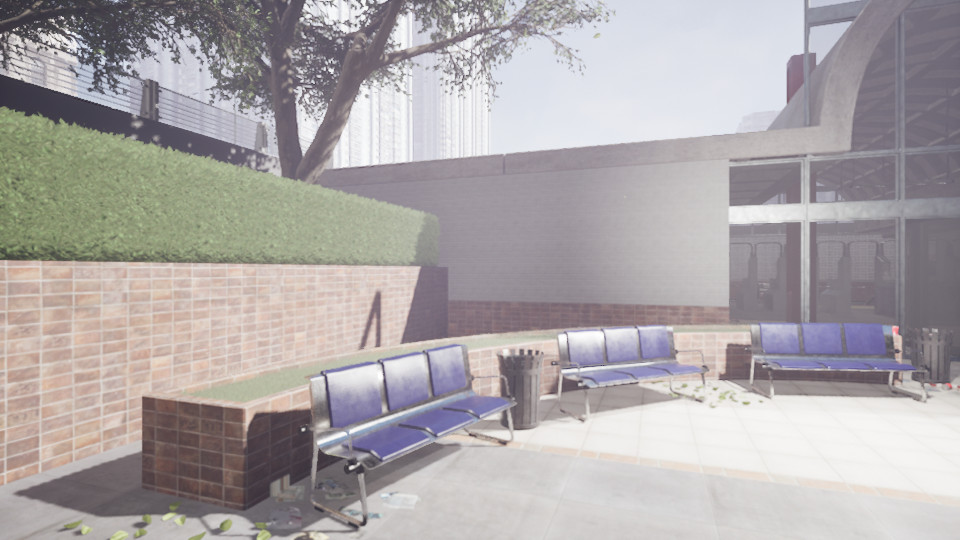}%
\else
\includegraphics[width=\linewidth]{assets/qualitative/S01/a015.jpg}%
\fi\\[-1pt]
{\scriptsize Step 15}\\[-1pt]
{\scriptsize Approach the seating area\strut}
\end{minipage}\hfill
\begin{minipage}[t]{0.323\linewidth}
\centering
\ifnum\pdfstrcmp{S01}{S16}=0
\includegraphics[width=\linewidth,trim=0 62 210 56,clip]{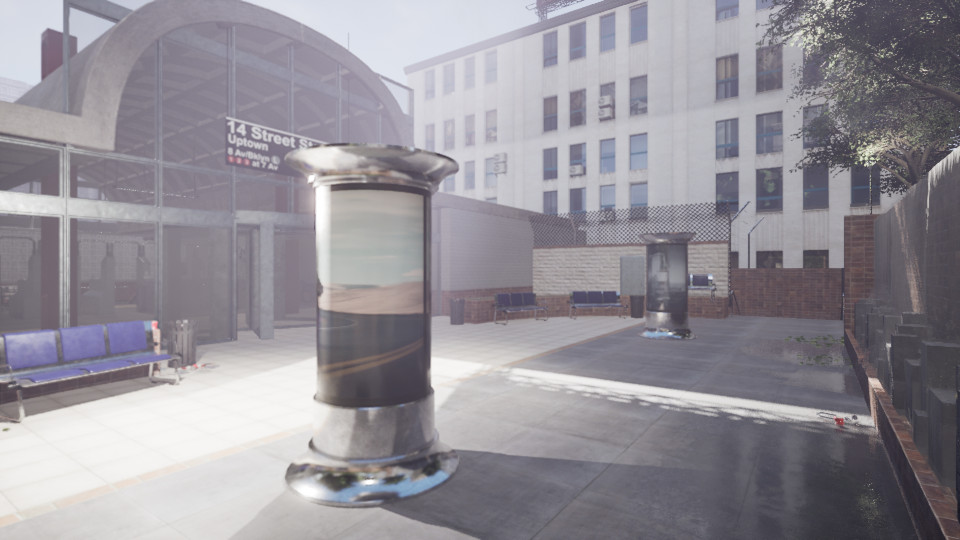}%
\else
\includegraphics[width=\linewidth]{assets/qualitative/S01/a017.jpg}%
\fi\\[-1pt]
{\scriptsize Step 17}\\[-1pt]
{\scriptsize Inspect the advertising column\strut}
\end{minipage}\hfill
\begin{minipage}[t]{0.323\linewidth}
\centering
\ifnum\pdfstrcmp{S01}{S16}=0
\includegraphics[width=\linewidth,trim=0 62 210 56,clip]{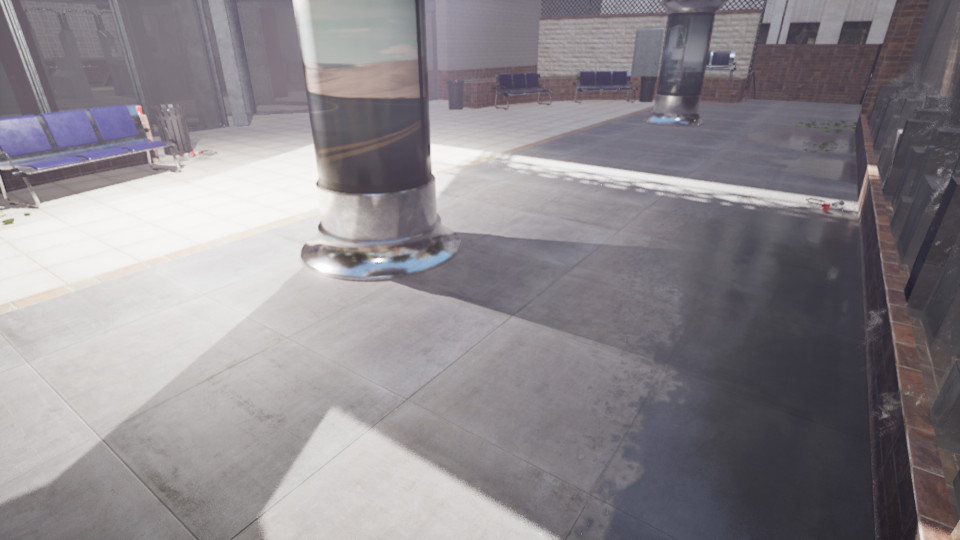}%
\else
\includegraphics[width=\linewidth]{assets/qualitative/S01/a018.jpg}%
\fi\\[-1pt]
{\scriptsize Step 18}\\[-1pt]
{\scriptsize Look down at its base\strut}
\end{minipage}\par\vspace{7pt}
\begin{minipage}[t]{0.323\linewidth}
\centering
\ifnum\pdfstrcmp{S01}{S16}=0
\includegraphics[width=\linewidth,trim=0 62 210 56,clip]{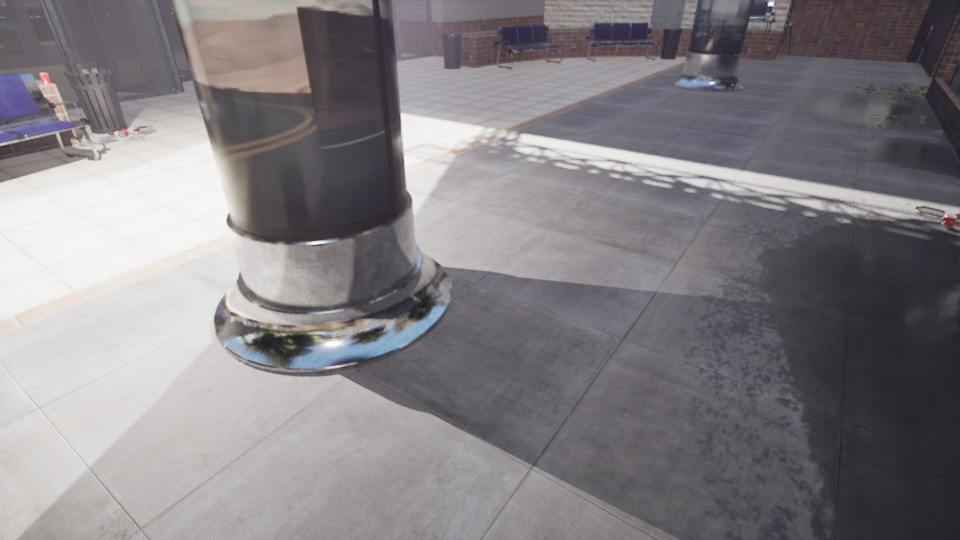}%
\else
\includegraphics[width=\linewidth]{assets/qualitative/S01/a021.jpg}%
\fi\\[-1pt]
{\scriptsize Step 21}\\[-1pt]
{\scriptsize Move closer to the base\strut}
\end{minipage}\hfill
\begin{minipage}[t]{0.323\linewidth}
\centering
\ifnum\pdfstrcmp{S01}{S16}=0
\includegraphics[width=\linewidth,trim=0 62 210 56,clip]{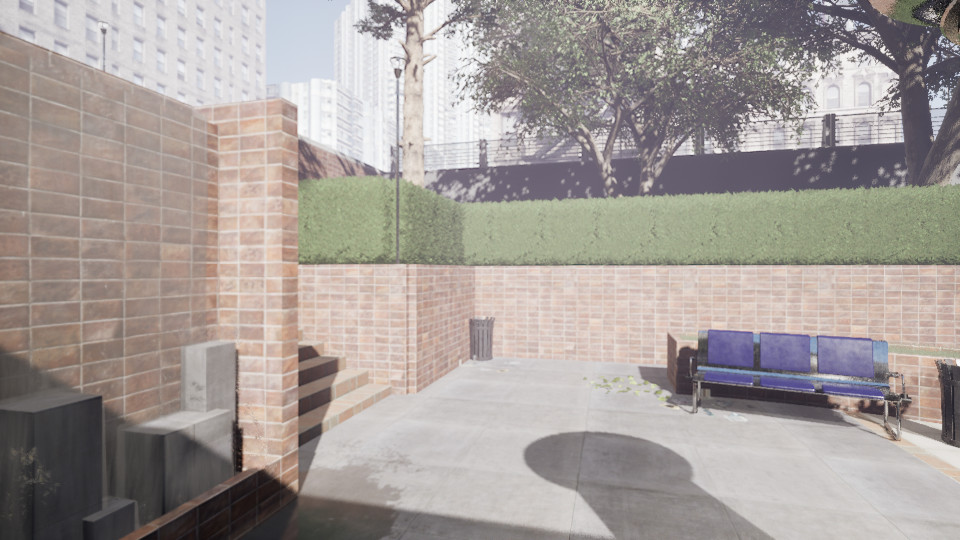}%
\else
\includegraphics[width=\linewidth]{assets/qualitative/S01/a025.jpg}%
\fi\\[-1pt]
{\scriptsize Step 25}\\[-1pt]
{\scriptsize View the nearby seating\strut}
\end{minipage}\hfill
\begin{minipage}[t]{0.323\linewidth}
\centering
\ifnum\pdfstrcmp{S01}{S16}=0
\includegraphics[width=\linewidth,trim=0 62 210 56,clip]{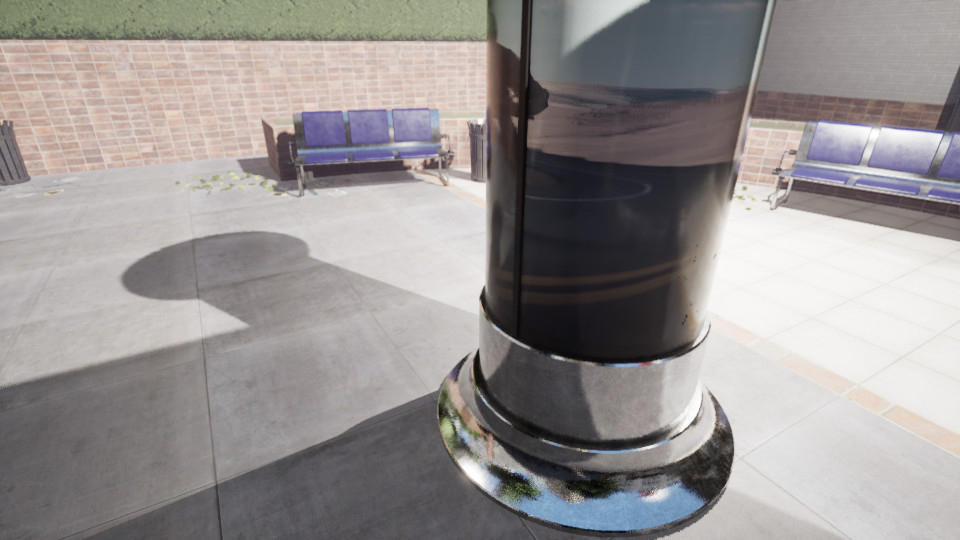}%
\else
\includegraphics[width=\linewidth]{assets/qualitative/S01/a027.jpg}%
\fi\\[-1pt]
{\scriptsize Step 27}\\[-1pt]
{\scriptsize Recheck the column base\strut}
\end{minipage}\par\vspace{7pt}
\begin{minipage}[t]{0.323\linewidth}
\centering
\ifnum\pdfstrcmp{S01}{S16}=0
\includegraphics[width=\linewidth,trim=0 62 210 56,clip]{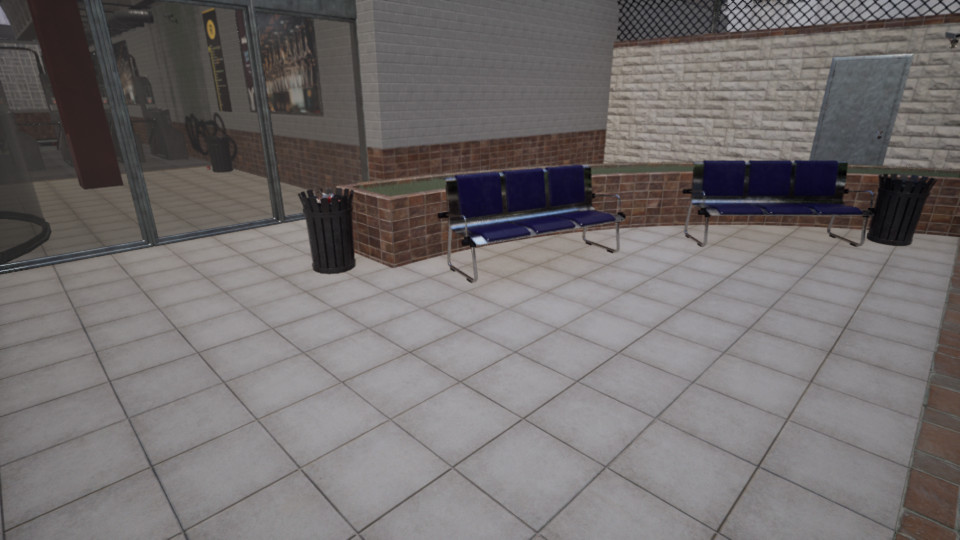}%
\else
\includegraphics[width=\linewidth]{assets/qualitative/S01/a035.jpg}%
\fi\\[-1pt]
{\scriptsize Step 35}\\[-1pt]
{\scriptsize Inspect the seating area\strut}
\end{minipage}\hfill
\begin{minipage}[t]{0.323\linewidth}
\centering
\ifnum\pdfstrcmp{S01}{S16}=0
\includegraphics[width=\linewidth,trim=0 62 210 56,clip]{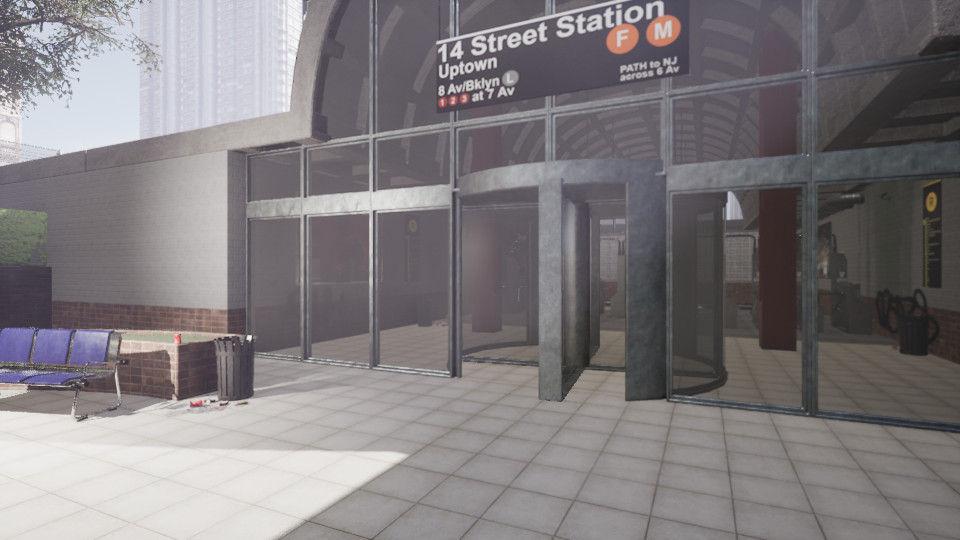}%
\else
\includegraphics[width=\linewidth]{assets/qualitative/S01/a037.jpg}%
\fi\\[-1pt]
{\scriptsize Step 37}\\[-1pt]
{\scriptsize View a nearby bench\strut}
\end{minipage}\hfill
\begin{minipage}[t]{0.323\linewidth}
\centering
\ifnum\pdfstrcmp{S01}{S16}=0
\includegraphics[width=\linewidth,trim=0 62 210 56,clip]{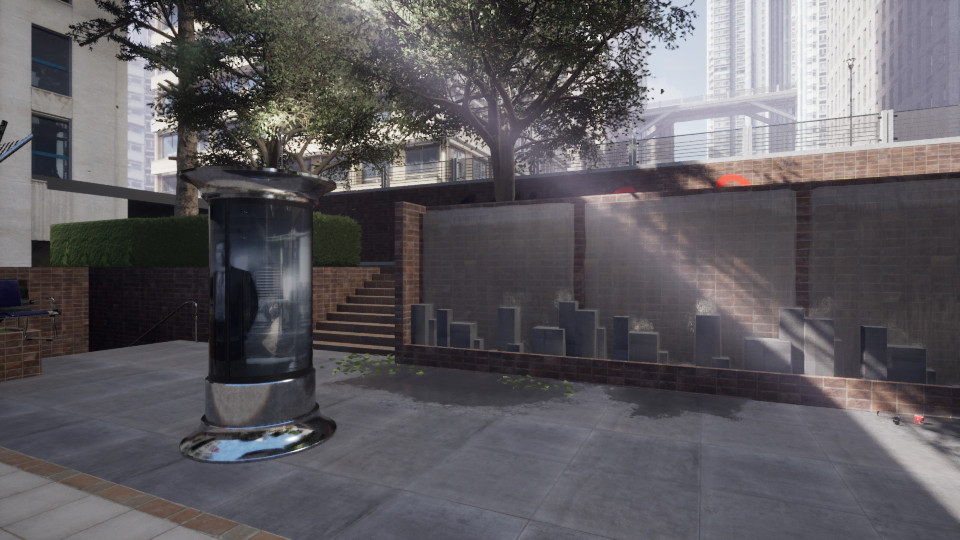}%
\else
\includegraphics[width=\linewidth]{assets/qualitative/S01/a040.jpg}%
\fi\\[-1pt]
{\scriptsize Step 40}\\[-1pt]
{\scriptsize Final observation\strut}
\end{minipage}\par\vspace{7pt}
\caption{Selected observations while inspecting the floating bench, ordered left to right and top to bottom.}\label{fig:qual-S01}
\end{figure}
\FloatBarrier

\noindent\textbf{Target anomaly.} An entire waiting bench floats approximately 60 centimetres above its support surface.

\noindent\textbf{Auditor report.} The auditor reports a floating cylindrical advertising column, citing views of its base in Steps 17--18, 21, and 27. It states that the other objects, including benches, rest properly on their surfaces.

\noindent\textbf{Interpretation.} The auditor misidentifies the advertising column as a floating object while overlooking the anomalous bench. Despite repeatedly inspecting the column base, it incorrectly reports the column and concludes that the benches are grounded. This case illustrates a false positive on the wrong object alongside a missed target anomaly.

\endgroup

\end{document}